\documentclass[12pt,a4paper,reqno,twoside,english]{book}
\pdfoutput=1
\usepackage[nohyperlinks]{acronym}
\usepackage{algorithm2e}
\usepackage{amsfonts}
\usepackage{amsmath}
\usepackage{amssymb}
\usepackage{appendix}
\usepackage{array}
\usepackage[catalan,english]{babel}
\usepackage{bigstrut}
\usepackage{blkarray}
\usepackage{booktabs}
\usepackage{cancel}
\usepackage{caption}
\usepackage{chngcntr}
\usepackage{comment}
\usepackage{courier}
\usepackage{enumitem}
\usepackage{epsfig}
\usepackage{epstopdf}
\usepackage{expl3}
\usepackage[shortcuts]{extdash}
\usepackage{fancyhdr}
\usepackage{float}
\usepackage[top=3.34cm, bottom=3.09cm, outer=2.47cm, inner=3.08cm]{geometry} 
\usepackage{graphicx}
\makeatletter
\let\c@lofdepth\relax
\let\c@lotdepth\relax
\makeatother
\usepackage{graphics,subfigure}
\usepackage{hhline}
\usepackage[utf8]{inputenc}
\usepackage{makecell}
\usepackage{mdframed}
\usepackage{multicol}
\usepackage{multirow}
\usepackage{natbib}
\usepackage[super]{nth}
\usepackage{pifont}
\usepackage{placeins}
\usepackage{ragged2e}
\usepackage[counterclockwise, figuresleft]{rotating}
\usepackage{soul}
\usepackage{tabularx}
\usepackage{tabu}
\usepackage[nottoc]{tocbibind}
\usepackage{url}
\usepackage[table,usenames,dvipsnames]{xcolor}
\usepackage{xfrac}
\usepackage{threeparttable}
\usepackage{tikz}
\usetikzlibrary{shapes.misc, fit,positioning}

\usepackage[pdftex,
			plainpages=true,
			pdfpagelabels,
            pdfauthor={Javier Juan Albarracín},
            pdftitle={},
            pdfsubject={Doctoral thesis},
            pdfkeywords={data quality, variability, statistics, probability, bioinformatics, medical informatics, data mining, information theory, information geometry, jensen-shannon, visual analytics, change detection, biomedical data},hidelinks]{hyperref}

\counterwithout{footnote}{chapter}

\newcolumntype{Y}{>{\centering\arraybackslash}X}

\newcolumntype{Z}{>{\raggedright\arraybackslash}X}
\newcolumntype{T}{>{\raggedleft\arraybackslash}X}

\fancypagestyle{plain}{%
  \fancyhf{}%
  \fancyfoot[RO, LE] {\thepage}
}

\newmdenv[topline=false, bottomline=false, linewidth=1pt]{siderules}

\hypersetup{pdfauthor={Javier Juan Albarracín}}

\DeclareMathOperator*{\argmin}{arg\,min}
\DeclareMathOperator*{\argmax}{arg\,max}
\renewcommand{\vec}[1]{\ensuremath{\mathbf{#1}}}
\newcommand{\Reals}{\mathbb{R}}
\newcommand{\Ti}{T$_1$}
\newcommand{\Tii}{T$_2$}
\newcommand{\Tiis}{T$_{2^*}$}
\newcommand{\Tic}{T$_{1CE}$}
\newcommand{\Tid}{T$_{1Diff}$}

\begin{document}

\begin{titlepage}
    \begin{center}
        \vspace*{1cm}
        
        \LARGE

        \textbf{Unsupervised learning for vascular\\heterogeneity assessment of glioblastoma\\based on magnetic resonance imaging:\\ The Hemodynamic Tissue Signature}
        
        \vspace{1.5cm}
        
		\includegraphics[width=0.2\textwidth]{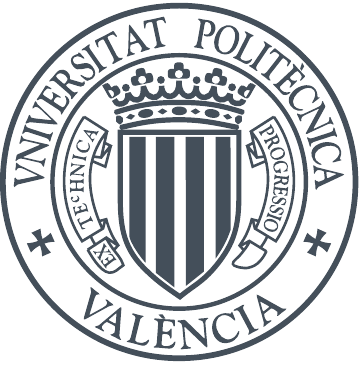}
        
        \normalsize
        {\sc Universitat Politècnica de València}\\
        {\sc Programa de Doctorado en \\Tecnologías para la Salud y el Bienestar}
        
        \vspace{1.5cm}
        {\sc DOCTORAL THESIS}
        
		\vfill		
		      
		Presented by\\
		\textbf{Javier Juan Albarracín}
        
        \vspace{0.8cm}
        
        Directed by\\
        Dr. Juan M García-Gómez\\
 		Dr. Elies Fuster i Garcia
 		
		\vspace{1.2cm}
		\normalsize
        Valencia, Spain\\
        March 2020
    \end{center}
\end{titlepage}

\thispagestyle{empty}

\frontmatter

\fancyhead[LE]{\slshape\small \nouppercase{Agradecimientos/Acknowledgements}}
\chapter*{Agradecimientos\\Acknowledgements}
\label{chapter:acknowledgements}
La finalización de esta tesis me brinda la oportunidad de reconocer a todas aquellas personas que, gracias al apoyo, el esfuerzo y la confianza incondicional que han depositado en mi, han hecho de este trabajo una realidad. A ellas quiero dedicar esta tesis.

\medskip

En primer lugar, quiero expresar mi más profundo agradecimiento a mis directores de tesis: el Dr. Juan Miguel García Gómez y el Dr. Elies Fuster i Garcia. Hace más de 8 años que me abrieron las puertas del grupo de investigación IBIME y apostaron por mí para desarrollar la línea de investigación de \emph{pattern recognition} en imagen médica. Hoy, los frutos del trabajo que llevamos realizando juntos durante todo este tiempo se ven reflejados en esta tesis que, de no haber sido por su certera dirección, su excelencia científica y su enorme calidad humana, no habría sido posible. Gracias por ofrecerme la oportunidad de trabajar a vuestro lado y de aprender de quienes considero referentes tanto a nivel profesional como personal.

\medskip

Quiero agradecer también a todos los compañeros y amigos del grupo IBIME que me han acompañado durante todos estos años en los que he tenido la oportunidad de desarrollar mi carrera científica y mi vida personal. Todos ellos han sido parte activa de este camino, contribuyendo con sus ideas, reflexiones y conocimiento al desarrollo esta tesis. Gracias a los que ahora formáis parte del grupo: el Dr. Carlos Sáez, el Dr. Jose Enrique Romero, Marta Durá, Vicent Blanes, Mari Alvarez, Pablo Ferri y Ángel Sánchez; y a los que formasteis parte en el pasado: el Dr. Salvador Tortajada, Miguel Esparza, el Dr. Adrián Bresó, Germán García y Alfonso Pérez, así como al Dr. Jose Vicente Manjón Herrera por su inestimable ayuda y su inagotable pasión por la ciencia. Gracias por servir de apoyo ante cualquier obstáculo y por los insuperables buenos momentos y risas que compartimos siempre en cualquier momento.

\medskip

La investigación de excelencia es hoy en día posible únicamente mediante la participación en proyectos de I+D+i tanto nacionales como internacionales, públicos o privados. En este sentido quiero agradecer a las diferentes instituciones y estructuras de financiación de investigación que han contribuido al desarrollo de esta tesis. En especial quiero agradecer a la Universitat Politècnica de València, donde he desarrollado toda mi carrera académica y científica, así como al Ministerio de Ciencia e Innovación, al Ministerio de Economía y Competitividad, a la Comisión Europea, al EIT Health Programme y a la fundación Caixa Impulse. También quiero agradecer a los hospitales y centros sanitarios que han participado en esta tesis aportando casos de estudio y acertados consejos para el buen desarrollo de este trabajo. Agradezco al Hospital Politècnico y Universitario La Fe, al Hospital de la Ribera, al Hospital de Manises, al Hospital Clínic de Barcelona, al Hospital Vall d'Hebrón, a l'Azienda Ospedaliero-Universitaria di Parma, al CHU de Liège y al Oslo University Hospital.

\medskip

Por último, quiero dedicar especialmente esta tesis a mi familia. A mi madre María Ángeles, que siempre me ha inculcado los valores de lucha y esfuerzo, de responsabilidad y constancia. Tu incansable empeño en educarme en estos valores han hecho de mi la persona que soy hoy. A mi hermano Eduardo, al que cuantos más años pasan más admiro. Gracias por estar a mi lado en los momentos más difíciles y guiarme en muchos aspectos de mi vida. Te has convertido en un espejo en el que mirarme cada día. Y por último a mi padre Rafael, que siempre me ha enseñado el valor de la honestidad y el respeto, de la excelencia y el sacrificio. A veces la vida es más dura con quien menos se lo merece, y aún en los peores momentos has demostrado ser siempre ser un referente, para mí y para todos. No puedo sentirme más orgulloso de tener esta familia.

\cleardoublepage

\fancyhead[LE]{\slshape\small \nouppercase{Abstract}}
\chapter{Abstract}
\label{chapter:abstract}
The future of medical imaging is linked to \ac{AI}. The manual analysis of medical images is nowadays an arduous, error-prone and often unaffordable task for humans, which has caught the attention of the \ac{ML} community. \ac{MRI}, which constitutes the standard imaging technique for the diagnosis of many lethal diseases, provides us with a wide variety of rich representations of the morphology and behavior of lesions completely inaccessible without a risky invasive intervention. Nevertheless, harnessing the powerful but often latent information contained in \ac{MRI} acquisitions is a very complicated task, which requires computational intelligent analysis techniques.

\medskip

Central nervous system tumors are one of the most critical diseases studied through \ac{MRI}. Specifically, glioblastoma represents a major challenge, as it remains a lethal cancer that, to date, lacks a satisfactory therapy. Of the entire set of characteristics that make glioblastoma so aggressive, a particular aspect that has been widely studied is its vascular heterogeneity. The strong vascular proliferation of glioblastomas, as well as their robust angiogenesis and extensive microvasculature heterogeneity have been claimed responsible for the high lethality of the neoplasm. Therefore, the study of these hallmarks is crucial to better understand the tumor's aggressiveness and design new effective therapies that improve patient prognosis.

\medskip

This thesis focuses on the research and development of the \ac{HTS} method: an unsupervised \ac{ML} approach to describe the vascular heterogeneity of glioblastomas by means of perfusion \ac{MRI} analysis. The \ac{HTS} builds on the concept of \emph{habitats}. A habitat is defined as a sub-region of the lesion with a particular \ac{MRI} profile describing a specific physiological behavior. The \ac{HTS} method delineates four habitats within the glioblastoma: the \ac{HAT} habitat, as the most perfused region of the enhancing tumor; the \ac{LAT} habitat, as the region of the enhancing tumor with a lower angiogenic profile; the potentially \ac{IPE} habitat, as the non-enhancing region adjacent to the tumor with elevated perfusion indexes; and the \ac{VPE} habitat, as the remaining edema of the lesion with the lowest perfusion profile. The research and development of the \ac{HTS} method has generated a number of contributions to this thesis.

\medskip

First, in order to verify that unsupervised learning methods are reliable to extract \ac{MRI} patterns to describe the heterogeneity of a lesion, a comparison among several structured and non-structured unsupervised learning methods was conducted for the task of high grade glioma segmentation. Additionally a generic postprocessing stage was also developed to automatically map each label of an unsupervised segmentation to a healthy or pathological tissue of the brain.

Second, a Bayesian unsupervised learning algorithm from the family of \acp{SVFMM} is proposed. The algorithm, named \ac{NLSVFMM}, successfully integrates a continuous Gauss-\ac{MRF} prior density weighted by the probabilistic \ac{NLM} weighting function, to codify the idea that neighboring pixels tend to belong to the same semantic object. The proposed prior simultaneously enforces local smoothness on the segmentations, while preserves the edges and the structure between classes.

Third, the \ac{HTS} method to describe the vascular heterogeneity of glioblastomas through the aforementioned habitats is presented. The \ac{HTS} method has been applied to real cases, both in a local cohort of patients from a single-center, and in an international retrospective cohort of more than 180 patients from 7 European centers. A comprehensive evaluation of the method was conducted to measure the prognostic potential of the \ac{HTS} habitats, as well as their stratification capabilities to identify populations with different prognosis. Statistically significant associations were found between most of \ac{HTS} habitats and \ac{OS} of patients, as well as significant differences were observed in survival rates of sub-populations divided according to \ac{HTS} derived measurements.

Finally, the methods and technology developed in this thesis have been integrated into an online public open-access platform for its academic use. The ONCOhabitats platform is hosted at \url{https://www.oncohabitats.upv.es}, and provides two main services: 1) glioblastoma tissue segmentation, and 2) vascular heterogeneity assessment of glioblastomas by means of the \ac{HTS} method. Both services, in addition to preprocessed images and segmentation maps, automatically generate a radiological report, summarizing the findings of the study. ONCOhabitats not only offers the scientific and medical community access to leading-edge algorithms for the analysis of these tumors, but gives access to its computational cluster capable to process about 300 cases per day.

\medskip

The results of this thesis have been published in ten scientific contributions, including top-ranked journals and conferences in the areas of Medical Informatics, Statistics and Probability, Radiology \& Nuclear Medicine, Machine Learning and Data Mining and Biomedical Engineering. An industrial patent registered in Spain (ES201431289A), Europe (EP3190542A1) and EEUU (US20170287133A1) was also issued, summarizing the efforts of this thesis to generate tangible assets besides the academic revenue obtained from research publications. Finally, the methods, technologies and original ideas conceived in this thesis led to the foundation of ONCOANALYTICS CDX, a company framed into the business model of companion diagnostics for pharmaceutical compounds, thought as a vehicle to facilitate the industrialization of the ONCOhabitats technology.

\cleardoublepage

\fancyhead[LE]{\slshape\small \nouppercase{Resumen}}
\chapter{Resumen}
El futuro de la imagen médica está ligado a la Inteligencia Artificial (IA). El análisis manual de imágenes médicas es hoy en día una tarea ardua, propensa a errores y a menudo inasequible para los humanos, que ha llamado la atención de la comunidad de Aprendizaje Automático (AA). La Imagen por Resonancia Magnética (IRM), que constituye la técnica de imagen estándar para el diagnóstico de muchas enfermedades letales, nos proporciona una amplia y rica variedad de representaciones de la morfología y el comportamiento de lesiones completamente inaccesibles sin una intervención invasiva arriesgada. Sin embargo, explotar la potente pero a menudo latente información contenida en las adquisiciones de IRM es una tarea muy complicada, que requiere técnicas de análisis computacional inteligente.

\medskip

Los tumores del sistema nervioso central son una de las enfermedades más críticas estudiadas a través de IRM. Específicamente, el glioblastoma representa un gran desafío, ya que, hasta la fecha, continua siendo un cáncer letal que carece de una terapia satisfactoria. De todo el conjunto de características que hacen del glioblastoma un tumor tan agresivo, un aspecto particular que ha sido ampliamente estudiado es su heterogeneidad vascular. La fuerte proliferación vascular de los glioblastomas, así como su robusta angiogénesis y la extensa heterogeneidad de su microvasculatura han sido consideradas responsables de la alta letalidad de esta neoplasia. Por lo tanto, el estudio de estos factores es crucial para entender mejor la agresividad del tumor y diseñar nuevas terapias efectivas que mejoren el pronóstico del paciente.

\medskip

Esta tesis se centra en la investigación y desarrollo del método \acf{HTS}: un método de aprendizaje no supervisado para describir la heterogeneidad vascular de los glioblastomas mediante el análisis de perfusión por IRM. El método \ac{HTS} se basa en el concepto de \emph{hábitats}. Un hábitat se define como una subregión de la lesión con un perfil particular de IRM, que describe un comportamiento fisiológico concreto. El método \ac{HTS} delinea cuatro hábitats dentro del glioblastoma: el hábitat \acf{HAT}, como la región más perfundida del tumor con captación de contraste; el hábitat \acf{LAT}, como la región del tumor con captación de contraste con un perfil angiogénico más bajo; el hábitat \acf{IPE}, como la región edematosa sin captación de contraste adyacente al tumor con índices de perfusión elevados; y el hábitat \acf{VPE}, como el edema restante de la lesión con el perfil de perfusión más bajo. La investigación y desarrollo del método \ac{HTS} ha originado una serie de contribuciones enmarcadas en esta tesis.

\medskip

En primer lugar, para verificar que los métodos de aprendizaje no supervisados son fiables a la hora de extraer patrones de IRM para describir la heterogeneidad de una lesión, se realizó una comparación entre varios métodos de aprendizaje estructurado y no estructurado no supervisados en la tarea de segmentación de gliomas de grado alto. Adicionalmente, se desarrolló un método genérico de postproceso para mapear automáticamente cada etiqueta de una segmentación no supervisada a un tejido sano o patológico del cerebro.

En segundo lugar, se ha propuesto un algoritmo de aprendizaje Bayesiano no supervisado dentro de la familia de los \acfp{SVFMM}. El algoritmo, llamado \acf{NLSVFMM}, integra con éxito un Gauss-\acf{MRF} continuo ponderado por la función probabilística \acf{NLM} como densidad a priori del modelo, para codificar la idea de que los píxeles vecinos tienden a pertenecer al mismo objeto semántico. La probabilidad a priori propuesta refuerza simultáneamente la suavidad local en las segmentaciones, a la vez que preserva los bordes y la estructura entre clases.

En tercer lugar, se presenta el método \ac{HTS} para describir la heterogeneidad vascular de los glioblastomas mediante los hábitats mencionados. El método \ac{HTS} se ha aplicado a casos reales, tanto en una cohorte local de pacientes de un solo centro, como en una cohorte retrospectiva internacional de más de 180 pacientes de 7 centros europeos. Se llevó a cabo una evaluación exhaustiva del método para medir el potencial pronóstico de los hábitats, así como las capacidades de estratificación de los mismos para identificar poblaciones con pronósticos diferentes. Se encontraron asociaciones estadísticamente significativas entre la mayoría de los hábitats \ac{HTS} y la supervivencia global de los pacientes, así como diferencias significativas en las tasas de supervivencia de subpoblaciones divididas según mediciones derivadas del \ac{HTS}.

Finalmente, los métodos y la tecnología desarrollados en esta tesis se han integrado en una plataforma web \emph{online} de acceso público para su uso académico. La plataforma ONCOhabitats se aloja en \url{https://www.oncohabitats.upv.es}, y ofrece dos servicios principales: 1) segmentación de tejidos de glioblastoma, y 2) evaluación de la heterogeneidad vascular de los glioblastomas mediante el método \ac{HTS}. Ambos servicios, además de las imágenes preprocesadas y los mapas de segmentación, generan automáticamente un informe radiológico resumiendo los hallazgos del estudio. ONCOhabitats no sólo ofrece a la comunidad científica y médica acceso a algoritmos del estado del arte para el análisis de estos tumores, sino que también permite acceder a su clúster computacional, capaz de procesar cerca de 300 casos al día.

\medskip

Los resultados de esta tesis han sido publicados en diez contribuciones científicas, incluyendo revistas y conferencias de primer nivel en las áreas de Informática Médica, Estadística y Probabilidad, Radiología y Medicina Nuclear, Aprendizaje Automático y Minería de Datos e Ingeniería Biomédica. También se emitió una patente industrial registrada en España (ES201431289A), Europa (EP3190542A1) y EEUU (US20170287133A1), que representa los esfuerzos de esta tesis para generar activos tangibles además de los méritos académicos obtenidos de las publicaciones de investigación. Finalmente, los métodos, tecnologías e ideas originales concebidas en esta tesis dieron lugar a la creación de ONCOANALYTICS CDX, una empresa enmarcada en el modelo de negocio de los \emph{companion diagnostics} de compuestos farmacéuticos, pensado como vehículo para facilitar la industrialización de la tecnología ONCOhabitats.

\cleardoublepage

\fancyhead[LE]{\slshape\small \nouppercase{Resum}}
\chapter{Resum}
\selectlanguage{catalan}
El futur de la imatge mèdica està lligat a la Intel·ligència Artificial (IA). L'anàlisi manual d'imatges mèdiques és hui dia una tasca àrdua, propensa a errors i sovint inassequible per als humans, que ha cridat l'atenció de la comunitat d'Aprenentatge Automàtic (AA). La Imatge per Ressonància Magnètica (IRM), que constitueix la tècnica d'imatge estàndard per al diagnòstic de moltes malalties letals, ens proporciona una àmplia i rica varietat de representacions de la morfologia i el comportament de lesions completament inaccessibles sense una intervenció invasiva arriscada. Tanmateix, explotar la potent però sovint latent informació continguda a les adquisicions de IRM esdevé una tasca molt complicada, que requereix tècniques d'anàlisi computacional intel·ligent.

\medskip

Els tumors del sistema nerviós central són una de les malalties més crítiques estudiades a través de IRM. Específicament, el glioblastoma representa un gran repte, ja que, fins hui, continua siguent un càncer letal que manca d'una teràpia satisfactòria. Dintre del conjunt de característiques que fan del glioblastoma un tumor tan agressiu, un aspecte particular que ha sigut àmpliament estudiat és la seua heterogeneïtat vascular. La forta proliferació vascular dels glioblastomes, així com la seua robusta angiogènesi i l'extensa heterogeneïtat de la seua microvasculatura han sigut considerades responsables de l'alta letalitat d'aquesta neoplàsia. Per tant, l'estudi d'aquests factors esdevé crucial per entendre millor l'agressivitat del tumor i dissenyar noves teràpies efectives que milloren el pronòstic del pacient.

\medskip

Aquesta tesi es centra en la recerca i desenvolupament del mètode \acf{HTS}: un mètode d'aprenentatge no supervisat per descriure l'heterogeneïtat vascular dels glioblastomas mitjançant l'anàlisi de perfusió per IRM. El mètode \ac{HTS} es basa en el concepte d'\emph{hàbitats}. Un hàbitat es defineix com una subregió de la lesió amb un perfil particular d'IRM, que descriu un comportament fisiològic concret. El mètode \ac{HTS} delinea quatre hàbitats dins del glioblastoma: l'hàbitat \acf{HAT}, com la regió més perfosa del tumor amb captació de contrast; l'hàbitat \acf{LAT}, com la regió del tumor amb captació de contrast amb un perfil angiogènic més baix; l'hàbitat \acf{IPE}, com la regió edematosa sense captació de contrast adjacent al tumor amb índexs de perfusió elevats, i l'hàbitat \acf{VPE}, com l'edema restant de la lesió amb el perfil de perfusió més baix. La recerca i desenvolupament del mètode \ac{HTS} ha originat una sèrie de contribucions emmarcades a aquesta tesi.

\medskip

En primer lloc, per verificar que els mètodes d'aprenentatge no supervisats són fiables a l'hora d'extraure patrons d'IRM per descriure l'he\-te\-ro\-ge\-ne\-ï\-tat d'una lesió, es va realitzar una comparació entre diversos mètodes d'aprenentatge estructurat i no estructurat no supervisats en la tasca de segmentació de gliomes de grau alt. Addicionalment, es va desenvolupar un mètode genèric de post-processament per mapejar automàticament cada etiqueta d'una segmentació no supervisada a un teixit sa o patològic del cervell.

En segon lloc, s'ha proposat un algorisme d'aprenentatge Bayesià no supervisat dintre de la família dels \acfp{SVFMM}. L'algorisme, anomenat \acf{NLSVFMM}, integra amb èxit un Gauss-\acf{MRF} continu ponderat per la funció probabilística \acf{NLM} com a densitat a priori del model, per a codificar la idea que els píxels veïns tendeixen a pertànyer al mateix objecte semàntic. La probabilitat a priori proposada reforça simultàniament la suavitat local en les segmentacions, alhora que preserva les vores i l'estructura entre classes.

En tercer lloc, es presenta el mètode \ac{HTS} per descriure l'heterogeneïtat vascular dels glioblastomas mitjançant els hàbitats esmentats. El mètode \ac{HTS} s'ha aplicat a casos reals, tant en una cohort local de pacients d'un sol centre, com en una cohort retrospectiva internacional de més de 180 pacients de 7 centres europeus. Es va dur a terme una avaluació exhaustiva del mètode per mesurar el potencial pronòstic dels hàbitats, així com les capacitats d'estratificació dels mateixos per identificar poblacions amb pronòstics diferents. Es van trobar associacions estadísticament significatives entre la majoria dels hàbitats \ac{HTS} i la supervivència global dels pacients, així com diferències significatives en les taxes de supervivència de sub-poblacions dividides segons mesuraments derivats de l'\ac{HTS}.

Finalment, els mètodes i la tecnologia desenvolupats en aquesta tesi s'han integrat en una plataforma web \emph{online} d'accés públic per al seu ús acadèmic. La plataforma ONCOhabitats s'allotja en \url{https://www.oncohabitats.upv.es}, i ofereix dos serveis principals: 1) segmentació dels teixits del glioblastoma, i 2) avaluació de l'heterogeneïtat vascular dels glioblastomes mitjançant el mètode \ac{HTS}. Ambdós serveis, a més de les imatges preprocessades i els mapes de segmentació, generen automàticament un informe radiològic resumint els descobriments de l'estudi. ONCOhabitats no sols ofereix a la comunitat científica i mèdica accés a algorismes de l'estat de l'art per l'anàlisi d'aquests tumors, sinó que també permet accedir al seu clúster computacional, capaç de processar prop de 300 casos al dia.

\medskip

Els resultats d'aquesta tesi han sigut publicats en deu contribucions científiques, incloent revistes i conferències de primer nivell a les àrees d'Informàtica Mèdica, Estadística i Probabilitat, Radiologia i Medicina Nuclear, Aprenentatge Automàtic i Mineria de Dades i Enginyeria Biomèdica. També es va emetre una patent industrial registrada a Espanya (ES201431289A), Europa (EP3190542A1) i els EUA (US20170287133A1), que representa els esforços d'aquesta tesi per generar actius tangibles a més dels mèrits acadèmics obtinguts de les publicacions d'investigació. Finalment, els mètodes, tecnologies i idees originals concebudes en aquesta tesi van donar lloc a la creació d'ONCOANALYTICS CDX, una empresa emmarcada en el model de negoci dels \emph{companion diagnostics} de compostos farmacèutics, pensat com a vehicle per facilitar la industrialització de la tecnologia ONCOhabitats.

\selectlanguage{english}

\cleardoublepage

{\footnotesize
\fancyhead[LE]{\slshape\small \nouppercase{Glossary}}
\chapter{Glossary}

\section*{Mathematical notation}

\begin{tabular}{lp{13cm}}
$X$ & Random variable \\
$x$ & Particular realization of the random variable $X$ \\
$p\left( X \right)$ & Marginal probability density function of the random variable $X$ \\
$p\left( X,Y \right)$ & Joint probability density function of two random variables $X$ and $Y$ \\
$p\left( X | Y \right)$ & Conditional probability density function of the random variable $X$ conditioned to $Y$ \\
$\vec{\Theta}$ & Vector of parameters of a probability density function \\
$\hat{\vec{\Theta}}$ & Optimal vector of parameters under some optimization criteria \\
$\tilde{\vec{\Theta}}$ & Initial guess of a vector of parameters \\
$p\left( X ; \vec{\Theta} \right)$ & Probability density function of the random variable $X$ subject to $\vec{\Theta}$ \\
$\mathcal{L} \left( \vec{\Theta}; X \right)$ & Likelihood function of random variable $X$ given parameter vector $\vec{\Theta}$ \\
$\vec{x}$ & Column vector $\vec{x}$ \\
$\vec{x}^T$ & Transpose of $\vec{x}$\\
$\mathbb{E} \left[ p(X) \right]$ & Expectation of probability density function $p\left( X \right)$ \\
$\mathbb{E}_{p\left(Y \right)} \left[ p(X) \right]$ & Conditional expectation of probability density function $p\left( X \right)$ subject to $p\left( Y \right)$ \\
$\mathbb{R}^D$ & $D$-dimensional space of real numbers \\
$\frac{\partial x}{\partial t}$ & Partial derivative of variable $x$ with respect to variable $t$ \\
$\mathcal{M}^i$ & Set of neighbors of the $i^{th}$ observation \\
$\otimes$ & Convolution product \\
$\delta \left( \cdot \right)$ & Dirac delta function \\ 
$\|\cdot\|$ & Euclidean norm \\
$\{\cdot\}$ & Set of elements \\
\end{tabular}

\clearpage

\section*{Acronyms}
{\small
\begin{acronym}
\acro{AI}{Artificial Intelligence}
\acro{AIF}{Arterial Input Function}
\acro{ANN}{Artificial Neural Network}
\acro{ANTs}{Advanced Normalization Tools}
\acro{ASL}{Arterial Spin Labeling}
\acro{BDSLab}{Biomedical Data Science Laboratory}
\acro{BRATS}{BRAin Tumor Segmentation}
\acro{CAE}{Convolutional AutoEnconder}
\acro{CBF}{Cerebral Blood Flow}
\acro{CBICA}{Center for Biomedical Image Computing and Analytics}
\acro{CBV}{Cerebral Blood Volume}
\acro{cdf}{cumulative distribution function}
\acro{CDSS}{Clinical Decision Support System}
\acro{CDx}{Companion Diagnostic}
\acro{CNN}{Convolutional Neural Network}
\acro{CSF}{Cerebro-Spinal Fluid}
\acro{CRF}{Conditional Random Field}
\acro{DCAGMRF}{Directional Class-Adaptive Gauss-Markov Random Field}
\acro{DCE}{Dynamic Contrast Enhanced}
\acro{DCM}{Dirichlet Compound Multinomial}
\acro{DL}{Deep Learning}
\acro{DSC}{Dynamic Susceptibility Contrast}
\acro{DTI}{Diffusion Tensor Imaging}
\acro{DWI}{Diffusion Weighted Imaging}
\acro{EGFR}{Epidermal Growth Factor Receptor}
\acro{EM}{Expectation-Maximization}
\acro{ET}{Enhancing Tumor}
\acro{FLAIR}{Fluid Attenuation Inversion Recovery}
\acro{FMM}{Finite Mixture Model}
\acro{FSE}{Fast Spin Echo}
\acro{GBCA}{Gadolinium-Based Contrast Agent}
\acro{GM}{Grey Matter}
\acro{GMM}{Gaussian Mixture Model}
\acro{HAC}{Hierarchical Agglomerative Clustering}
\acro{HAT}{High Angiogenic Tumor}
\acro{HMRF}{Hidden Markov Random Field}
\acro{HR}{Hazard Ratio}
\acro{HTS}{Hemodynamic Tissue Signature}
\acro{HUPLF}{Hospital Universitario y Politécnico La Fe}
\acro{ICBM}{International Consortium of Brain Mapping}
\acro{i.i.d.}{independent and identically distributed}
\acro{IPE}{Infiltrated Peripheral Edema}
\acro{LAT}{Low Angiogenic Tumor}
\acro{MAP}{Maximum A Posteriori}
\acro{MICCAI}{Medical Image Computing and Computer Assisted Intervention}
\acro{ML}{Machine Learning}
\acro{MLE}{Maximum Likelihood Estimation}
\acro{MLP}{Multi-Layer Perceptron}
\acro{MNI}{Montreal Neurological Institute}
\acro{MPRAGE}{Magnetization-Prepared Rapid Acquisition with Gradient Echo}
\acro{MR}{Magnetic Resonance}
\acro{MRF}{Markov Random Field}
\acro{MRI}{Magnetic Resonance Imaging}
\acro{MRSI}{Magnetic Resonance Spectroscopy Imaging}
\acro{MTT}{Mean Transit Time}
\acro{NLM}{Non Local Means}
\acro{NLSVFMM}{Non Local Spatially Varying Finite Mixture Model}
\acro{NMR}{Nuclear Magnetic Resonance}
\acro{OS}{Overall Survival}
\acro{PCA}{Principal Component Analysis}
\acro{PD}{Proton Density}
\acro{PFS}{Progression Free Survival}
\acro{PCT}{Patent Cooperation Treaty}
\acro{pdf}{probability density function}
\acro{pmf}{probability mass function}
\acro{PWI}{Perfusion Weighted Imaging}
\acro{qMRI}{Quantitative Magnetic Resonance Imaging}
\acro{rCBV}{relative Cerebral Blood Volume}
\acro{rCBF}{relative Cerebral Blood Flow}
\acro{RF}{Random Forest}
\acro{RI}{Rand Index}
\acro{RMSE}{Root Mean Squared Error}
\acro{ROI}{Region Of Interest}
\acro{SaaS}{Software as a Service}
\acro{SAR}{Simultaneous Auto-Regressive}
\acro{SOM}{Self-Organizing Map}
\acro{SVD}{Singular Value Decomposition}
\acro{SVFMM}{Spatially Varying Finite Mixture Model}
\acro{SVM}{Support Vector Machine}
\acro{TC}{Tumor Core}
\acro{TCIA}{The Cancer Imaging Archive}
\acro{TE}{Echo Time}
\acro{TR}{Repetition Time}
\acro{UAB}{University of Alabama at Birmingham}
\acro{UPGMA}{Unweighted Pair Group Method with Arithmetic mean}
\acro{UPV}{Universitat Polit\`ecnica de Val\`encia}
\acro{VPE}{Vasogenic Peripheral Edema}
\acro{WHO}{World Health Organization}
\acro{WM}{White Matter}
\acro{WT}{Whole Tumor}
\end{acronym}
}
}
\fancyhead[LE]{\slshape\small \nouppercase{Glossary}}

\cleardoublepage

\fancyhead[LE]{\slshape\small \nouppercase{\leftmark}}

\tableofcontents

\mainmatter


\chapter{Introduction}
\label{chapter:introduction}

\section{Motivation}
\label{section:motivation}
Medical imaging has widely proven to be an essential tool for modern medicine. The ability to non-invasively visualize in-vivo representations of the interior of the human body constituted an unprecedented breakthrough in the diagnosis, prognosis and follow-up processes of the diseases \citep{Mcrobbie2007}. The first medical image acquisition dates back to 1895 with the discovery of X-rays \citep{Rontgen1898}, however it was not until the end of the 20th century that medical imaging took a qualitative leap with the development of the \ac{MRI} \citep{Damadian1971, Lauterbur1973, Lauterbur1974, Mansfield1977}. Over the years, medical imaging has evolved rapidly, reaching sophisticated techniques capable of quantifying large amounts of information of the anatomy and functionality of human tissues. Consequently, the analysis of such complex information has become a specialized discipline that requires advanced computational techniques to make the most of the knowledge contained therein.

\medskip

In this context, \ac{ML} emerges as a solid candidate for medical image analysis. \ac{ML} is an application of \ac{AI} that provides systems the ability to learn and identify complex patterns in multi-dimensional data, and perform specific tasks without being explicitly programmed. The beginnings of \ac{ML} in medical imaging dates back to the middle of the 20th century with the birth of the expert systems \citep{Russell2016}. From that moment on, an unstoppable proliferation of \ac{ML} systems for multitude of clinical problems took place, reaching its first hype cycle peak at the end of 20th century. Today, with the advent of the \ac{DL} techniques, medical image analysis is undergoing a new deep revolution that is settling the \ac{ML} as an indisputable instrument in the modern clinical practice.

\medskip

However, \ac{ML} has historically addressed medical problems from the perspective of automatizing arduous complex tasks for humans. Tied to medical imaging, \ac{ML} has successfully addressed complicated problems such as identifying and delineating abnormal tissues in multiple images or classifying and grading lesions from medical acquisitions \citep{Azuaje2019, Levine2019, Kann2019}. Supervised learning, which is a family of techniques under the \ac{ML} umbrella, has led this approach with unquestionable effectiveness. However, due to its learning nature, supervised learning is only able to address problems where humans already know the answer \citep{Duda2000}. This family of techniques builds models by means of learning the relations between tuples of $\left< input~data,~desired~output \right>$, which necessarily requires that all algorithm's possible outputs are explicitly known. This approach, without underestimating its unquestionable usefulness, reduces the \ac{ML} to an instrument for solving and automatizing tasks with already well-known targets, with the added value of high accuracy, repeatability and reliability.

\medskip

On the contrary, we would like \ac{ML} to help us to discover new knowledge from the medical data beyond what humans can devise. At this point unsupervised learning arises as a tailor-made solution to this purpose. Unsupervised learning, unlike supervised learning, inspects unlabeled data for hidden patterns and inner relationships that describe the latent structure of the data \citep{Bishop2006}. This alternative approach has the intrinsic capability to discover new knowledge from the data in the form of new hypothesis about the structure and arrangement of the information. In this sense, unsupervised learning should assume a relevant role in medical imaging and must drive \ac{ML} to serve not only as a tool for automating complex processes, but also as an instrument for exploring and extracting hidden knowledge from medical images.

\medskip

A canonical example in which the convergence of medical imaging and \ac{ML} must be reoriented towards the discovery of new knowledge is the study of highly aggressive heterogeneous tumors \citep{Rudie2019, Tandel2019}. Specifically, glioblastoma tumor represents a major challenge, as it remains a lethal tumor that, to date, lacks a satisfactory therapy, presenting one of the poorest prognosis among all human cancers, with a median survival time of 12-15 months despite aggressive treatment \citep{Jain2018}. Since the introduction of the Stupp treatment \citep{Stupp2005} in 2005, there have been no significant changes in the therapies that have led to an improvement in patient prognosis. Therefore, the blockbuster model of \emph{``the same treatment for all"} is considered to be depleted. In this regard, efforts today must be placed into the extraction of new knowledge from medical data that allow us to move towards new personalized more effective therapies.

\medskip

One of the reasons believed to be behind this tumor's malignancy is its highly heterogeneous nature \citep{Soeda2015}. Glioblastomas are malignant masses characterized by hyper-cellularity, pleomorphism, micro-vascular proliferation and high necrosis mitotic activity \citep{Gladson2010}. Particularly, vascular heterogeneity has been largely studied since it is considered crucial for glioblastoma propagation and survival \citep{Das2013}. Glioblastoma presents strong abnormal vascular proliferation, robust angiogenesis, and extensive micro-vasculature heterogeneity \citep{Kargiotis2006}, which have been shown to have a direct effect on prognosis \citep{Hardee2012}. Therefore, the early assessment of the heterogeneous vascular architecture of the tumor is thought to provide important information to improve and design new therapies.

\medskip

However, measuring vascular heterogeneity from medical images is currently an uncertain task. There is no a consensus nor an accepted method to assess it. Consequently, there are no imaging protocols nor tools that includes such information in the clinical routine, hence ignoring this valuable knowledge in the management of the disease. Likewise, there are no expert manual annotations nor medical imaging datasets from where supervised \ac{ML} algorithms can learn models. Such an undefined problem provides an excellent opportunity to apply unsupervised learning as an instrument for exploring new knowledge about the vascular profile of the tumor.

\medskip

This thesis confronts all the factors mentioned above. The thesis focuses on the conjugation of unsupervised \ac{ML} techniques on medical imaging data to non-invasively measure and describe the vascular heterogeneity of glioblastomas. This premise has established the main motivation and goals of this thesis, leading to the following research questions and objectives.

\section{Research questions and objectives}
\label{section:research_questions_and_objectives}
The application of unsupervised \ac{ML} techniques to medical images for brain tumor analysis poses a number of challenges that need to be addressed. First of all, by definition, unsupervised learning is a more open and undefined task than supervised learning. Unlike the latter, unsupervised learning solutions generally lack of semantic meaning. On the contrary, they usually consist of an interpretation of the data structure in terms of subgroups that share similar patterns with each other. Therefore, these algorithms require intelligent strategies, both during training and in the posterior analysis of the solution, to guide them towards the extraction of useful, consistent and interpretable knowledge.

Tied to the characterization of the vascular heterogeneity of glioblastomas, unsupervised learning methods need to be guided to find hidden patterns in perfusion medical images that are consistent with physiologically plausible hypotheses. In this sense, these unsupervised methods must serve not only as a mere algorithms for the characterization of the vascular profile of the tumor, but to tools to measure advanced imaging biomarkers that could give clues about the underlying physiological process taking place in the tumor.

Under a more technical point of view, learning patterns from imaging data also poses important challenges that must take into a account. Imaging data presents patterns of local regularity and spatial redundancy that suggest that they cannot be assumed to be \ac{i.i.d.}. Robust image analysis algorithms - both supervised and unsupervised learning - must consider this structured nature of the images and must include mechanisms to take advantage from this latent information. 

As a consequence the next research questions are proposed in this thesis:

\begin{itemize}
\item[RQ1] Is unsupervised learning an adequate and reliable solution to extract valuable knowledge from complex \ac{MRI} data?

\item[RQ2] Can we contribute to the unsupervised learning family with new structured prediction algorithms to improve the extraction of knowledge from images?

\item[RQ3] Can we contribute to a better understanding of the vascular heterogeneity of the glioblastoma by means of an advanced analysis of \ac{MRI} through unsupervised learning methods?

\item[RQ4] Can useful measurements be obtained from the vascular heterogeneity description of glioblastoma that provide meaningful information on relevant patient's clinical outcomes?

\item[RQ5] Can the unsupervised learning method to describe the vascular heterogeneity of glioblastomas be part of a medical imaging software for a complete analysis of the tumor through \ac{MRI}?
\end{itemize}

The research work conducted in this thesis aims to provide solutions to these questions by means of theoretical and empirically validated scientific methods applied to the study of the vascular heterogeneity of glioblastomas by \ac{MRI}. To this end the following objectives were defined:

\begin{itemize}
\item[O1] Review of the state-of-the-art in unsupervised learning algorithms, especially focused on image oriented algorithms and paying special attention to those applied to the analysis of brain tumor \ac{MRI} images.

\item[O2] Evaluate the feasibility of unsupervised learning algorithms to analyze and extract knowledge from \ac{MRI} data.

\item[O3] Develop a new unsupervised learning algorithm to exploit the structured nature of the images and take advantage from the spatial redundancy and local information contained therein.

\item[O4] Develop a methodology based on image processing and unsupervised learning algorithms to detect and describe the vascular heterogeneity of glioblastoma through \ac{MRI}.

\item[O5] Validate the vascular heterogeneity assessment method on real clinical routine \ac{MRI}, by exploring associations between the proposed heterogeneity description and relevant clinical outcomes of the patient.

\item[O6] Implement a reliable tool capable of providing an advanced state-of-the-art analyses for glioblastoma \ac{MRI} studies, including the methods to describe the vascular heterogeneity profile of the tumor.
\end{itemize} 

The proposed objectives enclose the main goal of this thesis: \emph{the study of unsupervised \ac{ML} techniques for medical imaging data to non-invasively describe the vascular heterogeneity of glioblastomas}. Such goal can also be decomposed in two strands depending on the research scope: the \emph{technical goal}, which aims to consolidate the unsupervised learning as a reliable tool for the future of medical image analysis; and the \emph{clinical goal}, which intends to non-invasively describe the vascular heterogeneity of glioblastoma trough \ac{MRI} to improve tumor understanding and clinical decision making. In this sense, the following scientific contributions support the achievement of the proposed objectives.

\section{Thesis contributions}
\label{section:thesis_contributions}
This section presents the main contributions of this thesis. First, a summary of the most relevant aspects of each contribution is presented. Next, the scientific publications in high impact journals and conferences are listed. Finally, the technological and software results, as well as clinical studies, industrial patents and transfer actions are compiled.

\subsection{Main contributions}
\label{subsection:main_contributions}
\begin{description}
\item[\textbf{C1 -}] \textbf{Comparative study of unsupervised learning algorithms for glioblastoma segmentation} \\
In this study, a comparison of unsupervised learning algorithms, including structured and non-structured methods was performed for the task of high grade glioma segmentation. The study describes the statistical model underlying each algorithm and also proposes a general post-processing stage to identify which classes of an unsupervised segmentation correspond to pathological or healthy tissues. An independent evaluation of the performance of the unsupervised learning algorithms was carried out in a public real dataset, which demonstrated the capability of unsupervised learning to extract relevant knowledge from \ac{MRI} data. This work was published in the journal contribution \textbf{P1} \citep{JuanAlbarracin2015a} and presented in the conference \textbf{P2} \citep{JuanAlbarracin2015b}.

\item[\textbf{C2 -}] \textbf{An unsupervised learning algorithm for structured prediction} \\
A new variant of the \acfp{SVFMM} family is proposed in this thesis. The algorithm, named \acf{NLSVFMM}, successfully merges the \acsp{SVFMM} with the \acf{NLM} framework, proposing a continuous \acf{MRF} that simultaneously enforces smooth constraints in homogeneous regions of the image while preserves the edges and structures without degradation. This approximation improves the existing approaches in terms of complexity of the model, as the \ac{NLM} weighting function does not introduce additional parameters into the model to be estimated. Moreover, it outperforms current methods in terms of performance in a segmentation task of real world images. This work was published in the journal contribution \textbf{P3} \citep{JuanAlbarracin2019b}.

\item[\textbf{C3 -}] \textbf{A method for the vascular heterogeneity assessment of glioblastoma} \\
The \acf{HTS} method analyzes the perfusion \ac{MRI} of a glioblastoma using an unsupervised learning approach to delineate four \emph{habitats} within the lesion that exhibit different hemodynamic activity. The habitats describe the \acf{HAT} and \acf{LAT} regions of the glioblastoma, and the potentially \acf{IPE} and \acf{VPE} of the lesion. Such approximation establishes a conceptual frame for the description of the tumor heterogeneity by means of the detection of clinically relevant sub-regions, a.k.a habitats, with differentiated imaging biomarkers. The preliminar results of this work were first presented in the conference contribution \textbf{P4} \citep{JuanAlbarracin2016} and it was finally published in the journal contribution \textbf{P5} \citep{JuanAlbarracin2018}.

\item[\textbf{C4 -}] \textbf{Preliminary validation of the vascular heterogeneity assessment method on a local cohort of glioblastomas} \\
A preliminary validation study was performed to assess the association of the \ac{HTS} habitats with relevant clinical outcomes. Specifically, measurements on the distributions of the hemodynamic biomarkers confined at each \ac{HTS} habitat were explored for potential correlations and predictive capabilities with the overall survival of the patients. Additionally, a technical study was conducted to measure the degree of dissimilarity between these distributions, in order to confirm the physiological differences of the hemodynamic activity of the habitats. Results on a real cohort from a local hospital were published in the journal articles \textbf{P5} \citep{JuanAlbarracin2018} and \textbf{P6} \citep{FusterGarcia2018}.

\item[\textbf{C5 -}] \textbf{International retrospective multicenter clinical study for validating the vascular heterogeneity assessment method} \\
The relevant findings obtained in the experiments for the preliminary validation of the aforementioned vascular heterogeneity assessment method led us to initiate an international multi-center validation study of the technology. This constituted the first clinical study in which the \ac{UPV} was sponsor. The clinical study was formally registered in the \emph{ClinicalTrial.gov} platform from the U.S. National Library of Medicine with identifier NCT03439332, as seen in contribution \textbf{CS}. It consists of a multi-center observational retrospective study with data collected from 7 international hospitals, with a total of 305 patients enrolled since 1st of January of 2012 until February of 2018. Results obtained with this large heterogeneous cohort of untreated glioblastomas were published in the journal contribution \textbf{P7} \citep{Alvarez2019}, consolidating the previous findings about the predictive potential of the habitats.

\item[\textbf{C6 -}] \textbf{An online open-access system for glioblastoma \ac{MRI} analysis} \\
This contribution consists of the development of a web-based system for the analysis of glioblastomas by means of \ac{MRI}. The system, named ONCOhabitats (\url{https://www.oncohabitats.upv.es}), provides free access to all the methods developed and validated in this thesis, but also to other state-of-the-art algorithms in the field of medical image analysis, to offer a complete solution for the study of glioblastoma from raw unprocessed \ac{MRI}. ONCOhabitats implements two main services to describe the morphological and vascular heterogeneity of the glioblastoma, generating for each service an automated {\LaTeX}-based report summarizing all the findings of the study. The details of the system were presented in the journal contribution \textbf{P8} and conference contribution \textbf{P9}, and the software was registered in the technological catalogue of the \ac{UPV}, as shown in contributions \textbf{S1} and \textbf{S2}.

\item[\textbf{C7 -}] \textbf{An industrial patent for generating multi-parametric nosological images} \\
In addition to the scientific and academic contributions, the methods, technologies and original ideas conceived in this thesis were protected under the international patent mentioned in contribution \textbf{PT}. The patent issued \emph{``Method and system for generating multi-parametric nosological images"} was registered in Spain (ES201431289A), with the added value of being evaluated with previous exam; was extended to the European (EP3190542A1) territory and United States (US20170287133A1) through the \ac{PCT} programme. The patent protects a method to produce nosological images from multiple medical image acquisitions, with the aim of facilitating the diagnosis and treatment of diseases. In this sense, this thesis has contributed not only with advances in knowledge in the fields of \ac{ML} and medical imaging, but with a technological asset of high value for the \ac{UPV}, which opens the door to transfer actions for the creation of new business opportunities.

\item[\textbf{C8 -}] \textbf{Foundation of the ONCOANALYTICS CDX, S.L. company} \\
The issuance of the patent led us to participate in two of the most cutting-edge national programmes for the generation of business models and new start-ups in the field of healthcare technologies. The author of this thesis, together with the advisors, participated in the EIT Health Headstart Proof of Concept 2016 programmme, in which we were awarded the best Proof of Concept Spain for a technology-based start-up; and in the \emph{CaixaImpulse} acceleration programme for facilitating entrepreneurship in biomedicine. Such mentoring activities finally led to the foundation of ONCOANALYTICS CDX, S.L. in 2018, with the commercial name \emph{Texture CDx}, as shown in contribution \textbf{TR}. The company was framed into the business model of companion diagnostics for pharmaceutical compounds, with the aim of using the aforementioned vascular heterogeneity assessment technology to help in the stratification of patients affected by glioblastoma during the clinical trial of a drug. ONCOANALYTICS CDX was established by a multi-disciplinary team made up of computer scientists, physicists, oncologists, biomedical engineers and financial experts, including 6 \ac{UPV} graduates and 4 Phd, thus contributing to the generation of professional opportunities for highly qualified personnel.
\end{description}

The work developed in this thesis has been framed in several national research projects, one of which has obtained the A+ rating, i.e. the best rating available. This has made it possible to raise public funds, new research projects and doctoral grants that have consolidated the research line in the \ac{BDSLab} of the \ac{UPV}.

\subsection{Scientific publications}
The scientific contributions of this thesis have been published in six scientific top-ranked journals and five conference proceedings in the fields of \acl{ML}, Statistics and Probability, Radiology \& Nuclear Medicine, Medical Imaging and Biomedical Data Mining. The publications are listed as follows:

\begin{small}
\begin{itemize}
\item[P1 -] \textbf{Javier Juan-Albarracín}, Elies Fuster-Garcia, José V. Manjón, Montserrat Robles, F. Aparici-Robles, L. Martí-Bonmatí and Juan M. García-Gómez. \emph{`Automated Glioblastoma Segmentation Based on a Multiparametric Structured Unsupervised Classification'}. PLoS One; 2015; 10(5):e0125143. May 2015. \citep{JuanAlbarracin2015a}.
\item[] IF: 3.057 (JCR 2015): 11/63 Multi-disciplinary sciences (Q1).

\item[P2 -] \textbf{Javier Juan-Albarracín}, Elies Fuster-Garcia and Juan M. García-Gómez. \emph{`Hierarchical Tissue-Guided Glioblastoma Segmentation based on DCA-SVFMM'}. II International Symposium on Clinical and Basic Investigation in Glioblastoma. GBM2015. 3(1):101. Toledo, Spain. September 2015. \citep{JuanAlbarracin2015b}.

\item[P3 -] \textbf{Javier Juan-Albarracín}, Elies Fuster-Garcia and Juan M. García-Gómez. \emph{'Non-Local Spatially Varying Finite Mixture Models for Image Segmentation'}. Statistics and Computing; September 2019; Accepted for publication. \citep{JuanAlbarracin2019b}. 
\item[] IF: 2.383 (JCR 2018): 16/123 Statistics \& Probability (Q1), 31/105 Computer Science, Theory \& Methods (Q2).

\item[P4 -] \textbf{Javier Juan-Albarracín}, Elies Fuster-Garcia and Juan M. García-Gómez. \emph{`An online platform for the automatic reporting of multi-parametric tissue signatures: A case study in Glioblastoma'}. In: Crimi A., Menze B., Maier O., Reyes M., Winzeck S., Handels H. (eds) Brainlesion: Glioma, Multiple Sclerosis, Stroke and Traumatic Brain Injuries. BrainLes 2016. Lecture Notes in Computer Science, vol 10154. Springer, Cham. Athens, Greece. October 2016. \citep{JuanAlbarracin2016}.

\item[P5 -] \textbf{Javier Juan-Albarracín}, Elies Fuster-Garcia, Alexandre Pérez-Girbés, F. Aparici-Robles, Ángel Alberich-Bayarri, Antonio Revert-Ventura, L. Martí-Bonmatí and Juan M. García-Gómez. \emph{`Glioblastoma: Vascular Habitats Detected at Preoperative Dynamic Susceptibility-weighted Contrast-enhanced Perfusion MR Imaging Predict Survival'}. Radiology; 2018; 287(3):944-954. Jun 2018. \citep{JuanAlbarracin2018}.
\item[] IF: 7.608 (JCR 2018): 4/129 Radiology, Nuclear Medicine \& Magnetic Resonance Imaging (Q1).

\item[P6 -] Elies Fuster-Garcia, \textbf{Javier Juan-Albarracín}, Germán A. García-Ferrando, L. Martí-Bonmatí, F. Aparici-Robles and Juan M. García-Gómez. \emph{`Improving the estimation of prognosis for glioblastoma patients by MR based hemodynamic tissue signatures'}. NMR in Biomedicine; 2018; 31(12):e4006. December 2018. \citep{FusterGarcia2018}. 
\item[] IF: 3.414 (JCR 2018): 5/41 Spectroscopy (Q1), 30/129 Radiology, Nuclear Medicine \& Magnetic Resonance Imaging (Q1), 22/73 Biophysics (Q2).

\item[P7 -] María Del Mar Álvarez-Torres and \textbf{Javier Juan-Albarracín} and Elies Fuster-Garcia and Fuensanta Bellvís-Bataller and David Lorente and Gaspar Reynés and Jaime Font de Mora and Fernando Aparici-Robles and Carlos Botella and Jose Muñoz-Langa and Raquel Faubel and Sabina Asensio-Cuesta and Germán A. García-Ferrando and Eduard Chelebian and Cristina Auger and Jose Pineda and Alex Rovira and Laura Oleaga and Enrique Mollà-Olmos and Antonio J. Revert and Luaba Tshibanda and Girolamo Crisi and Kyrre E. Emblem and Didier Martin and Paulina Due-Tønnessen and Torstein R. Meling and Silvano Filice and Carlos Sáez and Juan M García-Gómez. \emph{`Robust association between vascular habitats and patient prognosis in glioblastoma: an international retrospective multicenter study'}. Journal of Magnetic Resonance Imaging; 2019; 31(12):e4006. October 2019. \citep{Alvarez2019}. 
\item[] IF: 3.732 (JCR 2018): 26/129 Radiology, Nuclear Medicine \& Magnetic Resonance Imaging (Q1).

\item[P8 -] \textbf{Javier Juan-Albarracín}, Elies Fuster-Garcia, Germán A. García-Ferrando and Juan M. García-Gómez. \emph{`ONCOhabitats: A system for glioblastoma heterogeneity assessment through MRI'}. International Journal of Medical Informatics; 2019; 128():53-61. August 2019. \citep{JuanAlbarracin2019a}. 
\item[] IF: 2.731 (JCR 2018): 57/155 Computer Science and Information Systems (Q2), 28/98 Healthcare Sciences \& Services (Q2), 11/26 Medical Informatics (Q2).

\item[P9 -] \textbf{Javier Juan-Albarracín}, Elies Fuster-Garcia and Juan M. García-Gómez. \emph{`MTSimaging: multiparametric image analysis services for vascular characterization of glioblastoma'}. The European Society of Magnetic Resonance in Medicine and Biology Congress. ESMRMB 2017. 30 (Suppl 1): S501–S692. Barcelona, Spain. October 2017. \citep{JuanAlbarracin2017}.

\item[P10 -] \textbf{Javier Juan-Albarracín}, Elies Fuster-Garcia and María del Mar Álvarez-Torres and Eduard Chelebian and Juan M. García-Gómez. \emph{`ONCOhabitats glioma segmentation model'}. In: Crimi A., Menze B., Maier O., Reyes M., Winzeck S., Handels H. (eds) Brainlesion: Glioma, Multiple Sclerosis, Stroke and Traumatic Brain Injuries. BrainLes 2019. Lecture Notes in Computer Science, vol 10154. Springer, Cham. Shenzhen, China. October 2019. \citep{JuanAlbarracin2019c}.

\item[P11 -] Juan Ortiz-Pla, Elies Fuster-Garcia, \textbf{Javier Juan-Albarracín} and Juan M. García-Gómez. \emph{`GBM Modeling with Proliferation and Migration Phenotypes: A Proposal of Initialization for Real Cases'}. In: Tsaftaris S., Gooya A., Frangi A., Prince J. (eds) Simulation and Synthesis in Medical Imaging. SASHIMI 2016. Lecture Notes in Computer Science, vol 9968. Springer, Cham. Athens, Greece. October 2016. \citep{OrtizPla2016}.
\end{itemize}
\end{small}

\subsection{Software}
The research conducted in this thesis has led to the creation of ONCOhabitats platform (\url{https://www.oncohabitats.upv.es}). ONCOhabitats is an online professional system for glioblastoma analysis using \ac{MRI}, which encapsulates all the original methods and algorithms developed in this thesis, and several state-of-the-art algorithms for medical image analysis. Preliminary versions of some methods of ONCOhabitats were first registered in the technological catalogue of the \ac{UPV} under the acronym CURIAM BT+, while the final updated complete version of the ONCOhabitats system has been recently registered as an asset of high value for the technological offer of the \ac{UPV}.

\begin{small}
\begin{itemize}
\item[S1 -] \textbf{Javier Juan-Albarracín}, Elies Fuster-Garcia, Juan M. García-Gómez, Carlos Sáez, Montserrat Robles and Miguel Esparza. \emph{`R-16874-2014 - Caracterización de firmas biológicas de glioblastomas (CURIAM BT+)'}. CARTA Registry of the Universitat Politècnica de València. 28/02/2014.

\item[S2 -] \textbf{Javier Juan-Albarracín}, Elies Fuster-Garcia, Juan M. García-Gómez. \emph{`R-XXXXX-2019 - Detección de hábitats para evaluación de heterogeneidad biológica de glioblastomas (ONCOhabitats)'}. CARTA Registry of the Universitat Politècnica de València. In process.

\end{itemize}
\end{small}

\subsection{Clinical studies}
The aforementioned ONCOhabitats platform is currently enrolled in an international multicenter observational restrospective clinical study registered at \emph{ClinicalTrials.gov} from the U.S. National Library of Medicine. The aim of the study is to validate the prognostic capabilities of the \ac{HTS} habitats for patients affected with glioblastoma. To this end, the primary and secondary outcomes fixed for the clinical study are: the \emph{``correlation between overall survival and progression-free survival (in days) of patients undergoing standard-of-care treatment and the tumor vascular heterogeneity described by the four habitats obtained by the \ac{HTS} biomarker"}. The clinical study involves data collected from 7 international hospitals, with a total of 305 patients recruited since 1st of January of 2012 until February of 2018.

\begin{small}
\begin{itemize}
\item[CS -] Multicenter Retrospective Observational Clinical Study \textbf{NCT03439332}. \emph{`Multicentre Validation of How Vascular Biomarkers From Tumor Can Predict the Survival of the Patient With Glioblastoma (ONCOhabitats)'}. \url{https://clinicaltrials.gov/ct2/show/NCT03439332}. \acf{UPV}. 20/02/2018.
\end{itemize}
\end{small}

\subsection{Patents}
The know-how in medical image analysis generated by the author in this thesis was protected under the international patent \emph{``Method and system for generating multi-parametric nosological images"}. The patent is currently registered in Spain (ES201431289A) with previous exam, and was extended to Europe (EP3190542A1) and United States (US20170287133A1) following the \ac{PCT} procedure for patent internationalization. It describes a procedure based on multi-parametric medical images to generate nosological masks capable of describing the underlying physiological processes taking place in the lesion. The patent materializes the interest that the scientific research conducted in this thesis has originated in both the academic and business spheres. Moreover, it represents the efforts of this thesis to generate tangible assets for a later phase of business development.

\begin{small}
\begin{itemize}
\item[PT -] \textbf{Javier Juan-Albarracín}, Elies Fuster-Garcia, Juan M. García-Gómez, Miguel Esparza-Manzano, Jose V. Manjón-Herrra, Monserrat Robles-Viejo, Carlos Sáez. \emph{`Method and system for generating multi-parametric nosological images'}. Asignee: \acf{UPV}.
\item[] ES201431289A: Oficina Española de Patentes y Marcas. 05/09/2014. Legal status: Active.
\item[] EP3190542A4: European Patent Office. PCT/ES2015/070584. 28/07/2015. Legal status: Active.
\item[] US9990719B2: United States Patent and Trademark Office. PCT/ES2015/070584. 28/07/2015. Legal status: Active.
\end{itemize}
\end{small}

\subsection{Transference}
The experience, knowledge and original ideas conceived in this thesis, together with the issuance of the patent, aroused the author's interest in taking a step beyond the academic field. This led the author and the thesis advisors to the conceptualization of a business plan to capitalize the results obtained in the thesis. In this sense, we participated in the EIT Health Headstart Proof of Concept 2016 programmme, in which we were awarded the best business plan for a biotech start-up; and in the \emph{CaixaImpulse} programme for facilitating entrepreneurship in biomedicine. The experience obtained in these programmes in conjunction with the background of the advisors in generating spin-offs of the \ac{UPV}, finally led to the foundation of ONCOANALYTICS CDX company. ONCOANALYTICS CDX is formed by a multi-disciplinary team made up of computer scientists, physicists, oncologists, biomedical engineers and financial experts, with a total of 6 \ac{UPV} graduates and 4 Phd. Supported by the aforementioned ONCOhabitats platform, the company is focused in developing image-based \acs{CDx} for glioblastoma, to facilitate patient stratification during the clinical trial of a drug. 

\begin{small}
\begin{itemize}
\item[PT -] \textbf{Javier Juan-Albarracín}, Elies Fuster-Garcia, Juan M. García-Gómez, Germán A. García-Ferrando, Carlos Vidal-Trujillo, José Muñoz-Langa, David Lorente-Estellés, Ana González-Segura, Fuensanta Bellvís-Bataller. \emph{`ONCOANALYTICS CDX S.L.'}. Commercial name: Texture CDx. CIF: B98981889. 01/03/2018.
\end{itemize}
\end{small}

\section{Projects and partners}
During the development of this thesis the author has actively participated in several national, European and private research projects in collaboration with several hospitals and clinical institutions. The projects related to this thesis are listed below:

\paragraph{\textbf{CURIAM BT+}} \emph{Caracterización de firmas biológicas de glioblastomas mediante modelos no-supervisados de predicción estructurada basados en biomarcadores de imagen.} Funded by the Spanish Ministry of Economy and Competitiveness (TIN2013-43457-R, 2014-2016).

\textbf{Objectives}: This project aims to develop a computational medical imaging system to obtain radiological profiles of the different areas of the tumor, to accurately measure the vascular properties of the glioblastoma. Such profiles will also provide information about the tumor grading and the expected survival of the patient.

\textbf{Partners}: \acf{BDSLab}-ITACA group of the Universitat Politècnica de Valencia (Valencia, Spain). \acf{HUPLF} (Valencia, Spain).

\hfill

\paragraph{\textbf{MTS4Up}} \emph{Biomarcadores dinámicos basados en firmas tisulares multiparamétricas para el seguimiento y evaluación de la respuesta a tratamiento de pacientes con glioblastoma y cáncer de próstata.} Funded by the Spanish National Research Agency (DPI2016-80054-R, 2017-2018).

\textbf{Objectives}: This project extends the TIN2013-43457-R project by improving the technology to obtain radiological signatures of the glioblastoma incorporating diffusion \ac{MRI} to describe not only tumor vascularity, but also the cell density properties of the tissues. Such improvements will allow an early evaluation of tumor progression and an accurate assessment of the patient's response to treatment. Finally, the technology will also be evaluated on other pathologies such as prostate tumor to measure the versatility of the methodology in other solid tumors.

\textbf{Partners}: \acf{BDSLab}-ITACA group of the Universitat Politècnica de Valencia (Valencia, Spain).

\hfill

\paragraph{\textbf{GLIO-MARKERS}} \emph{Estudio integrado de biomarcadores moleculares y de imagen en pacientes con glioblastoma.} Funded by the Universitat Politècnica de València and Hospital Universitario y Politécnico La Fe (Prueba de Concepto 2015, UPV-FE-15-B, 2015-2016).

\textbf{Objectives}: This project aims to combine and integrate the analysis of glioblastoma biomarkers from three different physiological areas: blood circulating proteins, immunohistologic biomarkers and \ac{MRI} biomarkers. The purpose is to develop predictive models for response to treatment assessment and measuring tumor progression, as well as finding correlations between imaging biomarkers and circulating proteins.

\textbf{Partners}: \acf{BDSLab}-ITACA group of the Universitat Politècnica de Valencia (Valencia, Spain), \acf{UPV} (Valencia, Spain).

\hfill

\paragraph{\textbf{DSSRADIOPLAN}} \emph{Inclusión de las tecnologías de firma tisular y modelos mutiescala para el soporte a la planificación de la radioterapia en el tratamiento del glioblastoma.} Funded by the Universitat Politècnica de València and Hospital Universitario y Politécnico La Fe (Prueba de Concepto 2016, UPV-FE-16-B, 2016-2017).

\textbf{Objectives}: The main purpose of this project is to plan and carry out the necessary actions to evaluate the applicability and added value of the \ac{HTS} technology to provide clinical decision support in the management and planning of radiotherapy in patients affected by glioblastoma.

\textbf{Partners}: \acf{BDSLab}-ITACA group of the Universitat Politècnica de Valencia (Valencia, Spain), \acf{UPV} (Valencia, Spain).

\hfill

\paragraph{\textbf{MULTIBIOIM}} \emph{Multiparametric nosological images for supporting clinical decisions in solid tumors.} Funded by the EIT Health E.V. (Proof of Concept 2016, POC-2016-SPAIN-07, 2016-2017).

\textbf{Objectives}: This project aims to develop a Proof of Concept (PoC) of the patented procedure ES201431289A for generation of multiparametric tissue signatures based on structural and functional \ac{MRI} for solid tumors. This PoC aims to solve specific technical, strategical, legal, and commercial barriers to generate a reliable business model for the medical image analysis software market and convert a cutting-edge technology into a clinically validable product.

\textbf{Partners}: \ac{BDSLab}-ITACA group of the Universitat Politècnica de Valencia (Valencia, Spain).

\hfill

The projects on which the author was involved in previously and in parallel to the development of this thesis are listed as follows:

\paragraph{\textbf{DQV-MINECO}} \emph{Servicio de evaluación y rating de la calidad de repositorios de datos biomédicos.} Funded by the Spanish Ministry of Economy and Competitiveness (Retos-Colaboración 2013 programme, RTC-2014-1530-1, 2013-2016).

\textbf{Objectives}: This project aims to define a data quality evaluation and rating service to assure the data value aimed to its reuse in clinical, strategic and scientific decision making. It will be based on two software services. The first will evaluate nine data quality dimensions. The second will generate a data quality rating positioning the evaluated datasets according to several reuse knowledge extraction purposes.

\textbf{Partners}: VeraTech for Health S.L. (Valencia, Spain) and IBIME-ITACA group of the Universitat Politècnica de Valencia, (Spain)

\hfill

\paragraph{\textbf{HELP4MOOD}} \emph{A Computational Distributed System to Support the Treatment of Patients with Major Depression.} Funded by the European Commission. VII Framework Program (FP7-ICT-2009-4; 248765, 2011-2013).

\textbf{Objectives}: This project focuses on major depression disease. Patients with major depression typically recover through antidepressant drugs, psychological therapy or hospitalization. However, it has been shown that in many situations such recovery is either slow or incomplete. Research shows that psychological therapies can be delivered effectively without face to face contact at individual's home by computerized cognitive behavioral therapy. The project aims to advance the state-of-the-art in computerized support for people with major depression by monitoring mood, thoughts, physical activity and voice characteristics, by means of intelligent systems based on virtual agent.

\textbf{Partners}: \ac{BDSLab}-ITACA group of the Universitat Politècnica de Valencia, (Spain), University of Edinburgh (United Kingdom), Fundació I2CAT (Spain), Universitatea Babes Bolyai (Romania), FVA SAS di Louis Ferrini (Italy), OBS Medical Ltd. (Italy), Universitat Politècnica de Catalunya, (Spain), Heriot-Watt University (United Kingdom).

\hfill

\section{Thesis outline}
The thesis is structured in eight chapters that thoroughly describe the research work carried out during the thesis. The Chapter \ref{chapter:introduction} has introduced the motivations, research objectives and main contributions. Chapter \ref{chapter:rationale} describes the thesis rationale, introducing the clinical problems addressed as well as the theoretical background needed to complement the description of the methods developed in the thesis. Chapter \ref{chapter:comparative_unsupervised_learning} presents a preliminary study on the viability of the unsupervised learning paradigm to identify and delineate pathological tissues in glioblastomas based on \ac{MRI} patterns. Chapter \ref{chapter:nlsvfmm} presents the mathematical development of a new unsupervised structured learning algorithm for image segmentation, and a comparison of its performance against alternative approaches. Chapter \ref{chapter:hts} introduces the \ac{HTS} method: an unsupervised learning method based on perfusion \ac{MRI} to delineate vascular habitats within the glioblastoma to assess its vascular heterogeneity. An study on the association of the vascular habitats and the patient \ac{OS} is presented. Chapter \ref{chapter:validation} describes the international multi-center validation of the \ac{HTS} method, under the framework of the observational retrospective clinical study \href{https://clinicaltrials.gov/ct2/show/NCT03439332}{NCT03439332}. The validation of the association of the vascular habitats and the patient \ac{OS}, as well as the stratification capabilities of the \ac{HTS} habitats is presented. Chapter \ref{chapter:oncohabitats} presents ONCOhabitats platform (\url{https://www.oncohabitats.upv.es}). ONCOhabitats encapsulates all the work conducted in this thesis in a public open-access platform, offering medical image analysis services to analyze both the morphological and vascular heterogeneity of the glioblastoma. Finally, chapter \ref{chapter:conclusion} ends this dissertation with the concluding remarks and recommendations to continue with the research developed in this thesis.

\begin{sidewaysfigure}[p]
\centering
\includegraphics[width=.825\textwidth]{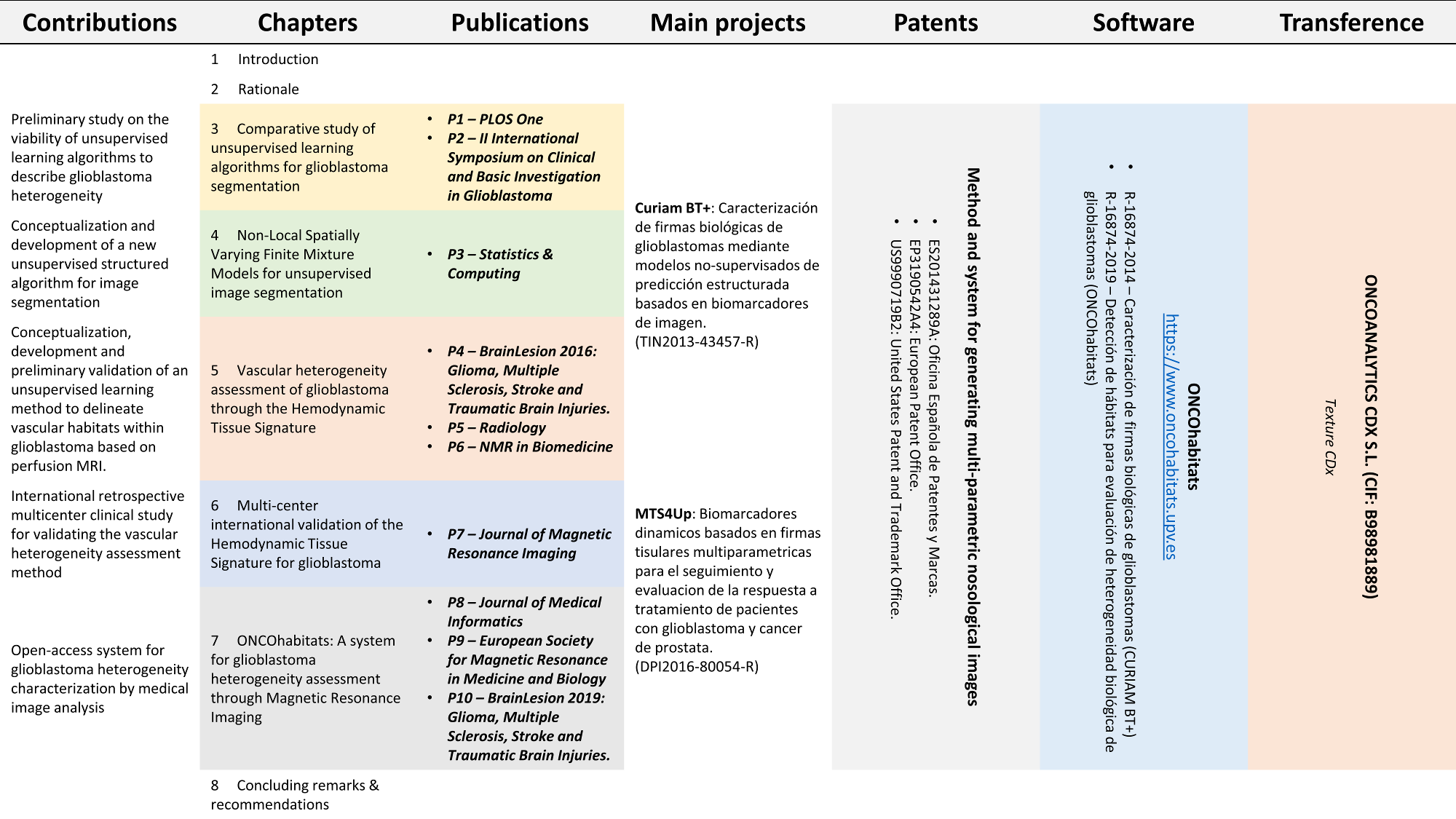}
\caption{Outline of the thesis contributions, chapters, publications, projects, transference actions, patents and software}
\label{figure:outline}
\end{sidewaysfigure}

Figure \ref{figure:outline} outlines the thesis contributions structured among the thesis chapters, along with the publications, research projects, transfer actions, patents and the software developed during this study.

\chapter{Rationale}
\label{chapter:rationale}
This chapter describes the thesis rationale divided in five sections. First, the glioblastoma tumor is introduced, describing its epidemiology, etiology, biologic behavior, morphological features, diagnosis and treatment. Second, \ac{MRI} technique is disclosed, illustrating their physical mechanisms, theoretical foundations and acquisition protocols and sequences. Third a general review on the theoretical background probability and statistical parameters estimation recommended for the understanding of the methods developed in the thesis is provided. Fourth, an in-depth explanation of \acl{FMM} and \acl{SVFMM} and their parameter estimation is presented to lay the foundation for many of the methods developed in this thesis. Finally, a general review of the \ac{DL} paradigm, as well as a revision of \aclp{ANN}, \aclp{CNN} and the back-propagation mechanics is provided. This review is intended to establish a common basis to complement the descriptions of background and methods explained in the following chapters.

\section{Glioblastoma}
\label{section:rationale_glioblastoma}
The first recorded clinical report identifying glioblastoma as a tumor originating from neuroglial cells dates back to 1863 by \cite{virchow2018}. Since then, enormous progress has been made in the understanding of this neoplasm thanks to an exhaustive multidisciplinary research in the clinical, pathological, radiological, molecular and genetic aspects of the tumor. Such efforts have led nowadays to a detailed description of the glioblastoma that has given us crucial, but still not sufficient, information to design successful treatments for this disease.

Glioblastoma is a grade IV \ac{WHO} deadly primary brain tumor considered the most aggressive neoplasm of the central nervous system. It is the most frequent and malignant astrocytoma in humans, accounting for more than 60\% of all brain tumors in adults. Glioblastoma has a global incidence of 4.67 to 5.73 per 100000 people and presents a poor prognosis of 14-15 months despite aggressive treatments. Although it can debute at any age, more than the 70\% of the cases are seen in patients between the ages of 45 and 70. Likewise, the incidence in males is 1.6 higher than in females and it is 2 times higher in Caucasians than in other races \citep{Tamimi2017}.

\medskip

Glioblastomas are infiltrative and deeply invasive heterogeneous masses characterized by hypercellularity, pleomorphism, microvascular proliferation and high necrosis mitotic activity \citep{Gladson2010}. Typically, glioblastoma exhibits diffuse margins with co-existence of different tissues including active tumor, cysts, necrosis and edema; all of them exhibiting a high variability related to the aggressiveness of the neoplasm \citep{Hardee2012}. Strong vascular proliferation, robust angiogenesis, and extensive microvasculature heterogeneity are major pathological hallmarks that differentiate glioblastomas from low-grade gliomas \citep{Kargiotis2006}.

\begin{figure}[htbp!]
\centering
\includegraphics[width=0.9\linewidth]{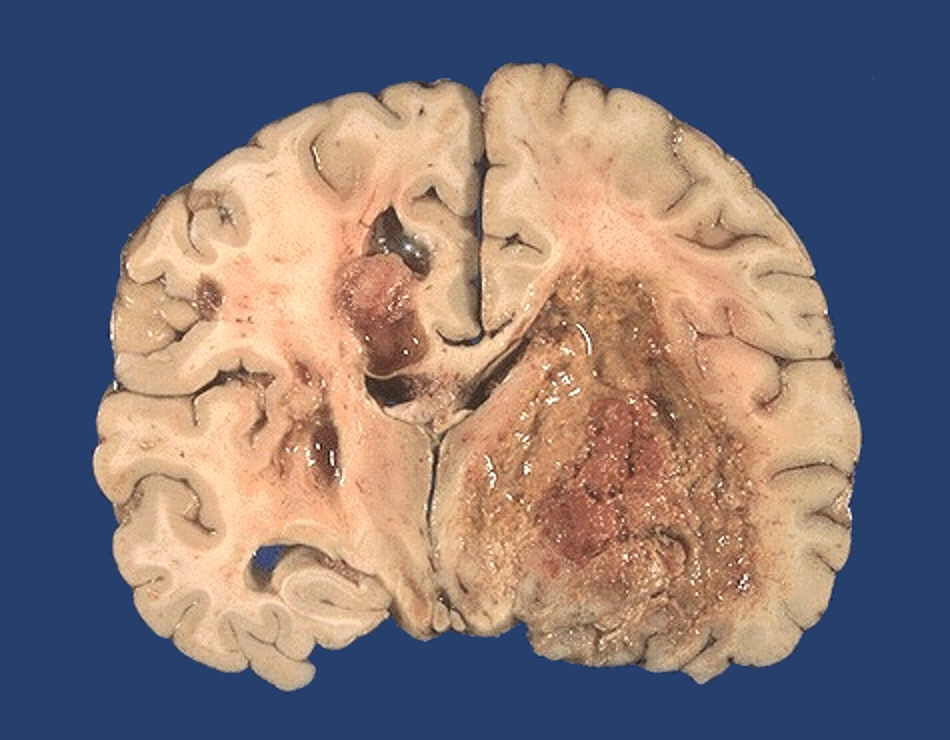}
\caption{Macroscopic example of glioblastoma. Image taken with kind permission from \url{https://webpath.med.utah.edu/CNSHTML/CNS136.html}}
\label{figure:rationale_glioblastoma_brain}
\end{figure}

Figure \ref{figure:rationale_glioblastoma_brain} shows an example of a brain affected by glioblastoma. Note the deep invasive ability of the tumor, evidenced by the tumor foci crossing the midline towards the opposite hemisphere of the largest mass.

Heterogeneity has therefore been considered crucial to understand the aggressiveness of this tumor and its resistance to effective therapies. Glioblastoma heterogeneity manifests itself at both macroscopic and microscopic levels. At the macroscopic level, co-existance of an amalgam of blended malignant tissues including enhancing tumor, non-enhancing tumor, hemorrhage, cyst, inflammation, necrosis or edema, results in a chaotic mass highly complicated to manage clinically. At the microscopic level, different glioblastoma molecular sub-types and genetic alterations have been discovered in the past years. In 2010, \cite{Verhaak2010} established a classification for glioblastoma into four sub-types associated to mutations in EGFR, TP53, NF1 and PDGFRA/IDH1 genes: the classical, mesenchymal, neural and proneural sub-types. The main characteristics of these tumor sub-types are summarized in Table \ref{table:rationale_glioblastoma_verhaak}.

\begin{table}[htbp!]
\caption{Summary of the most relevant characteristics of Verhaak glioblastoma subtypes classification. \textsuperscript{*} LOH: Loss of heterozygosity.}
\centering
\begin{tabular}{|p{0.46\textwidth}|p{0.46\textwidth}|}
	\hline
	\rowcolor{gray!25}
	\textbf{CLASSICAL} & \textbf{MESENCHYMAL} \\ \hline
	\begin{itemize}[itemsep=0.5pt, topsep=0pt]
		\item High EGFR (97\%)
		\item Lack of TP53 mutations
		\item Chromosome 7 amplification with LOH\textsuperscript{*} chromosome 10
		\item CDKNA2 deletion (94\%)
		\item High Notch and Sonic Hedgehog markers
		\item Patients survive longest given aggressive treatment
	\end{itemize} & \begin{itemize}[itemsep=0.5pt, topsep=0pt]
		\item Focal deletions at 17q11.2
		\item Mutated NF1 (70\%)
		\item Mutated TP53
		\item Mutated PTEN
		\item Expression of CH13L1 marker
		\item Expression of MET marker
		\item Higher activity of astrocytic markers (CD44 and MERTK)
		\item Increased NF-kB pathway
	\end{itemize} \\ \hline
	\rowcolor{gray!25}
	\textbf{PRONEURAL} & \textbf{NEURAL} \\ \hline
	\begin{itemize}[itemsep=0.5pt, topsep=0pt]
		\item Altered PDGFRA
		\item Point mutations at IDH1 (93\%)
		\item TP53 LOH (67\%)
		\item Lesser chromosome 7 amplification with LOH chromosome 10
		\item Focal amplifications at 4q12 (higher than other subtypes)
		\item Expression of oligodendrotytics genes
	\end{itemize} & \begin{itemize}[itemsep=0.5pt, topsep=0pt]
		\item Neuron markers (NEFL, GABRA1, SYT1, SLC12A5)
		\item Few (75\%) has normal cells in pathology slides
		\item Association with oligodendrocytic and astrocytic differentiation but mostly express neuron markers
	\end{itemize} \\
	\hline
\end{tabular}
\label{table:rationale_glioblastoma_verhaak}
\end{table}

In 2016, the \ac{WHO} revisited the official classification for glioblastoma sub-types, distinguishing between two groups: the IDH-wildtype (90\% of cases) and the IDH-mutant (10\% of cases), which are closely related to primary and secondary glioblastomas respectively. Table \ref{table:rationale_glioblastoma_who} summarizes the most relevant aspects of the IDH-wildtype and IDH-mutant glioblastomas.

\begin{table}[htbp!]
\caption{Summary of the most relevant aspects of IDH-wildtype and IDH-mutant glioblastomas.}
\centering
\resizebox{\textwidth}{!}{
\rowcolors{2}{gray!10}{white}
\begin{tabular}{lll}
	\hline
	\rowcolor{gray!25}
	& \textbf{IDH-wildtype glioblastomas} & \textbf{IDH-mutant glioblastomas} \\ \hline
	\rowcolor{white}
	Synonym & Primary glioblastoma & Secondary glioblastoma \\
	\rowcolor{gray!10}
	Precursor lesion & Not identifiable; develops & Diffuse astrocytoma, \\
	\rowcolor{gray!10}
	                 & \emph{de novo} & Anaplastic astrocytoma \\
	\rowcolor{white}
	Proportion of glioblastomas & $\sim$90\% & $\sim$10\% \\
	\rowcolor{gray!10}
	Median age at diagnosis & $\sim$62 years & $\sim$44 years  \\
	\rowcolor{white}
	Male-to-female ratio & 1.42 : 1 & 1.05 : 1 \\
	\rowcolor{gray!10}
	Mean length of clinical history & 4 months & 15 months \\
	\rowcolor{white}
	Median overall survival & & \\
	\rowcolor{white}
    \hspace*{1cm}Surgery + radiotherapy & 9.9 months & 24 months \\
    \rowcolor{white}
    \hspace*{1cm}\makecell[l]{Surgery + radiotherapy \\ + chemotherapy} & 15 months & 31 months \\
	\rowcolor{gray!10}
	Location & Supratentorial & Preferentially frontal \\
	\rowcolor{white}
	Necrosis & Extensive & Limited \\
	\rowcolor{gray!10}
	TERT promoter status & 72\% & 26\% \\
	\rowcolor{white}
	TP53 mutations & 27\% & 81\% \\
	\rowcolor{gray!10}	
	ATRX mutations & Exceptional & 71\% \\
	\rowcolor{white}	
	EGFR mutations & 35\% & Exceptional \\
	\rowcolor{gray!10}	
	PTEN mutations & 24\% & Exceptional \\
	\hline
\end{tabular}
}
\label{table:rationale_glioblastoma_who}
\end{table}

The study of these transcriptional subtypes has yielded relevant findings such as significant correlation with patient prognosis \citep{Parsons2008}. IDH-mutant glioblastomas show a significant improvement in \ac{OS} with a median survival of 31 months, with respect to IDH-wildtype glioblastomas that present a median survival of 15 months \cite{Louis2016}. However, although tumor subtypes tend to correlate with relevant clinical outcomes, the degree of correlation is often moderate and contradictory studies constantly appear with confronted conclusions \citep{Akagi2018}. Moreover, survival rates have shown no notable improvement since the last three decades \citep{Stupp2005}, so alternative approaches are required to study the glioblastoma heterogeneity and its association with the tumor evolution.

In this sense, significant interest has been placed recently in the analysis of glioblastoma heterogeneity through medical imaging. The ability to discover non-invasive markers associated with tumor sub-types, \ac{OS}, \ac{PFS} or response to treatment has received much attention as it may help in improving clinical decision making at an early stage of the disease. In this sense, \ac{MRI} emerged as one of the most reliable tool for quantifying in-vivo non-invasive imaging features capable to accurately describe the heterogeneity of the glioblastoma.

\section{\acl{MRI}}
\label{section:rationale_mri}
\acf{MRI} is a medical imaging technique used to provide in-vivo internal representations of the human body. This technique was developed in the decade of 1970 by the professors \cite{Damadian1971, Lauterbur1973, Lauterbur1974, Mansfield1977}. Although Damadian's work on the \ac{NMR} relaxation of different tissues laid the groundwork for many further developments in \ac{MRI}, it was Paul Lauterbur who finally developed a reliable technique based on gradient magnets to generate the first 2D and 3D \ac{MR} images of the interior of the human body. A few years later Peter Mansfield developed a mathematical formulation that dramatically accelerated the acquisition of \ac{MR} images (seconds rather than hours), making it a practical technique for the clinical routine. Paul Lauterbur and Peter Mansfield finally awarded the Nobel prize in 2003 for their contributions and advances in \ac{MRI}.

\medskip

\begin{figure}[htbp!]
\centering
\includegraphics[width=0.7\linewidth]{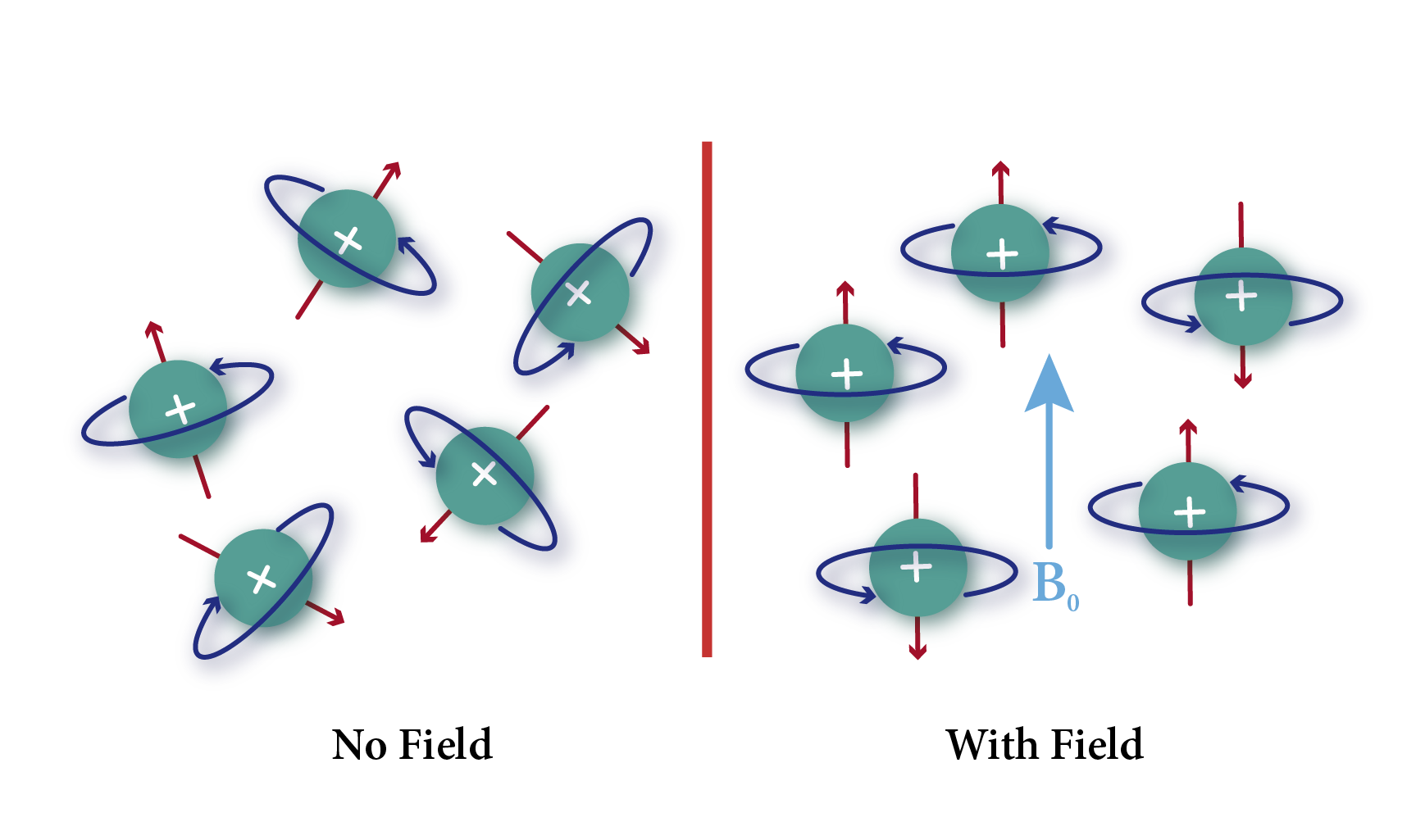}
\caption{Diagram of the spin angular momentum of $H^+$ nuclei at a resting natural state (left) and under the influence of an uniform magnetic field $B_0$ (right).}
\label{figure:rationale_mri_spins}
\end{figure}

\ac{MRI} is based on the magnetic properties of the atomic nuclei, specifically on the spin angular momentum of the hydrogen nucleus $\left( H^+ \right)$. At a resting natural state, all the hydrogen $H^+$ nucleus in the human body spin randomly, thus canceling the angular momentum each other and producing an overall zero spin magnetic momentum. Under the influence of an external uniform magnetic field $B_0$, the $H^+$ nuclei align their spin with $B_0$ in a parallel (low energy) or anti-parallel (high energy) state, producing an overall spin magnetic momentum $M_z$, with $z$ the direction of $B_0$ (see figure \ref{figure:rationale_mri_spins}). The influence of $B_0$ also makes the $H^+$ nuclei to precess at a specific frequency (denominated the \emph{Larmor} frequency), which depends on the strength of $B_0$ and the gyromagnetic properties of the hydrogen nucleus.

\begin{figure}[h]
\centering
\includegraphics[width=0.7\linewidth]{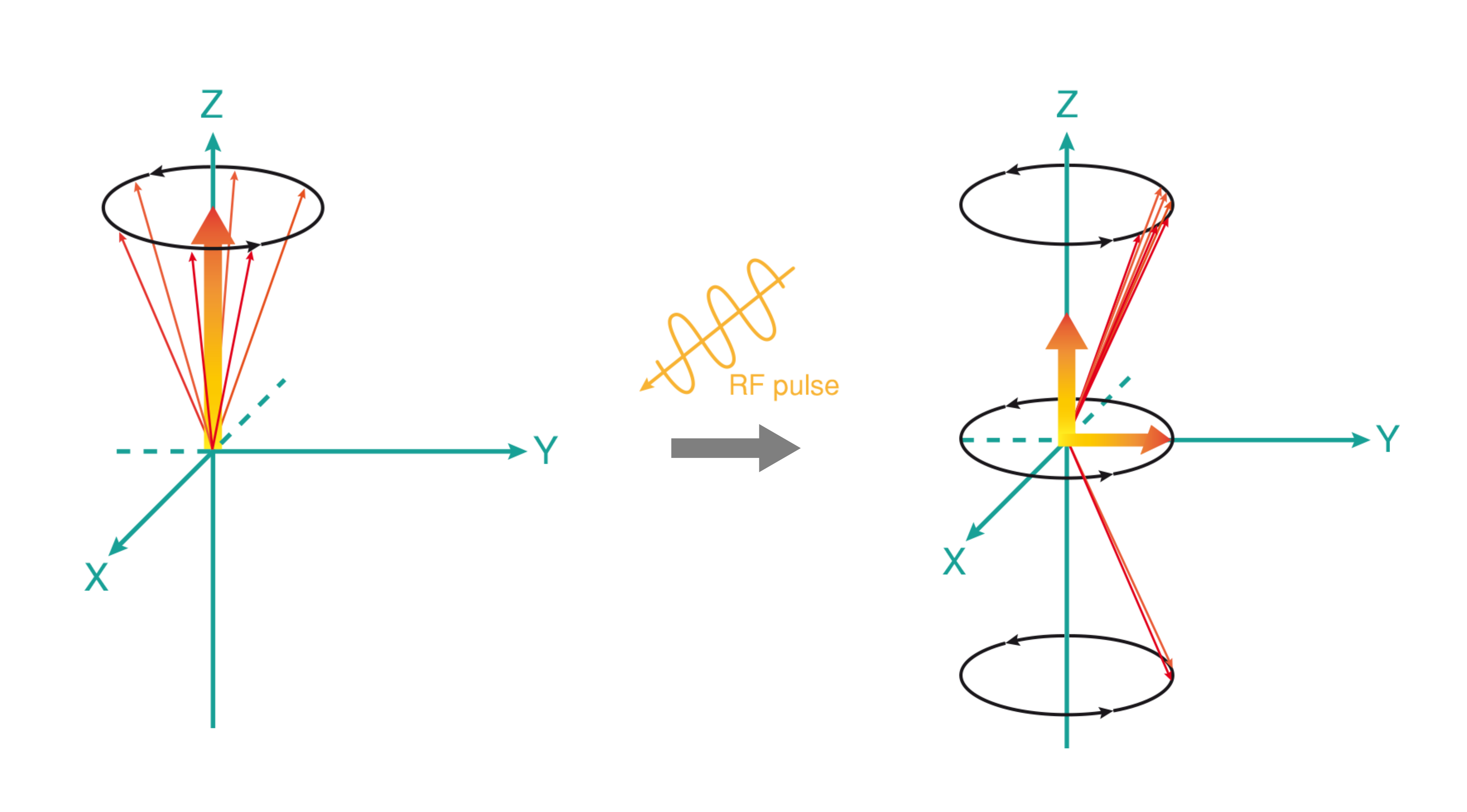}
\caption{Diagram of the effect produced by the radio-frequency pulse on the spin angular momentum of the $H^+$ nuclei. Some $H^+$ nuclei lift their state to the anti-parallel high energy mode and a precession synchronization is attained.}
\label{figure:rationale_mri_nmr1}
\end{figure}

The \ac{NMR} phenomena is related to the excitation produced to the $H^+$ nuclei by the influence of an additional temporal magnetic field (radio-frequency pulse) with a direction different from $B_0$. When a radio-frequency pulse at the Larmor frequency is triggered at $B_0$, an energy exchange occurs with some $H^+$ nuclei, lifting them to the high energy anti-parallel state. In addition, a synchronization of the precession of all the $H^+$ nuclei is induced, so that they all begin to precess in phase. This event causes two magnetic effects in the system: 1) a decrease in the overall spin magnetic momentum $M_z$ due to the lifted $H^+$ nuclei; and 2) the apparition of a spin magnetic momentum $M_{xy}$ transverse to the $B_0$ field due to the phase coherence precession. Such transverse magnetization $M_{xy}$ generates a magnetic signal called the \emph{Nuclear Magnetic Resonance Signal}. Figure \ref{figure:rationale_mri_nmr1} shows a diagram of the effect produced by the radio-frequency pulse on the spin angular momentum of the $H^+$ nuclei.

\begin{figure}[h]
\centering
\includegraphics[width=0.75\linewidth]{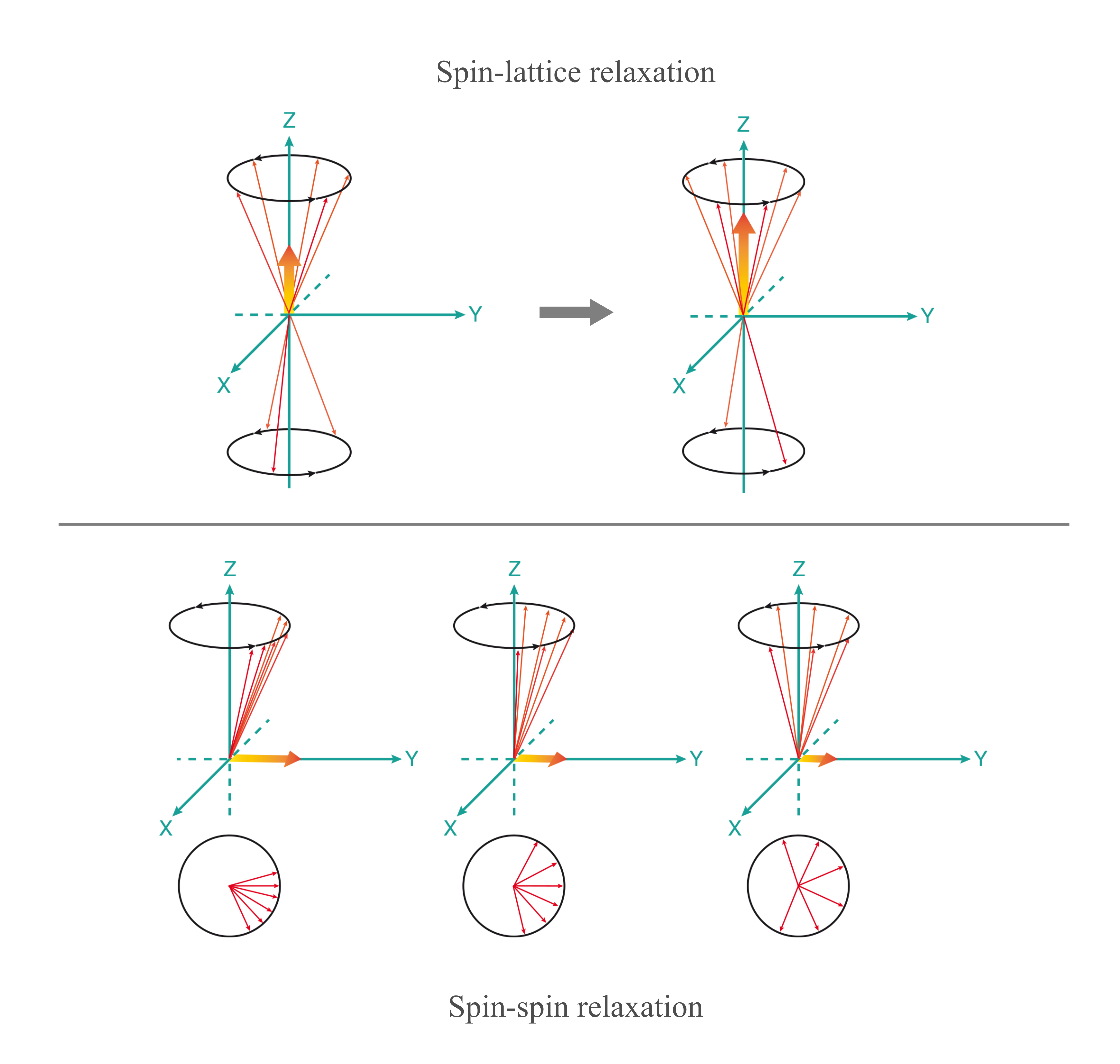}
\caption{Diagram of the relaxation process after the radio-frequency pulse ends. On top, the \Ti{} spin-lattice relaxation, showing a recovery of the original $M_z$ magnetization. On bottom, the \Tii{} spin-spin relaxation, demonstrating the lose of the $M_{xy}$ magnetization due to the phase de-coherence.}
\label{figure:rationale_mri_nmr2}
\end{figure}

Once the radio-frequency pulse has ended, the system begins to relax, recovering its initial state of equilibrium. The precession of the $H^+$ nuclei begins to lose the phase coherence due to the differences in the chemical context of each $H^+$ nucleus. Therefore, the transverse spin magnetic momentum $M_{xy}$ begins to disappear, leading to the so-called \emph{spin-spin} relaxation or \emph{transverse} \Tii{} relaxation process. Likewise, the previously lifted $H^+$ nuclei return to their original low energy parallel state, restoring the longitudinal magnetization $M_z$ to its original value. This process is called the \emph{spin-lattice} relaxation or \emph{longitudinal} \Ti{} relaxation (see figure \ref{figure:rationale_mri_nmr2}).

\Ti{} and \Tii{} relaxation times are significantly different between them, with \Ti{} a longer process than \Tii{}. Generally, \Ti{} relaxation time ranges from 300 to 2000 ms while \Tii{} relaxation is about 30 to 150 ms. Moreover, it is difficult to determine the end of the \Ti{} and \Tii{} relaxations exactly. Therefore, it is considered that the \Ti{} relaxation is completed once the signal recovers the 63\% of the original $M_z$ magnetization. Similarly, the \Tii{} relaxation is considered ended once the $M_{xy}$ signal falls under the 37\% of its original value.

As stated above, the \Ti{} and \Tii{} relaxation is different for each $H^+$ nucleus depending on its chemical context. Hence, different \ac{NMR} signals are emitted during the relaxation process, inducing dissimilar electric signals in the receiver coils of the \ac{MR} machine. Such disparities will in essence determine the intensities for each tissue in the \ac{MR} image. There are three different types of contrast in \ac{MR} images: \Ti{}-weighted, \Tii{}-weighted and \ac{PD} images; which are related to the so-called \ac{TR} and the \ac{TE} times. \ac{TR} is the time between successive radio-frequency pulses and affects the speed in which $H^+$ nuclei realigns to the $B_0$ field. The \ac{TE} refers to the time at which the electrical signal induced by the $H^+$ nuclei is measured in the magnetic coils and concerns the degree of dephase of the spins of the $H^+$ nuclei. Thus, the \ac{TR} is closely connected to \Ti{} relaxation effects, while \ac{TE} is more related to \Tii{} relaxation events.

\begin{figure}[h]
\centering
\includegraphics[width=0.7\linewidth]{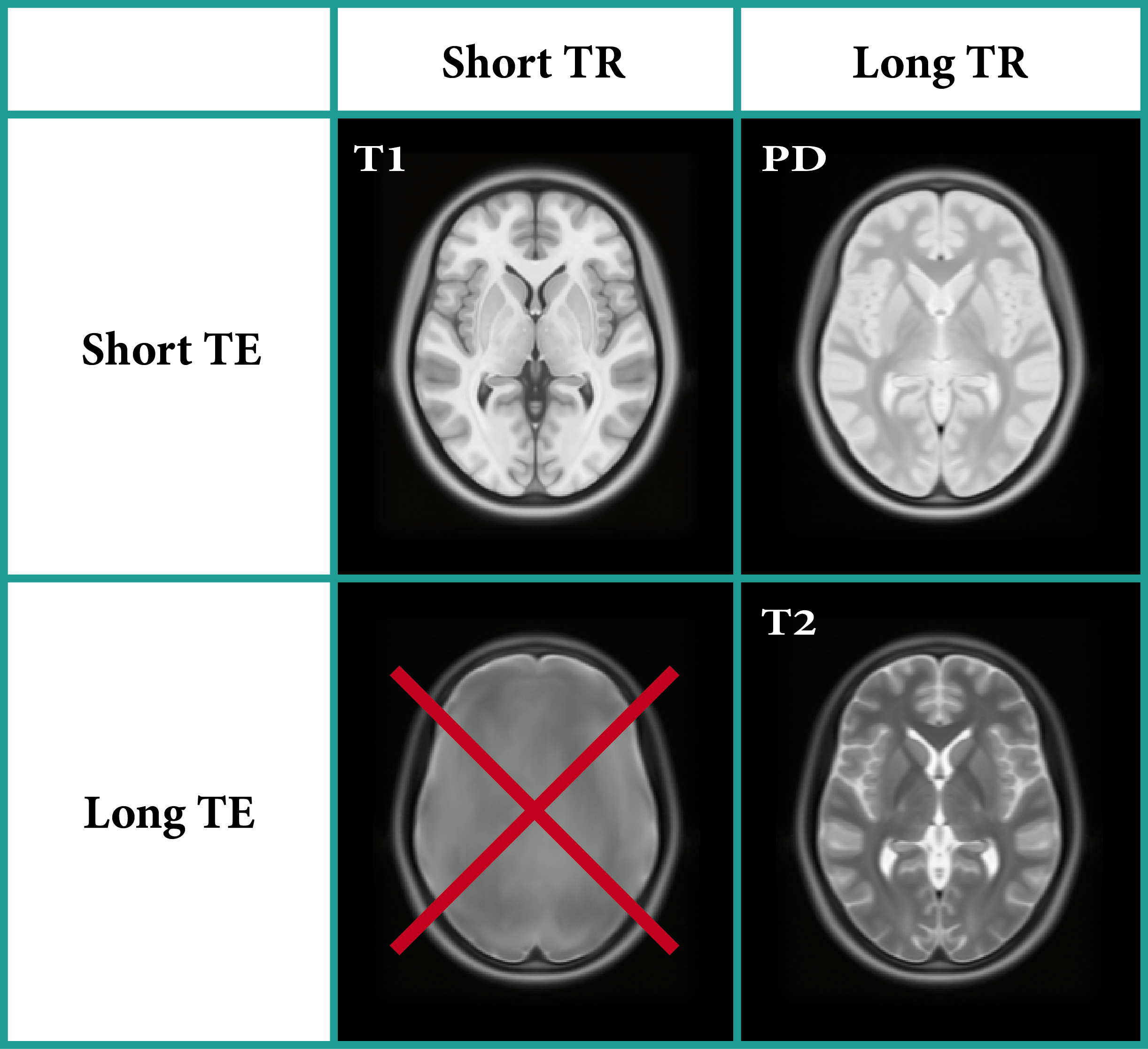}
\caption{Relation between the \ac{TR} and \ac{TE} and the intensity contrast produced in the \ac{MR} images.}
\label{figure:rationale_mri_TRTE}
\end{figure}

Figure \ref{figure:rationale_mri_TRTE} shows the relation between the \ac{TR}, \ac{TE} and the contrast produced on the \ac{MR} images.

\subsection{Anatomical \acs{MRI}}
\label{subsection:rationale_mri_anatomical_mri}
So far we have seen that \ac{NMR} relaxation process is different for each tissue depending on its chemical context. This property allows to visually differentiate human tissues via the measurement of their associated electric signals, generating images that depict the anatomical structures of the region under study.

In the case of central nervous system tumors such as glioblastoma, several images are typically acquired to best approach the lesion. These images intend to anatomically describe the morphology of the lesion and its associated structures.

A common protocol for a \ac{MRI} glioblastoma study should include the following sequences: 

\begin{description}
\item[\textbf{\Ti{}-weighted}]: typically a volumetric 3D \ac{FSE} or \ac{MPRAGE} sequence. The aim of the sequence is the anatomical overview of the lesion, including the soft tissues below the base of skull.
\item[\textbf{\Tic{}-weighted}]: typically a post-contrast volumetric 3D \ac{FSE} or \ac{MPRAGE} sequence. It is a \Ti{}-weighted image acquired with administration of \ac{GBCA}. Its main purpose is to assess vascular structures of the region under study. The list of approved contrast-agents for human use is: gadoterate meglumine (Dotarem\textsuperscript{\tiny{\textregistered}}), gadobutrol (Gadavist\textsuperscript{\tiny{\textregistered}}), gadopentetate dimeglumine (Magnevist\textsuperscript{\tiny{\textregistered}}), gadobenate dimeglumine (MultiHance\textsuperscript{\tiny{\textregistered}}), gadodiamide (Omniscan\textsuperscript{\tiny{\textregistered}}), gadoversetamide (OptiMARK\textsuperscript{\tiny{\textregistered}}), gadoteridol (ProHance\textsuperscript{\tiny{\textregistered}}).
\item[\textbf{\Tii{}-weighted}]: typically a 2D axial \ac{FSE} sequence. Its purpose is the evaluation of basal cisterns, ventricular system and subdural spaces, evaluation of vasogenic edema and good visualization of flow-voids in vessels.
\item[\textbf{FLAIR}]: typically a 2D axial \ac{FLAIR} \Tii{}-weighted sequence. It is a special inversion recovery sequence with a long inversion time. Its main characteristic is that it suppress the signal from the cerebrospinal fluid so that it appears similar to a \Tii{}-weighted image but with the cerebrospinal fluid dark instead of bright. Its purpose is the assessment of white-matter tumor involvement and related vasogenic edema.
\end{description}

\begin{figure}[htbp!]
\centering
\includegraphics[width=\linewidth]{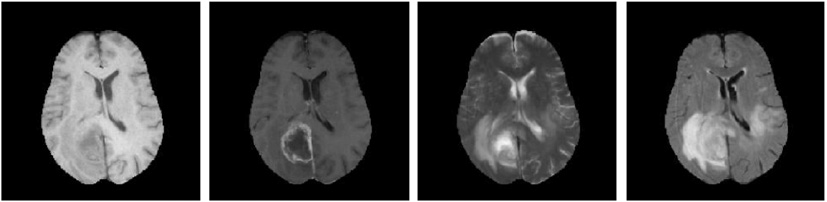}
\caption{Example of anatomical \ac{MR} images typically acquired in a glioblastoma \ac{MR} study. From left to right: \Ti{}-weighted, \Tic{}-weighted, \Tii{}-weighted and \ac{FLAIR} \ac{MRI}.}
\label{figure:rationale_mri_anatomicaltmri}
\end{figure}

Figure \ref{figure:rationale_mri_anatomicaltmri} shows an example of the aforementioned \ac{MR} images for a case affected by glioblastoma.

\subsection{Quantitative \acs{MRI}}
\label{subsection:rationale_mri_qmri}
\ac{qMRI} emerged in the 1980's in the attempt to measure the biological \ac{MR} properties of the tissues \citep{Tofts2004}. Therefore, its purpose is to turn the \ac{MRI} into a scientific measurement instrument and not only on an image acquisition system. This \emph{concept-shift} is the essence of \ac{qMRI}. \ac{qMRI} attempts to develop quantitative and reproducible techniques based on \ac{MR} data, able to measure biomarkers related to the underlying biological processes of the tissues \citep{Yankeelov2011}. \ac{MRI} Relaxometry \citep{Deoni2010}, \ac{DWI} \citep{Schaefer2000} and \acf{PWI} \citep{Svolos2014} composes some of the most important techniques of \ac{qMRI}.

Regarding glioblastoma tumors, one of the most relevant \ac{qMRI} techniques is \ac{PWI}. As previously mentioned, glioblastoma tumors are characterized by a strong vascular proliferation, robust angiogenesis, and extensive microvasculature heterogeneity. In this sense, \ac{PWI} allows the measurement of the kinetic properties of a paramagnetic contrast agent, which is intravenously injected to the patient. \ac{PWI} biomarkers are able to reveal the local vascular properties of the tissues and their hemodynamic behavior. There are three main \ac{PWI} techniques: \ac{DSC} perfusion, \ac{DCE} perfusion and \ac{ASL} perfusion. For reasons of simplicity, and given that only \ac{DSC} perfusion was used in the development of this thesis, \ac{DCE} and \ac{ASL} perfusion quantification will not be covered in this dissertation.  

\subsubsection{\ac{DSC} perfusion \ac{MRI}}
\label{subsubsection:rationale_mri_qmri_perfusion}
\ac{DSC} is the most frequently used technique for \ac{MRI} perfusion of the brain. It relies on the susceptibility-induced signal loss on \Tiis{}-weighted sequences caused by the pass of a \ac{GBCA} bolus through a capillary bed. A rapid repeated image acquisition is performed during the bolus injection, resulting into a series of images with the signal at each voxel representing the susceptibility-induced signal loss of the corresponding tissue, which is proportional to the amount of contrast present in the microvasculature. A mathematical model is then fit to the intensity-time signals to derive various perfusion parameters. The most commonly calculated parameters are \ac{rCBV}, \ac{rCBF} and \ac{MTT}.

Let $S$ the susceptibility-induced signal loss observed at a given voxel. To quantify the \ac{rCBV}, \ac{rCBF} and \ac{MTT} parameters, a \ac{GBCA} concentration-time conversion of $S$ must be performed. It is normally assumed that the tissue concentration of the contrast agent is proportional to the change in \Tiis{} relaxation rate, i.e. $\Delta R2^{\ast}$, by means of:

\begin{equation}
\label{eq:rationale_qpwi_AR2}
\Delta R2^{\ast}\left(t\right) = -\frac{1}{TE} \log \left(\frac{S\left(t\right)}{S_0}\right) = C\left(t\right)
\end{equation}

\noindent where $t$ indicates the time, $TE$ refers to the echo time of the \ac{DSC} sequence and $S_0$ refers to the intensity baseline of $S$.

Hence, \ac{DSC} quantification allows to calculate the amount of \ac{GBCA} tracer remaining in the tissue at each time step at which the $\Delta R2^{\ast}\left(t\right)$ is measured. This is described by the so-called \emph{tissue impulse residue} function, which is denoted as $R\left(t\right)$. Therefore, $R\left(t\right)$ represents the fraction of the tracer still present in the tissue at time $t$ after an instantaneous infinitesimal injection (i.e. a Dirac-delta function) of tracer into a tissue-feeding artery \citep{Knutsson2010}:

\begin{equation}
R\left(t\right) = 1 - \int_0^t h\left(t\right)dt
\end{equation}

\noindent where $h\left(t\right)$ is the distribution of the capillary transit time. By definition, it is considered that $R\left(0\right) = 1$, meaning that all the tracer is present in the tissue at $t=0$, and $R\left(\infty\right) = 0$, meaning that the tracer lefts the tissue after a sufficiently long time (assuming an intact blood–brain barrier with no extravasation).

In practice, however, the arterial tracer bolus arrives at the tissue with a delay dependent on the circulatory path from the injection site to the tissue of interest. Consequently, the measured $\Delta R2^{\ast}\left(t\right)$ signal in the tissue does not reflect the response to an instantaneous arterial input bolus, but to the convolution of a kernel given by $ CBF \cdot R\left(t\right)$ and the $\Delta R2^{\ast}\left(t\right)$ signal in the tissue-feeding artery, i.e. the so-called \ac{AIF} \citep{Rempp1994}:

\begin{equation}
\label{eq:cbf}
\begin{split}
C\left(t\right) &= CBF \cdot R\left(t\right) \otimes AIF\left(t\right) \\
                &= CBF \int_0^t AIF\left(\tau\right) R\left(t-\tau\right)d\tau
\end{split}
\end{equation}

\noindent where $\otimes$ stands for the convolution product and CBF is the Cerebral Blood Flow.

From equation \ref{eq:cbf} we can rapidly observe that knowing the \ac{AIF}, the CBF can be determined by deconvolution of the peak height of the $CBF \cdot R\left(t\right)$ signal, given that $R \in \left[ 0, 1 \right]$.

Dozens of models has been proposed in the literature to successfully estimate the CBF by deconvolution: fast truncated \ac{SVD} family (e.g. s\ac{SVD}, c\ac{SVD}, o\ac{SVD} \citep{Ostergaard1996, Wu2003, Zanderigo2009}), nonlinear stochastic regularization \citep{Zanderigo2009, Pillonetto2010}, Fourier-Hunt frequency-domain deconvolution \citep{Ostergaard1996, Chen2005}, wavelet thresholding \citep{Connelly2006}, classical Tikhonov regularization \citep{Ostergaard1996, Zanderigo2009, Calamante2003}, maximum likelihood estimation \citep{Vonken1999}, maximum entropy deconvolution \citep{Drabycz2005}, Gaussian process deconvolution \citep{Andersen2002}, or hierarchical Bayesian models \citep{Schmid2007, Schmid2011}.

In this thesis, the delay-insensitive o\ac{SVD} method was employed to estimate the CBF, as it has demonstrated to be robust to differences in tracer arrival times and to \ac{AIF} selection variability. Following this approach, the deconvolution problem can be formulated as a matrix equation of the form $c = Ab$:

\begin{equation}
\begin{bmatrix}
C\left(t_0\right) \\
C\left(t_1\right) \\
\vdots \\
C\left(t_{N-1}\right) \\
\end{bmatrix}
= \Delta t
\begin{bmatrix}
AIF\left(t_0\right) & 0 & \dots & 0 \\
AIF\left(t_1\right) & AIF\left(t_0\right) & \dots & 0 \\
\vdots & \vdots & \ddots & \vdots \\
AIF\left(t_{N-1}\right) & AIF\left(t_{N-2}\right) & \dots & AIF\left(t_0\right) \\
\end{bmatrix}
\begin{bmatrix}
R\left(t_0\right) \\
R\left(t_1\right) \\
\vdots \\
R\left(t_{N-1}\right) \\
\end{bmatrix}
CBF
\end{equation}

The above equation can be solved for $b$, with $b = R\left(t\right) \cdot CBF$. By decomposing $A = U \cdot S \cdot V^T$, with $U$ and $V$ orthogonal matrices and $S$ a non-negative square diagonal matrix, the inverse of $A$ can be expressed as $A^{-1} = V \cdot W \cdot U^T$ where $W = 1/S$ along the diagonal only where values are higher than a threshold (truncation of the diagonal for numerical stability reasons), and zero elsewhere. Therefore, we can estimate $CBF$ by solving for $b$ and calculating its maximum value:

\begin{equation}
\label{eq:rationale_mri_qmri_perfusion_cbf}
CBF = \max \left( V \cdot W \cdot U^T \cdot C \right)
\end{equation}

This procedure assumes that the $AIF$ always arrive before the voxel signal $C$, which in some situations may not be true (for example if the $AIF$ is not the true $AIF$ of the corresponding tissue). In such situations, the computation of the $CBF$ results in an incorrect estimation, typically an underestimation of the $CBF$, due to a shifting in the $R\left(t \right)$ function.

As proposed by \cite{Wu2003} circular deconvolution can be employed to solve this problem. By zero-padding the $N$ point-time series $AIF\left(t\right)$ and $C\left(t\right)$ to length $L$, with $L \geq 2N$, time aliasing can be avoided. Thus, replacing also the matrix $A$ by a block-circulant matrix $D$ of the form:

\begin{equation}
d_{i,j} = 
\begin{cases}
a_{i,j}, & \text{for} j \leq i \\
a_{L+i-j,0}, & \text{otherwise} \\
\end{cases}
\end{equation}

\noindent and solving using the equation \ref{eq:rationale_mri_qmri_perfusion_cbf}, the $CBF$ can be estimated safely also even if delay effects exist in the $AIF$. 

\bigskip

The \ac{MTT} can be calculated by Zierler's area-to-height relationship \citep{Zierler1962}, by the equation:

\begin{equation}
\label{eq:mtt}
MTT = \frac{\int_0^\infty R\left(t\right)dt}{\max\left[ R\left(t \right)\right]} = \int_0^\infty R\left(t\right)dt
\end{equation}

According to the central volume theorem \citep{Meier1954, Weisskoff1993}, the CBV can be calculated following the relation:

\begin{equation}
CBV = CBF \cdot MTT
\end{equation}

Following this relation and combining equations \ref{eq:cbf} and \ref{eq:mtt}, the CBV can also be estimated by:

\begin{equation}
\label{eq:cbv}
CBV = \frac{\int_0^\infty C\left(t\right)dt}{\int_0^\infty AIF\left(t\right)dt}
\end{equation}

\bigskip
\noindent \emph{\textbf{Leakage correction}}
\medskip

Computational kinetic models for \ac{DSC} quantification explicitly assume that the \ac{GBCA} remains in the intra-vascular space for the duration of the perfusion acquisition. However, this assumption is typically not valid for glioblastomas, since they often present blood-brain barrier breakdown. Therefore, \ac{GBCA} extravasates to the extra-vascular interstitial space producing a distortion in the $\Delta R2^{\ast}\left(t\right)$ signal that leads to an overestimation or underestimation of perfusion parameters if non leakage correction is performed. \ac{GBCA} leakage can be manifested in $\Delta R2^{\ast}\left(t\right)$ signals in generally two ways: the \Tiis{}-dominant extravasation and the \Ti{}-dominant extravasation effects.

Numerous methods have been proposed in the literature to correct the leakage effect in perfusion, and the deviation it produces in the estimation of the perfusion markers: use of pre-dosing \citep{Donahue2000}, double-echo acquisitions \citep{Vonken2000, Uematsu2001} or parametric modeling ($\gamma$-variate fit of the $\Delta R2^{\ast}\left(t\right)$ signals) \citep{Thompson1964}. The most common and reference technique is the one proposed by \cite{Weisskoff1994} and later elaborated by \cite{Boxerman2006}, which corrects the leaky signal by estimating its deviation from a non-leaky reference signal of the same patient.

Following Boxerman's method, the $\Delta R2^{\ast}\left(t\right)$ observed at a given voxel can be determined as:

\begin{equation}
\Delta R2^{\ast}\left(t\right) \approx K_1 \overline{\Delta R2^{\ast}}\left(t\right) - K_2 \int_0^t \overline{\Delta R2^{\ast}}\left(t^{\prime}\right) dt^{\prime}	
\end{equation}

\noindent where $\overline{\Delta R2^{\ast}}\left(t\right)$ stands for a whole-brain average signal in non-enhancing voxels, $K_1$ is a scale-constant of the whole-brain average signal to fit the observed signal at the corresponding voxel, and $K_2$ is the term that reflects the effects of the leakage (both \Ti{} and \Tii{} dominant effects). Therefore, by simple manipulation (please refer to \cite{Boxerman2006} for more details), the leakage corrected signal can be determined as:

\begin{equation}
\Delta R2_{corr}^{\ast}\left(t\right) = \Delta R2^{\ast}\left(t\right) + K_2 \int_0^t \overline{\Delta R2^{\ast}}\left(t^{\prime}\right) dt^{\prime}	
\end{equation}

\begin{figure}[htbp!]
\centering
\includegraphics[width=\linewidth]{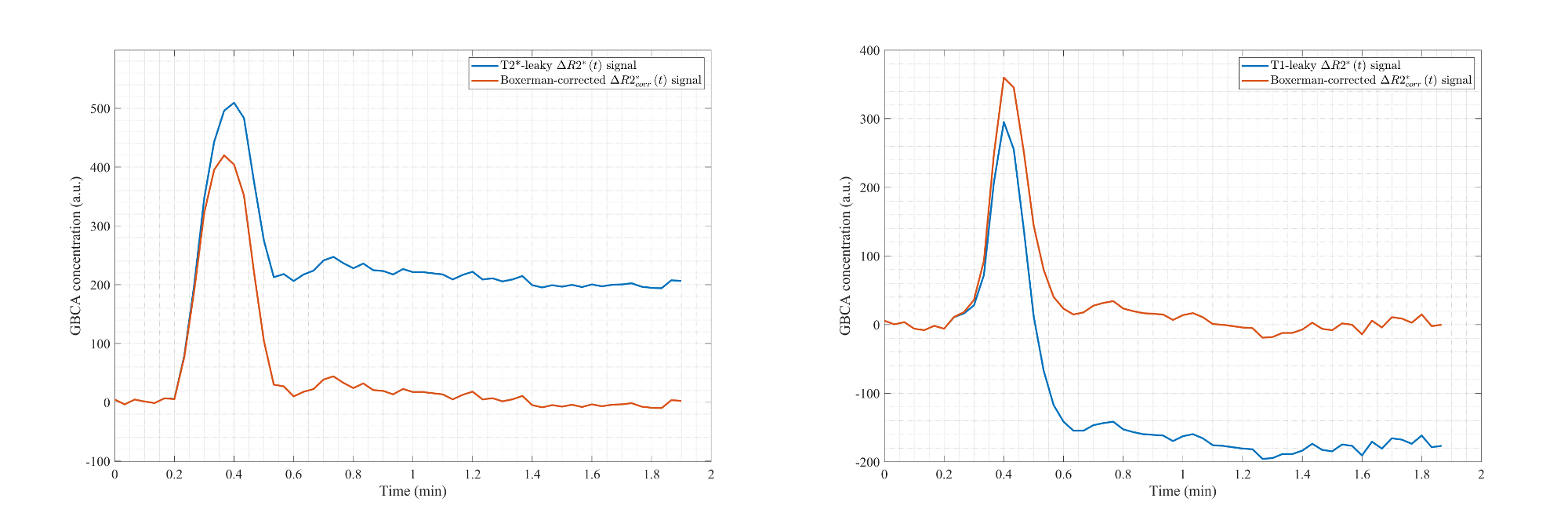}
\caption{Example of a \Tiis{}- and \Ti{}-leaky curves with their associated Boxerman's corrected concentration curves.}
\label{figure:rationale_mri_leakage_corrected_curves}
\end{figure}

Both $K_1$ and $K_2$ can be determined by simple least squares fitting, and then corrected $\Delta R2_{corr}^{\ast}\left(t\right)$ signals can be computed to quantify $CBV$, $CBF$ and $MTT$ as stated above. Figure \ref{figure:rationale_mri_leakage_corrected_curves} shows an example of the Boxerman's correction of a \Tiis{}- and a \Ti{}-leaky curve.

\bigskip
\noindent \emph{\textbf{Recirculation correction}}
\medskip

\ac{DSC} quantification involves the kinetic analysis of the first pass of a intravenously injected \ac{GBCA}. However, the $\Delta R2^{\ast}\left(t\right)$ signal observed at each voxel primarily reflects the first pass of the tracer, but also the second wave of \ac{GBCA} that recirculates into the tissue, after it has been shunted through the renal and coronary circulations and back into the heart. This second wave, affects the $\Delta R2^{\ast}\left(t\right)$, preventing the signal from returning to its original baseline.

Hence, in order not to overestimate the perfusion parameters, typically the $CBV$, which is obtained from the integral or area under the $\Delta R2^{\ast}\left(t\right)$ signal, the recirculation phase must be corrected. One of the most popular methods is to use a $\gamma$-variate fitting.

The $\gamma$-variate function commonly employed to describe the first pass of the \ac{GBCA} is written as:

\begin{equation}
S\left(t\right) = \mathcal{K} \cdot \left(t - t_0\right)^{\alpha} \exp^{-\left(t - t_0\right) / \beta} 
\end{equation}

\noindent where $\mathcal{K}$ is a scaling factor, $t_0$ is the bolus arrival time, and $\alpha$ and $\beta$ determine the shape and scale of the distribution.

Levenberg-Marquardt algorithm was employed in this thesis to perform the $\gamma$-variate fitting of all the $\Delta R2^{\ast}\left(t\right)$ signals of the \ac{DSC} perfusions. Levenberg-Marquardt algorithms searches for minimizing parameters in the form:

\begin{equation}
\left(\hat{\mathcal{K}}, \hat{t_0}, \hat{\alpha}, \hat{\beta} \right) = \argmin_{\mathcal{K}, t_0, \alpha, \beta} \sum_{t=0}^N \left[ \Delta R2^{\ast}\left(t\right) - S\left(t | \mathcal{K}, t_0, \alpha, \beta \right)  \right]^2
\end{equation}

\begin{figure}[htbp!]
\centering
\includegraphics[width=0.7\linewidth]{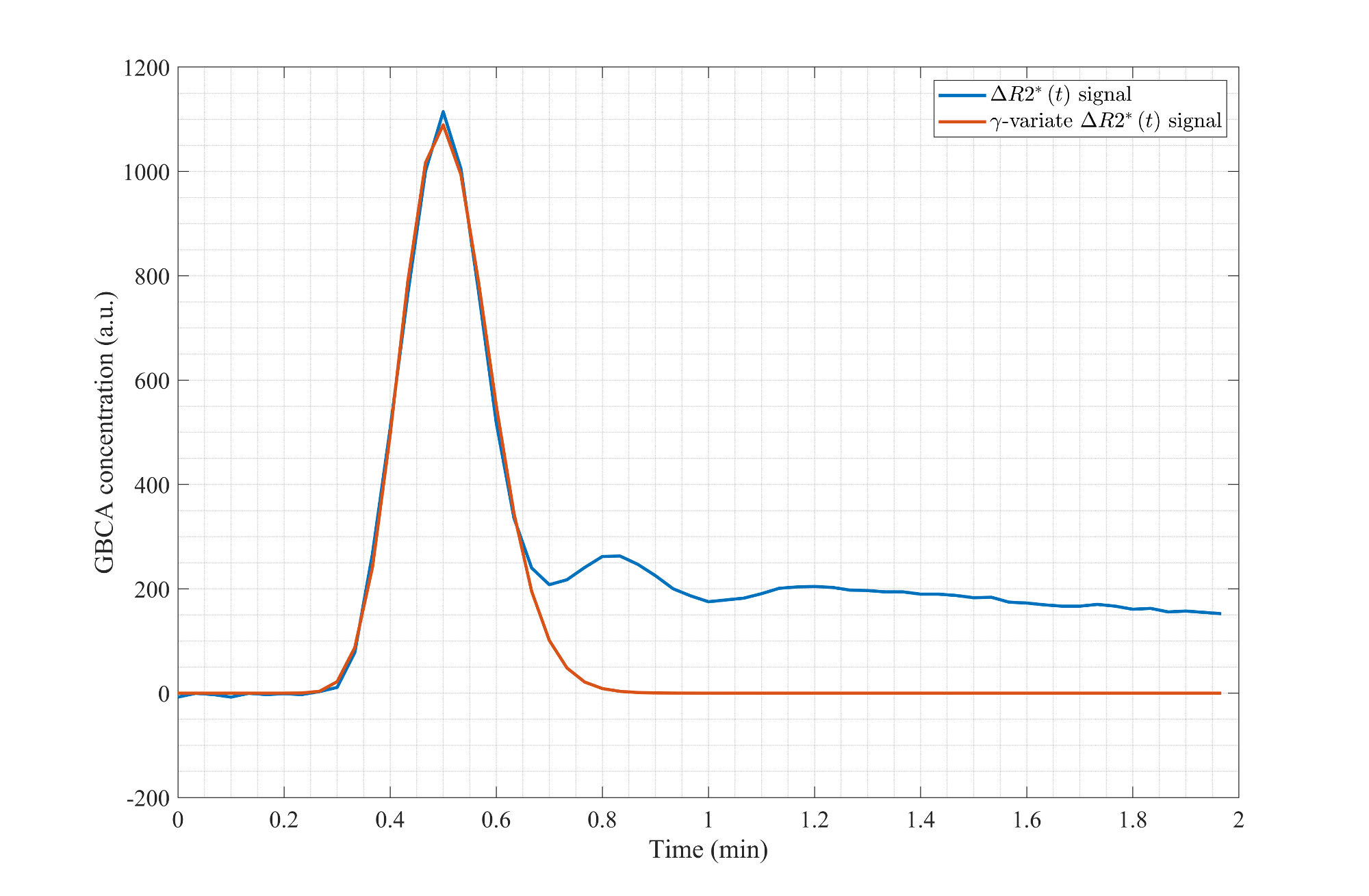}
\caption{Example of a $\gamma$-variate fitting of a $\Delta R2^{\ast}\left(t\right)$ signal.}
\label{figure:rationale_mri_dsc_gammafit}
\end{figure}

Figure \ref{figure:rationale_mri_dsc_gammafit} shows an example of a $\gamma$-variate fitting of a $\Delta R2^{\ast}\left(t\right)$ signal.

\bigskip
\noindent \emph{\textbf{Contralateral white matter normalization}}
\medskip

Finally, the \ac{rCBV} and \ac{rCBF} parameters are obtained by normalizing the CBV and CBF quantities against the contralateral unaffected white matter value. This procedure has shown to successfully standardize perfusion values among individuals, regardless of their \ac{MRI} protocol, hence allowing inter-patient multi-center studies \citep{Alvarez2019}.

\section{Probability theory}
\label{section:rationale_probability_theory}

\subsection{Basic concepts}
\label{subsection:rationale_probability_basic_concepts}
In the following sections, the basic concepts of probability theory will be described to facilitate the reading and understanding of subsequent chapters and contributions of this thesis.

\subsubsection{Random variables}
\label{subsubsection:rationale_probability_random_variables}
A random variable can be informally defined as a variable whose possible values are the outcomes of a random phenomenon. Thus, random variables conceptually represent an abstraction that allows addressing uncertainty in the measurement of quantities in real world scenarios.

More formally, a random variable $X: \Omega \rightarrow E$ is mathematically defined as a measurable function from a possible set of outcomes $\Omega$ to a measurable space $E$. Random variables can be of two types: \textbf{discrete} random variables, whose domain is a countable number of distinct values (typically $E = \mathbb{N}$); and \textbf{continuous} random variables, whose domain takes an infinite number of values in a predefined interval (typically $E = \mathbb{R}$). Usually, random variables are denoted by uppercase letters, e.g. $X$ and their realizations by lowercase letters, e.g. $x$. Hence, for example, a random variable that represents the volume of the brain in cubic centimeters in a population could denoted as $X$, and the particular measurement of a patient within this population could be denoted as $x=1273.6$ cm\textsuperscript{3}.

Random variables are usually governed by probability distributions that represent the likelihood that any of the possible values that the variable can take would occur. Such distributions constitute a fundamental principle to \ac{ML} as they represent the mathematical approximation to deal with the uncertainty in the process of learning patterns from real data.


\subsubsection{Probability distribution}
\label{subsubsection:rationale_probability_marignal_distribution}
A probability distribution is a mathematical function that describes the likelihood of obtaining the possible values that a random variable can assume. Given a random variable $X$, the probability that it takes the value $x$ is denoted as:

\begin{equation}
p\left( X=x \right)
\end{equation}

\noindent where for the shake of simplicity we can assume $p\left( X=x \right) \equiv p\left( x \right)$.

Probability distributions are generally divided into two classes depending on the type of the random variable. For discrete random variables, the probability distribution is known as \emph{\ac{pmf}}, denoted as $f\left( x \right)$, and gives the probability mass that the random variable $X$ takes exactly the value $x$.

\begin{equation}
p\left( x \right) = f\left( x \right) ~ \rightarrow ~ \left[ 0,1 \right]
\end{equation}

Since the probability mass is distributed among all the possible outcomes that the discrete random variable can take, then:

\begin{equation}
\sum\limits_{x \in E} f\left( x \right) = 1
\end{equation}

The \emph{\ac{cdf}}, denoted as $F\left( x \right)$, is defined as the probability that the random variable $X$ takes values less than or equal to $x$, and takes the form:

\begin{equation}
F\left( x \right) = p\left( X \leq x \right) = \sum\limits_{\substack{y \leq x \\ y \in E}} f\left( y \right)
\end{equation}

On the other hand, the probability distribution of a continuous random variable is known as \emph{\ac{pdf}}, denoted also as $f\left( x \right)$. Since for continuous random variables there is an infinite number of values in any interval, the \ac{pdf} for an exact specific value is not meaningful. Instead, the \ac{pdf} for a continuous random variable is commonly defined over an interval $\left[ a,b \right]$ in the form:

\begin{equation}
p\left( a \leq X \leq b \right) = \int_a^b f\left( x \right)dx ~ \rightarrow ~ \left[ 0,1 \right]
\end{equation}

For the shake of simplicity, we can assume that for continuous random variables $p\left( x \right)$ refers to the \ac{pdf} of $x$ bounded to the infinitesimal interval $\left[ x, x+dx \right]$:

\begin{equation}
p\left( x \right) = \int_x^{x+dx} f\left( y \right)dy
\end{equation}

Likewise the discrete case, the probability density is distributed among all the interval of possible values that the continuous random variable can take, which in the most general case is $\left[ -\infty, +\infty \right]$, thus:

\begin{equation}
\int_{-\infty}^{+\infty} f\left( x \right) dx = 1
\end{equation}

The \ac{cdf} for a continuous random variable is finally defined as:

\begin{equation}
F\left( x \right) = \int_{-\infty}^x f \left( y \right)dy 
\end{equation}

\subsubsection{The joint, marginal and conditional probability distributions}
\label{subsubsection:rationale_probability_joint_conditional_distributions}
In many situations we will manage more than one random variable that operates on the same probability space. In that situations, the joint, marginal and conditional probability distributions arise to model uncertain events from different interrelated events.

The joint probability distribution, denoted as $p\left( x, y \right)$, is defined as the probability distribution of two random variables operating in the same probability space whose outcomes occur simultaneously. The conditional distribution, denoted as $p \left( x | y \right)$, is defined as the probability distribution of a random variable when another random variable is known to have a particular value. Both distributions are closely related through the product rule, which takes the form:

\begin{equation}
p \left( x, y \right) = p\left( x | y \right) p \left( y \right) = p \left( y | x \right) p \left( x \right)
\end{equation}

We can also relate these distributions with the marginal probability distribution through the sum rule, i.e. by integrating out a random variable that takes all the values in its subset. For discrete random variables the marginal probability distribution is defined as:

\begin{equation}
p\left( x \right) = \sum\limits_y p \left( x, y \right) = \sum\limits_y p \left( x | y \right) p \left( y \right)
\end{equation}

\noindent while for continuous random variables it is expressed as:

\begin{equation}
p \left( x \right) = \int_y p \left( x, y \right) dy = \int_y p \left( x | y \right) p \left( y \right) dy
\end{equation}

\subsubsection{The Bayes' theorem and Bayes decision rule}
\label{subsubsection:rationale_probability_bayes_theorem}
The Bayes' theorem plays a central role in \ac{ML} as it provides a relationship between conditional and joint probability distributions of two random variables, which defines the so-called \emph{minimum-risk decision rule} \citep{Duda2000, Bishop2006}. Therefore, the Bayes' theorem allows to take \emph{optimal} decisions for uncertain events based on prior knowledge and conditions related to that events.

The Bayes' theorem, also named Bayes' rule or Bayes' law, is defined as:

\begin{equation}
\label{eq:bayestheorem}
p \left( y | x \right) = \frac{p \left( x | y \right ) p \left( y \right)}{p \left( x \right)}
\end{equation} 

Under the context of a pattern recognition classification problem, these distributions have specific interpretations related to the degree of belief they give to the occurrence of different events. In this sense, the Bayes' formula can be informally expressed as:

\begin{equation}
\text{posterior} = \frac{\text{likelihood} \times \text{prior}}{\text{evidence}}
\end{equation}

\noindent where each term semantically means:
\begin{description}[leftmargin=0cm]
\item[$p \left( y | x \right)$]: \emph{posterior probability}, that is the quantity that should be estimated and represents the degree of belief of observing the event (or \emph{class} in a classification problem) $y$, after taking into account the evidence $x$.
\item[$p \left( x | y \right)$]: \emph{likelihood distribution}, it acts as a model of the random variable $Y$ that specifies the most probable outcome of $X$ given a specific state of $Y$.
\item[$p \left( y \right)$]: \emph{prior distribution}, it provides the degree of initial belief of observing the event $y$ of the random variable $Y$.
\item[$p \left( x \right)$]: \emph{evidence distribution}, it acts as a normalization term that provides the total degree of belief of the evidences $X$.
\end{description}

Note that the equation \ref{eq:bayestheorem} can also be expressed in terms of joint probability distributions in the form:

\begin{equation}
p \left( y | x \right) = \frac{p \left( x,y \right)}{\sum\limits_z p \left( x,z \right)}
\end{equation} 

\noindent from where it can be observed that the denominator acts as a normalization constant to ensure the conditional probability to sum 1.

Therefore, assuming $Y$ to be a discrete random variable that models the different classes of a pattern recognition classification problem, the Bayes' minimum-risk decision rule, sometimes also called minimum error-rate classifier, can be expressed as:

\begin{equation}
\text{Decide}~y~\text{if}~p \left( y | x \right) > p \left( z | x \right)~~\forall z \neq y
\end{equation}

\noindent or more formally defined as:

\begin{equation}
\label{eq:bayesdecisionrule}
\hat{y} = \argmax_y p \left( y | x \right)
\end{equation}

\noindent where $\hat{y}$ refers to the optimal class for the decision problem.

Following the Bayes' theorem, we can reformulate the equation \ref{eq:bayesdecisionrule} to:

\begin{equation}
\begin{split}
\hat{y} &= \argmax_y p \left( y | x \right) \\
        &= \argmax_y \frac{p\left( x | y \right) p \left( y \right)}{p \left( x \right)} \\
        &= \argmax_y p \left( x | y \right) p \left( y \right) \\
        &= \argmax_y p \left( x,y \right)
\end{split}
\end{equation}

\noindent from where it can be observed that we can drop out the $p \left( x \right)$ term as it is constant for the maximization (which is only dependent on $y$).

\medskip

The previous result directly leads to the definition of the two main families of models within the \ac{ML} classification problem: the \emph{generative} models and the \emph{discriminative} models. Generative models explicitly model the \emph{joint distribution} $p \left( x,y \right)$, which semantically means that they learn a model to describe each class. By the opposite, the discriminative models directly learn the \emph{posterior distribution} $p \left( y | x \right)$, which just models the decision boundaries between the different classes of the problem, ignoring the properties or characteristics of the classes.

\begin{figure}[h]
\centering
\includegraphics[width=0.9\linewidth]{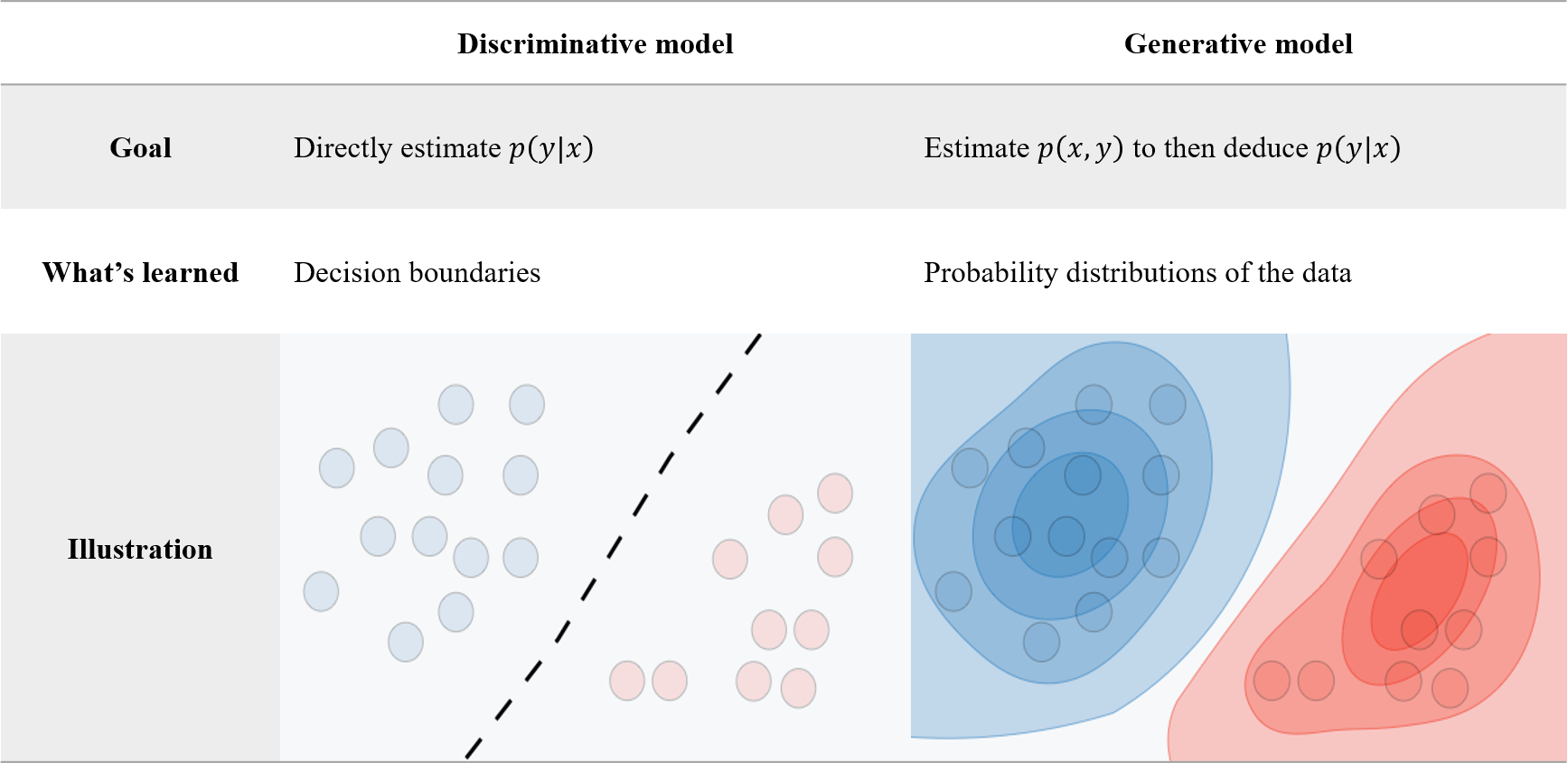}
\caption{Summary of the goals and learning objectives of the generative and discriminative models for a pattern recognition classification problem. Image taken with kind permission from \url{https://stanford.edu/~shervine/teaching/cs-229/cheatsheet-supervised-learning}.}
\label{figure:GenerativeDiscriminativeModels}
\end{figure}

Figure \ref{figure:GenerativeDiscriminativeModels} illustrates both approaches, with the corresponding associated probability distributions and learning goals for each model.

\subsection{Probability distributions}
\label{subsection:rationale_probability_distributions}

\subsubsection{The exponential family}
\label{subsubsection:rationale_probability_exponential_family}
The exponential family of probability distributions is an unified set of distributions that can be expressed in the form:

\begin{equation}
p \left( x; \Theta \right) = h \left( x \right)\exp \left( \eta \left( \Theta \right) \cdot T \left( x \right) - A \left( \Theta \right) \right)
\end{equation}

\noindent for some functions $h \left( x \right)$, $\eta \left( \theta \right)$, $T \left( x \right)$ and $A \left( \Theta \right)$; where $\Theta$ refers to the set of \emph{parameters} of the distribution.

The exponential family represents one of the most important families of distributions in statistics as it covers a wide range of distributions that naturally arise in many natural phenomena. Among others, the most important probability distributions of the exponential family are: the \emph{Normal}, \emph{t-Student}, \emph{Gamma}, \emph{Multinomial} and \emph{Dirichlet} distributions, which are employed in some contributions of this thesis.

\subsubsection{The Normal distribution}
\label{subsubsection:rationale_probability_normal}
The most important distribution of the exponential family, and in statistics in general, is the Normal distribution. The Normal distribution, also called Gaussian distribution, is a continuous symmetric probability distribution defined in the range $\left[-\infty, +\infty \right]$, with parameters $\Theta = \left\lbrace \mu, \sigma^2 \right\rbrace$, that takes the form:

\begin{equation}
p \left( x; \mu, \sigma^2 \right) = \frac{1}{\sqrt{2\pi\sigma^2}}\exp\left( -\frac{\left( x - \mu \right)^2}{2\sigma^2} \right)
\end{equation}

\noindent where $\mu$ is referred as the \emph{mean} and represents the central tendency or \emph{expected value} of the random variable governed by the distribution; and $\sigma^2$ is referred as the \emph{variance} and represents the degree of divergence of the realizations of the random variable from the central tendency.

The Normal distribution is also a fundamental distribution for probability theory as it arises from the \emph{central limit theorem}, which states that the summation of independent random variables (properly normalized) tends to be normally distributed. This theorem has important implications since it allows to employ statistical methods designed for normal distributions to other problems involving non-normal distributions.

The generalization of the one-dimensional Normal distribution to the $d$-dimensional multivariate case with $d > 1$ is defined as:

\begin{equation}
p \left( \vec{x}; \vec{\mu}, \Sigma \right) = \frac{1}{{ \left( 2\pi \right)}^{n/2} |\Sigma|^{1/2}}\exp\left( -\frac{1}{2}\left( \vec{x} - \vec{\mu} \right)^T \Sigma^{-1} \left( \vec{x} - \vec{\mu} \right) \right)
\end{equation}

\noindent where $\Sigma$ refers to the covariance matrix, and models the variance of each independent variable and the pair-wise covariance interactions between them.

The Normal distribution is typically denoted as $\mathcal{N} \left( \mu, \sigma^2 \right)$ for the univariate case and $\mathcal{N} \left( \vec{\mu}, \Sigma \right)$ for the multivariate case.

\subsubsection{The t-Student distribution}
\label{subsubsection:rationale_probability_tstudent}
The t-Student distribution is another continuous distribution of the exponential family that arises from the estimation of the mean of a population that is normally distributed. It is a symmetric zero-centered distribution, also defined in the range $\left[-\infty, +\infty \right]$, with parameters $\Theta = \left\lbrace \nu \right\rbrace$, that takes the form:

\begin{equation}
p \left( x; \nu \right) = \frac{\Gamma\left( \frac{\nu + 1}{2} \right)}{\Gamma\left(\frac{\nu}{2}\right)\sqrt{\nu\pi}}\left( 1 + \frac{x^2}{\nu} \right)^{-\frac{\nu + 1}{2}}
\end{equation}

\noindent with $\nu > 0$ referring to the \emph{degrees of freedom} of the distribution.

The t-Student distribution is very similar to the Normal distribution. In the limit $\nu \rightarrow +\infty  $ the t-Student distribution exactly converges to a Normal typified distribution, while for values of $\nu \in [1, 30]$ the distribution presents heavier and thicker tails, making it a natural choice for robust data modeling in presence of outliers. 

\begin{figure}[h]
\centering
\includegraphics[width=0.75\linewidth]{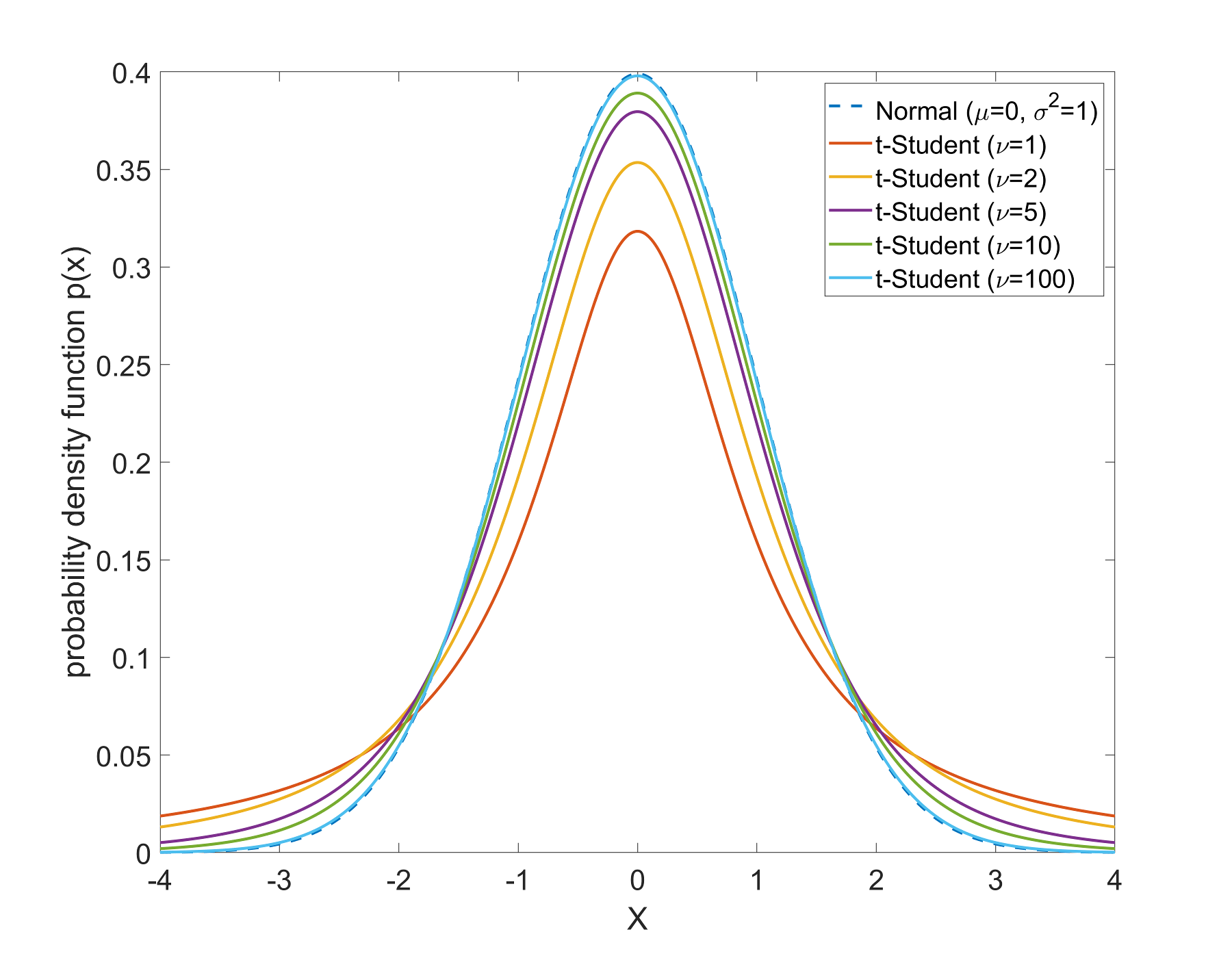}
\caption{Comparison between probability density functions of a typified Normal distribution $\left( \mu=0, \sigma^2=1 \right)$ and different instances of t-Student distributions with parameters $\nu = \left\lbrace 1, 2, 5, 10, 100 \right\rbrace$}
\label{figure:tStudent}
\end{figure}

The distribution can be generalized to a location-scale distribution \citep{Bishop2006} by compounding a Normal distribution of mean $\mu$ and unknown variance with an inverse Gamma distribution placed over the variance with parameters $\alpha=\nu/2$ and $\beta=\nu\sigma^2 / 2$. The resulting density has the form:
 
\begin{equation}
p \left( x; \mu, \sigma^2, \nu \right) = \frac{\Gamma\left( \frac{\nu + 1}{2} \right)}{\Gamma\left(\frac{\nu}{2}\right)\sqrt{\nu\pi\sigma^2}}\left( 1 + \frac{1}{\nu} \frac{\left( x - \mu \right) ^ 2}{\sigma^2} \right)^{-\frac{\nu + 1}{2}}
\end{equation}

The generalization of the one-dimensional t-Student distribution to the $d$-dimensional multivariate case is defined as:

\begin{equation}
p \left( \vec{x}; \vec{\mu}, \Sigma, \nu \right) = \frac{\Gamma\left( \frac{\nu + d}{2} \right)}{\Gamma\left(\frac{\nu}{2}\right)\left(\nu\pi\right)^{d/2}|\Sigma| ^ {1/2}}\left( 1 + \frac{1}{\nu} \left( \vec{x} - \vec{\mu} \right)^T \Sigma^{-1} \left( \vec{x} - \vec{\mu} \right)\right)^{-\frac{\nu + d}{2}}
\end{equation}

\noindent where $d$ refers to the dimensions of the random variable and $\Sigma$ refers to the scale or shape matrix, which in general is not equivalent to the covariance matrix of the Normal distribution.

The t-Student distribution is typically denoted as $\mathcal{S}t \left( \nu \right)$,  or $\mathcal{S}t \left( \mu, \sigma^2, \nu \right)$ and $\mathcal{S}t \left( \vec{\mu}, \Sigma, \nu \right)$ for the location-scale univariate and multivariate versions respectively.

\subsubsection{The Gamma distribution}
\label{subsubsection:rationale_probability_gamma}
The Gamma distribution is another continuous probability distribution of the exponential family, which is very common in the biomedical field as many biological phenomena follow a Gamma distribution. It is a positive-only right-skewed distribution defined in the range $\left(0, +\infty \right]$, with parameters $\Theta = \left\lbrace k, \theta \right\rbrace$, that takes the form:

\begin{equation}
p \left( x; k, \theta \right) = \frac{1}{\Gamma\left( k \right)\theta^k} x^{k-1}\exp\left( -\frac{x}{\theta} \right)
\end{equation}

\noindent where $k$ is referred to as the \emph{shape} parameter and $\theta$ is referred to as the \emph{scale} parameter.

Such distribution is widely employed to model time-dependent biological phenomena that have a natural minimum of 0, such as the \ac{MR} signal decay produced by the first pass of the bolus of a paramagnetic contrast agent intravenously injected to a patient.

The generalization of the univariate Gamma distribution to the multivariate case is the Wishart distribution. This distribution is defined for positive-definite $ \left( d \times d \right)$ matrices that represents the \emph{scatter} matrix of two $d$-variate Normal-distributed random variables, which falls far outside the scope of this thesis.

The Gamma distribution is typically denoted as $\mathcal{G} \left( k, \theta \right)$.

\subsubsection{The Multinomial distribution}
\label{subsubsection:rationale_probability_multinomial}
The Multinomial distribution is a discrete probability distribution that models the probability of observing each of $K$ different outcomes in an experiment involving $n$ repeated trials. Therefore, the distribution governs a discrete random variable $X$, whose realizations are vectors of the form $\vec{x} = \left( x_1, \ldots, x_K \right)$, with $x_k \in \left\lbrace 0, \ldots, n \right\rbrace$ representing the number of times the $k^{th}$ outcome has been observed after the $n$ trials.

The Multinomial distribution has the parameters $\Theta = \left\lbrace \vec{p}, n \right\rbrace$, with $\vec{p} = \left( p_1, \ldots, p_K \right)$, representing the prior probability of observing each of the $K$ possible outcomes before the trials, and $n$ indicating the number of trials. The \ac{pmf} of the distribution takes the form:

\begin{equation}
\label{eq:multinomial}
p \left( \vec{x}; \vec{p}, n \right) = 
    \begin{cases}
      \frac{n!}{x_1!\ldots x_K!}p_1^{x_1} \ldots p_K^{x_K} & \text{when} \sum\limits_{i=1}^K=n \\
      0 & \text{otherwise}
    \end{cases}
\end{equation}

The Multinomial distribution is the generalization of several discrete distributions, those ones taking specific values for $k$ and $n$ parameters. Thus, the Multinomial distribution converts to the \emph{Bernoulli} distribution when $k = 2$ and $n = 1$; to the \emph{Binomial} distribution when $k = 2$ and $n > 1$; and to the \emph{Categorical} distribution when $k > 1$ and $n = 1$.

The Multinomial distribution can also be expressed using the Gamma function as:

\begin{equation}
p \left( \vec{x}; \vec{p} \right) = \frac{\Gamma \left( \sum\limits_k x_k + 1 \right)}{\prod\limits_k \Gamma \left( x_k + 1 \right)}\prod\limits_{k=1}^K p_k^{x_k}
\end{equation}

\noindent whose form is very similar to the Dirichlet distribution, which is its conjugate prior.

The Multinomial distribution is typically denoted as $\text{Mult} \left( \vec{p}, n \right)$.

\subsubsection{The Dirichlet distribution}
\label{subsubsection:rationale_probability_dirichlet}
The Dirichlet distribution is a continuous probability distribution typically used in Bayesian statistics as conjugate prior of the Multinomial distribution. It is the multivariate generalization of the Beta distribution, which is commonly used to model prior knowledge about the probability of an event. Therefore, the Dirichlet distribution governs a random variable $X$ whose outcomes are the realizations of a Multinomial distribution in the form $\vec{x} = \left( x_1, \ldots, x_K \right)$, with $x_k \in \left[ 0, 1 \right]$. The distribution has the parameters $\Theta = \left\lbrace \vec{\alpha} \right\rbrace$ and is defined as:

\begin{equation}
p \left( \vec{x}; \vec{\alpha} \right) = \frac{\Gamma \left( \sum\limits_k \alpha_k \right)}{\prod\limits_k \Gamma \left( \alpha_k \right)}\prod\limits_{k=1}^K x_k^{\left( \alpha_k - 1 \right)}
\end{equation}

\noindent with $\vec{\alpha}$ called the \emph{concentration} parameter.

As stated above, the Dirichlet distribution is the \emph{conjugate prior} of the Multinomial distribution, which means that if a random variable $X$ follows a Multinomial distribution, and we assume a prior Dirichlet distribution over its parameter $\vec{p}$, then the posterior distribution of $\vec{p}$ is also Dirichlet distributed. Under the Bayesian statistics, this relationship leads to a powerful mechanism to, given a new observation $\vec{x}$, update the parameter $\vec{p}$ in a pure algebraic manner, without recalculating the joint probability distribution $p \left( \vec{x}, \vec{p} \right)$, which normally remains intractable.

The Dirichlet distribution is typically denoted as $\text{Dir} \left( \vec{\alpha} \right)$.

\subsection{Parameter estimation}
\label{subsection:rationale_probability_parameter_estimation}
Statistical inference is the process of deducing properties and making generalizations about a population, by estimating the parameters of an underlying probability distribution used to model observations taken from that population. Therefore, statistical modeling typically consists of two steps: 1) choosing the distribution that best fits the data, and 2) determining the parameters $\Theta$ of the distribution.

However, $\Theta$ is generally considered unobservable, as the whole population cannot be observed either. Therefore, an estimation of the parameters $\hat{\Theta}$ is usually calculated, based on a random sample drawn from the population. The most typical analytical methods of estimating the parameters of a probability distribution given its \ac{pdf} (or \ac{pmf}) are: \ac{MLE}, \ac{MAP} estimation and Bayesian inference, of which the first two are briefly described below as they are extensively used in this thesis. On the other hand, numerical optimization methods such as the \ac{EM} algorithm are also typically employed to estimate the parameters of complex models that do not have closed-form analytical solution. The \ac{EM} algorithm is also discussed below as it is extensively used in this thesis for several contributions.

\subsubsection{\acl{MLE}}
\label{subsubsection:rationale_probability_mle}
The \acf{MLE} focuses on obtaining the best distribution parameters $\hat{\Theta}$ by maximizing the likelihood function $\mathcal{L} \left( \Theta;x \right)$. This approach can be semantically interpreted as \emph{``maximizing the probability of observing the random sample $X$ given a current guess about the parameters $\Theta$ of the model"}.

Let $X = \left( \vec{x}^1, \ldots , \vec{x}^N \right)$ a set of observations of a random variable. Formally, the \ac{MLE} estimate is defined as:

\begin{equation}
\begin{split}
\hat{\Theta}_{MLE} &= \argmax_\Theta \mathcal{L} \left( \Theta; X \right) \\
                   &= \argmax_\Theta p \left( X; \Theta \right) \\
                   &= \argmax_\Theta \prod\limits_i p \left( \vec{x}^i; \Theta \right)
\end{split}
\end{equation}

Differentiation is employed to maximize $\mathcal{L} \left( \Theta;X \right)$, by setting the partial derivatives of the function with respect to each parameter to zero. However, maximizing $\mathcal{L} \left( \Theta;X \right)$ is hard as it involves a product over all realizations of $X$. Typically, natural logarithm is taken to simplify the expression in various senses: first, natural logarithm is a monotonically increasing function so the maximum value of $\mathcal{L} \left( \Theta;X \right)$ occurs at the same point as the maximum value of $\log \mathcal{L} \left( \Theta;X \right)$; next, the logarithm of the product converts to the sum of the logarithms, which is far easier to maximize than the product since the derivative of a sum is the sum of derivatives; additionally, probability distributions involving the exponential function take benefit from the logarithm to simplify the expression to be maximized; finally, computing a sum of log-derivatives is computationally more stable than calculating a product of probabilities, which quickly tend to zero leading to numerical representation problems.

Therefore, the log-likelihood is usually maximized in \ac{MLE}, which is defined as:

\begin{equation}
\begin{split}
\hat{\Theta}_{MLE} &= \argmax_\Theta \log \mathcal{L} \left( \Theta; X \right) \\
                   &= \argmax_\Theta \log p \left( X; \Theta \right) \\
                   &= \argmax_\Theta \log \prod\limits_i p \left( x^i; \Theta \right) \\
                   &= \argmax_\Theta \sum\limits_i \log p \left( x^i; \Theta \right)
\end{split}
\end{equation}

\subsubsection{\acl{MAP}}
\label{subsubsection:rationale_probability_map}
\acf{MAP} estimation is closely related to \ac{MLE} in the sense that both look for the parameters that equal to the mode of a distribution (the maximum of the distribution). However, while \ac{MLE} maximizes the likelihood (or log-likelihood) distribution of the parameters $\Theta$, \ac{MAP} estimation maximizes the posterior distribution of $\Theta$ with respect to the observations $X$. Therefore, in this case, \ac{MAP} estimate can be semantically interpreted as \emph{``maximizing the probability of a specific setting of the $\Theta$ parameters of the model given the current observations of the random variable $X$"}. In this sense, \ac{MAP} is also closely related to Bayesian estimation, since it considers $\Theta$ parameters as a random variable that in turn is governed by some distribution with its own parameters. Therefore, assuming $\Theta$ as a random variable and taking the Bayes' rule:

\begin{equation}
\begin{split}
p \left( \Theta | X \right) &= \frac{p \left( X | \Theta \right) p \left( \Theta \right)}{p \left( X \right)} \\
                            &\propto p \left( X | \Theta \right) p \left( \Theta \right)
\end{split}
\end{equation}

\noindent where we can ignore the normalizing constant as we are strictly speaking about normalization, so proportionality is sufficient. Hence, the \ac{MAP} estimate is defined as:

\begin{equation}
\begin{split}
\hat{\Theta}_{MAP} &= \argmax_\Theta \log \mathcal{L} \left( X | \Theta \right) \\
                   &= \argmax_\Theta \log p \left( \Theta | X \right) \\
                   &= \argmax_\Theta \log \left( p \left( X | \Theta \right) p \left( \Theta \right) \right) \\
                   &= \argmax_\Theta \log p \left( X | \Theta \right) + \log p \left( \Theta \right) \\
                   &= \argmax_\Theta \sum\limits_i \log p \left( x^i| \Theta \right) + \log p \left( \Theta \right)
\end{split}
\end{equation}

\noindent from where it can be clearly observed that \ac{MAP} is an augmentation of \ac{MLE} that allows to introduce a density $p \left( \Theta \right)$ over the parameters of the model, to inject initial beliefs and/or prior knowledge about them. Moreover, it can also be quickly deduced that \ac{MLE} is a particular case of \ac{MAP} estimation when $p \left( \Theta \right)$ is assumed to be constant. Thus, \ac{MAP} can also be thought as a regularization/generalization of the \ac{MLE} method.

As in \ac{MLE}, differentiation is employed to maximize $ \log \mathcal{L} \left( X|\Theta \right)$ and obtain closed-form solutions for the estimation of the parameters of the model. At this point, conjugate priors, briefly introduced in section \ref{subsubsection:rationale_probability_dirichlet}, become important when modeling $p \left( \Theta \right)$, since assuming an adequate distribution for $p \left( \Theta \right)$ may be crucial to update the parameters in a straightforward manner.

\subsubsection{\acl{EM} algorithm}
\label{subsubsection:rationale_probability_em}
As stated above, \ac{MLE} and \ac{MAP} estimation consist in the maximization of the likelihood and posterior functions respectively, by mathematical differentiation. However, setting partial derivatives for the parameters of many complex models usually does not find closed-form expressions, hence preventing direct analytical solutions.

A common way to deal with the complexity in optimizing parameters of models that do not have closed form solution is to introduce a set of latent variables $Z$ that allow the model to be formulated in a more tractable way. Therefore, defining a joint distribution $p \left( X, Z; \Theta \right)$ over the augmented space of observed and latent variables allows to estimate the parameters of the model $\Theta$ in a more straightforward manner. Then, once $\Theta$ is estimated, the distribution of the observed variables $p \left(X; \Theta \right)$ can be obtained by marginalization. The \ac{EM} algorithm takes advantage of such idea.

The \ac{EM} algorithm, formalized in 1977 by Arthur Dempster, Nan Laird and Donald Rubin \citep{Dempster1977}, provides an iterative numerical optimization framework for the \ac{MLE} and \ac{MAP} estimate of models involving latent variables, when a closed-form analytical solution is not possible. The algorithm relies on the introduction of an \emph{informative} latent variables that allow the model to be reformulated in such a way that a closed-form solution for the estimation of the parameters can be found.

Of course, these latent variables $Z$ are unknown quantities that prevent a direct estimation of the model, so maximizing $\log \mathcal{L} \left( \Theta; X,Z \right)$ (for the \ac{MLE} example case) is not possible. However, some state of knowledge about the value of $Z$ can be obtained by computing its posterior distribution given the observations $X$ and a guess about the parameters of the model $\tilde{\Theta}$, i.e. $p \left( Z|X; \tilde{\Theta} \right)$. Therefore, instead of maximizing $\log \mathcal{L} \left( \Theta; X,Z \right)$ the \ac{EM} algorithm maximizes an auxiliary function, typically referred as the $\mathcal{Q}$-function, which is the expected value of $\log \mathcal{L} \left( \Theta; X,Z \right)$ under this posterior distribution. That is:

\begin{equation}
\begin{split}
\hat{\Theta}_{MLE} &= \argmax_\Theta \mathcal{Q} \left( \Theta; \tilde{\Theta} \right) \\
                   &= \argmax_\Theta \mathbb{E}_{p \left( Z|X; \tilde{\Theta} \right)} \left[ \log \mathcal{L}\left( \Theta; X,Z \right) \right] \\
                   &= \argmax_\Theta \sum\limits_Z p \left( Z|X; \tilde{\Theta} \right) \log p \left( X,Z ;\Theta \right) \\
                   &= \argmax_\Theta \sum\limits_i p \left( z^i| x^i; \tilde{\Theta} \right) \log p \left( x^i, z^i ;\Theta \right)
\end{split}
\end{equation}

This formulation quickly suggest an iterative scheme based on alternating between computing the conditional expectations of the latent variables given the observations, i.e. computing $p \left( Z|X; \tilde{\Theta} \right)$, and updating the parameters of the model via the closed-form solutions obtained thanks to the augmented formulation $p \left( X,Z; \Theta \right)$.

Therefore, the general scheme of the \ac{EM} algorithm (for the \ac{MLE} case) is:

\begin{siderules}
\begin{description}
\item[Initialization:] Choose an initial setting for $\tilde{\Theta}$.
\item[Expectation step:] Evaluate the $\mathcal{Q}$-function:
\begin{equation*}
\mathcal{Q} \left( \Theta; \tilde{\Theta} \right) = \mathbb{E}_{p \left( Z|X; \tilde{\Theta} \right)} \left[ \log \mathcal{L}\left( \Theta; X,Z \right) \right]
\end{equation*}
\noindent which actually means estimating the unknown quantity $p\left( Z|X; \tilde{\Theta} \right)$
\item[Maximization step:] Maximize the $\mathcal{Q}$-function:
\begin{equation*}
\hat{\Theta}_{MLE} = \argmax_\Theta \mathcal{Q} \left( \Theta; \tilde{\Theta} \right)
\end{equation*}
\noindent which consist of updating the parameters $\Theta$ of the model based on $p \left( Z|X; \tilde{\Theta} \right)$
\item[Convergence:] Stop if $\mathcal{L} \left( \hat{\Theta};X \right) - \mathcal{L} \left( \tilde{\Theta};X \right) \leq \epsilon$; otherwise $\tilde{\Theta} = \hat{\Theta}$ and go to \textbf{Expectation step}.
\end{description}
\end{siderules}

Parameter estimation via \ac{EM} algorithm is only guaranteed to converge to a local maxima or to a saddle point of the function under maximization. Convergence to the global maxima is not assured since the algorithm is only guaranteed to iteratively increase the likelihood (or posterior) function after each iteration. Hence, optimizing multi-modal functions can easily lead to local maximas, also conditioned by the initialization of $\tilde{\Theta}$. An in-depth dissertation on the convergence properties of the \ac{EM} was made by professor \cite{Wu1983}.

In the following chapters, several applications of \ac{MLE} and \ac{MAP} parameter estimation of complex models via \ac{EM} will be presented.

\pagebreak

\section{Mixture Models}
\label{section:rationale_mixture_models}
So far we have seen that probability distributions give us a mathematical form to describe the properties of a population represented by a random variable. However, assuming that each observation is drawn from a single unimodal distribution leads to a very simplistic scenario that generally does not apply in real-world situations. In fact, real-world data typically present more complex patterns such as multi-modality, sparsity, noise or presence of outliers. In this sense, models capable of capturing such variability are required to describe the data in a more reliable sense.

In the following sections, three different families of mixture models grouped by their structured or non-structured nature will be discussed.

\subsection{\acl{FMM}}
\label{subsection:rationale_mixture_models_finite}
\acp{FMM} are probabilistic models that model a random variable as a convex combination of \acp{pdf}. Therefore, \ac{FMM} provides a statistical formal framework to describe heterogeneous data trough a weighted sum of single distributions, each one representing a sub-population within a dataset.

Let $X = \left( \vec{x}^1, \ldots , \vec{x}^N \right)$ a set of observations of a random variable, where $\vec{x}^i \in \Reals^d$. A \ac{FMM} of $K$ components assumes a \ac{pdf} over $X$ in the form:

\begin{equation}
\label{eq:fmm}
\begin{split}
p \left( X; \Theta \right) &= \prod\limits_{i=1}^N p \left( \vec{x}^i; \Theta \right) \\
                           &= \prod\limits_{i=1}^N \sum\limits_{j=1}^K \pi_j \phi \left( \vec{x}^i ; \Theta_j \right)
\end{split}
\end{equation}

\noindent where $\Theta = \left\lbrace \Theta_1, \ldots, \Theta_K, \pi_1, \ldots, \pi_K \right\rbrace$ are the parameters of the model, with $\left\lbrace \Theta_j, \pi_j \right\rbrace$ the parameters of the $j^{th}$ component of the mixture modeled by the \ac{pdf} $\phi \left( \vec{x}^i ; \Theta_j \right)$. The parameters $\left\lbrace \pi_1,\ldots, \pi_K \right\rbrace $ are typically called \emph{mixing coefficients} and can be seen as the prior probability of each component of the mixture describing the data. Therefore, it follows that $0 \leq \pi_j \leq 1$ and:

\begin{equation}
\sum\limits_{j=1}^K \pi_j = 1
\end{equation}

\begin{figure}[h]
\centering
\includegraphics[width=0.8\linewidth]{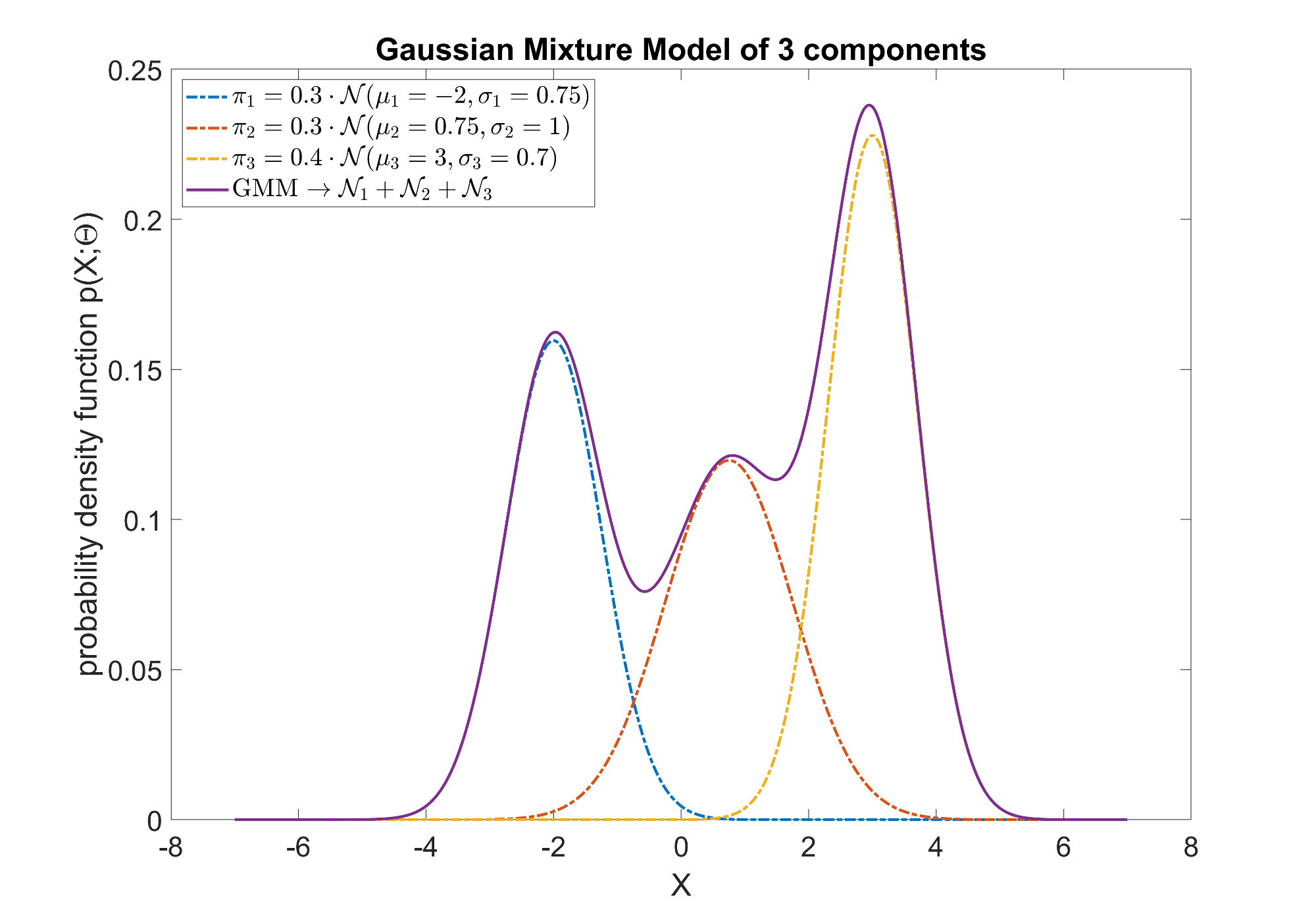}
\caption{Probability density functions of a 3-component \ac{GMM} with their associated Normal distributions.}
\label{figure:finitemixturemodel}
\end{figure}

Due to its mathematical tractability, Gaussian (or t-Student) distributions are typically employed to model the data, thus $\phi \left( \vec{x}^i ; \Theta_j \right) \sim \mathcal{N} \left( \vec{x}^i ; \mu_j, \Sigma_j \right)$. Figure \ref{figure:finitemixturemodel} illustrates the \ac{pdf} of a 3-component \ac{GMM} (in purple), with their associated Normal distributions (blue, red and yellow).

A common way to estimate the parameters of a \ac{FMM} could be via \ac{MLE}. From \ref{eq:fmm} we can set:

\begin{equation}
\begin{split}
\hat{\Theta}_{MLE} &= \argmax_\Theta \log \mathcal{L}\left( \Theta;X \right) \\
                   &= \argmax_\Theta \log p \left( X;\Theta \right) \\
                   &= \argmax_\Theta \log \prod\limits_{i=1}^N \sum\limits_{j=1}^K \pi_j \phi \left( \vec{x}^i ; \Theta_j \right) \\
                   &= \argmax_\Theta \sum\limits_{i=1}^N \log \sum\limits_{j=1}^K \pi_j \phi \left( \vec{x}^i ; \Theta_j \right)
\end{split}
\end{equation}

\noindent from where we can immediately see that the maximization is very complex due to the summation over $K$ inside the logarithm. In fact, there is no closed-form analytical solution for the \ac{MLE} (nor \ac{MAP} estimate) of a \ac{FMM}. In this sense, alternative methods are required to estimate the parameters of these models in a reliable but affordable manner.

\subsubsection{\acs{EM} inference for \acs{FMM}}
\label{subsubsection:rationale_mixture_models_finite_em}
As stated in chapter \ref{subsubsection:rationale_probability_em}, a common way to deal with the complexity in optimizing parameters of models that do not have closed form solution is to introduce a set of latent variables $Z$ that allow the model to be formulated in a more tractable way. Under the \ac{FMM}'s point of view, a latent variable $Z$ is typically introduced indicating the component of the mixture to which the observation belongs to. Formally, let $Z = \left( \vec{z}^1, \ldots , \vec{z}^N \right)$ a random variable where $\vec{z}^i \in \left\lbrace 0,1 \right\rbrace ^K$ represents a binary $K$-dimensional one-hot encoding variable, where $z_k^i=1$ indicates that the $i^{th}$ observation has originated by the $k^{th}$ component of the mixture, and hence $z_j^i=0,~\forall j \neq k$. Under this assumption, we can reformulate the \ac{FMM} by defining a joint density over the latent and observed variables in the form:

\begin{equation}
p \left( X,Z; \Theta \right) = \prod\limits_{i=1}^N \prod\limits_{j=1}^K \left( \pi_j \phi \left( \vec{x}^i ; \Theta_j \right) \right)^{z_j^i}
\end{equation}

\noindent where we assume that $Z$ follows a Multinomial unit-length distribution given by equation \ref{eq:multinomial} with $n=1$ (i.e. Categorical), i.e.:

\begin{equation}
z^i \sim Cat_K(\vec{\pi}^i)
\end{equation}

The \ac{MLE} estimate of this augmented \ac{FMM} can be expressed as:

\begin{equation}
\label{eq:mlefmm}
\begin{split}
\hat{\Theta}_{MLE} &= \argmax_\Theta \log \mathcal{L}\left( \Theta; X,Z \right) \\
                   &= \argmax_\Theta \log p \left( X,Z ;\Theta \right) \\
                   &= \argmax_\Theta \log \prod\limits_{i=1}^N \prod\limits_{j=1}^K \left( \pi_j \phi \left( \vec{x}^i ; \Theta_j \right) \right)^{z_j^i} \\
                   &= \argmax_\Theta \sum\limits_{i=1}^N \sum\limits_{j=1}^K \log \left( \pi_j \phi \left( \vec{x}^i ; \Theta_j \right) \right)^{z_j^i} \\
                   &= \argmax_\Theta \sum\limits_{i=1}^N \sum\limits_{j=1}^K z_j^i \log \left( \pi_j \phi \left( \vec{x}^i ; \Theta_j \right) \right) \\
                   &= \argmax_\Theta \sum\limits_{i=1}^N \sum\limits_{j=1}^K z_j^i \left( \log \pi_j + \log  \phi \left( \vec{x}^i ; \Theta_j \right) \right)
\end{split}
\end{equation}

\noindent which is far easier to optimize than the original \ac{FMM}. In fact, closed-form solutions can be found for all the parameters of the model depending on the form of $\phi \left(x^i; \Theta_j \right)$.

Optimizing $\pi_j$ requires considering the constraint that $\sum_k \pi_k = 1$, which can be easily achieved by using Lagrange multipliers. Therefore, setting the derivatives of $\log \mathcal{L}\left( \Theta; X,Z \right)$ with respect $\pi_j$ and $\lambda$ Lagrange multiplier yield:

\begin{equation}
\begin{split}
\frac{1}{\pi_j} \sum\limits_{i=1}^N z_j^i + \lambda &= 0 \\
\sum\limits_{k=1}^K \pi_k - 1 &= 0 
\end{split}
\end{equation}

Solving the system of equations and rearranging for $\pi_j$ provides:

\begin{equation}
\label{eq:piestimation}
\pi_j = \frac{\sum\limits_{i=1}^N z_j^i}{\sum\limits_{i=1}^N \sum\limits_{k=1}^K z_k^i}
\end{equation}

Note that the denominator of the above expression (\ref{eq:piestimation}) is actually equal to $N$, since it is a summation of N one-hot encoding variables $\vec{z}^i$, which take a value 1 for one of their components and 0 for the rest.

For the most typical case of \ac{FMM} where $\phi \left(x^i; \Theta_j \right) \sim \mathcal{N} \left(x^i; \mu_j, \Sigma_j \right)$ (i.e. \ac{GMM}), means and covariance matrices must be optimized. Setting the derivatives of $\log \mathcal{L}\left( \Theta; X,Z \right)$ with respect to the mean yields:

\begin{equation}
\sum\limits_{i=1}^N z_j^i \Sigma_j^{-1} \left( \vec{x}^i - \vec{\mu}_ j \right) = 0
\end{equation}

\noindent where rearranging:

\begin{equation}
\label{eq:muestimation}
\vec{\mu}_j = \frac{1}{\sum\limits_{i=1}^N z_j^i} \sum\limits_{i=1}^N z_j^i \vec{x}^i
\end{equation}

\noindent and solving for the covariance matrix yields:

\begin{equation}
\sum\limits_{i=1}^N z_j^i \left( -\frac{1}{2} \Sigma_j^{-1} + \frac{1}{2} \left( \vec{x}^i - \vec{\mu}_ j \right)^T \Sigma_j^{-2} \left( \vec{x}^i - \vec{\mu}_ j \right) \right)  = 0
\end{equation}

\noindent where rearranging:

\begin{equation}
\label{eq:sigmaestimation}
\Sigma_j = \frac{1}{\sum\limits_{i=1}^N z_j^i} \sum\limits_{i=1}^N z_j^i \left( \vec{x}^i - \vec{\mu}_ j \right) \left( \vec{x}^i - \vec{\mu}_ j \right)^T
\end{equation}

\medskip

Up to this point we have seen that assuming that we have both the observations and latent variables $\left\lbrace X, Z \right\rbrace$, typically referred as the \emph{complete} dataset, the \ac{MLE} (and also the \ac{MAP}) estimate find closed-form solutions for the parameters of the \ac{FMM}. However, actually, we only have the observations $X$ (i.e. the \emph{incomplete} dataset), so, as previously described in section \ref{subsubsection:rationale_probability_em}, the only information we can get from the latent variable $Z$ is given by its posterior distribution conditioned to the observations $X$ and to a current estimate of the parameters of the model $\tilde{\Theta}$, i.e. $p \left( Z|X; \tilde{\Theta} \right)$. Following the Bayes' rule it is given by:

\begin{equation}
p \left( z_j^i = 1 | \vec{x}^i; \tilde{\Theta} \right) = \gamma_j^i = \frac{\pi_j \phi \left( \vec{x}^i; \tilde{\Theta}_j \right)}{\sum\limits_k \pi_k \phi \left( \vec{x}^i; \tilde{\Theta}_k \right)}
\end{equation}

\medskip

Therefore, because it is not possible to compute the \ac{MLE} of $\log \mathcal{L}\left( \Theta; X,Z \right)$ (due to we only can observe $X$), we just can compute its expected value under the posterior distribution $p \left( Z | X; \tilde{\Theta} \right)$, which yields the so-called $Q$-function:

\begin{equation}
\begin{split}
\hat{\Theta}_{MLE} &= \argmax_\Theta \mathcal{Q} \left( \Theta; \tilde{\Theta} \right) \\
                   &= \argmax_\Theta \mathbb{E}_{p\left(Z|X; \tilde{\Theta} \right)} \log \mathcal{L}\left( \Theta; X,Z \right) \\
                   &= \argmax_\Theta \sum\limits_Z p\left(Z|X; \tilde{\Theta} \right) \log p \left( X,Z ;\Theta \right) \\
                   & \ldots \\
                   &= \argmax_\Theta \sum\limits_{i=1}^N \sum\limits_{j=1}^K \gamma_j^i \left( \log \pi_j + \log  \phi \left( \vec{x}^i ; \Theta_j \right) \right)
\end{split}
\end{equation}

\noindent which yields the same equation than \ref{eq:mlefmm} but substituting the latent variable $z_j^i$ by its posterior quantity $\gamma_j^i$. Setting derivatives of this expression obviously gives the same results as \ref{eq:piestimation}, \ref{eq:muestimation} and \ref{eq:sigmaestimation}, but replacing $z_j^i$ by $\gamma_j^i$.

This formulation hence suggest an iterative scheme for \ac{MLE} or \ac{MAP} estimate, based on alternating between computing $p \left( Z|X; \tilde{\Theta} \right)$, referred as the \emph{Expectation}-step or \emph{E}-step; and updating the parameters of the model based on this distribution, referred as the \emph{Maximization}-step or \emph{M}-step. Therefore, the \ac{EM} algorithm can be finally summarized as:

\begin{siderules}
\begin{description}
\item[Initialization:] Choose an initial setting for $\Theta^{\left( 0 \right)}$.
\item[Expectation step:] Estimate $p \left( Z|X; \Theta^{\left( t \right)} \right)$
\begin{equation*}
p \left( z_j^i = 1 | \vec{x}^i; \Theta^{\left( t \right)} \right) = {\gamma_j^i}^{\left( t \right)} = \frac{\pi_j^{\left( t \right)} \phi \left( \vec{x}^i; \Theta_j^{\left( t \right)} \right)}{\sum\limits_k \pi_k^{\left( t \right)} \phi \left( \vec{x}^i; \Theta_k^{\left( t \right)} \right)}
\end{equation*}
\item[Maximization step:] Update the parameters of the model given $p \left( Z|X; \Theta^{\left( t \right)} \right)$
\begin{equation*}
\hat{\Theta}_{MLE}^{\left( t+1 \right)} = \argmax_\Theta \mathbb{E}_{p\left(Z|X; \Theta^{\left( t \right)} \right)} \log \mathcal{L}\left( \Theta; X,Z \right)
\end{equation*}
\noindent where considering the \ac{GMM} of the example:
\begin{equation*}
\begin{split}
\pi_j^{\left( t+1 \right)} &= \frac{\sum\limits_{i=1}^N {\gamma_j^i}^{\left( t \right)}}{\sum\limits_{i=1}^N \sum\limits_{k=1}^K {\gamma_k^i}^{\left( t \right)}} \\
\vec{\mu}_j^{\left( t+1 \right)} &= \frac{1}{\sum\limits_{i=1}^N {\gamma_j^i}^{\left( t \right)}} \sum\limits_{i=1}^N {\gamma_j^i}^{\left( t \right)} \vec{x}^i \\
\Sigma_j^{\left( t+1 \right)} &= \frac{1}{\sum\limits_{i=1}^N {\gamma_j^i}^{\left( t \right)}} \sum\limits_{i=1}^N {\gamma_j^i}^{\left( t \right)} \left( \vec{x}^i - \vec{\mu}_ j^{\left( t+1 \right)} \right) \left( \vec{x}^i - \vec{\mu}_ j^{\left( t+1 \right)} \right)^T
\end{split}
\end{equation*}
\item[Convergence:] Stop if $\mathcal{L} \left( \Theta^{\left( t+1 \right)} ; X \right) - \mathcal{L} \left( \Theta^{\left( t \right)} ; X \right) \leq \epsilon$; otherwise $t = t + 1$ and go to \textbf{Expectation step}.
\end{description}
\end{siderules}

\subsection{\acl{SVFMM}}
\label{subsection:rationale_mixture_models_svfmm}
So far, we have seen that \acp{FMM} provide a rigorous statistical framework for modeling heterogeneous data in a precise and formal manner. The optimization of the likelihood function of a \ac{FMM} provides a computationally feasible yet powerful solution to fit and evaluate complex models to model random variables.

However, \acp{FMM} make an important assumption that can turn into a severe drawback when it comes to structured data, such as images. \acp{FMM} assume observations $X$ to be independent and identically distributed, hence implying no correlations between them. This assumption does not hold for imaging data in which observations, i.e. pixels in 2D images or voxels in 3D volumes, are strictly arranged in a structure that inherently defines explicit correlations between them. Therefore, ignoring the prior knowledge that adjacent observations are more likely to belong to the same class wastes a highly useful information that allows images to be described in a more concise, adequate and realistic manner.

The \acf{SVFMM} is an extension of the \ac{FMM} for structured data, that aims to facilitate the modeling of the spatial correlations inherent in images. The \ac{SVFMM} mainly differs from the \ac{FMM} in the definition of the mixing coefficients which, as we saw, can also be interpreted as the prior probabilities of the components of the mixture. Specifically, the \ac{SVFMM} assumes that each observation $\vec{x}^i$ has its own different vector of mixing coefficients $\vec{\pi}^i$, denoted as \emph{contextual mixing coefficients}, rather than sharing a global vector for all observations. Additionally, the \ac{SVFMM} assumes $\Pi = \left( \vec{\pi}^1, \ldots, \vec{\pi}^N \right)$ to be a vector of random variables rather than parameters, allowing a prior density to be defined to introduce statistical correlations among them.

Let $X = \left( \vec{x}^1, \ldots , \vec{x}^N \right)$ a set of observations of a random variable, where $\vec{x}^i \in \Reals^d$. A \ac{SVFMM} of $K$ components assumes a \ac{pdf} over $X$ in the form:

\begin{equation}
\label{eq:svfmm}
p \left( X| \Theta, \Pi \right) = \prod\limits_{i=1}^N \sum\limits_{j=1}^K \pi_j^i \phi \left( \vec{x}^i ; \Theta_j \right)
\end{equation}

\noindent where $\Theta = \left\lbrace \Theta_1, \ldots, \Theta_K \right\rbrace$ and $\Pi = \left\lbrace \vec{\pi}^1, \ldots, \vec{\pi}^N \right\rbrace$ are the parameters of the model. Note that $\Theta$ and $\Pi$ are actually treated as random variables, and that each observation $\vec{x}^i$ has its own associated vector of contextual mixing coefficients $\vec{\pi}^i$. Likewise \ac{FMM}, contextual mixing coefficients must also satisfy $0 \leq \pi_j^i \leq 1$ and:

\begin{equation}
\sum\limits_{j=1}^K \pi_j^i = 1
\end{equation}

A \ac{MAP} estimate of the model is usually employed to introduce a proper density over $\Pi$, typically in the form of a \ac{MRF}, to establish dependencies between adjacent contextual mixing coefficients. Therefore, the \ac{MAP} estimate of the \ac{SVFMM} is defined as:

\begin{equation}
\label{eq:mapsvfmm}
\begin{split}
\left( \hat{\Theta}, \hat{\Pi} \right)_{MAP} &= \argmax_{\left( \Theta, \Pi \right)} \log \mathcal{L} \left( X | \Theta, \Pi \right) \\
                   &= \argmax_{\left( \Theta, \Pi \right)} \log p \left( \Theta, \Pi | X \right) \\
                   &= \argmax_{\left( \Theta, \Pi \right)} \log \left\lbrace \frac{p \left( X | \Theta, \Pi \right) p \left( \Theta, \Pi\right)}{p \left( X \right)} \right\rbrace \\
                   &= \argmax_{\left( \Theta, \Pi \right)} \log \left\lbrace p \left( X | \Theta, \Pi \right) p \left( \Theta, \Pi\right) \right\rbrace \\
                   &= \argmax_{\left( \Theta, \Pi \right)} \log p \left( X | \Theta, \Pi \right) + \log p \left( \Theta, \Pi\right) \\
                   &= \argmax_{\left( \Theta, \Pi \right)} \log p \left( X | \Theta, \Pi \right) + \cancelto{const}{\log p \left( \Theta \right)} + \log p \left( \Pi \right) \\
                   &= \argmax_{\left( \Theta, \Pi \right)} \log p \left( X | \Theta, \Pi \right) + \log p \left( \Pi \right)
\end{split}
\end{equation}

\noindent where we have assumed independence in $p \left( \Theta, \Pi \right)$ and also assumed a constant uniform distribution for $p \left( \Theta \right)$ that gets rid out from the maximization.

\medskip

Several densities for $p \left( \Pi \right)$ has been proposed in the literature \citep{Sanjay1998, Woolrich2005, Blekas2005, Sfikas2008} to codify the concept that neighboring observations tend to share the same component of the mixture. A family of densities that has been widely used with successful results are the Gauss-\acp{MRF} \citep{Nikou2007}. This family of priors encodes the general idea that the estimator of the contextual mixing coefficients can be defined as the average of its spatial neighbors:

\begin{equation}
\hat{\vec{\pi}}^i = \frac{1}{\left| \mathcal{M}^i \right|} \sum\limits_{m \in \mathcal{M}^i} \vec{\pi}^m
\end{equation}

\noindent where $\mathcal{M}^i$ indicates the set of neighbors of the $i^{th}$ observation. In other words, this prior assumes that differences between adjacent contextual mixing coefficients for a given component $j$ of the mixture are Gaussian distributed in the form:

\begin{equation}
\pi_j^i - \pi_j^m \sim \mathcal{N} \left( 0,\beta^2 \right)~~~~~~m \in \mathcal{M}^i
\end{equation}

This approach is somewhat naïve and can be refined to capture variability among different components of the mixture and spatial directions of the images. The \ac{DCAGMRF} prior, which lies into the family of \ac{SAR} models, has been used to regularize ill-posed inverse problems with successful results. The \ac{DCAGMRF} takes the form:

\begin{equation}
\label{eq:dcagmrf}
p \left( \Pi \right) = \prod\limits_{i=1}^N \prod\limits_{j=1}^K \prod\limits_{d=1}^D \prod\limits_{m\in\mathcal{M}_d^i} \frac{1}{\sqrt{2 \pi \beta_{j,d}^2}} \exp \left( - \frac{ \left( \pi_j^i - \pi_j^m \right)^2}{2 \beta_{j,d}^2}\right)
\end{equation}

\noindent where sub-index $d$ refers to the different spatial adjacency directions (i.e. horizontals, verticals or diagonals), and $\mathcal{M}_d^i$ indicates the set of neighbors of the $i^{th}$ observation that lies in the $d^{th}$ spatial direction.

Merging \ref{eq:mapsvfmm} with \ref{eq:svfmm} finally yields:

\begin{equation}
\label{eq:svfmmbadoptimization}
\left( \hat{\Theta}, \hat{\Pi} \right)_{MAP} = \argmax_{\left( \Theta, \Pi \right)} \sum\limits_{i=1}^N \log \sum\limits_{j=1}^K \pi_j^i \phi \left( \vec{x}^i | \Theta_j \right) + \log p \left( \Pi \right)
\end{equation}

\noindent which again is analytically intractable, so numerical approximations should be employed.

\subsubsection{\acs{EM} inference for \acs{SVFMM}}
\label{subsubsection:rationale_mixture_models_svfmm_em}
As in the \ac{FMM} case, a binary $K$-dimensional one-hot encoding latent variable $Z$ is introduced to simplify the estimation of the model, by assuming knowledge about the component of the mixture to which the observation belongs to. The formulation of the \ac{SVFMM} assuming the existence of the latent variables results in:

\begin{equation}
p \left( X,Z| \Theta, \Pi \right) = \prod\limits_{i=1}^N \prod\limits_{j=1}^K \left( \pi_j^i \phi \left( \vec{x}^i ; \Theta_j \right) \right)^{z_j^i}
\end{equation}

The \ac{MAP} estimate of the augmented \ac{SVFMM} can be expressed as:

\begin{equation}
\label{eq:mapsvfmmlatent}
\begin{split}
\left( \hat{\Theta}, \hat{\Pi} \right)_{MAP} &= \argmax_{\left( \Theta, \Pi \right)} \log \mathcal{L} \left( X,Z | \Theta, \Pi \right) \\
                   &= \argmax_{\left( \Theta, \Pi \right)} \log p \left( \Theta, \Pi | X,Z \right) \\
                   & \ldots \\
                   &= \argmax_{\left( \Theta, \Pi \right)} \log p \left( X,Z | \Theta, \Pi \right) + \log p \left( \Pi \right) \\
                   &= \argmax_{\left( \Theta, \Pi \right)} \log \prod\limits_{i=1}^N \prod\limits_{j=1}^K \left( \pi_j^i \phi \left( \vec{x}^i ; \Theta_j \right) \right)^{z_j^i} + \log p \left( \Pi \right) \\
                   & \ldots \\
                   &= \argmax_{\left( \Theta, \Pi \right)} \sum\limits_{i=1}^N \sum\limits_{j=1}^K z_j^i \left( \log \pi_j^i + \log  \phi \left( \vec{x}^i ; \Theta_j \right) \right) + \log p \left( \Pi \right)
\end{split}
\end{equation}

\noindent where again is far easier to optimize than \ref{eq:svfmmbadoptimization} and from where we can quickly observe that the maximization of $\Theta$ will be exactly the same as in the classic \ac{FMM}. Only the estimation of $\pi_j^i$ will be affected by the density $p \left( \Pi \right)$ as expected.

\medskip

Following the same reasoning than in classic \ac{FMM}, the only information we can get from the latent variable $Z$ is given by its posterior density $p \left( Z | X, \tilde{\Theta}, \tilde{\Pi} \right)$. Therefore, the conditional expectation of $\log \mathcal{L} \left( X,Z | \Theta, \Pi \right)$ under this density is used as auxiliar function for the optimization. Rearranging \ref{eq:dcagmrf} into \ref{eq:mapsvfmmlatent}, substituting $z_j^i$ by their posterior quantity $\gamma_j^i$ and dropping constant terms that do not affect the maximization yields the following $Q$-function:

\begin{equation}
\label{eq:qfunctionsvfmm}
\resizebox{\textwidth}{!} 
{
$
\mathcal{Q} \left( \Theta, \Pi | \tilde{\Theta}, \tilde{\Pi} \right) = \sum\limits_{i=1}^N \sum\limits_{j=1}^K \gamma_j^i \left( \log \pi_j^i + \log  \phi \left( \vec{x}^i ; \Theta_j \right) \right) - \sum\limits_{d=1}^D \log \frac{\beta_{j,d}^2}{2} + \frac{\sum\limits_{m\in\mathcal{M}_d^i} \left( \pi_j^i - \pi_j^m \right)^2}{2 \beta_{j,d}^2}
$
}
\end{equation} 

Optimizing $\pi_j^i$ from expression \ref{eq:qfunctionsvfmm} must take into account several considerations. First, note that $\pi_j^i$ appears in the term $\sum_{m\in\mathcal{M}_d^i} \left( \pi_j^i - \pi_j^m \right)^2$ once as the central individual and $D \cdot \left| \mathcal{M}_d^i \right|$ times as the neighbor of $\pi_j^m$ individuals. Second, no closed-form solution can be obtained when introducing the constraint $\sum_{j=1}^K \pi_j^i = 1$ in the maximization. Instead, \emph{reparatory} methods such as the quadratic programming algorithm proposed in \citep{Blekas2005}, must be employed to project the solutions onto the constraints to ensure probabilities sum up to 1.

Taking all this into consideration, the updates for $\pi_j^i$ are obtained as the roots of the following second degree equation obtained by setting partial derivatives of \ref{eq:qfunctionsvfmm} with respect $\pi_j^i$ (without restrictions):

\begin{equation}
\left( \pi_j^i \right)^2 \sum\limits_{d=1}^D \frac{ \left| \mathcal{M}_d^i \right|}{\beta_{j,d}^2} - \left( \pi_j^i \right) \sum\limits_{d=1}^D \frac{\sum\limits_{m \in \mathcal{M}_d^i} \pi_j^m}{\beta_{j,d}^2} - \frac{\gamma_j^i}{2} = 0
\end{equation}

\noindent which can easily demonstrated that always have a real non negative solution.

The \ac{DCAGMRF} prior also introduces a new set of parameters $\beta_{j,d}^2$ that govern the variances of the Gauss-\ac{MRF}. Setting derivatives of \ref{eq:qfunctionsvfmm} with respect $\beta_{j,d}^2$ yields:

\begin{equation}
\label{eq:betaoptimization}
\beta_{j,d}^2 = \frac{1}{N} \sum\limits_{i=1}^N \sum\limits_{m \in \mathcal{M}_d^i} \frac{\left( \pi_j^i - \pi_j^m \right)^2}{\left| \mathcal{M}_d^i \right|}
\end{equation}

\medskip

Therefore, the \ac{MAP} \ac{EM} algorithm for the \ac{SVFMM} with the \ac{DCAGMRF} prior can be finally summarized as:

\begin{siderules}
\begin{description}
\item[Initialization:] Choose an initial setting for $\Theta^{\left( 0 \right)}$ and $\Pi^{\left( 0 \right)}$.
\item[Expectation step:] Estimate $p \left( Z | X, \Theta^{\left( t \right)}, \Pi^{\left( t \right)} \right)$
\begin{equation*}
p \left( z_j^i = 1 | \vec{x}^i, \Theta^{\left( t \right)}, \Pi^{\left( t \right)} \right) = {\gamma_j^i}^{\left( t \right)} = \frac{{\pi_j^i}^{\left( t \right)} \phi \left( \vec{x}^i; \Theta_j^{\left( t \right)} \right)}{\sum\limits_k {\pi_k^i}^{\left( t \right)} \phi \left( \vec{x}^i; \Theta_k^{\left( t \right)} \right)}
\end{equation*}
\item[Maximization step:] Update the parameters of the model given $p \left( Z|X, \Theta^{\left( t \right)} \Pi^{\left( t \right)} \right)$
\begin{equation*}
\left( \hat{\Theta}, \hat{\Pi} \right)_{MAP}^{\left( t+1 \right)} = \argmax_{\left( \Theta, \Pi \right)} \mathbb{E}_{p\left(Z | X, \Theta^{\left( t \right)}, \Pi^{\left( t \right)} \right)} \log \mathcal{L}\left( X,Z | \Theta, \Pi \right)
\end{equation*}
\noindent where considering the \ac{GMM} example:
\begin{equation*}
\begin{split}
\left( {\pi_j^i}^{\left( t+1 \right)} \right)^2 & \sum\limits_{d=1}^D \frac{ \left| \mathcal{M}_d^i \right|}{{\beta_{j,d}^2}^{\left( t \right)}} - \left( {\pi_j^i}^{\left( t+1 \right)} \right) \sum\limits_{d=1}^D \frac{\sum\limits_{m \in \mathcal{M}_d^i} {\pi_j^m}^{\left( t \right)}}{{\beta_{j,d}^2}^{\left( t \right)}} - \frac{{\gamma_j^i}^{\left( t \right)}}{2} = 0~~~~~~~\left( \ast \right)\\
\vec{\mu}_j^{\left( t+1 \right)} &= \frac{1}{\sum\limits_{i=1}^N {\gamma_j^i}^{\left( t \right)}} \sum\limits_{i=1}^N {\gamma_j^i}^{\left( t \right)} \vec{x}^i \\
\Sigma_j^{\left( t+1 \right)} &= \frac{1}{\sum\limits_{i=1}^N {\gamma_j^i}^{\left( t \right)}} \sum\limits_{i=1}^N {\gamma_j^i}^{\left( t \right)} \left( \vec{x}^i - \vec{\mu}_ j^{\left( t+1 \right)} \right) \left( \vec{x}^i - \vec{\mu}_ j^{\left( t+1 \right)} \right)^T \\
{\beta_{j,d}^2}^{\left( t+1 \right)} &= \frac{1}{N} \sum\limits_{i=1}^N \sum\limits_{m \in \mathcal{M}_d^i} \frac{\left( {\pi_j^i}^{\left( t+1 \right)} - {\pi_j^m}^{\left( t+1 \right)} \right)^2}{\left| \mathcal{M}_d^i \right|}
\end{split}
\end{equation*}
\begin{small}
(*) Choose the real non-negative solution and project onto the constraints using the quadratic programming algorithm \citep{Blekas2005} to ensure $\sum\limits_{j=1}^K \pi_j^i = 1$
\end{small}
\item[Convergence:] Stop if $\mathcal{L} \left( X | \Theta^{\left( t+1 \right)}, \Pi^{\left( t+1 \right)} \right) - \mathcal{L} \left( X | \Theta^{\left( t \right)}, \Pi^{\left( t \right)} \right) \leq \epsilon$; otherwise $t = t + 1$ and go to \textbf{Expectation step}.
\end{description}
\end{siderules}

\subsection[DCM-Spatially Varying Finite Mixture Models]{\acl{DCM}-\acl{SVFMM}}
\label{subsection:rationale_mixture_models_dcmsvfmm}
An important limitation of the \ac{SVFMM} is that it does not inherently preserves the condition of $\sum_j \pi_j^i = 1$. Instead, reparatory projections must be employed to accomplish with this constraint, thereby compromising the assumed Bayesian framework.

An interesting alternative is to consider that $\Pi$ follows a \acf{DCM} distribution \citep{Nikou2010}. The \ac{DCM} distribution is a hierarchical discrete multivariate distribution, where an observation $\vec{z}^i$ is drawn from a Multinomial distribution governed by a parameter vector $\vec{\pi}^i$, which in turn is drawn from a Dirichlet distribution governed by parameter vector $\vec{\alpha}^i$.

\begin{equation}
\begin{split}
z^i \sim Cat_K \left( \vec{\pi}^i \right) \\
\vec{\pi}^i \sim Dir_K \left( \vec{\alpha}^i \right)
\end{split}
\end{equation}

Such scheme presents two important advantages for the \ac{SVFMM}. First, the Dirichlet distribution draws probability vectors that intrinsically satisfies the constraint $\sum_j \pi_j^i = 1$, so no reparatory corrections will be required. Next, the $\vec{\alpha}^i$ parameters of the Dirichlet distribution are not subject to any restriction (rather than $\alpha_j^i \geq 0$) so a Gauss-\ac{MRF} can be imposed over them to introduce the spatial correlations and local regularity in a more straightforward manner.

In this sense, the \ac{pdf} of the $\vec{z}^i$ variable given the parameters of the Categorical distribution (Multinomial unit-length with $ n = 1$) is:

\begin{equation}
p \left( \vec{z}^i | \vec{\pi}^i \right) = \frac{n!}{\prod\limits_{j=1}^K \left( z_j^i \right) !} \prod\limits_{j=1}^K \left( \pi_j^i \right)^{z_j^i} = \prod\limits_{j=1}^K \left( \pi_{j}^i \right)^{z_j^i}
\end{equation}

The \ac{pdf} of the $\vec{\pi}^i$ variable given the parameters of the Dirichlet distribution is

\begin{equation}
p \left( \vec{\pi}^i | \vec{\alpha}^i \right) = \frac{\Gamma \left( \sum\limits_{j=1}^K \alpha_{j}^i \right)}{\prod\limits_{j=1}^K \Gamma \left( \alpha_j^i\right)} \prod\limits_{j=1}^K {\pi_j^i}^{\left( \alpha_j^i - 1 \right)}
\end{equation}

\noindent with $\Gamma(\cdot)$ the Gamma function. Integrating out both \acp{pdf} we obtain the following density for the $\vec{z}^i$ variables conditioned to $\vec{\alpha}^i$:

\begin{equation}
p \left( \vec{z}^i | \vec{\alpha}^i \right) = \frac{\Gamma\left( \sum\limits_{j=1}^K \alpha_j^i \right)}{\Gamma \left( \sum\limits_{j=1}^K \alpha_j^i + z_j^i\right)} \prod\limits_{j=1}^K \frac{\Gamma \left( \alpha_j^i + z_j^i \right)}{\Gamma \left( \alpha_j^i \right)}
\end{equation}

Taking into account that $\Gamma \left( x + 1 \right) = x\Gamma(x)$, the vector $\vec{\pi}^i$ of label prior probabilities can be finally computed as:

\begin{equation}
\pi_j^i = \frac{\alpha_j^i}{\sum\limits_{j=1}^K \alpha_k^i}
\end{equation}

Therefore, the previously proposed \ac{DCAGMRF} of equation \ref{eq:dcagmrf} can be imposed over the parameters of the Dirichlet distribution in the form:

\begin{equation}
\label{eq:dcagmrfdirichlet}
p \left( A \right) = \prod\limits_{i=1}^N \prod\limits_{j=1}^K \prod\limits_{d=1}^D \prod\limits_{m\in\mathcal{M}_d^i} \frac{1}{\sqrt{2 \pi \beta_{j,d}^2}} \exp \left( - \frac{ \left( \alpha_j^i - \alpha_j^m \right)^2}{2 \beta_{j,d}^2}\right)
\end{equation}

Arranging this density into the \ac{SVFMM} and substituting $\vec{\pi}^i$ by its conditional density under the Dirichlet process yields:

\begin{equation}
\label{eq:svfmmdirichletbadoptimization}
\left( \hat{\Theta}, \hat{A} \right)_{MAP} = \argmax_{\left( \Theta, \Pi \right)} \sum\limits_{i=1}^N \log \sum\limits_{j=1}^K \frac{\alpha_j^i}{\sum\limits_{k=1}^K \alpha_k^i} \phi \left( \vec{x}^i | \Theta_j \right) + \log p \left( A \right)
\end{equation}

\noindent which, as expected, is again analytically intractable.

\subsubsection{\acs{EM} inference for \acs{DCM}-\acs{SVFMM}}
\label{subsubsection:rationale_mixture_models_dcmsvfmm_em}
As in the previous cases, a binary $K$-dimensional one-hot encoding latent variable $Z$ is introduced to simplify the estimation of the model. The formulation of the \ac{DCM}-\ac{SVFMM} assuming the existence of this latent variable results in:

\begin{equation}
p \left( X,Z| \Theta, A \right) = \prod\limits_{i=1}^N \prod\limits_{j=1}^K \left( \frac{\alpha_j^i}{\sum\limits_{k=1}^K \alpha_k^i} \phi \left( \vec{x}^i ; \Theta_j \right) \right)^{z_j^i}
\end{equation}

The \ac{MAP} estimate of the augmented \ac{DCM}-\ac{SVFMM} is expressed as:

\begin{equation}
\label{eq:mapdcmsvfmmlatent}
\begin{split}
\left( \hat{\Theta}, \hat{A} \right)_{MAP} &= \argmax_{\left( \Theta, A \right)} \log \mathcal{L} \left( X,Z | \Theta, \Pi \right) \\
                   &= \argmax_{\left( \Theta, A \right)} \log p \left( \Theta, A | X,Z \right) \\
                   & \ldots \\
                   &= \argmax_{\left( \Theta, A \right)} \log p \left( X,Z | \Theta, A \right) + \log p \left( A \right) \\
                   &= \argmax_{\left( \Theta, A \right)} \log \prod\limits_{i=1}^N \prod\limits_{j=1}^K \left( \frac{\alpha_j^i}{\sum\limits_{k=1}^K \alpha_k^i} \phi \left( \vec{x}^i ; \Theta_j \right) \right)^{z_j^i} + \log p \left( A \right) \\
                   & \ldots \\
                   &= \argmax_{\left( \Theta, A \right)} \sum\limits_{i=1}^N \sum\limits_{j=1}^K z_j^i \left( \log \frac{\alpha_j^i}{\sum\limits_{k=1}^K \alpha_k^i} + \log  \phi \left( \vec{x}^i ; \Theta_j \right) \right) + \log p \left( A \right)
\end{split}
\end{equation}

\noindent which, as usual, is easier to optimize than the original model.

The conditional expectation of $\log \mathcal{L} \left( X,Z | \Theta, A \right)$ under the posterior density $p \left( Z | X, \tilde{\Theta}, \tilde{A} \right)$ is again used as auxiliar function for the optimization, yielding the following $Q$-function:

\begin{equation}
\label{eq:qfunctiondcmsvfmm}
\resizebox{\textwidth}{!} 
{
$
\mathcal{Q} \left( \Theta, A | \tilde{\Theta}, \tilde{A} \right) = \sum\limits_{i=1}^N \sum\limits_{j=1}^K \gamma_j^i \left( \log \frac{\alpha_j^i}{\sum\limits_{k=1}^K \alpha_k^i} + \log  \phi \left( \vec{x}^i ; \Theta_j \right) \right) - \sum\limits_{d=1}^D \log \frac{\beta_{j,d}^2}{2} + \frac{\sum\limits_{m\in\mathcal{M}_d^i} \left( \alpha_j^i - \alpha_j^m \right)^2}{2 \beta_{j,d}^2}
$
}
\end{equation} 

As in the \ac{SVFMM} case, the optimization of $\Theta$ is exactly equal than in classic \acp{FMM}. The optimization of $\alpha_j^i$, however, does not require to introduce any constraint to preserve the condition $\sum_j \pi_j^i = 1$. Then, it is only necessary to consider that $\alpha_j^i$ appears in the term $\sum_{m\in\mathcal{M}_d^i} \left( \alpha_j^i - \alpha_j^m \right)^2$ once as the central individual and $D \cdot \left| \mathcal{M}_d^i \right|$ times as the neighbor of $\alpha_j^m$ individuals.

Setting partial derivatives of \ref{eq:qfunctiondcmsvfmm} with respect $\alpha_j^i$ yields the following third degree equation:

\begin{equation}
\left( \alpha_j^i \right)^3 + \left( \alpha_j^i \right)^2 \left( A_{-j}^i - \frac{C_j^i}{B_j^i} \right) - \left( \alpha_j^i \right) \left( \frac{A_{-j}^i C_j^i}{B_j^i} \right) - \frac{z_j^i A_{-j}^i}{2 B_j^i} = 0
\end{equation}

\noindent where

\begin{equation}
\begin{split}
A_{-j}^i &= \sum\limits_{\substack{k=1 \\ k \neq j}}^K \alpha_k^i \\
B_j^i &= \sum\limits_{d=1}^D \frac{\left| \mathcal{M}_d^i \right|}{\beta_{j,d}^2} \\
C_j^i &= \sum\limits_{d=1}^D \frac{\sum\limits_{m \in \mathcal{M}_d^i} \alpha_j^m}{\beta_{j,d}^2}
\end{split}
\end{equation}

\noindent where, under polynomial theory, it can be demonstrated that there is always a real non negative solution that satisfies $\alpha_j^i \geq 0$. The Cardano's or Vieta's methods can be employed to algebraically obtain the roots of the proposed third degree equation. 

The optimization of $\beta_{j,d}^2$ yields the same equation than \ref{eq:betaoptimization} but expressed in terms of $\alpha$ variables:

\begin{equation}
\beta_{j,d}^2 = \frac{1}{N} \sum\limits_{i=1}^N \sum\limits_{m \in \mathcal{M}_d^i} \frac{\left( \alpha_j^i - \alpha_j^m \right)^2}{\left| \mathcal{M}_d^i \right|}
\end{equation}

\medskip

Therefore, the \ac{MAP} \ac{EM} algorithm for the \ac{DCM}-\ac{SVFMM} with the \ac{DCAGMRF} prior can be finally summarized as:

\begin{siderules}
\begin{description}
\item[Initialization:] Choose an initial setting for $\Theta^{\left( 0 \right)}$ and $A^{\left( 0 \right)}$.
\item[Expectation step:] Estimate $p \left( Z | X, \Theta^{\left( t \right)}, A^{\left( t \right)} \right)$
\begin{equation*}
p \left( z_j^i = 1 | \vec{x}^i, \Theta^{\left( t \right)}, A^{\left( t \right)} \right) = {\gamma_j^i}^{\left( t \right)} = \frac{{\pi_j^i}^{\left( t \right)} \phi \left( \vec{x}^i; \Theta_j^{\left( t \right)} \right)}{\sum\limits_k {\pi_k^i}^{\left( t \right)} \phi \left( \vec{x}^i; \Theta_k^{\left( t \right)} \right)}
\end{equation*}
\noindent where
\begin{equation*}
{\pi_j^i}^{\left( t \right)} = \frac{{\alpha_j^i}^{\left( t \right)}}{\sum\limits_{j=1}^K {\alpha_k^i}^{\left( t \right)}}
\end{equation*}
\item[Maximization step:] Update the parameters of the model given $p \left( Z|X, \Theta^{\left( t \right)} A^{\left( t \right)} \right)$
\begin{equation*}
\left( \hat{\Theta}, \hat{A} \right)_{MAP}^{\left( t+1 \right)} = \argmax_{\left( \Theta, A \right)} \mathbb{E}_{p\left(Z | X, \Theta^{\left( t \right)}, A^{\left( t \right)} \right)} \log \mathcal{L}\left( X,Z | \Theta, A \right)
\end{equation*}

\noindent where considering the \ac{GMM} example:
\begin{equation*}
\resizebox{.95 \textwidth}{!} 
{
$
\left( {\alpha_j^i}^{\left( t+1 \right)} \right)^3 + \left( {\alpha_j^i}^{\left( t+1 \right)} \right)^2 \left( {A_{-j}^i}^{\left( t \right)} - \frac{{C_j^i}^{\left( t \right)}}{{B_j^i}^{\left( t \right)}} \right) - \left( {\alpha_j^i}^{\left( t+1 \right)} \right) \left( \frac{{A_{-j}^i}^{\left( t \right)} {C_j^i}^{\left( t \right)}}{{B_j^i}^{\left( t \right)}} \right) - \frac{z_j^i {A_{-j}^i}^{\left( t \right)}}{2 {B_j^i}^{\left( t \right)}} = 0~~\left( \ast \right)
$
}
\end{equation*}
\begin{equation*}
\begin{split}
\vec{\mu}_j^{\left( t+1 \right)} &= \frac{1}{\sum\limits_{i=1}^N {\gamma_j^i}^{\left( t \right)}} \sum\limits_{i=1}^N {\gamma_j^i}^{\left( t \right)} \vec{x}^i \\
\Sigma_j^{\left( t+1 \right)} &= \frac{1}{\sum\limits_{i=1}^N {\gamma_j^i}^{\left( t \right)}} \sum\limits_{i=1}^N {\gamma_j^i}^{\left( t \right)} \left( \vec{x}^i - \vec{\mu}_ j^{\left( t+1 \right)} \right) \left( \vec{x}^i - \vec{\mu}_ j^{\left( t+1 \right)} \right)^T \\
{\beta_{j,d}^2}^{\left( t+1 \right)} &= \frac{1}{N} \sum\limits_{i=1}^N \sum\limits_{m \in \mathcal{M}_d^i} \frac{\left( {\alpha_j^i}^{\left( t+1 \right)} - {\alpha_j^m}^{\left( t+1 \right)} \right)^2}{\left| \mathcal{M}_d^i \right|}
\end{split}
\end{equation*}
\begin{small}
(*) Choose the real non-negative solution.
\end{small}
\item[Convergence:] Stop if $\mathcal{L} \left( X | \Theta^{\left( t+1 \right)}, A^{\left( t+1 \right)} \right) - \mathcal{L} \left( X | \Theta^{\left( t \right)}, A^{\left( t \right)} \right) \leq \epsilon$; otherwise $t = t + 1$ and go to \textbf{Expectation step}.
\end{description}
\end{siderules}

\section{Deep Learning}
\label{section:deeplearing}
At the beginning of the 2000's the \ac{ML} started an important revolution with the birth of the \acf{DL} techniques \citep{Lecun2015}. Until that time, \ac{ML} was generally constituted by two disjoint and clearly differentiated stages: the feature extraction step and the classification/regression step. The feature extraction step was typically addressed manually, computing hand-crafted features from the raw data guided by the expertise, knowledge and intuition of the researcher. These features were then fed to the classifier/regressor, expecting to be discriminant informational inputs to successfully solve the corresponding task. Therefore, a crucial step for the good performance of classic \ac{ML} methods typically relied in a powerful feature extraction process.

\begin{figure}[htbp!]
\centering
\includegraphics[width=0.8\linewidth]{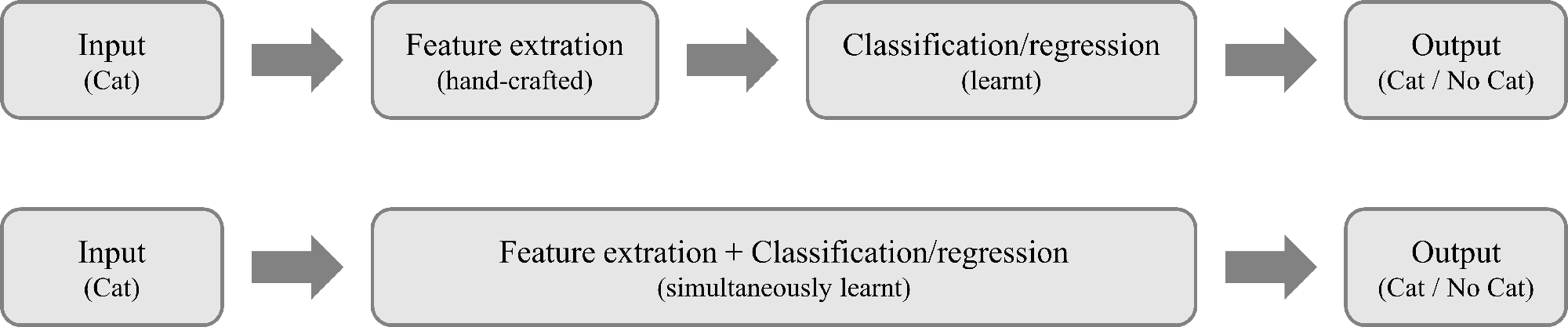}
\caption{Conceptual schema of the differences between classic \ac{ML} and \ac{DL}.}
\label{figure:rationale_dl}
\end{figure}

The qualitative leap made by \ac{DL} techniques concerns mainly the feature extraction stage. Instead of computing hand-crafted features, \ac{DL} models attempt to simultaneously learn a discriminant representation of the raw input data at the same time as the classifier (regressor) is trained. This concept shift allows machines to learn the optimal set of features appropriate for the task. Therefore, \ac{DL} models not only learn the decision boundaries (or regression model) of a discriminative classifier, but they also learn a manifold to optimally project the data into it in such a way that the classification task becomes as trivial as possible: the so-called \emph{representation learning}. Figure \ref{figure:rationale_dl} depicts a general conceptual schema of the concept shift between classic \ac{ML} approach and the current \ac{DL} paradigm.

\medskip

\begin{figure}[htbp!]
\centering
\includegraphics[width=0.9\linewidth]{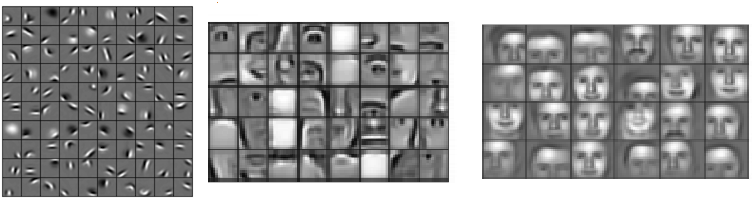}
\caption{Example of hierarchical features extracted during the learning process of a \ac{DL} model. Image taken with kind permission from \url{https://devblogs.nvidia.com/deep-learning-nutshell-core-concepts/hierarchical_features/}}
\label{figure:rationale_hierarhical_features}
\end{figure}

\ac{DL} addresses the representation learning problem by building concepts out of simpler concepts. This kind of hierarchical learning perfectly fits into the nature of \acp{ANN}. \acp{ANN} are \ac{ML} models that process the input hierarchically by passing it through a nested set of layers, each of which learns a deeper level of abstraction of the input. Therefore, concepts inside a \ac{ANN} are built by concatenating the output of a layer, which represents a simpler abstract representation of the input, and the input of the successive layer, which learns a more complex and richer composite representation from those previous outputs. Figure \ref{figure:rationale_hierarhical_features} shows an example of the hierarchical concept representation of images showing faces by a \ac{DL} model.

The following sections briefly describe the basics and fundamental principles of \acp{ANN}, and present the \acp{CNN}: the current state-of-the-art classifier for most of the supervised visual computing related tasks.

\subsection{\aclp{ANN}}
\label{subsections:rationale_dl_ann}
\acp{ANN} are connectionist computational models in which a number of processors, named \emph{artificial neurons} or \emph{perceptrons} are interconnected in a manner inspired by the connections between neurons in a human brain. \acp{ANN} were born in the mid-twentieth century, in a combined effort of multiple researchers. In 1943 \cite{Mcculloch1943} created the first computational model of a biological neuron. Few years later \cite{Hebb2002} designed the first learning rule for \acp{ANN} based on the concept of \emph{neural plasticity}, whose premise was that if two neurons were active simultaneously, then the strength of the connection between them should increase. In 1958, \cite{Rosenblatt1958} created the perceptron, which is the basis of the \acp{ANN}, and \cite{Ivakhnenko1967} developed the first multi-layer \ac{ANN}. In subsequent years, \cite{Kelley1960} laid the basis of the backpropagation algorithm, but it was not until the work of \cite{Rumelhart1986} where the first \ac{ANN} was effectively trained using the backpropagation algorithm by adjusting the weights of the network proportional to the gradient computed from the error.

Today, \acp{ANN} are the most powerful and widely used \ac{ML} algorithms, as they have proven to largely outperform other approaches in many tasks such as visual object recognition, object detection and segmentation and many other domains such as speech recognition, handwritten text recognition or financial time series prediction. The following sections provides an overview of the basics of \acp{ANN} to facilitate the reading and understanding of subsequent chapters and contributions of this thesis.

\subsubsection{The Perceptron}
\label{subsubsections:rationale_dl_ann_perceptron}
\begin{figure}[htbp!]
\centering
\includegraphics[width=0.65\linewidth]{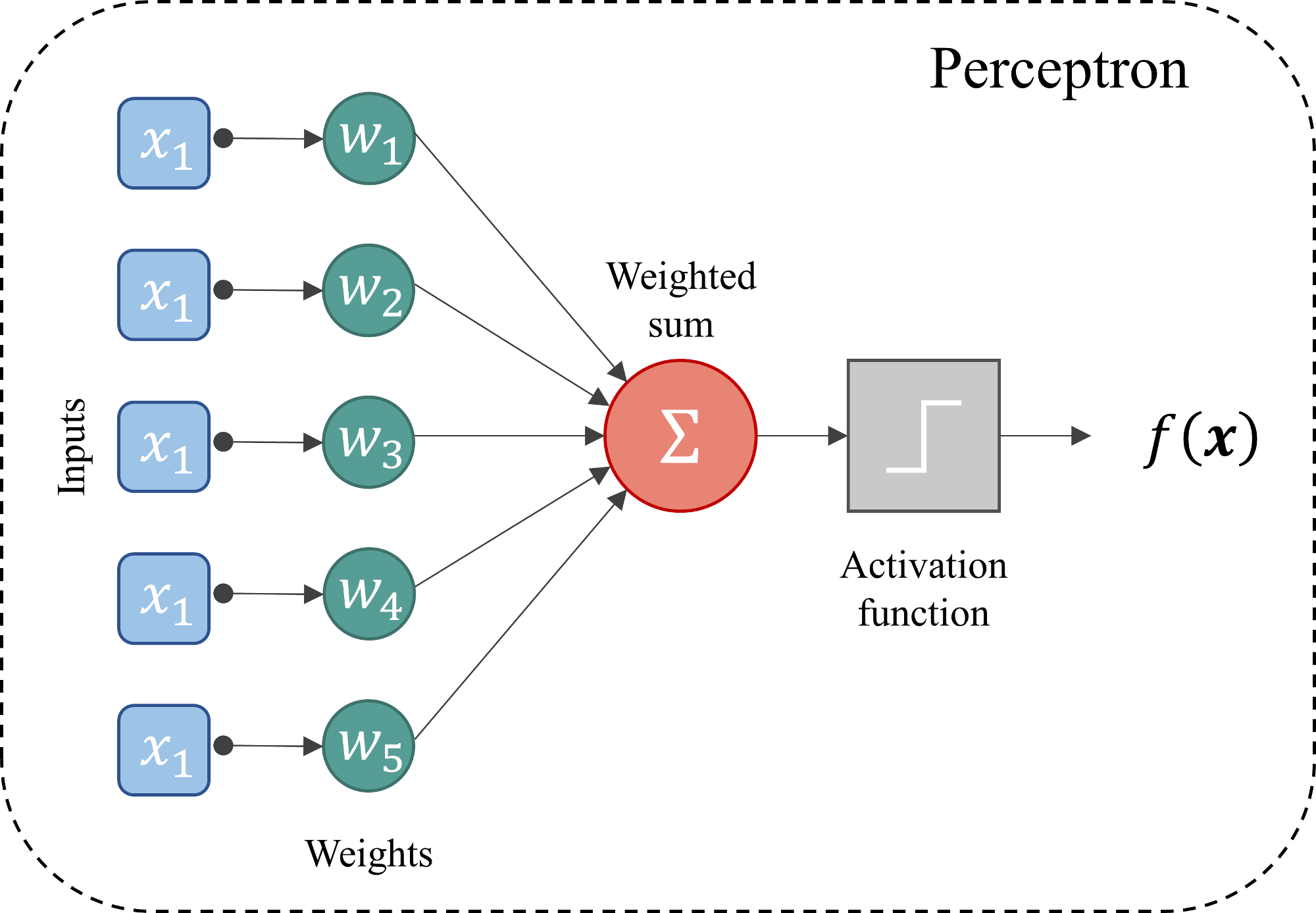}
\caption{The perceptron.}
\label{figure:rationale_dl_perceptron}
\end{figure}

\acp{ANN} are composed by basic units called \emph{perceptrons} or \emph{artificial neurons}. The perceptron (see Figure \ref{figure:rationale_dl_perceptron}), in its most elementary form, is a simplified model of a biological neuron, approximated as a mathematical function for learning a binary linear classifier. Let $\vec{x} = \left\lbrace x_1, \ldots , x_D \right\rbrace$ a $\mathbb{R}^D$ vector of features representing an observation. Mathematically, a perceptron is defined as:

\begin{equation}
f \left( \vec{x} \right) = 
\begin{cases}
1 & \text{if } \sum_i w_i x_i + b > 0 \\
0 & \text{otherwise} \\
\end{cases}
\end{equation}

\noindent where $w_i$ represents a weight associated to the $x_i$ feature, and $b$ stands for the bias of the linear model. The Heaviside step activation function is employed in the most basic perceptron algorithm, setting the output of the neuron to 1 ("activated") if $\sum_i w_i x_i + b$ is greater than a threshold (typically 0); or to 0 ("deactivated") otherwise.

Perceptrons are only able to solve linearly separable problems regardless of the activation function employed. Therefore, they are only guaranteed to converge if the training set is linearly separable. Otherwise, the perceptron will fail and no approximate solution is returned.

\subsubsection{The Multi-Layer Perceptron}
\begin{figure}[htbp!]
\centering
\includegraphics[width=0.60\linewidth]{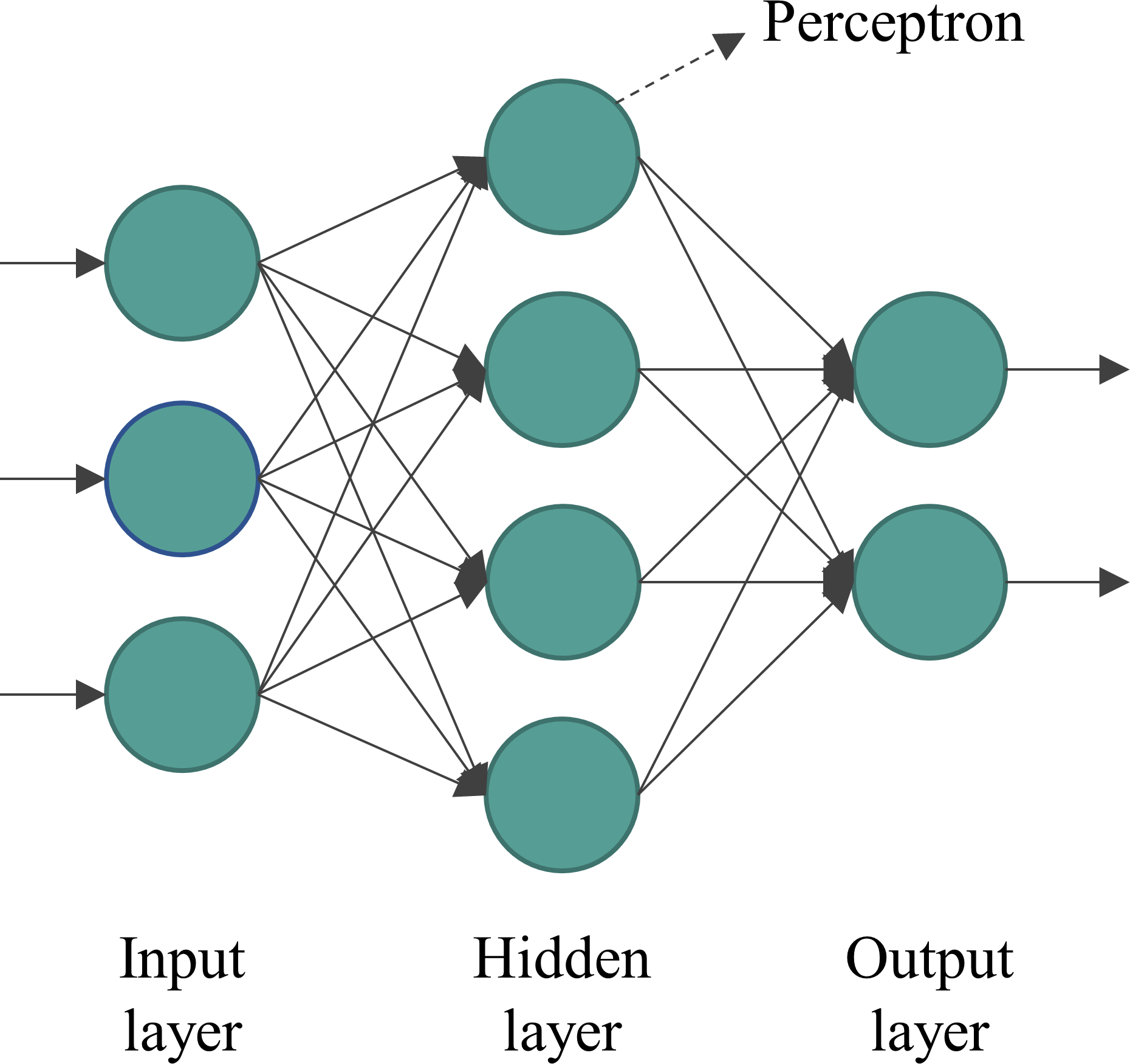}
\caption{A multi-layer perceptron.}
\label{figure:rationale_dl_mlp}
\end{figure}

A \ac{MLP} is a class of feedforward \ac{ANN} composed of at least three layers of interconnected perceptrons. Figure \ref{figure:rationale_dl_mlp} shows an example of a simple \ac{MLP} with an input layer, a hidden layer and an output layer. Except of the input layer, all the neurons in the other layers are perceptrons.

The arrangement of multiple layers of perceptrons in combination with the non-linear activation functions allow \acp{MLP} to solve non-linearly separable problems. Indeed, it is demonstrated that \acp{MLP} are universal mathematical function approximators through superposition of non-linearly activated perceptrons.

\subsubsection{Activation functions}
\label{subsubsections:rationale_dl_ann_activation_functions}
Activation functions are determinant for \acp{ANN} since they are fundamental to allow the network to solve non-linear problems. An \ac{ANN} of multiple layers, each one activated by a linear function passed to the next layer, results in a final function that is a combination of linear functions in a linear form, which by definition can be replaced by a single linear function. In other words, no matter how many layers are stacked in an \ac{ANN}, if only linear activation functions are used, the network is still equivalent to a single layer with linear activation, and therefore is only capable of solving linear problems.

To mitigate this effect multitude of activation functions are proposed in the literature. The most important are summarized below:

\paragraph{\emph{Linear activation}}

The most simple activation function. The output of the perceptron is proportional to the input. It is defined as:

\begin{equation}
f \left( x \right) = cx
\end{equation}

It allows a continuous output range, which is more powerful than a binary output, however its derivative with respect to $x$ is constant. This means that the gradient has no relation to $x$ and, therefore, regardless of the magnitude of the prediction error, the changes made by backpropagation are constant. Additionally, as said above, a multi-layer \ac{ANN} fully activated by linear functions is equivalent to a single-layer \ac{ANN} only capable of solving linear problems.

\paragraph{\emph{Heaviside step activation}}

A binary activation function. The output of the perceptron is "activated" if $\sum_i w_i x_i + b$ is greater than a threshold, or is "deactivated" otherwise. It is typically defined as:

\begin{equation}
f \left( x \right) = 
\begin{cases}
1 & \text{if } \sum_i w_i x_i + b > 0 \\
0 & \text{otherwise} \\
\end{cases}
\end{equation}

\noindent where "activated" is typically represented by 1, and "deactivated" fires the 0 value. Likewise, the typical threshold employed to determine the output is the 0 value.

The Heaviside function outputs a binary discrete value, which presents some disadvantages. By definition it can only properly represent binary classification problems, in which the \ac{ANN} has only one output neuron. For multi-label  problems where \acp{ANN} have multiple output neurons firing Heaviside step activations, it is not possible to correctly identify the corresponding class, hence preventing their training. Moreover, continuous activation functions allow a smoother training process, less prone to fall into bad local minima.

\paragraph{\emph{Sigmoid / Logistic activation}}

One of the most historically used activation functions. It represents the continuous soft approximation of the Heaviside step function. It is mathematically defined as:

\begin{equation}
f \left( x \right) = \frac{1}{1 + \exp\left( -x \right)}
\end{equation}

Its non-linearity allows the network to learn complex decision boundaries by stacking layers sequentially. Moreover, the gradient is related to the input, so it propagates the magnitude of the error through the entire network. Finally, the output is continuous and bounded to the $\left[ 0,1 \right]$ range, allowing a smoother training process without gradient explosion.

The main disadvantage of this activation function is that it originates the so-called \emph{gradient vanishing} problem. The gradient vanishing problem is related to the update process of the $w_i$ weights of a network trained through gradient-based learning algorithms such as \emph{backpropagation}. In these algorithms, the weights of the network are updated proportionally to the partial derivative of the error function with respect to the weights. Depending on the nature of the activation function, it is possible for the gradient to become extremely small, barely modifying the weights of the network, and therefore slowing or even preventing the network from further training.

In the specific case of the sigmoid activation function, it can be seen that for $x$ values above $x=2$ and below $x=-2$, the slope of the function is almost near-horizontal. Therefore, there is no effectively change in the gradient, resulting in the network refusing to learn further. For small networks, this does not represent a big problem, however, the more layers stacked in the network, the problem grows exponentially, eventually collapsing the training of the model.

\paragraph{\emph{Tahn activation}}

A similar activation function than the sigmoid function. Indeed, the tanh activation is a scaled version of the sigmoid function:

\begin{equation}
f \left( x \right) = \tanh \left( x \right) = \frac{2}{1 + \exp\left(-2x \right)}-1 = 2 sigmoid \left( 2x \right) - 1
\end{equation}

It has similar properties than the sigmoid function. Its gradient is stronger than the sigmoid but it also suffers from the vanishing gradient problem.

\paragraph{\emph{Softmax activation}}

Typically an activation function only employed in the final layer of a network to convert the activations into probabilities (posteriors). It is defined as:

\begin{equation}
f \left( \vec{x} \right) = \frac{\exp\left(x_i\right)}{\sum_j \exp\left( x_j \right)}
\end{equation}

The aim of this activation function is to normalize the outputs so that each neuron triggers a value in the range $\left[ 0,1 \right]$ and all of them add up to 1, giving the probability of the input value being in a specific class.

\paragraph{\emph{ReLU activation}}

First presented in 2009, the Rectified Linear Unit (ReLU) activation function can be considered a milestone in \ac{DL}. It is mathematically defined as:

\begin{equation}
f \left( x \right) = \max \left( 0, x\right)
\end{equation}

This simple function provides a set of benefits that have made ReLU as the most widely used activation function nowadays. First, ReLU is a non-linear function, so it allows the network to model complex functions by stacking layers. Moreover, the function is ranged between $\left[ 0, +\infty \right)$, hence allowing the gradient to not vanish as there is no saturation in any range of the function. It also provides sparsity activation since multiple neurons can fire 0 activation because of a negative input of the ReLU, therefore becoming a lighter network.

However, ReLU activation suffers from the so-called \emph{dying ReLU} problem. This problem arises when a neuron continuously trigger negative values to the ReLU. In these cases, the gradient become 0, preventing the neuron from responding to changes in error, and updating its weights. If the problem affects multiple neurons in the network, it could lead to a substantial part of the network passive.

\paragraph{\emph{Leaky-ReLU activation}}

It consist of a simple modification of the ReLU function to avoid the dying ReLU problem:

\begin{equation}
f \left( x \right) = \max \left( \alpha x, x\right)
\end{equation}

\noindent with $\alpha$ a parameter, typically adopting small values $\left( \alpha = 0.01 \right)$, to allow a small positive slope for the negative range of the function. Such modification allows the neuron to prevent its paralysis, eventually reactivating it during the training process.

\subsubsection{Backpropagation algorithm}
The backpropagation algorithm was originally introduced in the 1970s, but its important was not fully recognized until the 1986 paper by \cite{Rumelhart1986}. Backpropagation is an algorithm to efficiently train an \ac{ANN} by adjusting its weights so that the network output minimizes a given loss function. More formally, the backpropagation is an iterative optimization algorithm based on a gradient descent technique, to update the weights of a network by computing the gradient of the loss function with respect to each weight following the chain rule.

\begin{figure}[htbp!]
\centering
\includegraphics[width=0.60\linewidth]{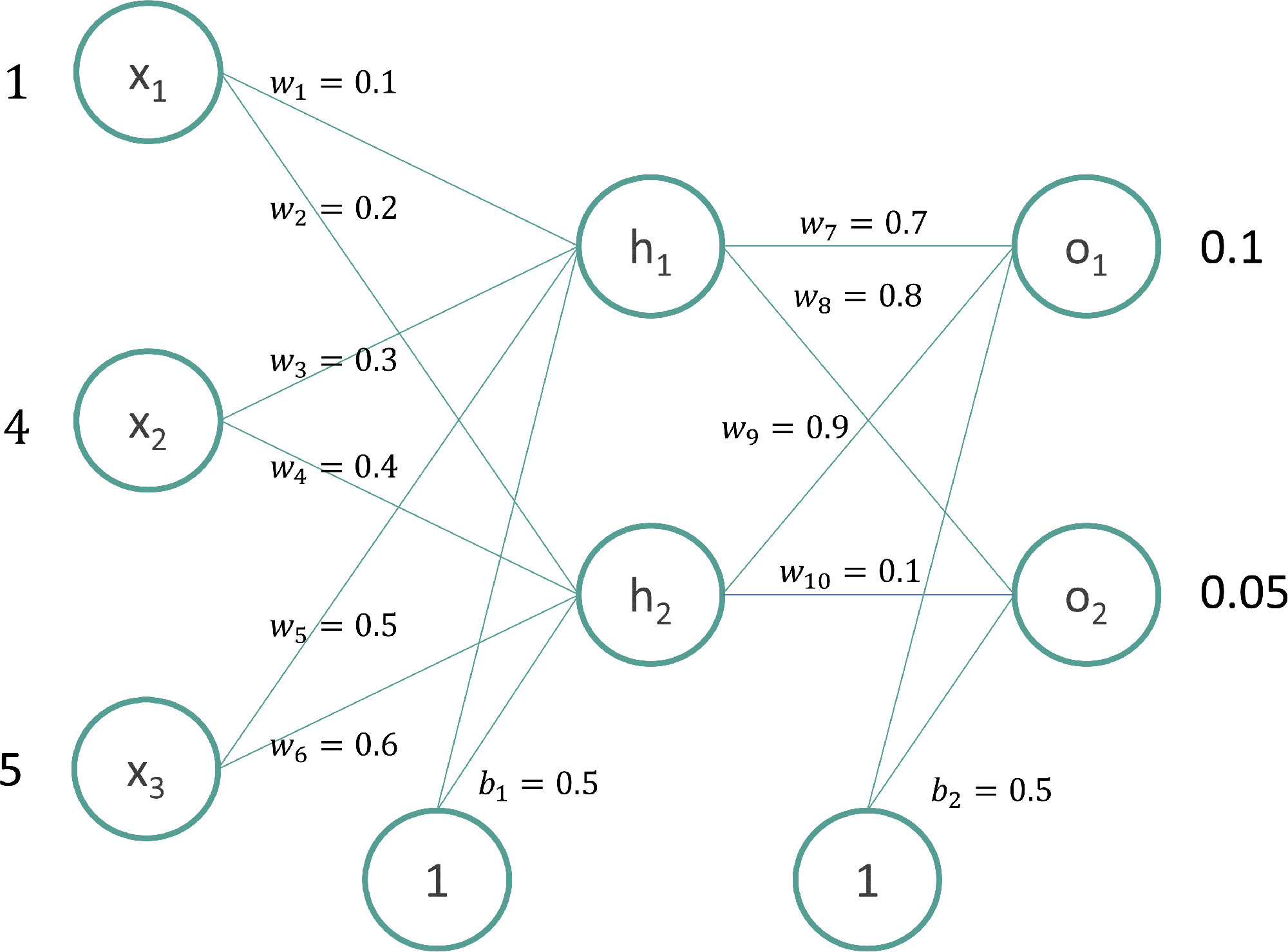}
\caption{\ac{ANN} example.}
\label{figure:rationale_dl_ann}
\end{figure}

To illustrate the mechanics of the backpropagation algorithm, lets consider the simple \ac{ANN} shown in figure \ref{figure:rationale_dl_ann} (example taken from \url{https://www.anotsorandomwalk.com/backpropagation-example-with-numbers-step-by-step/} with kind permission).

The following relations are established between the neurons of the network:

\begin{equation*}
\begin{split}
y_{h_1} &= w_1 x_1 + w_3 x_2 + w_5 x_3 + b_1 \\
y_{h_2} &= w_2 x_1 + w_4 x_2 + w_6 x_3 + b_1 \\
z_{h_1} &= f \left( y_{1h_1} \right) \\
z_{h_2} &= f \left( y_{h_2} \right) \\
\\
y_{o_1} &= w_7 z_{h_1} + w_9 z_{h_2} + b_2 \\
y_{o_2} &= w_8 z_{h_1} + w_{10} z_{h_2} + b_2 \\
z_{o_1} &= \sigma\left( y_{o_1} \right) \\
z_{o_2} &= \sigma\left( y_{o_2} \right) \\
\end{split}
\end{equation*}

\noindent where $y_{\cdot}$ refers to the input of the corresponding neuron, and $z_{\cdot}$ indicates the output of the neuron after the activation function $f\left(y\right)$ is applied, which in the example will be the sigmoid function.

Let assume the input values of the network $x_1 = 1, x_2 = 4, x_3=5$ and the desired target $t_1 = 0.1, t_2 = 0.05$. The forward pass of the network yields the following results:

\begin{equation*}
\begin{split}
y_{h_1} &= 0.1 \left( 1 \right) + 0.3 \left( 4 \right) + 0.5 \left( 5 \right) + 0.5 = 4.3 \\
y_{h_2} &= 0.2 \left( 1 \right) + 0.4 \left( 4 \right) + 0.6 \left( 5 \right) + 0.5 = 5.3 \\
z_{h_1} &= \sigma\left( 4.3 \right) = 0.9866 \\
z_{h_2} &= \sigma\left( 5.4 \right) = 0.9950 \\
\\
y_{o_1} &= 0.7 \left( 0.9866 \right) + 0.9 \left( 0.9950 \right) + 0.5 = 2.0862 \\
y_{o_2} &= 0.8 \left( 0.9866 \right) + 0.1 \left( 0.9950 \right) + 0.5 = 1.3888 \\
z_{o_1} &= \sigma\left( 2.0862 \right) = 0.8896 \\
z_{o_2} &= \sigma\left( 1.3888 \right) = 0.8004 \\
\end{split}
\end{equation*}

\medskip

Once the forward pass is completed, weights are updated by setting partial derivatives of the error function $E$ with respect to each weight, following the chain rule. As an example, setting partial derivatives to update the weight $w_7$ yield:

\begin{equation}
\label{eq:rationale_partial_derivatives}
\frac{\partial E}{\partial w_7} = \frac{\partial E}{\partial z_{o_1}} \cdot \frac{\partial z_{o_1}}{\partial y_{o_1}} \cdot \frac{\partial y_{o_1}}{\partial w_7}
\end{equation}

For the shake of simplicity lets consider $E$ the sum of squared errors of the predictions with respect to the targets:

\begin{equation}
E = \frac{1}{2} \left[ \left( z_{o_1} - t_1 \right)^2 + \left( z_{o_2} - t_2 \right)^2 \right]
\end{equation}

Solving for the first term $\frac{\partial E}{\partial z_{o_1}}$ of the partial derivatives in equation \ref{eq:rationale_partial_derivatives} yields:

\begin{equation}
\frac{\partial E}{\partial z_{o_1}} = z_{o_1} - t_1 = 0.8896 - 0.1 = 0.7896
\end{equation}

The next term to solve is $\frac{\partial z_{o_1}}{\partial y_{o_1}}$, which only engages the activation function of the neuron. The activation function chosen for the example is the sigmoid function, whose derivative is $x \left( 1 - x \right)$. Applying to the update process of the weights yields:

\begin{equation}
\frac{\partial z_{o_1}}{\partial y_{o_1}} = z_{o_1} \left( 1 - z_{o_1} \right) = 0.8896 \left( 1 - 0.8896 \right) = 0.0982
\end{equation}

Finally, the partial derivative $\frac{\partial y_{o_1}}{\partial w_7}$ involves the derivation of $w_7 z_{h_1} + w_9 z_{h_2} + b_2$ with respect $w_7$, which is trivial:

\begin{equation}
\frac{\partial y_{o_1}}{\partial w_7} = z_{h_1} = 0.9866
\end{equation}

Therefore, the gradient of the error with respect to the weight $w_7$ is:

\begin{equation}
\frac{\partial E}{\partial w_7} = 0.7896 \cdot 0.0982 \cdot 0.9866 = 0.0765
\end{equation}

\medskip

Once the gradient of the loss function with respect to the corresponding weight is known, the final update of the weight is typically performed by:

\begin{equation}
\hat{w}_7 = w_7 - \eta \frac{\partial E}{\partial w_7} = 0.7 - \left( 0.01 \cdot 0.0765 \right) = 0.6992
\end{equation}

\noindent with $\eta$ referring to the \emph{learning rate} typically set to small values $\left( \eta = 0.01 \right)$.

\medskip

The updates of the remaining weights follow the same mechanics, with a slightly more cumbersome computation due to the increase in the number of neurons interconnected in deeper layers. The results are:

\begin{equation*}
\begin{split}
\hat{w}_1    &= w_1 - \eta \frac{\partial E}{\partial w_1} = 0.1000 \\
\hat{w}_2    &= w_2 - \eta \frac{\partial E}{\partial w_2} = 0.2000 \\
\hat{w}_3    &= w_3 - \eta \frac{\partial E}{\partial w_3} = 0.2999 \\
\hat{w}_4    &= w_4 - \eta \frac{\partial E}{\partial w_4} = 0.4000 \\
\hat{w}_5    &= w_5 - \eta \frac{\partial E}{\partial w_5} = 0.4999 \\
\hat{w}_6    &= w_6 - \eta \frac{\partial E}{\partial w_6} = 0.6000 \\
\end{split}
\quad\quad\quad\quad\quad
\begin{split}
\hat{w}_7    &= w_7 - \eta \frac{\partial E}{\partial w_7} = 0.6992 \\
\hat{w}_8    &= w_8 - \eta \frac{\partial E}{\partial w_8} = 0.7988 \\
\hat{w}_9    &= w_9 - \eta \frac{\partial E}{\partial w_9} = 0.8992 \\
\hat{w}_{10} &= w_{10} - \eta \frac{\partial E}{\partial w_{10}} = 0.0988 \\
\hat{b}_1    &= b_1 - \eta \frac{\partial E}{\partial b_1} = 0.5000 \\
\hat{b}_2    &= b_2 - \eta \frac{\partial E}{\partial b_2} = 0.4980 \\
\end{split}
\end{equation*}

After weight updating, computing the output of the network in the iteration $t=1$ yields:

\begin{equation*}
\begin{split}
z_{o_1}^{\left( 1 \right)} &= 0.8892 \\
z_{o_2}^{\left( 1 \right)} &= 0.7997 \\
\end{split}
\end{equation*}

Comparing the errors produced by the network in the first $\left( t = 0 \right)$ and second iterations $\left( t = 1 \right)$:

\begin{equation*}
\begin{split}
E^{\left( 0 \right)} &= 0.5933 \\
E^{\left( 1 \right)} &= 0.5925 \\
\end{split}
\end{equation*}

\noindent which initially may not seem too much, but after repeating the process 100.000 times, the error decreases to 0.0000351085, and the output of the network becomes:

\begin{equation*}
\begin{split}
z_{o_1}^{\left( 100k \right)} &= 0.1175 \quad\rightarrow\quad \left( \text{vs. } t_1 = 0.1 \right) \\
z_{o_1}^{\left( 100k \right)} &= 0.0582 \quad\rightarrow\quad \left( \text{vs. } t_2 = 0.05 \right) \\
\end{split}
\end{equation*}

\noindent concluding the training of the network. For an in-depth dissertation on \acp{ANN}, training procedures, numerical optimizers and convergence properties of \ac{DL} models please refer to \cite{Bengio2017}.

\subsection{\aclp{CNN}}
\acp{CNN} are a specialized type of \acp{ANN} inspired in the human visual cortex. In the 1950s, experiments conducted by Dr. Hubel and Dr. Wiesel identified that sets of different specific neurons in a cat's visual cortex responded very quickly when observing images with lines at specific angles, with light and dark patterns, or by observing movement in a certain direction. These results laid the foundation for further research into animal's visual understanding.

Human visual cortex process the information by passing it from one cortical area to another, where each cortical area is more specialized than the last one. For example, the primary visual cortex focuses on preserving spatial location of visual information, i.e. orientation of edges and lines. The secondary visual cortex feeds on the response of the primary visual cortex and focuses on collecting spatial frequencies, size, colors and shape of the object, while third visual cortex process the global motion and provides a complete visual representation. \acp{CNN} adopts this philosophy and tries to act like a computational visual cortex.

\acp{CNN} are a type of \ac{ANN} mainly oriented for image processing, that arose in the attempt to solve the limitations that classic \acp{MLP} have when dealing with images. First, fully-connected \acp{ANN} do not scale well for images. Consider the example of a small image of $100 \times 100$ pixels. Each neuron of the second layer of a fully-connected \ac{ANN} has $10000$ weights to learn. Therefore, assuming a simple \ac{MLP} with a hidden layer of $1000$ neurons has $10$M of weights only in the second layer, making it unaffordable to optimize. Second, \acs{MLP} also ignore the spatial information and local redundancy inherent in images. In a \ac{MLP}, the pixels of an image are independently connected to the layers of the network, thus loosing the semantic meaning of the position and its relation with their neighbors. This wastes a highly useful information that allows images to be described in a more concise, adequate and realistic manner. Finally, classic \acp{MLP} are not translation-invariant. Since the input image is connected in a static manner to the layers of the network, the patterns learned during the training process are statically linked to the position in which they appeared. Therefore, same objects appearing at different positions will change the weights associated to the corresponding pixels, intrinsically learning the position in which they appeared.

\acp{CNN}, by the opposite, are non-fully connected \acp{ANN}, primarily made of a set of stacked convolution layers, where each layer has a small set of shared weights that intelligently adapt to the properties of images. They are by definition translation invariant and context-sensitive, allowing for learning patterns related to objects of interest appearing anywhere in the image.

The following sections overall introduce the main components and state-of-the-art architectures for image segmentation with \acp{CNN}.

\subsubsection{Convolution layer}
\begin{figure}[htbp!]
\centering
\includegraphics[width=0.65\linewidth]{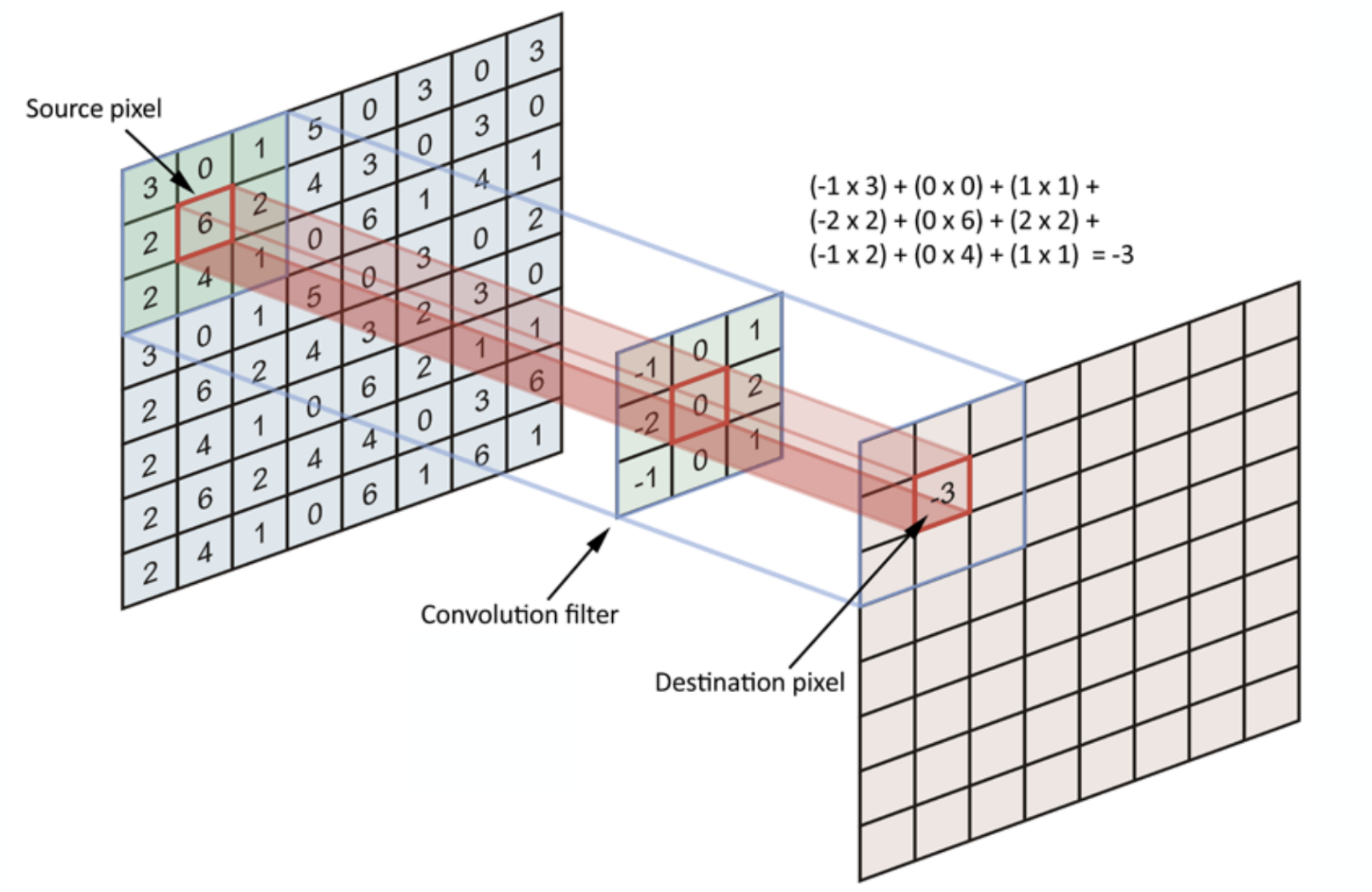}
\caption{The convolution operation. Image taken with kind permission from \url{https://towardsdatascience.com/simple-introduction-to-convolutional-neural-networks-cdf8d3077bac}.}
\label{figure:rationale_dl_convolution}
\end{figure}

A convolution layer is a special arrangement of neurons, typically in a 2D or 3D grid, that acts as a bank of learnable \emph{filters} (sometimes also called \emph{kernels}). These filters are used to perform a sliding dot product with an image to extract the set of optimal features to solve the classification/regression task. The dot product operation, also named \emph{convolution}, is defined as:

\begin{equation}
I \otimes K = \sum_i I_i \cdot K_i
\end{equation}

\noindent where $I$ and $K$ are arrays of the same dimension, and $i$ represents the index to traverse such arrays. Figure  \ref{figure:rationale_dl_convolution} shows an example of a convolution between a kernel of size $3 \times 3$ and a region of an image of the same size.

Therefore, unlike the classic fully-connected layers of an \ac{ANN}, a convolution layer is not fully connected to its input, but it contains small banks of filters of shared weights that traverse the different locations of the image. This introduces several benefits: first, the number of weights to train in a \ac{CNN} is drastically lower than on a classic \ac{ANN}. Considering the previous aforementioned example, a \ac{CNN} with a first layer of $100$ learnable filters of size $3 \times 3$ has only $900$ weights to train, regardless of the image size. Moreover, the $3 \times 3$ weights of each filter are shared across the entire image, making the pattern learned by the filter invariant to location and translation. Finally, the 2D or 3D arrangement of the filter weights inherently captures the spatial information and the semantic meaning of the positions in which the intensities appear in the images.

In addition to the kernel size, convolution operation also involves two parameters that must be taken into account: the \emph{stride} and the \emph{padding} methods. The stride controls how the filter convolves around the input image. In other words, the stride fixes how many pixels the kernel is moved in one direction between two successive convolutions. Therefore, a stride of $1$ means that, after a convolution, the kernel must be moved one pixel in only one direction and then perform the next convolution. The padding controls the behavior of the convolution in terms of the size of the resulting map after the convolution. The convolution operation, by definition, shrinks the image into a factor related to the size of the kernel used for the operation. Additionally, pixels in the corners and edges of the images are not visited the same number of times as central pixels, thus, giving more importance to the latter. To compensate for this effect, a padding method can be employed to pad the input image so that the resulting map after the convolution has the same size than the input image. There are different types of padding: zero-padding, mirror-padding and reflect-padding; with zero-padding the most widely used in \acp{CNN}.

\subsubsection{Batch Normalization layer}
Batch normalization is a technique employed to normalize the inputs of a layer, with the main purpose of mitigating the \emph{internal covariate shift} problem. The covariate shift problem refers to the continuous shift that the internal activations of the network undergo during the training process, due to continuous changes in the distribution of the inputs of the layers. Since the inputs of a layer are the activations of the previous layer, a significant change in the distribution of these activations forces the intermediate layers to continuously adapt itself to these numerical fluctuations. This leads to a situation where each layer in the \ac{ANN} wastes training iterations in this adaptation rather than learning the relations between the inputs and the desired target.

The batch normalization technique prevents this situation by normalizing the activations of a layer, before feeding them as the input of the next layer. Let $X = \left( \vec{x}^1, \ldots , \vec{x}^N \right)$ a batch of $N$ inputs of a layer, the batch normalization performs as follows:

\begin{equation*}
\begin{split}
\vec{\mu}_{\mathcal{B}} &= \frac{1}{N}\sum_{i=1}^N \vec{x}^i \\
\vec{\sigma}_{\mathcal{B}}^2 &= \frac{1}{N}\sum_{i=1}^N \left( \vec{x}^i - \vec{\mu}_{\mathcal{B}} \right)^2 \\
\hat{\vec{x}}^i &= \gamma \frac{\vec{x}^i - \vec{\mu}_{\mathcal{B}}}{\sqrt{\vec{\sigma}_{\mathcal{B}}^2}} + \beta \\
\end{split}
\end{equation*}

\noindent where $\vec{\mu}_{\mathcal{B}}$ is the mean of the batch, $\vec{\sigma}_{\mathcal{B}}^2$ is the variance of the batch and $\hat{\vec{x}}^i$ is the $i^{th}$ normalized sample of the batch. Batch normalization technique introduces two learnable parameters, $\gamma$ and $\beta$, to allow the network to perform a scale and shift of the batch if the training process requires it.

Besides mitigating the internal covariate shift, batch normalization also provides other related benefits to the training process of an \ac{ANN}. First, batch normalization allows each layer to learn a little bit more independently of other layers. Since the inputs are always normalized in the same numerical range, the layer can focus on learning the relations between the inputs rather than the relations between the adjacent layers. Second, batch normalization also significantly accelerates the training process. Higher learning rates can be employed when using batch normalization since it ensures that there are no a outlier activations. Finally, it reduces overfitting because it has a slight regularization effect over the entire network.

\subsubsection{Pooling layer}
The function of the pooling layers is to progressively reduce the size of the activation maps in order to also reduce the number of weights and the computational cost of the network, while retaining the most discriminant information. Additionally, pooling also allows to increase the receptive field of the network by compressing the information in smaller activation maps.

\begin{figure}[htbp!]
\centering
\includegraphics[width=0.65\linewidth]{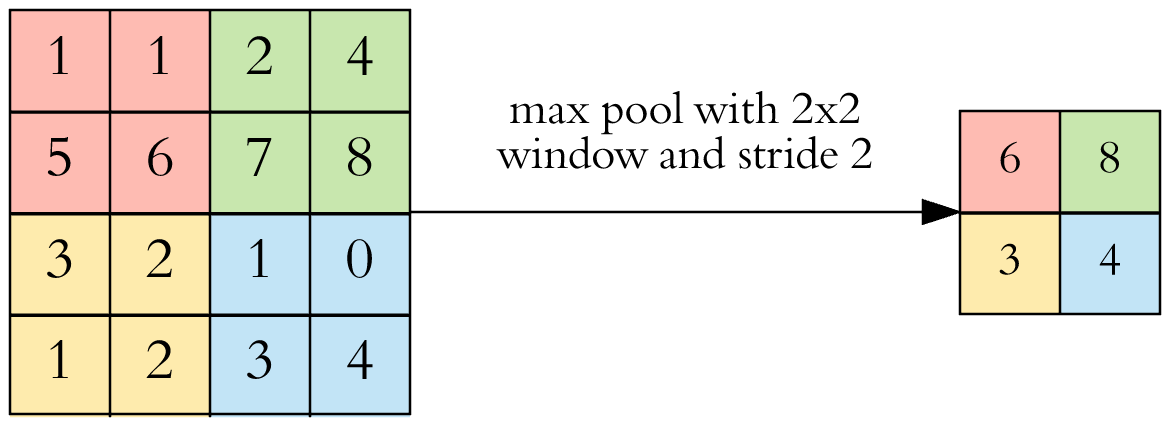}
\caption{Max pooling with filters of size $2 \times 2$ and stride $2$. Image taken with kind permission from \url{https://towardsdatascience.com/applied-deep-learning-part-4-convolutional-neural-networks-584bc134c1e2}.}
\label{figure:rationale_dl_maxpooling}
\end{figure}

There are three main types of pooling layers: max pooling, average pooling and sum pooling. Max pooling - the most frequently pooling layer used in \acp{CNN} - uses filters to take the largest elements of the activation maps in the pooling window. Figure \ref{figure:rationale_dl_maxpooling} shows an example of the result of a max pooling layer with filters of $2 \times 2$ and stride $2$.

As expected, average pooling computes the average value within the pooling layer, while sum pooling computes the summation of the values inside each pooling layer.

Pooling operation also provides benefits for preventing overfitting. In a general sense, pooling is a feature selection method by reducing the dimensions of input. Therefore, selecting a subset of activations from the output of the layers drop useless information and introduce slight variations in the new activation maps that helps in overfitting to specific patterns.

\subsubsection{U-Net architecture}
The U-Net architecture was first proposed by \cite{Ronneberger2015} as a modification of the classic \emph{Fully convolutional network} presented in \citep{Long2015}. It consist of a \emph{contracting path} similar than the fully convolutional network, followed by and \emph{expanding path} where pooling layers are replaced by up-sampling layers. This gives the architecture a ``U" shape that justifies its name.

The main innovation of this architecture was the expanding path made up of up-sampling convolutions that allowed to reconstruct the low-dimensional activation maps into high-dimensional maps of the same size of input image. This enabled to address image segmentation problems through \acp{CNN} in a complete naturally manner, since the output activation maps could directly represent the \emph{logits} of each class for each pixel of the image.

\begin{figure}[htbp!]
\centering
\includegraphics[width=0.75\linewidth]{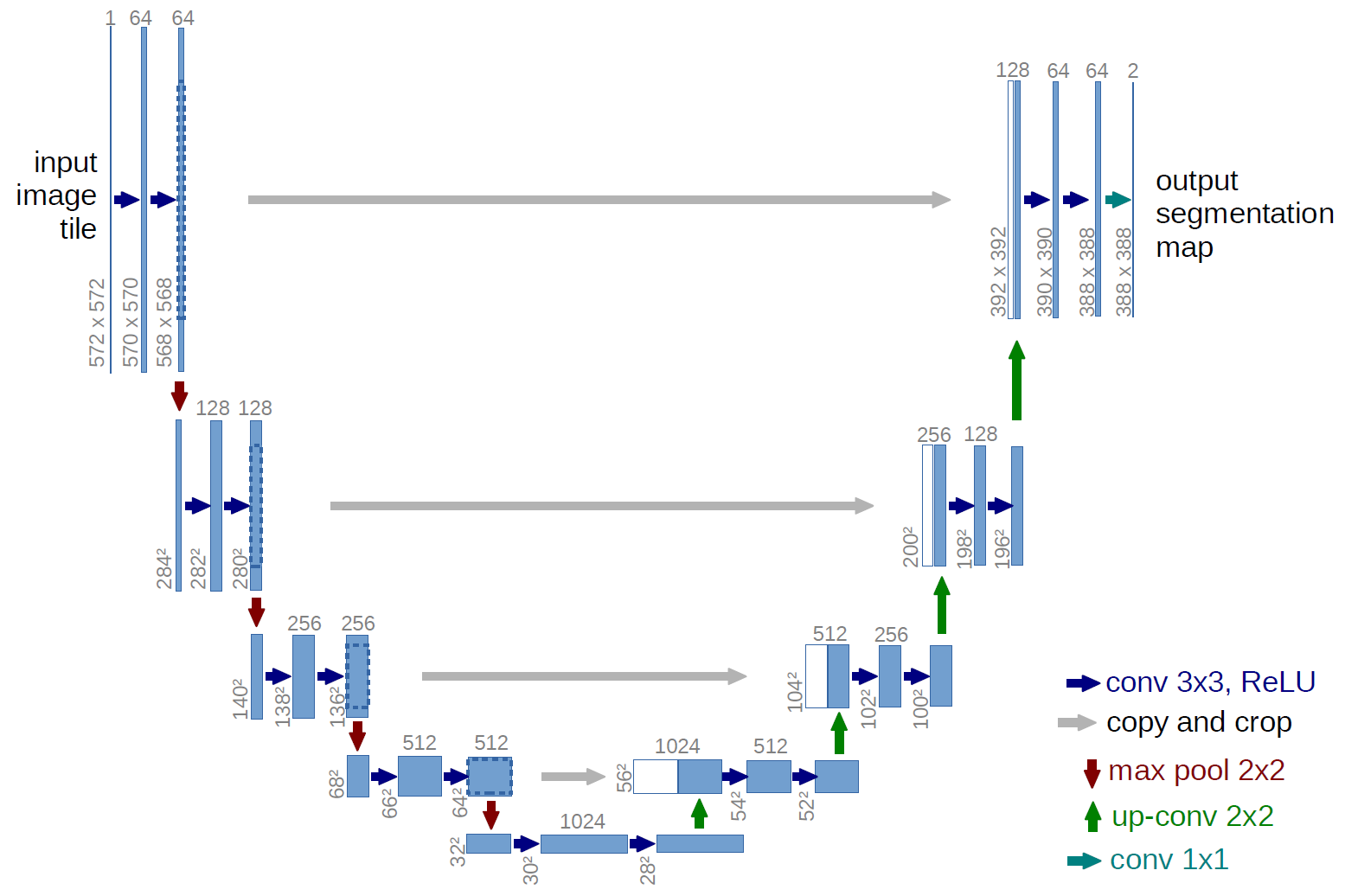}
\caption{U-Net original architecture.}
\label{figure:rationale_dl_unet}
\end{figure}

Figure \ref{figure:rationale_dl_unet} shows the original architecture proposed in the article of \cite{Ronneberger2015}. The network consisted in 5 levels of depth, with 64 activation maps in the first level, 128 in the second, 256 in the third, 512 in the fourth and 1024 in the final level. The contracting path was formed by blocks of two sequential $3 \times 3$ convolutions layers followed by max pooling layers with filters of size $2 \times 2$ at each level. The expanding path performed the opposite job by gradually projecting the low-dimensional activation maps into the space of the inputs, so that the final activation map had as many channels as the number of classes in which each pixel could be classified.

Another important contribution was the loss employed to train the network. The work of \cite{Ronneberger2015} was the first that proposed to apply a pixel-wise softmax to the final activation map, in order to convert the activations into probabilities that represent the degree of confidence of each pixel to belong to each class. Following this approach, the network was finally trained by minimizing the cross-entropy loss function between the softmax predictions and the groundthruth segmentation.

The final significant contribution of the paper was the so-called \emph{long-term skip connections} between same levels in the contracting and expanding paths of the U-Net architecture (denoted as gray arrows in the figure \ref{figure:rationale_dl_unet}). These connections are also biologically inspired in the human visual cortex, where neurons of the first areas of the cortex are directly connected with neurons in deeper regions, skipping intermediary cortex areas. This allows to directly propagate useful signals between different regions of the cortex to properly understand the scene. The same mechanism is mimicked in U-Net architectures with the long-term skip connections. These connections introduce two main benefits in the network: first, they allow to easily propagate the gradient to the first layers, where it is more difficult to adjust weights (due to gradient vanishing problem). Second, these connections bring the ability to directly pass useful information captured in the first layers that is later required to reconstruct the activation maps during the up-sampling.

Since 2012, as a result of the breakthrough brought about by AlexNet \citep{Krizhevsky2012}, \acp{CNN}, and specifically U-Net architectures, clearly dominate most of the computer vision and image understanding challenges, positioning them as the de facto standard \ac{DL} classifiers for addressing these tasks.

\chapter{Comparative study of unsupervised learning algorithms for glioblastoma segmentation}
\label{chapter:comparative_unsupervised_learning}
Unsupervised learning constitutes one of the most important roles in automated image understanding. Specifically, it has historically played a key role in the medical image segmentation task, as it provided the first approaches to identify tissues in \ac{MRI} acquisitions without requiring manual human intervention. However, currently, medical image segmentation is often dominated by supervised learning methods because of their superior performance over unsupervised learning methods. Nevertheless, supervised learning approaches have several limitations. The performance of supervised learning models is directly conditioned by the size and quality of the training corpus, whose collection is often tedious, time-consuming and sometimes even unaffordable. Though, most importantly, supervised learning can only learn tasks for which humans already know the solution. Because of their learning scheme, supervised approaches are only able to solve problems with well-known defined outputs. Unsupervised learning, on the contrary, is capable of recognizing patterns within the data in completely unexplored and unknown tasks. Regarding the image segmentation problem, unsupervised learning is able to delineate regions within the image with common \ac{MRI} properties, typically representing the same tissue or physiological area. However, due to the unguided blind learning approach of unsupervised methods, these generally do not reach as accurate results as those of a supervised learning model in well-known tasks.

In this sense, the aim of this chapter is to demonstrate that unsupervised learning can achieve competitive results comparable to those obtained by supervised learning methods in a well-known task. The purpose is to ratify if unsupervised learning is capable to detect consistent patterns in \ac{MRI}, discriminant enough to solve a segmentation task. For that purpose, we performed a comparative study of several unsupervised learning algorithms in the task of automated high grade glioma segmentation. A postprocessing stage was also developed to automatically map each label of an unsupervised segmentations to a specific tissue of the brain. The comparative was performed with the public reference \ac{BRATS} 2013 Test and Leaderboard datasets.

\medskip

\emph{The contents of this chapter were published in the publications \citep{JuanAlbarracin2015a, JuanAlbarracin2015b}---thesis contributions C1, P1 and P2.}

\section{Introduction}
\label{section:comparative_unsupervised_learning_intro}
Medical imaging plays an indisputable key role in the diagnosis and management of brain tumors. The intracranial location of these lesions and the unspecificity of clinical symptoms make medical imaging a necessary tool to diagnose and monitor the evolution of these diseases \citep{Wen2010}. Specifically, the early identification and delineation of the different tissues composing the tumor becomes crucial to take decisions that can improve the patient survival. However, the manual segmentation of these tissues constitutes a complex, time-consuming and biased task, which caught the attention of the 	\ac{ML} community \citep{Bauer2013}. Particularly, glioblastoma tumor has received most of this attention, as it is the most common and aggressive malignant tumor of the central nervous system \citep{Dolecek2012, Deimling2009}. Glioblastomas are heterogeneous lesions that present different areas of active tumor, necrosis and edema, all of them exhibiting a high variability related to the aggressiveness of the tumor. Hence, the automated segmentation of these lesions becomes a desired solution from the clinical standpoint and an interesting challenge to address from the \ac{ML} community.

Extensive reviews of brain tumor segmentation have been presented by \cite{Wadhwa2019, Saman2019, Gordillo2013, Bauer2013}. Most of these techniques fall into the supervised learning approach. \cite{Verma2008} and  \cite{Ruan2011} employed \acp{SVM} to segment healthy and pathological tissues, and additionally subcompartiments inside these areas. \cite{Jensen2009} used neural networks to detect brain tumor invasion. \cite{Lee2008} used a combination of \ac{CRF} and \ac{SVM} to perform tumor segmentation. \cite{Bauer2011} also used \ac{SVM} and Hierarchical \ac{CRF} to segment both healthy and tumor tissues including sub-compartments. Recently, \ac{RF} techniques have shown high success in the supervised brain tumor segmentation task. \cite{Meier2013, Festa2013, Reza2013, Tustison2013} proposed several approaches based on \ac{RF} variants for the \ac{BRATS} challenge of \ac{MICCAI} 2013 Conference, reaching the first positions in the competition. Nowadays, with the advent of novel deep learning techniques, the current state-of-the art is mostly dominated by \ac{CNN} classifiers. \acp{CNN} are a class of deep feed-forward neural networks whose architecture is particularly well suited for computer vision recognition tasks. \acp{CNN} have outperformed most of algorithms in many medical image segmentation problems, arising as the winner technique in most challenges such as \ac{BRATS}, ISLES or PROMISE12 challenges. \cite{BRATS2015, BRATS2016, BRATS2017, BRATS2018} summarize the most relevant contributions to the \ac{BRATS} challenge from 2015 to 2018, demonstrating the superiority of \acp{CNN} in the glioma segmentation task.

However, supervised learning performance depends directly on the quality and size of the training dataset, which often requires an expensive, time-consuming and biased task to collect \citep{Gordillo2013}. Moreover, changes in \ac{MRI} protocols, acquisition routines or clinical environments may distort the data and hence affect the performance of the supervised models \citep{Duda2000}. However, the major drawback of supervised learning is that it is only capable to address tasks that humans have already solved before. The mimic nature of supervised learning limits the paradigm to the set of problems with known solution, which, without underestimating its unquestionable usefulness, reduces the \ac{ML} to an instrument for automatizing well-known tasks.

On the contrary, unsupervised learning address these limitations in a more straightforward manner. Unsupervised learning does not require a training corpus from which to learn a model to solve the task, but it analyzes unlabeled data searching for hidden patterns and inner relationships that describes their latent structure \citep{Wittek2014}. Therefore, besides solving already known tasks, this approach has the innate ability to discover new knowledge from the data by exploring the arrangement of the information. Sticking to the glioblastoma segmentation task, unsupervised learning typically fits a customized segmentation model to the patient's \ac{MRI}, describing their own imaging patterns. This makes it possible to characterize the lesion in each case just by discriminating its own patterns and not by reinterpreting them under the knowledge drawn from other examples. By the opposite, the absence of manual segmentations to guide the learning process makes segmentation more challenging and often lead to a worse performance with respect to supervised approaches.

Some attempts for brain tissue segmentation have been made under the unsupervised paradigm. \cite{Fletcher2001} proposed an approach based on fuzzy clustering and domain knowledge for multi-parametric non-enhancing tumor segmentation. Domain knowledge and parenchymal tissue detection was based on heuristics related to geometric shapes and lesion locations, which may not be robust when high deformation is presented. Moreover, several assumptions such as prior knowledge about the number of existing tumor foci or the minimum required thickness of the \ac{MRI} slices introduced significant limitations to the method. \cite{Nie2009} used Gaussian clustering with a spatial accuracy-weighted \ac{HMRF} that allowed them to deal with images at different resolutions without interpolation. Nowadays, advanced reconstruction techniques such as super-resolution enables to work in a high resolution voxel space by reconstructing the low resolution images without interpolation. Moreover, no automated method was provided to differentiate between tumor labels and normal tissue labels after the unsupervised segmentation ends. \cite{Zhu2012} developed a software based on the segmentation approach proposed by \cite{Zhang2001}, which performs an \ac{GMM} clustering combined whit \acp{HMRF}. Zhu et al extended Zhang's approach through a sequence of additionally morphological and thresholding operations to refine the segmentation. Such operations are not fully specified and only overall commented, so the reproducibility of their results is not possible. \cite{Vijayakumar2007} proposed a method based on \acp{SOM} to segment tumor, necrosis, cysts, edema and normal tissues using \ac{MRI}. Although the learning process of \acp{SOM} was performed in an unsupervised manner, the dataset from which to infer the net structure was determined manually, similar than in a supervised approach. In their work, 700 pattern observations, corresponding to 7 different tissues, where selected manually, hence converting the process in a supervised task. \cite{Prastawa2003} proposed a similar approach than the followed in this study. They performed an unsupervised classification based on \acp{FMM} and also used a brain atlas to characterize the normal tissue labels. However, they made important simplifying assumptions to allow them to use the atlas without registration. Moreover, they also simplify the segmentation task in 2 labels (tumor and edema), ignoring other important tissues such as necrosis or non-enhancing tumor. \cite{Doyle2013} also proposed an approach based on \ac{GMM} clustering and \ac{MRF} priors. They defined different penalizations in the \ac{MRF} depending on the adjacency of the labels to prevent incoherent segmentations, however they did not clearly specify how they related the glioblastoma tissues with these labels before running the unsupervised segmentation. Furthermore, all the unsupervised approaches described above applied their algorithms on its own datasets, making difficult a general comparison of the methods. 

In this work, we propose a fully automated pipeline for unsupervised glioblastoma segmentation. Our contributions concern the assessment of the performance of several unsupervised segmentation methods, including both structured and non-structured classification algorithms, on a real public and reference dataset; and we also provide a generalized method to automatically identify pathological labels in an unsupervised segmentation that represent abnormal tissues in the brain. Our aim is to demonstrate that unsupervised segmentation algorithms can achieve competitive results, comparable to supervised approaches, by detecting imaging patterns that describe the tissue's \ac{MRI} profiles. 

We evaluated our unsupervised segmentation method using the public \ac{BRATS} 2013 Leaderboard and Test datasets provided for the International Image Segmentation Challenge of \ac{MICCAI} Conference. The proposed method with the \ac{GMM} algorithm improves the results obtained by most of the supervised approaches evaluated with the Leaderboard \ac{BRATS} 2013 set, reaching the 2nd position in the rank. Our variant using the Gauss-\ac{HMRF} improves the results obtained by the best unsupervised segmentation methods evaluated with the \ac{BRATS} 2013 Test set, and also reaches the 7th position in the general Test rank, mainly against supervised approaches.

\section{Materials}
\label{section:comparative_unsupervised_learning_materials}
In order to make our results comparable, we have used the public multi-modal \ac{BRATS} dataset 2013 \citep{Menze2015}, provided for the international \ac{BRATS} 2013 challenge in image segmentation of \ac{MICCAI} conference. We have evaluated our method with the Test set and the Leaderboard set, and we have made a comparison between our proposed method and the best algorithms that participated in the challenge.

The \ac{BRATS} 2013 Test set consists of multi-contrast \ac{MR} scans of 10 high-grade glioma patients without the manual expert labeling. The Leaderboard set consists of $11+10$ multi-contrast \ac{MR} scans of high-grade glioma patients, also without the manual expert labeling. The first 11 Leaderboard patients come from to the Test set of \ac{BRATS} 2012 Challenge, while the next 10 cases refer to the new Leaderboard cases for 2013 Challenge.

For each patient of the datasets, \Ti{}-weighted, \Tii{}-weighted, \ac{FLAIR} and post-gadolinium \Tic{}-weighted \ac{MR} images were provided. All images were linearly co-registered to the post-gadolinium \Tic{}-weighted sequence, skull stripped, and interpolated to 1 mm\textsuperscript{3} isotropic resolution. No inter-patient registration was made to put all the images in a common reference space.

Expert manual annotations of the \ac{MRI} studies were performed, considering five possible labels for the lesion:

\begin{description}[itemsep=0pt]
\item[Label 0:] background, brain and everything else not corresponding to labels 1, 2, 3 and 4.
\item[Label 1:] non-brain, non-tumor, necrosis, cyst and hemorrhage.
\item[Label 2:] surrounding edema.
\item[Label 3:] non-enhancing tumor.
\item[Label 4:] enhancing tumor.
\end{description}

An evaluation web page was provided to assess the quality of the segmentations, computing different metrics such as Dice coefficient, positive predictive value, sensitivity and Kappa indices. Furthermore, the evaluation was made over different sub-compartments of the lesion, to properly measure the performance of the different segmentation methods. The set of labels composing each sub-compartment are described in section \ref{subsection:comparative_unsupervised_learning_Evaluation}. 

\section{Methods}
\label{section:comparative_unsupervised_learning_methods}
This section describes the proposed pipeline for the automated unsupervised segmentation of glioblastoma, including a generalized postprocessing designed to identify the pathological classes of an unsupervised segmentation that represent abnormal tissues in the brain.

\subsection{\acs{MRI} preprocessing}
\label{subsection:comparative_unsupervised_learning_MRI_preprocessing}
\ac{MRI} preprocessing is an active field of research that attempts to enhance and correct \ac{MR} images for their posterior analysis. In an unsupervised segmentation approach there is no reference nor manual labeling from which to learn tumor tissue models. Therefore, common artifacts such as noise, \ac{NMR} inhomogeneities or registration missalignments can introduce undesired patterns into images that could led to erroneous classifications, describing \ac{MRI} artifacts rather than physiological tissues. This clearly increases the importance of an accurate and effective preprocess of the \ac{MRI} if an unsupervised segmentation is going to be performed. We propose the following scheme for the \ac{BRATS} 2013 data: 1) Denoising, 2) Skull stripping, 3) Bias field correction and 4) Super-resolution.

\subsubsection{Denoising}
\label{subsubsection:comparative_unsupervised_learning_denoising}
Denoising is a standard \ac{MRI} preprocessing task that aims to reduce or ideally remove the noise from an \ac{MR} image. Although \ac{MRI} noise has been usually modeled as Gaussian distributed, by definition \ac{MRI} noise follows a Rician distribution \citep{Gudbjartsson1995}. \cite{Diaz2011} presented a comprehensive analysis of different denoising methods, discussing their weaknesses and strengths. Recent filters such as the \ac{NLM} introduced by \cite{Buades2005a} has improved the existing techniques for \ac{MR} data. Based on this approach, \cite{Manjon2010a} introduced a variant of the \ac{NLM} filter, called Adaptive-\ac{NLM} filter, which does not assume an uniform distribution of the noise over the image, thereby adapting the strength of the filter depending on a local estimation of the noise. The filter also deals with both correlated Gaussian and Rician noise. We used the Adaptive-\ac{NLM}filter to remove the noise from the \ac{BRATS} images.

\subsubsection{Brain extraction}
\label{subsubsection:comparative_unsupervised_learning_brain_extraction}
\begin{figure}[h]
\centering
\includegraphics[width=0.90\linewidth]{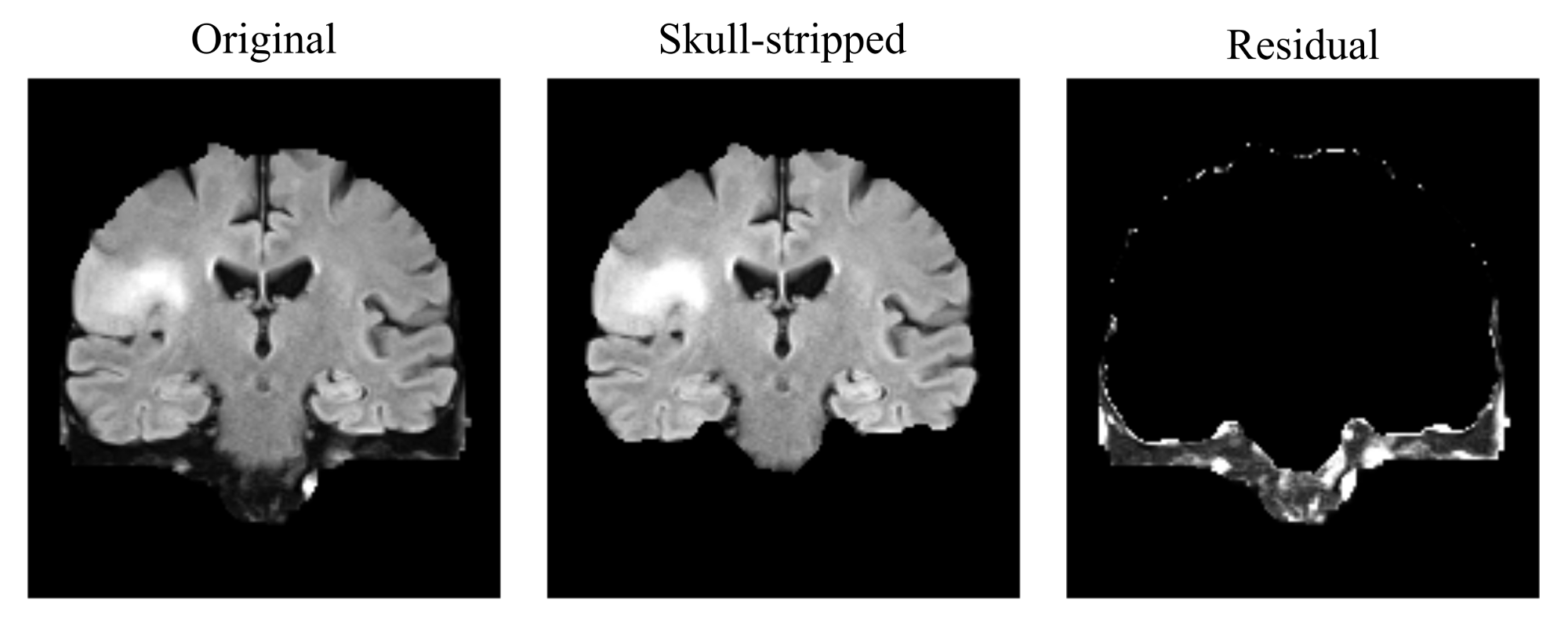}
\caption{Example of the proposed skull stripping. From left to right column: original \ac{BRATS} 2013 patient image, resultant image after the proposed skull stripping and the remaining residual.}
\label{figure:comparative_unsupervised_learning_skull_stripping}
\end{figure}

Brain extraction, also called skull-stripping, comprises the process of removing skull, extra-meningeal and non-brain tissues from the \ac{MRI} sequences. Although \ac{BRATS} 2013 dataset is already skull stripped, we detected several cases that include partial areas of the cranium and extra-meningeal tissues. In order to improve the preprocessing of the data, we recomputed the skull stripping masks for all patients using the \href{http://brainsuite.org/}{Brain Suite Software}, removing the non desired tissues of the \ac{MR} images. Figure \ref{figure:comparative_unsupervised_learning_skull_stripping} shows an example of a patient of the \ac{BRATS} 2013 dataset with the original skull stripping, the resultant image after our new skull stripping and the remaining residual.

\subsubsection{Bias field correction}
\label{subsubsection:comparative_unsupervised_learning_bias_correction}
Intensity inhomogeneity is another common artifact present in \ac{MRI} acquisitions. Magnetic field inhomogeneities are unavoidable effects consisting on low frequency signals that corrupt the images and affect their intensity levels. Typically, automated segmentation approaches based on \ac{MRI} are built upon the assumption that tissues have the same distribution of intensity across the image. Therefore, intensity inhomogeneities must be corrected to ensure a correct coherent segmentation. The popular non-parametric non-uniform intensity normalization N3 algorithm was proposed in 1998 by \cite{Sled1998}, becoming a reference technique for bias field correcting. \cite{Tustison2010} proposed in 2010 a new implementation of N3, called N4, which improves the N3 algorithm with a better B-spline fitting function and a hierarchical optimization scheme for the bias field correction. N4 was used in our study to correct \ac{MRI} inhomogeneities.

\subsubsection{Super-resolution}
\label{subsubsection:comparative_unsupervised_learning_super_resolution}
In a brain tumor lesion protocol, several \ac{MR} sequences are commonly acquired normally at different resolutions, thereby introducing spatial inconsistencies when a multi-modal \ac{MR} study is performed. In these cases, an upsampling or interpolation is needed to set a common voxel space for all sequences. Classic interpolation such as linear, cubic or splines interpolation could rise as a solution, but at the cost of introducing artifacts such as partial volume effects or stair-case artifacts. In contrast, more powerful and sophisticated methods such as super resolution could improve classic interpolation by reconstructing the low resolution images recovering its high frequency components. Several super resolution schemes for \ac{MRI} are available in the literature \citep{Plenge2013, Manjon2010b, Rousseau2010, Protter2009}.

\begin{figure}[h]
\centering
\includegraphics[width=\linewidth]{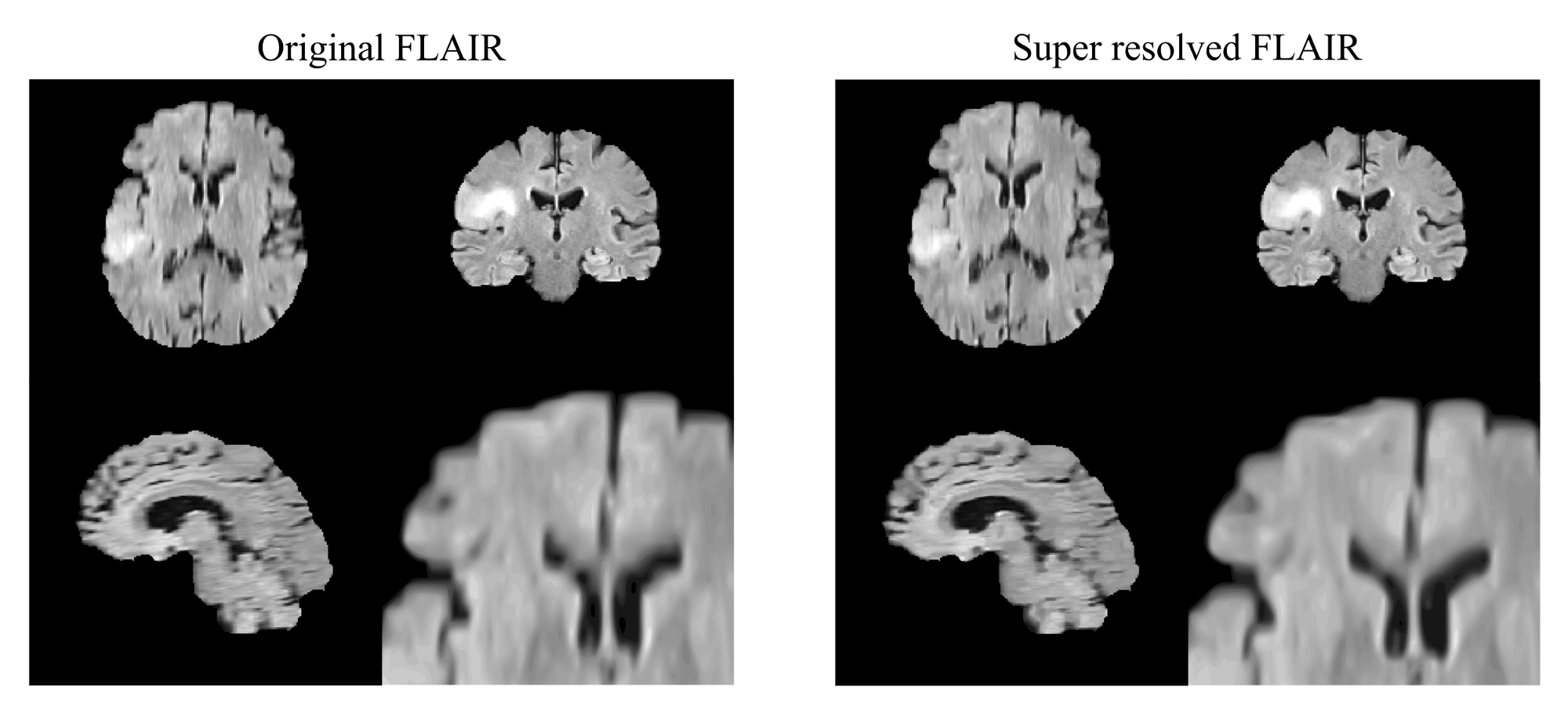}
\caption{Example of super resolution using Non-local Upsampling method of a \ac{FLAIR} sequence of the BRATS 2013 dataset.}
\label{figure:comparative_unsupervised_learning_super_resolution}
\end{figure}

\ac{BRATS} 2013 dataset comes with a 1mm\textsuperscript{3} isotropic voxel size resolution achieved through classic interpolation. In order to improve the resolution of these images, we employed the super resolution algorithm proposed by \cite{Manjon2010c}, which exploits the self-similarity present in \ac{MR} images through a patch-based non-local reconstruction process. Such method iteratively reconstructs a high resolution image by applying a \ac{NLM} filter with different strengths, aimed to increase image regularity while constraining intensity ranges so that they be coherent among scales through a local back-projection approach. Figure \ref{figure:comparative_unsupervised_learning_super_resolution} shows an example of a super resolved \ac{FLAIR} sequence of a patient of the BRATS 2013 dataset with a detailed zoom of an axial slice.

\subsection{Feature Extraction and Dimensionality Reduction}
\label{subsection:comparative_unsupervised_learning_fe_dr}
Feature extraction comprises the process of obtaining new features from the \ac{MR} images to improve discrimination between different labels. Although \ac{MRI} intensities are the most common used features to differentiate between the brain tissues, it has been shown that including texture features in combination with \ac{MR} intensities increases the performance of the segmentation algorithms \citep{Kassner2010, Ahmed2011}. In this sense, we have implemented the first order statistical texture features, also called histogram derived metrics or first order central moments.

For each patient we initially obtained an additional image, named \Tid{}, which consists on the absolute difference between the \Tic{} and the \Ti{} images \citep{Prastawa2003}. This image highlights the contrast enhanced areas of the patient, such as the active tumor, helping in their discrimination. Next, for each \ac{MR} image of the patient (\Ti{}, \Tic{}, \Tii{}, \ac{FLAIR} and \Tid{}), we computed their first order texture features. Such features consist on the computation of local histograms in 3D patches centered at each voxel of the image, and calculate the mean, skewness and kurtosis of these histograms. We used local 3D patches of $5 \times 5 \times 5$ voxels.

Thus, a set $X$ of 20 images is obtained for each patient, consisting on the following images:
$$ X = \left( T_1, T_{1CE}, T_2, FLAIR, T_{1Diff}, \mu T_1, ..., \gamma T_1, ..., \kappa T_{1Diff} \right) $$
where $\mu$, $\gamma$, and $\kappa$ prefixes refers to the mean, skewness and kurtosis features of the corresponding image.

In order to reduce the complexity and number of parameters to estimate in the models, a dimensionality reduction process was performed. Dimensionality reduction seeks for an efficient representation of the original high dimensional data into a lower dimensional space, retaining or increasing the most relevant information. In our study, we used \ac{PCA} for dimensionality reduction. We run \ac{PCA} on the $X$ set and selected the principal components, which together explained at least the 99\% of the variance of the data, reducing in most cases from 20 dimensions to 5 dimensions. These images make up the final stack of imaging data used for the posterior unsupervised segmentations.

\begin{figure}[h]
\centering
\includegraphics[width=\linewidth]{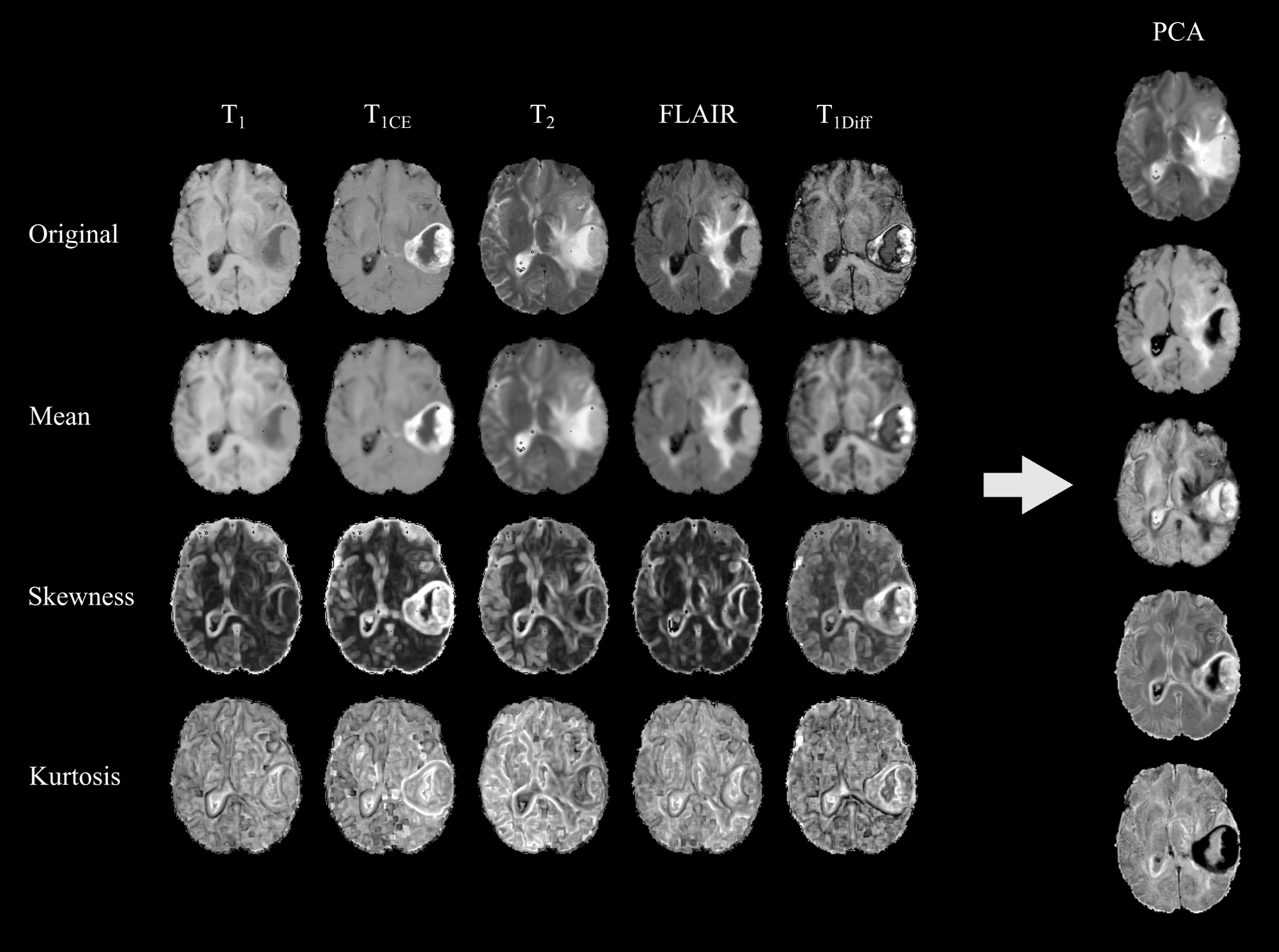}
\caption{Example of feature extraction and dimensionality reduction of a \ac{MRI} study of a patient of the \ac{BRATS} 2013 dataset.}
\label{figure:comparative_unsupervised_learning_feature_extraction}
\end{figure}

An slice example of the feature extraction and \ac{PCA} dimensionality reduction process of a \ac{MR} study is shown in Figure \ref{figure:comparative_unsupervised_learning_feature_extraction}.

\subsection{Unsupervised voxel classification}
\label{subsection:comparative_unsupervised_learning_segmentation}
The \ac{BRATS} 2013 dataset comprises 5 labels to be segmented, which in some cases a single label encloses different brain tissues (for example 0 or 1 label). This intra-class heterogeneity can severely affect the performance of unsupervised methods, since heterogeneity is naturally explained in unsupervised learning through the definition of different clusters. Hence, the same \emph{semantic class} in a problem can be modeled through a set of different clusters.

In this sense, in order to increase the expression capacity of the unsupervised models to capture such heterogeneity, we assumed that each tissue can be initially represented by 2 clusters. That is, we will assume that there exist 7 tissues in the brain, which are labels 1, 2, 3 and 4 proposed in \ac{BRATS} 2013 Challenge plus \ac{GM}, \ac{WM} and \ac{CSF}; each one of them with initially 2 clusters to represent their heterogeneity. Note that, therefore, we will estimate an unsupervised clustering of 14 classes throughout the brain, but we do not require each tissue to use exactly 2 classes. Hence, a tissue showing a high degree of heterogeneity can be modeled using a set of 3 or 4 clusters, while an homogeneous tissue can be segmented by a single cluster. Such assumption provided us a balance between the number of parameters to estimate to the models and the degrees of freedom required to explain the heterogeneity of the tissues.

We evaluated the most popular unsupervised classification algorithms. We divided the algorithm comparison in two groups: structured and non-structured methods. Non-structured algorithms classify data assuming an \ac{i.i.d.} condition between the voxels of the images. Structured prediction covers the range of algorithms that involve the classification of data with a specific structure, such as an image. Under the non-structured paradigm, we evaluated three methods: K-means, Fuzzy K-means and \ac{GMM} clustering. In the structured prediction case we evaluated the Gauss-\ac{HMRF} as the archetype of unsupervised structured learning models.

Let $ X = \left\lbrace \vec{x}^1, \vec{x}^2, \dots, \vec{x}^N \right\rbrace$ the set of voxels to be classified, where $ \vec{x}^i \in \Reals^D $ represents a feature vector of $D$ dimensions for the $i^{th}$ voxel. Let $Y = \left\lbrace y^1, y^2, \dots, y^N \right\rbrace$ the set of labels associated to each voxel, where $y^i \in \left\lbrace 1, \dots, K \right\rbrace$.

\subsubsection{K-means}
\label{subsubsection:comparative_unsupervised_learning_Kmeans}
K-means \citep{Lloyd1982, Macqueen1967} is an unsupervised non-structured iterative partitional clustering based on a distance minimization criterion. Its aim is to divide the data space $X$ into $K$ clusters $\Pi = \left\lbrace \Pi_1, \Pi_2, \dots, \Pi_K \right\rbrace $, so that each observation of $X$ belongs to the cluster with nearest centroid. The distance criterion minimized by K-means is:

\begin{equation}
\argmin_\Pi \sum_{j=1}^K \sum_{\vec{x}^i \in \Pi_j} \parallel \vec{x}^i - \vec{\mu}_j \parallel^2
\end{equation}

\noindent where $\vec{\mu}_j$ represents the centroid of the $j^{th}$ cluster.

\medskip

The K-means algorithm performs as follow:

\begin{siderules}
\begin{description}
\item[Initialization:] Make an initial guess about the cluster centroids $\vec{\mu}_j$
\item[Until] there are no changes in centroids $\vec{\mu}_j$ (or in clusters $\Pi$)
\begin{description}
\item[Assign] $y^i$ to the cluster with nearest centroid $\vec{\mu}_j$:
\begin{equation*}
y^i = \argmin_j \parallel \vec{x}^i - \vec{\mu}_j \parallel^2
\end{equation*}
\item[Update] cluster centroids $\vec{\mu}_j$
\begin{equation*}
\vec{\mu}_j = \frac{1}{\left| \Pi_j \right|} \sum_{\vec{x}^i \in \Pi_j} \vec{x}^i
\end{equation*}
\end{description}
\end{description}
\end{siderules}

From a statistical point of view, the iterative distance minimization criterion followed by K-means is equivalent to find the most likelihood parameters of a \ac{GMM} \citep{Duda2000}, assuming shared identity covariance matrices and uniform prior probabilities for all classes. The iterative approach followed by K-means is also demonstrated a special limit of the \ac{EM} algorithm \citep{Dempster1977, Bishop1995}, called \emph{Hard-\ac{EM}}, where each observation is uniquely assigned to a class with posterior probability equal to 1.

\subsubsection{Fuzzy K-means}
\label{subsubsection:comparative_unsupervised_learning_Fuzzy}
Likewise K-means, Fuzzy K-means \citep{Dunn1973, Bezdek1981} is a non-structured iterative partitional clustering base on a distance minimization criterion. Under a probabilistic paradigm, it is also equivalent to a \ac{GMM} assuming shared identity covariance matrices and uniform prior probabilities for all classes. However, Fuzzy K-means differs from K-means in which the assignment of an observation to a cluster is not \emph{hard} but \emph{fuzzy}. This means that each observation keeps a degree of membership to each cluster (equivalent to the posterior probability of a \ac{GMM}) rather than a unique assignment of the observation to a class with posterior probability equal to 1.

In the same manner as K-means, Fuzzy K-means aims to divide the data space $X$ into $K$ clusters $\Pi = \left\lbrace \Pi_1, \Pi_2, \dots, \Pi_K \right\rbrace $, but it also keeps a vector $\vec{u}^i$ for each observation that determines the degree of membership of the $i^{th}$ observation to the different clusters. The distance minimization criterion followed by Fuzzy K-means is:

\begin{equation}
\argmin_\Pi \sum_{j=1}^K \sum_{i=1}^N u_j^i \parallel \vec{x}^i - \vec{\mu}_j \parallel^2
\end{equation}

\noindent where $\vec{\mu}_j$ represents the centroid of the $j^{th}$ cluster.

\medskip 

The membership variable $u_j^i$ is typically defined as:

\begin{equation}
u_j^i = \frac{1}{\sum_{k=1}^K \left( \frac{\parallel \vec{x}^i - \vec{\mu}_j \parallel^2}{\parallel \vec{x}^i - \vec{\mu}_k \parallel ^2} \right)^{\frac{2}{m-1}}}
\end{equation} 

\noindent where $m$ controls the degree of fuzziness of the $j^{th}$ cluster, with $1 \leq m < \infty$ and typically set to 2 in absence of domain knowledge.

\medskip

The Fuzzy K-means algorithm performs as follow:

\begin{siderules}
\begin{description}
\item[Initialization:] Make an initial guess about the cluster centroids $\vec{\mu}_j$
\item[Until] there are no changes in $u_j^i$ greater than $\epsilon$
\begin{description}
\item[Estimate] the membership coefficients $u_j^i$ for each observation and cluster:
\begin{equation*}
u_j^i = \frac{1}{\sum_{k=1}^K \left( \frac{\parallel \vec{x}^i - \vec{\mu}_j \parallel^2}{\parallel \vec{x}^i - \vec{\mu}_k \parallel ^2} \right)^{\frac{2}{m-1}}}
\end{equation*}
\item[Update] cluster centroids $\vec{\mu}_j$ as the expected value of the observations given $u_j^i$:
\begin{equation*}
\vec{\mu}_j = \frac{\sum_{i=1}^N u_j^i \vec{x}^i}{\sum_{i=1}^N u_j^i}
\end{equation*}
\end{description}
\end{description}
\end{siderules}

\subsubsection{Gaussian Mixture Model}
\label{subsubsection:comparative_unsupervised_learning_GMM}
A \ac{GMM} is probabilistic model that model a random variable as a convex combination of Gaussian \acp{pdf}. Thus, it provides a statistical framework to describe the heterogeneity of a dataset trough a weighted sum of single distributions, each one representing a sub-population within a data. An in-depth explanation of \acp{FMM} and particularly \acp{GMM} is described in section \ref{section:rationale_mixture_models}. Please refer to this section for more details.

As a short remainder, the \ac{GMM} is defined as:

\begin{equation}
p \left( X; \Theta \right) = \prod\limits_{i=1}^N \sum\limits_{j=1}^K \pi_j \mathcal{N} \left( \vec{x}^i ; \vec{\mu}_j, \Sigma_j \right)
\end{equation}

\noindent where $\Theta = \left\lbrace \vec{\mu}_1, \ldots, \vec{\mu}_K, \Sigma_1, \ldots, \Sigma_K, \pi_1, \ldots, \pi_K \right\rbrace$ are the parameters of the model. The parameters $\left\lbrace \pi_1,\ldots, \pi_K \right\rbrace $ are typically called \emph{mixing coefficients} and can be seen as the prior probability of each component of the mixture describing the data.

Estimation on \acp{GMM} is performed via \ac{EM} algorithm \citep{Dempster1977} because direct \ac{MLE} estimate does not yield closed-form solutions for the parameters of the model. For a complete derivation of the \ac{MLE} estimate and the \ac{EM} procedure in \acp{FMM} please refer to section \ref{section:rationale_mixture_models}. The algorithm for \ac{GMM} clustering performs as follow:

\begin{siderules}
\begin{description}
\item[Initialization:] Choose an initial setting for $\Theta^{\left( 0 \right)}$.
\item[Expectation step:] Estimate $p \left( Z|X; \Theta^{\left( t \right)} \right)$
\begin{equation*}
p \left( z_j^i = 1 | \vec{x}^i; \Theta^{\left( t \right)} \right) = {\gamma_j^i}^{\left( t \right)} = \frac{\pi_j^{\left( t \right)} \mathcal{N} \left( \vec{x}^i; \vec{\mu}_j^{\left( t \right)}, \Sigma_j^{\left( t \right)} \right)}{\sum\limits_k \pi_k^{\left( t \right)} \mathcal{N} \left( \vec{x}^i; \vec{\mu}_k^{\left( t \right)}, \Sigma_k^{\left( t \right)} \right)}
\end{equation*}
\item[Maximization step:] Update the parameters of the model given $p \left( Z|X; \Theta^{\left( t \right)} \right)$
\begin{equation*}
\hat{\Theta}_{MLE}^{\left( t+1 \right)} = \argmax_\Theta \mathbb{E}_{p\left(Z|X; \Theta^{\left( t \right)} \right)} \log \mathcal{L}\left( \Theta; X,Z \right)
\end{equation*}
\noindent where sticking to \acp{GMM}:
\begin{equation*}
\begin{split}
\pi_j^{\left( t+1 \right)} &= \frac{\sum\limits_{i=1}^N {\gamma_j^i}^{\left( t \right)}}{\sum\limits_{i=1}^N \sum\limits_{k=1}^K {\gamma_k^i}^{\left( t \right)}} \\
\vec{\mu}_j^{\left( t+1 \right)} &= \frac{1}{\sum\limits_{i=1}^N {\gamma_j^i}^{\left( t \right)}} \sum\limits_{i=1}^N {\gamma_j^i}^{\left( t \right)} \vec{x}^i \\
\Sigma_j^{\left( t+1 \right)} &= \frac{1}{\sum\limits_{i=1}^N {\gamma_j^i}^{\left( t \right)}} \sum\limits_{i=1}^N {\gamma_j^i}^{\left( t \right)} \left( \vec{x}^i - \vec{\mu}_ j^{\left( t+1 \right)} \right) \left( \vec{x}^i - \vec{\mu}_ j^{\left( t+1 \right)} \right)^T
\end{split}
\end{equation*}
\item[Convergence:] Stop if $\mathcal{L} \left( \Theta^{\left( t+1 \right)} ; X \right) - \mathcal{L} \left( \Theta^{\left( t \right)} ; X \right) \leq \epsilon$; otherwise $t = t + 1$ and go to \textbf{Expectation step}.
\end{description}
\end{siderules}

\medskip
 
As stated above, \ac{GMM} clustering can be seen as the generalization of K-means and Fuzzy K-means algorithms, where the hard constraints related to the shared covariance matrices and the uniform prior probabilities are dropped.

\subsubsection{Gauss-Hidden Markov Random Field}
\label{subsubsection:comparative_unsupervised_learning_GHMRF}
\acp{MRF} are probabilistic undirected graphical models that define a family of joint probability distributions by means of an undirected graph \citep{Hammersley1971}. These graphs are used to introduce conditional dependencies between random variables of the model, which in the case of the image segmentation task, allows capturing the self-similarity and local redundancy present in the images. These dependencies are explicitly denoted via an undirected and cyclic graph, whose vertices represent the pixels/voxels of the images and whose edges represent the dependencies between them. In this sense, a generative model with a given set of parameters $\Theta$ incorporating a \ac{HMRF} can be defined as:

\begin{equation}
\begin{split}
p\left(X; \Theta \right) &= \sum_Y p\left( X, Y; \Theta \right) \\
                         &= \sum_Y p\left( Y \right) p\left( X | Y; \Theta \right)
\end{split}
\end{equation}

\acp{MRF} are usually used to model the prior density $p\left( Y \right)$ of a probabilistic generative model. According to the Hammersley–Clifford theorem, assuming that the prior density is strictly positive, a \ac{MRF} can be defined in terms of local energy functions and therefore it can be expressed as a Gibbs measure in the form:

\begin{equation}
p \left( Y \right) = \frac{1}{Z}\exp\left( -U\left(Y\right)\right)
\end{equation} 

\noindent where $U\left( Y \right)$ is an energy function that defines the conditional dependencies between the random variables via the graphical model, and $Z$ is called the \emph{partition function} and acts as a normalizer to ensure the density to sum 1:

\begin{equation}
Z = \sum_{Y^\prime} \exp \left( -U\left(Y^\prime \right)\right)
\end{equation}

Hammersey-Clifford theorem also states that $p\left( Y \right)$ can be factorized over the \emph{cliques} of the undirected graphical model. A clique is defined as a subset of vertices in the graph such that there exist an edge between all pairs of vertices in the subset. Let $\mathcal{Q}$ the set of all cliques of the graph, the energy function $U\left( Y \right)$ is defined as:

\begin{equation}
U\left( Y \right) = \sum_{q \in \mathcal{Q}} \Psi\left(Y\left(q\right)\right)
\end{equation}

Nowadays, if complexity is considered, the inference algorithms for \acp{MRF} are only able to optimize undirected graphs with cliques of order 2 (pairwise cliques), i.e. $q \rightarrow \left( y^n, y^m \right)\quad\forall q \in \mathcal{Q}$. Hence, the most widely used graphical model is the \emph{Ising model}. The Ising model defines a graph lattice with as many vertices as pixels/voxels exist in the image, where conditional dependencies of each variable are expressed in terms of its orthogonal adjacent neighborhood. The clique factorization for the Ising model is then performed in the form:

\begin{equation}
\begin{split}
U\left( Y \right) &= \sum_{\left(y^n, y^m \right) \in Q} \Psi(y^n, y^m) \\
                  &= \sum_{\left(y^n, y^m \right) \in Q} \beta^{\left(n,m\right)} \delta(y^n, y^m)
\end{split}
\end{equation}

\noindent where $\beta^{\left(n,m\right)}$ is a weight defined for the corresponding clique, and $\delta$ is a function that measures the dissimilarity between the clique. Typically, in absence of domain knowledge, $\beta^{\left(n,m\right)}$ is usually set to 1 and $\delta$ is defined as:

\begin{equation}
\delta(y^n, y^m) = \begin{cases} 0 & \mbox{if } y^n = y^m \\ 1, & \mbox{otherwise} \end{cases}
\end{equation}

\medskip

Typically, the class conditional density $p\left(X | Y; \Theta\right)$ is also expressed in terms of Gibbs measures to take advantage from \ac{MRF} solvers, so it is usually rewritten as:

\begin{equation}
p(X | Y; \Theta) = \frac{1}{Z} \exp \left( -U\left(X | Y ; \Theta \right) \right)
\end{equation}

\noindent where $U\left(X | Y ; \Theta\right)$ can be assumed \ac{i.i.d.} and is proportional to the class conditional $p\left(X | Y; \Theta\right)$ density, and $Z$ is again a partition function to ensure the distribution to sum 1. Under a Gaussian model, $U\left( X | Y; \Theta \right)$ can be defined as:

\begin{equation}
U\left( X | Y; \Theta \right) = \sum_{i=0}^N U\left( \vec{x}^i | y^i = j ; \Theta \right) = \sum_{i=0}^N -\log \mathcal{N}\left(\vec{x}^i ; \mu_j, \Sigma_j \right)
\end{equation}

\medskip

As a results, the complete structured model is defined as:

\begin{equation}
p\left(X ; \Theta \right) =  \sum_Y \frac{1}{Z} \exp \left(-U\left(X | Y; \Theta \right) - U\left(Y\right) \right)
\end{equation}

\medskip

Exact inference on this model is intractable due to the sum over all possible set of labels in $Z$, which becomes a $\# P-complete$ problem. However, approximate efficient algorithms to compute the best labeling $\hat{Y}$ are proposed in the literature. Iterated Conditional Modes \citep{Birchfield1998}, Monte Carlo Sampling \citep{Geyer1992}, Loopy Belief Propagation \citep{Yedidia2003}, Mean Field Approximation \citep{Parisi1998} or Graph cuts \citep{Boykov2001} are some of the available algorithms to solve pairwise \acp{MRF}-based models. In our study we used the \emph{FastPD} algorithm proposed by \cite{Komodakis2007, Komodakis2008}, which has been demonstrated to be superior to most of the aforementioned algorithms. This algorithm is based on a combination of Graph cuts with primal-dual strategies. Therefore the best labeling is computed as:

\begin{equation}
\hat{Y} = \argmax -U\left(X | Y; \Theta \right) - U\left(Y\right)
\end{equation}

In this sense, we can think of $\hat{Y}$ as a binary one-hot encoding variable holding the posterior probability of each label for each pixel/voxel, with $\hat{y}_j^i = 1$ for the winner class and $\hat{y}_k^i = 0,\quad k \neq j$ for the remaining classes.

Therefore, likewise \ac{GMM}, Gauss-\ac{HMRF} also finds the \ac{MLE} parameters of a mixture of Gaussian distributions that better fits the data, but imposing the structured \ac{MRF} prior. A Hard-\ac{EM} version (given that exact inference is not possible) of the \ac{EM} algorithm is employed to estimate the parameters of the model in the form:

\begin{siderules}
\begin{description}
\item[Initialization:] Choose an initial setting for $\Theta^{\left( 0 \right)}$.
\item[Expectation step:] Estimate $\hat{Y}$ by using \emph{FastPD} (or similar) algorithm
\begin{equation*}
\begin{split}
U\left( Y \right) &= \sum_{\left(y^n, y^m \right) \in Q} \beta^{\left(n,m\right)} \delta(y^n, y^m) \\
U\left( X | Y; \Theta^{\left(t\right)} \right) &= \sum_{i=0}^N -\log \mathcal{N}\left(\vec{x}^i ; \mu_j^{\left(t\right)}, \Sigma_j^{\left(t\right)} \right) \\
\hat{Y} &= \argmax -U\left(X | Y; \Theta^{\left(t\right)} \right) - U\left(Y\right)
\end{split}
\end{equation*}
\item[Maximization step:] Update the parameters of the model given $\hat{Y}$
\begin{equation*}
\hat{\Theta}_{MLE}^{\left( t+1 \right)} = \argmax_\Theta \mathbb{E}_{\hat{Y}} \log \mathcal{L}\left( \Theta; X,Z \right)
\end{equation*}
\noindent where sticking to \acp{GMM}:
\begin{equation*}
\begin{split}
\pi_j^{\left( t+1 \right)} &= \frac{\sum\limits_{i=1}^N {\hat{y}_j^i}}{\sum\limits_{i=1}^N \sum\limits_{k=1}^K {\hat{y}_k^i}} \\
\vec{\mu}_j^{\left( t+1 \right)} &= \frac{1}{\sum\limits_{i=1}^N {\hat{y}_j^i}} \sum\limits_{i=1}^N {\hat{y}_j^i} \vec{x}^i \\
\Sigma_j^{\left( t+1 \right)} &= \frac{1}{\sum\limits_{i=1}^N {\hat{y}_j^i}} \sum\limits_{i=1}^N {\hat{y}_j^i} \left( \vec{x}^i - \vec{\mu}_ j^{\left( t+1 \right)} \right) \left( \vec{x}^i - \vec{\mu}_ j^{\left( t+1 \right)} \right)^T
\end{split}
\end{equation*}
\item[Convergence:] Stop if $\mathcal{L} \left( \Theta^{\left( t+1 \right)} ; X \right) - \mathcal{L} \left( \Theta^{\left( t \right)} ; X \right) \leq \epsilon$; otherwise $t = t + 1$ and go to \textbf{Expectation step}.
\end{description}
\end{siderules}

\subsubsection{Algorithms initialization}
\label{subsubsection:comparative_unsupervised_learning_initialization}
A well-known requirement of unsupervised learning is the good initial seeding. Although the global minima is not usually reached even if a good initialization is provided, a bad initialization can lead the model to a very sub-optimal local minimum, thereby providing a poor segmentation. Several strategies such as multiple replications or intelligent initial seeding are proposed to palliate this effect. In our study, we implemented the \emph{K-means++} algorithm \citep{Arthur2007}, which provides an initialization that attempts to avoid local sub-optimal minimums.

We propose the following procedure to ensure a competitive unsupervised segmentation: First, generate 100 different initializations using \emph{K-means++} algorithm. Next, automatically select the 10 most promising initializations by minimizing the average intra-cluster sums of point-to-centroid distances of the initializations. Finally, run each unsupervised segmentation algorithm with the 10 most promising initializations and choose the best solution considering the following criteria:

\begin{description}[itemsep=0pt]
\item[\textbf{K-means:}] choose the solution with lowest intra-cluster sums of point-to-centroid distances.
\item[\textbf{Fuzzy K-means:}] choose the solution with lowest intra-cluster sums of point-to-centroid distances.
\item[\textbf{\ac{GMM}:}] choose the solution with highest log-Likelihood value.
\item[\textbf{Gauss-\ac{HMRF}:}] choose the solution with highest log-Likelihood value.
\end{description}

\subsection{Automated pathological label identification}
\label{subsection:comparative_unsupervised_learning_label_identification}
Unsupervised segmentation produces a partitioning of the data space into several classes, each one without \emph{semantic} sense. In other words, in the brain tumor unsupervised approach, labels do not directly identify a specific tissue but only distinguish between \ac{MRI} data different enough from each other to be considered equal. Moreover, labelings among different patient segmentations may not always represent the same tissue, complicating its biological interpretation. Hence, an automated pathological label identification is mandatory to provide a powerful and competitive unsupervised segmentation method. We propose the following method to automatically isolate pathological labels:

\begin{enumerate}[itemsep=0pt]
\item Identify and remove \ac{WM}, \ac{GM} and \ac{CSF} labels.
\item Remove outlier and partial volume labels.
\item Merge labels by statistical distribution similarities.
\end{enumerate}

\subsubsection{Identify and remove WM, GM and CSF labels}
\label{subsubsection:comparative_unsupervised_learning_WMGMCSFRemoval}
Under the \ac{ICBM} project, an unbiased standard \ac{MR} brain atlas was provided by the McConnell Brain Imaging Centre in 2009 \citep{Fonov2011, Fonov2009}. The \ac{ICBM} atlas include a \Ti{}, \Tii{} and Proton density \ac{MR} images, with the associated \ac{WM}, \ac{GM} and \ac{CSF} tissue probability maps. Such tissue probability maps indicate the probability for each voxel $v$ of the brain to belong to a normal tissue $ T = \left\lbrace WM, GM, CSF \right\rbrace$, with:

\begin{equation}
\sum_{t \in T} p(t | v) = 1
\end{equation}

\begin{figure}[h]
\centering
\includegraphics[width=0.90\linewidth]{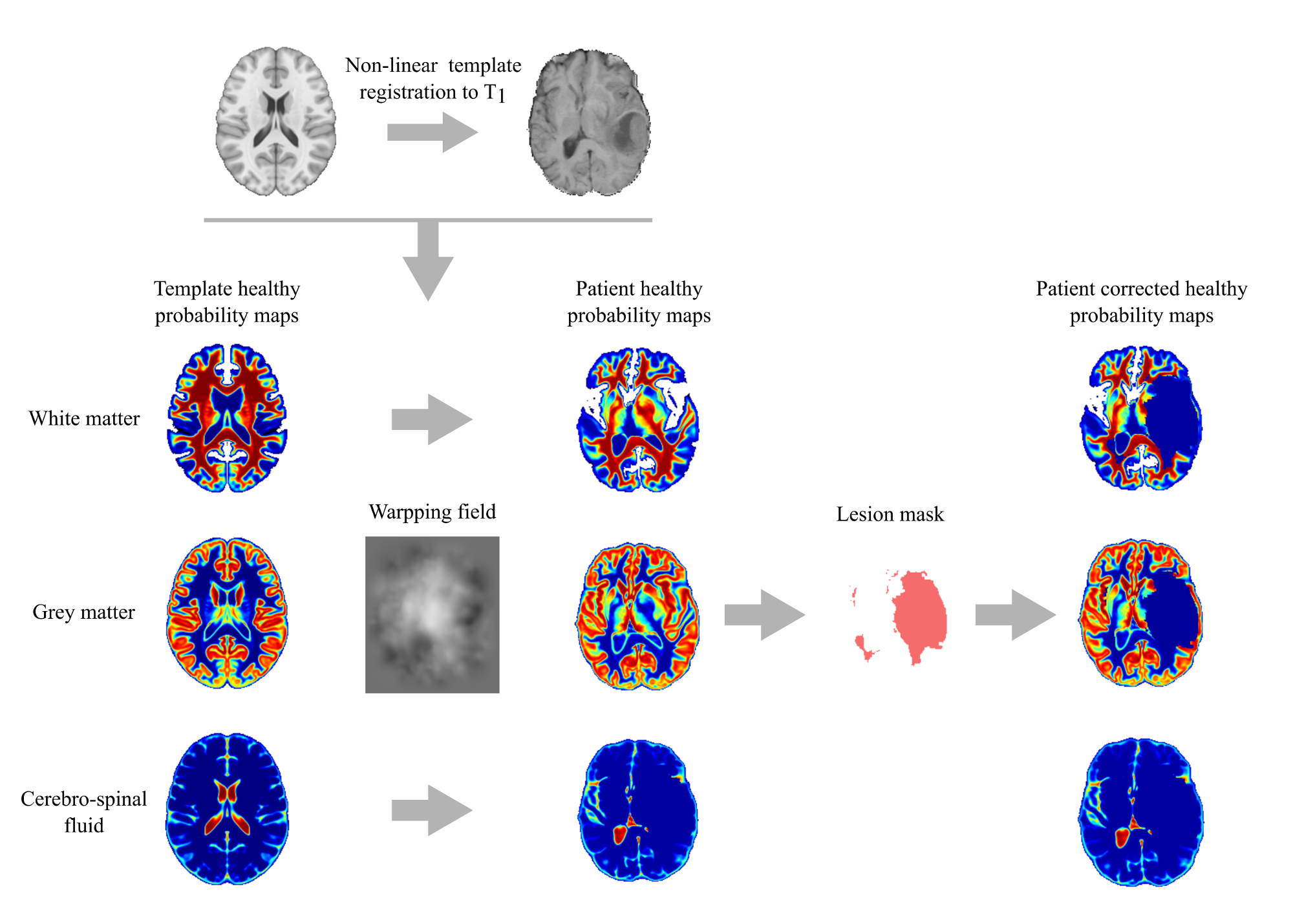}
\caption{Illustration of the algorithm to compute the patient healthy-tissue probability maps and lesion area correction.}
\label{figure:comparative_unsupervised_learning_tissue_maps_correction}
\end{figure}

In our study we used these tissue probability maps to detect which labels of a segmentation explain the \ac{WM}, \ac{GM} and \ac{CSF} tissues. However, taking into account that the \ac{ICBM} template represents a healthy brain, it is necessary to corrected these maps by setting to zero (or a smaller $\epsilon$ value) the probability of any voxel $v$ in the area of the lesion. Therefore, we first performed a non-linear registration of the \ac{ICBM} \Ti{} template to the \Ti{} sequence of the patient and applied the warp transformation to the tissue probability maps. Following the study conducted by \cite{Klein2009}, we used the SyN algorithm \citep{Avants2008} implemented in the ANTS software with cross-correlation similarity metric. Once we obtained the patient aligned healthy-tissue probability maps, a roughly approximate mask of the lesion of each patient was computed to correct the probability maps. The typical delineation of the lesion performed by expert radiologists is usually based on the hyper-intensity areas in the \Tii{} and \Tic{} sequences \citep{Bauer2013}. Following a similar criteria, we computed an approximate mask of the lesion by selecting the voxels whose value in the \ac{FLAIR} and \Tic{} images were higher than the median plus the standard deviation of the corresponding image. Next, we automatically filled the holes of the computed masks and removed the voxels that fell in the perimeter of the brain (typically showing hyper-intensities due to cranial traces). Finally, we corrected the healthy-tissue probability maps of each patient by setting an $\epsilon$ probability in the area determined by their corresponding lesion masks. Figure \ref{figure:comparative_unsupervised_learning_tissue_maps_correction} shows an illustration of the algorithm to compute the corrected tissue probability maps for a patient.

\medskip

Based on these patient healthy-tissue probability maps, we identified which classes of an unsupervised segmentation mostly represent a normal tissue. Let $S$ a segmentation obtained through any unsupervised method, and let $v_l = \left\lbrace v : S\left(v\right) = l \right\rbrace$ the set of voxels $v$ of $S$ classified with label $l$. Let $t$ a normal tissue where $t \in T$. To perform the pathological tissue identification we computed the following probability mass:

\begin{equation}
p(l|t,S) = \frac{\sum_{u \in v_l} p(t|u)}{\sum_v p(t|v)}
\end{equation}

Simplifying, the $p(c|t,S)$ determines how much of the normal tissue $t$ is explained by the label $l$ in the segmentation $S$. 

Therefore, for a given tissue $t$, and based on these probabilities, we constructed two vectors: one with the $p(c|t,S)$ values sorted in descending order, denoted as $\mathcal{P}_t$, and the other with the corresponding label codes sorted in the same manner, denoted as $\mathcal{K}_t$. 

\begin{equation}
\begin{split}
\mathcal{K}_t &= \left\lbrace l : p(l|t,S) \geq p(l^{\prime}|t,S) \right\rbrace \\
\mathcal{P}_t &= \left\lbrace p(l|t,S) : p(l|t,S) \geq p(l^{\prime}|t,S) \right\rbrace
\end{split}
\end{equation}

Then, we computed the cumulative sum of $\mathcal{P}_t$, denoted as $\mathcal{S}_t$, and finally choose the set of labels of $\mathcal{K}_t$ whose $\mathcal{S}_t$ value exceed a threshold $\tau$.

\begin{equation}
\mathcal{Z}_t = \left\lbrace \mathcal{K}_t \left( j \right) : \mathcal{S}_t \left( j \right) > \tau, \quad 1 \leq j < K \right\rbrace
\end{equation} 

The $\mathcal{Z}_t$ set contains the classes of $S$ that have a very low probability of explain the normal tissue $t$. Hence, we repeated the same procedure for each normal tissue $t$ and computed the intersection between the $\mathcal{Z}_t$ sets to finally isolate the labels that do not explain any normal tissue, i.e. the pathological classes:

\begin{equation}
\mathcal{Z} = \mathcal{Z}_{WM} \cap \mathcal{Z}_{GM} \cap \mathcal{Z}_{CSF}
\end{equation}

\medskip

Given that our aim is to evaluate the performance of each unsupervised segmentation algorithm, all of them in the same conditions, we do not carried out any particular optimization of the $\tau$ threshold for each algorithm. Instead, we fixed a general threshold for all the methods, set to $\tau = 0.8$, as a reasonably, high confidence and compatible value to perform the pathological labels identification. Note that $\tau = 1$ is not possible since this implies that all the classes of the segmentation will be required to explain only a single normal tissue.

\subsubsection{Remove outlier and partial volume labels}
\label{subsubsection:comparative_unsupervised_learning_OutlierClassesRemoval}
The above process of identifying and removing the healthy-tissue labels may leave some spurious labels that must be deleted. We found that these labels frequently appear in the perimeter of the brain or in a very low proportion compared to the other labels of the segmentation. The labels located at the perimeter of the brain typically correspond to remaining traces of cranium, intensity artifacts between brain and background or partial volume effects that super resolution cannot resolve. The low proportion labels often match to outlier voxels in terms of abnormal intensity values, usually produced by artifacts in the \ac{MR} acquisition.

In order to remove the perimeter unwanted labels, we deleted all the connected components of all the remaining labels of the segmentation, which overlapped more than the 50\% of its area with a binary dilated mask of the perimeter of the brain. To remove the smaller low proportion labels we removed those ones with a prevalence less than the 1\% the remaining segmentation after the above processes.

\subsubsection{Merge labels by statistical distribution similarities}
\label{subsubsection:comparative_unsupervised_learning_MergeSimilarClasses}
The heterogeneity of the pathological labels led us to assume that each tissue was initially modeled by two clusters. However, this was a general but not strict assumption, i.e. we did not enforce to use exactly two cluster per tissue. Instead, we estimated a general clustering of 14 components for each case. Hence, a tissue may have been represented by two or more labels in the segmentation or, conversely, by a single label depending on its homogeneity. Therefore, it was mandatory to design a mechanism to find which labels were explaining the same semantic concept, i.e. the same pathological tissue.

Based on the work proposed by \cite{Saez2017}, we analyzed the \ac{MRI} intensity distributions of the remaining labels to find potential clusters representing the similar information. To do so, we estimated the \ac{pdf} of each label through a kernel smoothing density estimation, and used the Jensen-Shannon divergence to measure the pairwise distances among them. Therefore, we constructed a pairwise matrix of statistical distribution distances and used \ac{HAC} with average link (\ac{UPGMA}) to merge the similar labels.

\begin{figure}[h]
\centering
\includegraphics[width=0.90\linewidth]{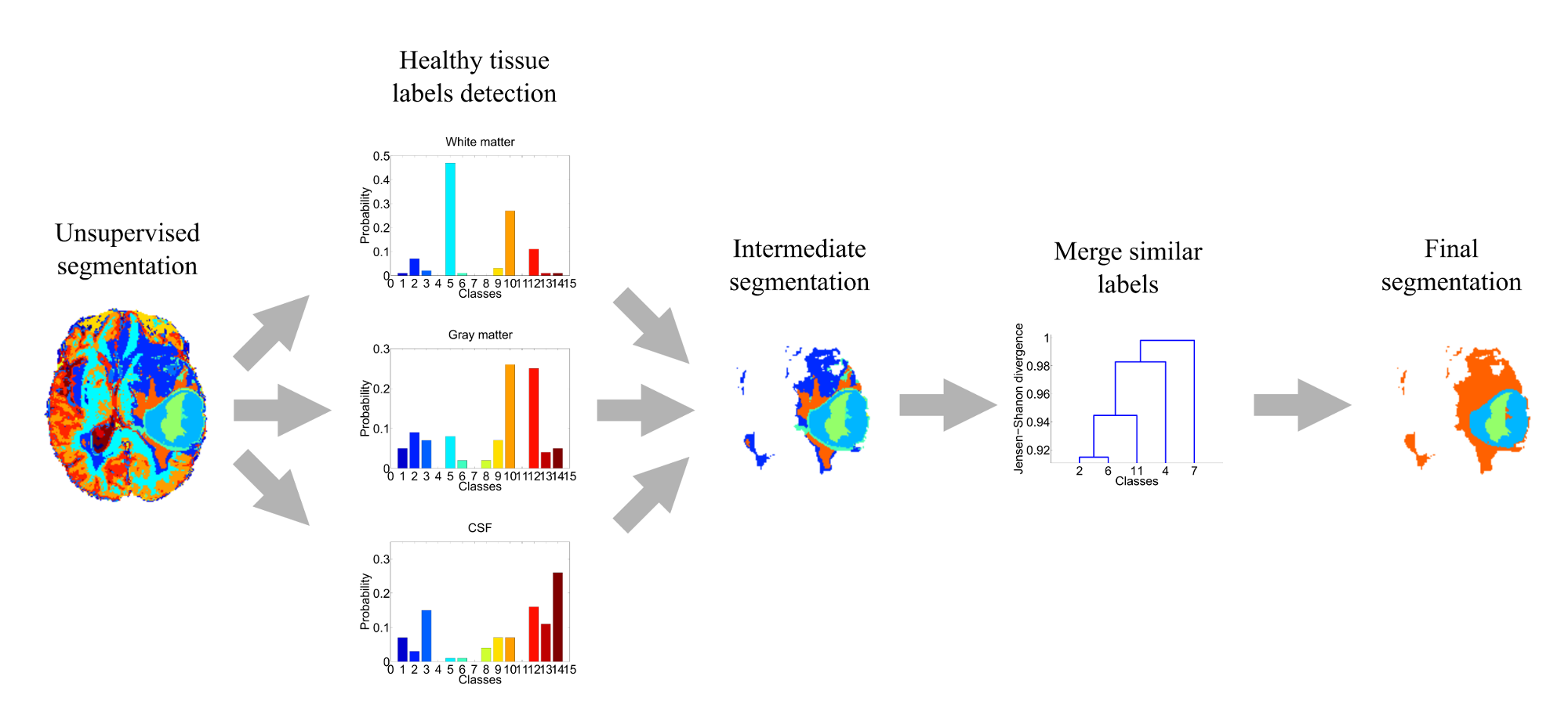}
\caption{Automated pathological label identification process.}
\label{figure:comparative_unsupervised_learning_pathological_classes_identification}
\end{figure}

Due to the \ac{BRATS} 2013 labeling considers 4 pathological labels to be segmented, we enforced the clustering to return a maximum of 4 classes. Note that the method is able to return less than 4 classes if the \ac{HAC} finds enough similarities between them, however, in any other case the method is enforced to return a maximum of 4 labels. Figure \ref{figure:comparative_unsupervised_learning_pathological_classes_identification} shows and example of the pathological labels isolation procedure.

\subsection{Evaluation}
\label{subsection:comparative_unsupervised_learning_Evaluation}
We evaluated our unsupervised brain tumor segmentation framework with the \ac{BRATS} 2013 Leaderboard and Test datasets. Segmentations provided by the different unsupervised methods in combination with the proposed preprocessing and postprocessing pipelines were sent to the \ac{BRATS} evaluation web page. The figures of merit provided to assess the quality of the segmentations were:

\begin{itemize}[itemsep=0.5pt]
\item Dice: $ \frac{2(TP + TN)}{P+N+\hat{P}+\hat{N}} $
\item PPV: $ \frac{TP}{TP + FP} $
\item Sensitivity: $ \frac{TP}{TP + FN} $
\item Kappa: $ \frac{P_A - P_E}{1 - P_E} $
\end{itemize}

\noindent where $TP$ refers to the true positives in the segmentation, $TN$ to the true negatives, $FP$ to the false positives, $FN$ to the false negatives, $P$ to the real positives of the ground truth, $N$ to the real negatives of the ground truth, $\hat{P}$ to the estimated positives of the proposed segmentation, $\hat{N}$ to the estimated negatives of the proposed segmentation, $P_A$ to the accuracy of the segmentation and $P_E$ to a term that measures the probability of success by chance, defined as: $P_E = \left( \frac{P}{P + N} \cdot \frac{\hat{P}}{\hat{P} + \hat{N}} \right) + \left( \frac{N}{P + N} \cdot \frac{\hat{N}}{\hat{P} + \hat{N}} \right)$.

Furthermore, three different sub-compartments of the lesion were evaluated to properly assess the quality of the segmentation methods. Table \ref{table:comparative_unsupervised_learning_subcompartments} describes the labels involved in each sub-compartment considered in the evaluation.

\begin{table}[h]
\caption{Labels composing each sub-compartment evaluated in the \ac{BRATS} 2013 challenge.}
\centering
\rowcolors{2}{gray!10}{white}
\begin{tabular}{lccccc}
	\hline
	\rowcolor{gray!25}
	\cellcolor{gray!25} & Label 0 & Label 1 & Label 2 & Label 3 & Label 4 \\
	\cellcolor{gray!25}\ac{WT} & & \ding{53} & \ding{53} & \ding{53} & \ding{53} \\
	\cellcolor{gray!25}\ac{TC} & & \ding{53} & & \ding{53} & \ding{53} \\
	\cellcolor{gray!25}\ac{ET} & & & & & \ding{53} \\ \hline
\end{tabular}
\label{table:comparative_unsupervised_learning_subcompartments}
\end{table}

\section{Results}
\label{section:comparative_unsupervised_learning_Results}
Tables \ref{table:comparative_unsupervised_learning_results_test} and \ref{table:comparative_unsupervised_learning_results_leaderboard} show the results obtained in the Test and Leaderboard datasets respectively, grouped by the unsupervised algorithms tested in this study.

\begin{table}[h]
\centering
\caption{Summary of average results obtained by the different unsupervised algorithms in combination with the proposed preprocess and postprocess over the \ac{BRATS} 2013 Test set.}	
\resizebox{\textwidth}{!}{
\begin{tabular}{lccccccccccccccc}
	\hline
	\cellcolor{gray!25} & \multicolumn{3}{c}{\cellcolor{gray!25}Dice} & \multicolumn{3}{c}{\cellcolor{gray!25}PPV} & \multicolumn{3}{c}{\cellcolor{gray!25}Sensitiviy} & \cellcolor{gray!25} \\
	\multirow{-2}{*}{\cellcolor{gray!25}Classifier} & \cellcolor{gray!25}\ac{WT} & \cellcolor{gray!25}\ac{TC} & \cellcolor{gray!25}\ac{ET} & \cellcolor{gray!25}\ac{WT} & \cellcolor{gray!25}\ac{TC} & \cellcolor{gray!25}\ac{ET} & \cellcolor{gray!25}\ac{WT} & \cellcolor{gray!25}\ac{TC} & \cellcolor{gray!25}\ac{ET} & \multirow{-2}{*}{\cellcolor{gray!25}Kappa} \\
	\cellcolor{gray!25}K-means       & \cellcolor{white}0.69 & \cellcolor{white}0.49 & \cellcolor{white}0.57 & \cellcolor{white}0.66 & \cellcolor{white}0.48 & \cellcolor{white}0.68 & \cellcolor{white}0.76 & \cellcolor{white}0.57 & \cellcolor{white}0.51 & \cellcolor{white}0.98  \\
	\cellcolor{gray!25}Fuzzy K-means & \cellcolor{gray!10}0.70 & \cellcolor{gray!10}0.46 & \cellcolor{gray!10}0.39 & \cellcolor{gray!10}0.73 & \cellcolor{gray!10}0.47 & \cellcolor{gray!10}0.51 & \cellcolor{gray!10}0.71 & \cellcolor{gray!10}0.54 & \cellcolor{gray!10}0.35 & \cellcolor{gray!10}0.98  \\
	\cellcolor{gray!25}GMM           & \cellcolor{white}0.69 & \cellcolor{white}0.60 & \cellcolor{white}0.55 & \cellcolor{white}0.63 & \cellcolor{white}0.60 & \cellcolor{white}0.64 & \cellcolor{white}0.78 & \cellcolor{white}0.68 & \cellcolor{white}0.55 & \cellcolor{white}0.98  \\
	\cellcolor{gray!25}Gauss-HMRF    & \cellcolor{gray!10}0.72 & \cellcolor{gray!10}0.62 & \cellcolor{gray!10}0.59 & \cellcolor{gray!10}0.68 & \cellcolor{gray!10}0.58 & \cellcolor{gray!10}0.67 & \cellcolor{gray!10}0.81 & \cellcolor{gray!10}0.75 & \cellcolor{gray!10}0.60 & \cellcolor{gray!10}0.98  \\ \hline
\end{tabular}
}
\label{table:comparative_unsupervised_learning_results_test}
\end{table}

\begin{table}[h]
\caption{Summary of average results obtained by the different unsupervised algorithms in combination with the proposed preprocess and postprocess over the \ac{BRATS} 2013 Leaderboard set.}
\resizebox{\textwidth}{!}{
\begin{tabular}{lccccccccccccccc}
	\hline
	\cellcolor{gray!25} & \multicolumn{3}{c}{\cellcolor{gray!25}Dice} & \multicolumn{3}{c}{\cellcolor{gray!25}PPV} & \multicolumn{3}{c}{\cellcolor{gray!25}Sensitiviy} & \cellcolor{gray!25} \\
	\multirow{-2}{*}{\cellcolor{gray!25}Classifier} & \cellcolor{gray!25}\ac{WT} & \cellcolor{gray!25}\ac{TC} & \cellcolor{gray!25}\ac{ET} & \cellcolor{gray!25}\ac{WT} & \cellcolor{gray!25}\ac{TC} & \cellcolor{gray!25}\ac{ET} & \cellcolor{gray!25}\ac{WT} & \cellcolor{gray!25}\ac{TC} & \cellcolor{gray!25}\ac{ET} & \multirow{-2}{*}{\cellcolor{gray!25}Kappa} \\
	\cellcolor{gray!25}K-means       & \cellcolor{white}0.76 & \cellcolor{white}0.49 & \cellcolor{white}0.53 & \cellcolor{white}0.75 & \cellcolor{white}0.44 & \cellcolor{white}0.66 & \cellcolor{white}0.82 & \cellcolor{white}0.56 & \cellcolor{white}0.48 & \cellcolor{white}0.99 \\
	\cellcolor{gray!25}Fuzzy K-means & \cellcolor{gray!10}0.77 & \cellcolor{gray!10}0.46 & \cellcolor{gray!10}0.25 & \cellcolor{gray!10}0.81 & \cellcolor{gray!10}0.46 & \cellcolor{gray!10}0.27 & \cellcolor{gray!10}0.77 & \cellcolor{gray!10}0.51 & \cellcolor{gray!10}0.27 & \cellcolor{gray!10}0.99  \\
	\cellcolor{gray!25}GMM           & \cellcolor{white}0.74 & \cellcolor{white}0.59 & \cellcolor{white}0.60 & \cellcolor{white}0.71 & \cellcolor{white}0.55 & \cellcolor{white}0.60 & \cellcolor{white}0.81 & \cellcolor{white}0.71 & \cellcolor{white}0.66 & \cellcolor{white}0.99 \\
	\cellcolor{gray!25}Gauss-HMRF    & \cellcolor{gray!10}0.77 & \cellcolor{gray!10}0.63 & \cellcolor{gray!10}0.32 & \cellcolor{gray!10}0.72 & \cellcolor{gray!10}0.61 & \cellcolor{gray!10}0.33 & \cellcolor{gray!10}0.84 & \cellcolor{gray!10}0.71 & \cellcolor{gray!10}0.50 & \cellcolor{gray!10}0.99  \\ \hline
\end{tabular}
}
\label{table:comparative_unsupervised_learning_results_leaderboard}
\end{table}

As expected, Gauss-\ac{HMRF} and \ac{GMM} demonstrate their superiority with respect the other algorithms. Almost all the metrics reveal that both algorithms obtain the best results in all the sub-compartments segmentations. Only the enhancing tumor sub-compartment in the Leaderboard set yielded worse results for the Gauss\ac{HMRF} compared to the results obtained in the other sub-compartments and datasets. Such effect typically occurs because of the smoothing prior of the Gauss-\ac{HMRF}, which is later discussed in the Discussion section.

Tables \ref{table:comparative_unsupervised_learning_brats_test} and \ref{table:comparative_unsupervised_learning_brats_leaderboard} show the published ranking of the \ac{BRATS} competition grouped by the learning paradigm adopted by each method and the metrics and sub-compartments evaluated in the Challenge. As shown in Table \ref{table:comparative_unsupervised_learning_brats_test}, we achieved the $1^{st}$ position in the ranking of the unsupervised methods of the Test set, and the $7^{th}$ position in the general ranking, mostly against supervised approaches. Table \ref{table:comparative_unsupervised_learning_brats_leaderboard} shows the Leaderboard ranking and the results achieved by our method. The proposed approach in combination with the \ac{GMM} algorithm reaches the $2^{nd}$ position of the Leaderboard ranking, improving the results obtained by many supervised methods, mainly in the enhancing tumor sub-compartment.

\begin{table}[h]
\caption{Ranking of the \emph{BRATS} 2013 Test set and the position occupied by our proposed unsupervised segmentation framework with the Gauss-\ac{HMRF} algorithm.}
\resizebox{\textwidth}{!}{
\begin{tabular}{clccccccccccccccc}
	\hline
	\cellcolor{gray!25} & \cellcolor{gray!25} & \multicolumn{3}{c}{\cellcolor{gray!25}Dice} & \multicolumn{3}{c}{\cellcolor{gray!25}PPV} & \multicolumn{3}{c}{\cellcolor{gray!25}Sensitiviy} & \cellcolor{gray!25} \\
	\cellcolor{gray!25} & \multirow{-2}{*}{\cellcolor{gray!25}User} & \cellcolor{gray!25}\ac{WT} & \cellcolor{gray!25}\ac{TC} & \cellcolor{gray!25}\ac{ET} & \cellcolor{gray!25}\ac{WT} & \cellcolor{gray!25}\ac{TC} & \cellcolor{gray!25}\ac{ET} & \cellcolor{gray!25}\ac{WT} & \cellcolor{gray!25}\ac{TC} & \cellcolor{gray!25}\ac{ET} & \multirow{-2}{*}{\cellcolor{gray!25}Kappa} \\
	\cellcolor{gray!25} & \cellcolor{white}Nick Tustison & \cellcolor{white}0.87 & \cellcolor{white}0.78 & \cellcolor{white}0.74 & \cellcolor{white}0.85 & \cellcolor{white}0.74 & \cellcolor{white}0.69 & \cellcolor{white}0.89 & \cellcolor{white}0.88 & \cellcolor{white}0.83 & \cellcolor{white}0.99  \\
	\cellcolor{gray!25} & \cellcolor{gray!10}Raphael Meier & \cellcolor{gray!10}0.82 & \cellcolor{gray!10}0.73 & \cellcolor{gray!10}0.69 & \cellcolor{gray!10}0.76 & \cellcolor{gray!10}0.78 & \cellcolor{gray!10}0.71 & \cellcolor{gray!10}0.92 & \cellcolor{gray!10}0.72 & \cellcolor{gray!10}0.73 & \cellcolor{gray!10}0.99  \\
	\cellcolor{gray!25}Supervised & \cellcolor{white}Syed Reza & \cellcolor{white}0.83 & \cellcolor{white}0.72 & \cellcolor{white}0.72 & \cellcolor{white}0.82 & \cellcolor{white}0.81 & \cellcolor{white}0.70 & \cellcolor{white}0.86 & \cellcolor{white}0.69 & \cellcolor{white}0.76 & \cellcolor{white}0.99  \\
	\cellcolor{gray!25}methods & \cellcolor{gray!10}Liang Zhao & \cellcolor{gray!10}0.84 & \cellcolor{gray!10}0.70 & \cellcolor{gray!10}0.65 & \cellcolor{gray!10}0.80 & \cellcolor{gray!10}0.67 & \cellcolor{gray!10}0.65 & \cellcolor{gray!10}0.89 & \cellcolor{gray!10}0.79 & \cellcolor{gray!10}0.70 & \cellcolor{gray!10}0.99  \\
	\cellcolor{gray!25} & \cellcolor{white}Nicolas Cordier & \cellcolor{white}0.84 & \cellcolor{white}0.68 & \cellcolor{white}0.65 & \cellcolor{white}0.88 & \cellcolor{white}0.63 & \cellcolor{white}0.68 & \cellcolor{white}0.81 & \cellcolor{white}0.82 & \cellcolor{white}0.66 & \cellcolor{white}0.99  \\
	\cellcolor{gray!25} & \cellcolor{gray!10}Joana Festa & \cellcolor{gray!10}0.72 & \cellcolor{gray!10}0.66 & \cellcolor{gray!10}0.67 & \cellcolor{gray!10}0.77 & \cellcolor{gray!10}0.77 & \cellcolor{gray!10}0.70 & \cellcolor{gray!10}0.72 & \cellcolor{gray!10}0.60 & \cellcolor{gray!10}0.70 & \cellcolor{gray!10}0.98  \\
	\cellcolor{gray!25} Unsupervised & \cellcolor{white}\textbf{This work} & \cellcolor{white}\textbf{0.72} & \cellcolor{white}\textbf{0.62} & \cellcolor{white}\textbf{0.59} & \cellcolor{white}\textbf{0.68} & \cellcolor{white}\textbf{0.58} & \cellcolor{white}\textbf{0.67} & \cellcolor{white}\textbf{0.81} & \cellcolor{white}\textbf{0.75} & \cellcolor{white}\textbf{0.60} & \cellcolor{white}\textbf{0.98}  \\
	\cellcolor{gray!25} methods & \cellcolor{gray!10}Senan Doyle & \cellcolor{gray!10}0.71 & \cellcolor{gray!10}0.46 & \cellcolor{gray!10}0.52 & \cellcolor{gray!10}0.66 & \cellcolor{gray!10}0.38 & \cellcolor{gray!10}0.58 & \cellcolor{gray!10}0.87 & \cellcolor{gray!10}0.70 & \cellcolor{gray!10}0.55 & \cellcolor{gray!10}10.98  \\ \hline
\end{tabular}
}
\label{table:comparative_unsupervised_learning_brats_test}
\end{table}

\begin{table}[h]
\caption{Ranking of the \emph{BRATS} 2013 Leaderboard set and the position occupied by our proposed unsupervised segmentation framework with the \ac{GMM} algorithm.}
\resizebox{\textwidth}{!}{
\begin{tabular}{clccccccccccccccc}
	\hline
	\cellcolor{gray!25} & \cellcolor{gray!25} & \multicolumn{3}{c}{\cellcolor{gray!25}Dice} & \multicolumn{3}{c}{\cellcolor{gray!25}PPV} & \multicolumn{3}{c}{\cellcolor{gray!25}Sensitiviy} & \cellcolor{gray!25} \\
	\cellcolor{gray!25} & \multirow{-2}{*}{\cellcolor{gray!25}User} & \cellcolor{gray!25}\ac{WT} & \cellcolor{gray!25}\ac{TC} & \cellcolor{gray!25}\ac{ET} & \cellcolor{gray!25}\ac{WT} & \cellcolor{gray!25}\ac{TC} & \cellcolor{gray!25}\ac{ET} & \cellcolor{gray!25}\ac{WT} & \cellcolor{gray!25}\ac{TC} & \cellcolor{gray!25}\ac{ET} & \multirow{-2}{*}{\cellcolor{gray!25}Kappa}  \\
	\cellcolor{gray!25}Supervised & \cellcolor{white}Nick Tustison & \cellcolor{white}0.79 & \cellcolor{white}0.65 & \cellcolor{white}0.53 & \cellcolor{white}0.83 & \cellcolor{white}0.70 & \cellcolor{white}0.51 & \cellcolor{white}0.81 & \cellcolor{white}0.73 & \cellcolor{white}0.66 & \cellcolor{white}0.99  \\
	\cellcolor{gray!25}method & \cellcolor{white} & \cellcolor{white} & \cellcolor{white} & \cellcolor{white} & \cellcolor{white} & \cellcolor{white} & \cellcolor{white} & \cellcolor{white} & \cellcolor{white} & \cellcolor{white} & \cellcolor{white} \\
	\cellcolor{gray!25}Unsupervised & \cellcolor{gray!10}\textbf{This work} & \cellcolor{gray!10}\textbf{0.74} & \cellcolor{gray!10}\textbf{0.59} & \cellcolor{gray!10}\textbf{0.60} & \cellcolor{gray!10}\textbf{0.71} & \cellcolor{gray!10}\textbf{0.55} & \cellcolor{gray!10}\textbf{0.60} & \cellcolor{gray!10}\textbf{0.81} & \cellcolor{gray!10}\textbf{0.71} & \cellcolor{gray!10}\textbf{0.66} & \cellcolor{gray!10}\textbf{0.99} \\
	\cellcolor{gray!25}method & \cellcolor{gray!10} & \cellcolor{gray!10} & \cellcolor{gray!10} & \cellcolor{gray!10} & \cellcolor{gray!10} & \cellcolor{gray!10} & \cellcolor{gray!10} & \cellcolor{gray!10} & \cellcolor{gray!10} & \cellcolor{gray!10} & \cellcolor{gray!10} \\
	\cellcolor{gray!25} & \cellcolor{white}Liang Zhao & \cellcolor{white}0.79 & \cellcolor{white}0.59 & \cellcolor{white}0.47 & \cellcolor{white}0.77 & \cellcolor{white}0.55 & \cellcolor{white}0.50 & \cellcolor{white}0.85 & \cellcolor{white}0.77 & \cellcolor{white}0.53 & \cellcolor{white}0.99 \\
	\cellcolor{gray!25}Supervised & \cellcolor{gray!10}Raphael Meier & \cellcolor{gray!10}0.72 & \cellcolor{gray!10}0.60 & \cellcolor{gray!10}0.53 & \cellcolor{gray!10}0.65 & \cellcolor{gray!10}0.62 & \cellcolor{gray!10}0.48 & \cellcolor{gray!10}0.88 & \cellcolor{gray!10}0.69 & \cellcolor{gray!10}0.64 & \cellcolor{gray!10}0.99 \\
	\cellcolor{gray!25}methods & \cellcolor{white}Syed Reza & \cellcolor{white}0.73 & \cellcolor{white}0.56 & \cellcolor{white}0.51 & \cellcolor{white}0.68 & \cellcolor{white}0.64 & \cellcolor{white}0.48 & \cellcolor{white}0.79 & \cellcolor{white}0.57 & \cellcolor{white}0.63 & \cellcolor{white}0.99  \\
	\cellcolor{gray!25} & \cellcolor{gray!10}Nicolas Cordier & \cellcolor{gray!10}0.75 & \cellcolor{gray!10}0.61 & \cellcolor{gray!10}0.46 & \cellcolor{gray!10}0.79 & \cellcolor{gray!10}0.61 & \cellcolor{gray!10}0.43 & \cellcolor{gray!10}0.78 & \cellcolor{gray!10}0.72 & \cellcolor{gray!10}0.52 & \cellcolor{gray!10}1.00  \\ \hline
\end{tabular}
}
\label{table:comparative_unsupervised_learning_brats_leaderboard}
\end{table}

Table \ref{table:comparative_unsupervised_learning_times} shows the average time in minutes required to obtain a segmentation for a single patient, including the preprocessing and postprocessing stages. Segmentations were computed in an Intel Xeon E5-2620 with 64GB of RAM using multi-threading. The preprocessing stage includes the denoising, bias field correction, skull-stripping and super resolution steps. The unsupervised classification time involves the parallel computation of the 10 different segmentations starting from the K-means++ initialization, and the posterior selection of the best solution. As expected, the more complex and sophisticated the algorithm is, the longer it takes to reach the solution. The postprocessing stage consist in the automated pathological label identification method, the outlier label removal and the merging process of similar statistical distribution labels. Such process includes the non-linear registration of the \ac{ICBM} template to the patient \Ti{} image, which practically covers the entire time of the postprocessing stage. It is worth noting that the non-linear \ac{ICBM} registration is performed only once for all the unsupervised segmentation algorithms.

\begin{table}[h]
\caption{Average computational times in minutes for the whole segmentation pipeline for a single patient.}
\resizebox{\textwidth}{!}{
\begin{tabular}{lcccc}
	\hline
	\cellcolor{gray!25} Algorithm & \cellcolor{gray!25} Preprocess & \cellcolor{gray!25} Unsupervised classification & \cellcolor{gray!25} Postprocess & \cellcolor{gray!25} Total \\
	\cellcolor{gray!25} K-means & \cellcolor{white} & \cellcolor{white}9 $\pm$ 5 & \cellcolor{white} & \cellcolor{white}110 $\pm$ 27 \\
	\cellcolor{gray!25} Fuzzy K-means & \cellcolor{white} & \cellcolor{gray!10}29 $\pm$ 3 & \cellcolor{white} & \cellcolor{gray!10}130 $\pm $ 25 \\
	\cellcolor{gray!25} GMM & \cellcolor{white}  & \cellcolor{white}41 $\pm$ 7 & \cellcolor{white} & \cellcolor{white}142 $\pm $ 29 \\
	\cellcolor{gray!25} Gauss-\ac{HMRF} & \multirow{-4}{*}{\cellcolor{white}13 $\pm$ 3} & \cellcolor{gray!10}39 $\pm$ 10 & \multirow{-4}{*}{\cellcolor{white}88 $\pm$ 19} & \cellcolor{gray!10}140 $\pm $ 32 \\ \hline
\end{tabular}
}
\label{table:comparative_unsupervised_learning_times}
\end{table}

\begin{figure}[h]
\centering
\includegraphics[width=\linewidth]{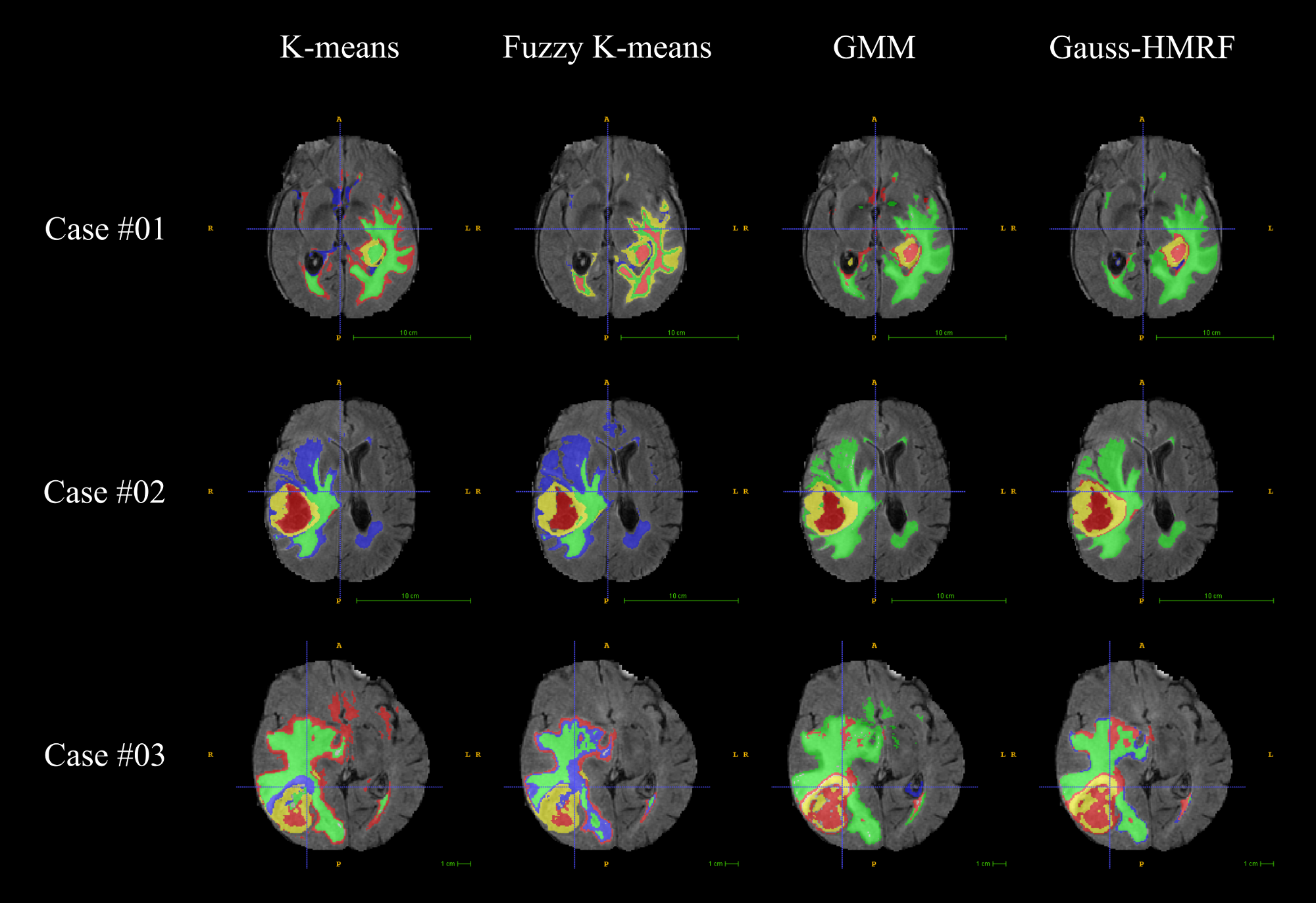}
\caption{Examples of final segmentations of 3 patients of \ac{BRATS} 2013 dataset computed by the different unsupervised algorithms.}
\label{figure:comparative_unsupervised_learning_segmentations}
\end{figure}

Finally, examples of segmentations achieved by the different unsupervised segmentation algorithms evaluated are shown in Figure \ref{figure:comparative_unsupervised_learning_segmentations}.

\section{Discussion}
\label{section:comparative_unsupervised_learning_Discussion}
In this study we have conducted an evaluation of the performance of several unsupervised learning algorithms for the glioblastoma segmentation task. In addition to the comparative, we have proposed a complete pipeline for automated brain tumor segmentation, including a postprocessing stage to automatically identify labels corresponding to pathological tissues in an unsupervised segmentation.

The proposed method is confirmed as a viable alternative for glioblastoma segmentation, as it has demonstrated to achieve competitive results in a public real reference dataset for brain tumor segmentation. The method improved the results obtained by the other unsupervised segmentation approaches evaluated in the \ac{BRATS} 2013 Challenge, and obtained competitive results with respect to supervised methods. Moreover, this study confirmed the capability of unsupervised learning to detect consistent patterns in medical imaging data related to \ac{MRI} properties of the tissue, which will serve as basis for the next work on this thesis.

The proposed unsupervised segmentation pipeline comprises four stages: \ac{MRI} preprocessing, feature extraction and dimensionality reduction, unsupervised voxel classification and automatic pathological label identification. Concerning the preprocessing stage, consolidated state of the art techniques that provide efficient solutions to enhance the information of the \ac{MR} images were employed. However, some preprocessing techniques are primarily oriented to non-pathological brains. This is the case of bias field correction. In our experiments, we found that the estimation of the magnetic field inhomogeneities with the N4 algorithm presented problems primarily with \ac{FLAIR} sequences. The hyper-intensity presented in the \ac{FLAIR} sequence related to the edema was confused frequently with inhomogeneities of the magnetic field, thereby reducing its intensity. In order to overcome this problem we reduced the number of iterations of the algorithm to remove as much inhomogeneities as possible, while keeping the intensities of the lesion. Such solution assumed a non optimal removal of the magnetic field inhomogeneities, but allowed to save the information contained in the lesion area, which becomes more important to the segmentation task. We empirically set a maximum of 10 iterations at each scale of the multi-scale approach of the N4 algorithm.

Several unsupervised classification algorithms were evaluated to assess its pros and cons, ranging from the most restrictive algorithms in terms of class-conditional probabilistic models (K-means and Fuzzy K-means) to more sophisticated models with more degrees of freedom such as \ac{GMM} or Gauss-\ac{HMRF}. The last one, also introduces statistical dependencies between adjacent variables of the model, that penalizes neighboring voxels with different labels. Hence, this structured prior aims to model the self similarity of the images, leading the algorithm to a more homogeneous segmentation than the non-structured classification techniques.

Therefore, the less restrictive algorithms were expected to achieve better results based on the hypothesis that these algorithms learn a more flexible model that best fits the data to be classified. Moreover, structured algorithms were also expected to obtain better results based on the hypothesis that these algorithms introduce mechanisms to model the self similarity of the images. Tables \ref{table:comparative_unsupervised_learning_results_test} and \ref{table:comparative_unsupervised_learning_results_leaderboard} confirm such hypotheses. Both \ac{GMM} and Gauss-\ac{HMRF} rose as the best algorithm tested in almost all the metrics returned by the evaluation web page. Only the results obtained by the Gauss-\ac{HMRF} model in the enhancing tumor sub-compartment of the Leaderboard set were not comparable with the other sub-compartments and datasets results. This effect was due to the smoothing prior imposed by the Gauss-\ac{HMRF}, which was too strong in some cases. We revised the cases that achieved low results in the enhancing tumor sub-compartment and realized that most of them had a large necrotic core with a thin low-brightness enhancing tumor ring. We also revised the K-means++ initializations and realized that the enhancing tumor was partially segmented in some cases but finally lost in the final segmentation due to the hard smoothing prior in the necrotic class. We are currently working on the introduction of different penalizations for the labels, depending on their statistical distribution similarities to avoid this over-smoothing.

It is worth noting that we obtained better results on the Leaderboard set (Table \ref{table:comparative_unsupervised_learning_brats_leaderboard}) than in the Test set (Table \ref{table:comparative_unsupervised_learning_brats_test}), in contrast with the rest of participants. This effect may have been produced by the fact that the Leaderboard set may include more heterogeneities and differences with respect to the Training set than to the Test set, thereby directly affecting the supervised approaches performance. Unsupervised paradigm avoids this possible overfitting by building a particular model for each patient considering only its own data, therefore achieving better results in the Leaderboard set against most of the supervised approaches evaluated.

In future work, we plan to improve our feature extraction process by analyzing the influence of the texture images in the final segmentations and including more sophisticated textures such as the Haralick texture features. Furthermore, we plan to extend our unsupervised methodology to the analysis and segmentation of \ac{PWI} in combination with anatomical images. The biomarkers obtained from \ac{PWI} might discover relevant segmentations by adding additional valuable functional information about the tissues. We consider that research efforts should be aligned with quantitative \ac{MRI} by providing powerful systems that leverage the information contained in these images.

\chapter{Non Local Spatially Varying Finite Mixture Models for unsupervised image segmentation}
\label{chapter:nlsvfmm}
As stated in the previous chapter, image segmentation is one of the most important core problems in computer vision. It constitutes one of the basic and fundamental steps for automated image understanding since its purpose is to delineate objects in the image with semantic meaning. Innumerable approaches have been proposed in the literature to address this problem, ranging from supervised to unsupervised \ac{ML} approaches. The latter are indispensable to the image understanding and segmentation task as they provide a robust and reliable solution to all the problems that do not have manually annotated datasets, which ultimately represent the vast majority of real-life image segmentation problems.

Images are structured arrangements of data in which, in addition to the pixel intensities, the location of the pixels provides important information to properly understand its content. Structured learning models capable to properly capture the patterns of local regularity and spatial redundancy of the images have demonstrated their superiority in the image segmentation task.

In this chapter a Bayesian model for unsupervised image segmentation based on a combination of the \acp{SVFMM} and the \ac{NLM} framework is presented. Such model successfully integrates a gauss-markov random field into a classic \ac{FMM}, to simultaneously codify the idea that neighboring pixels tend to belong to the same semantic object, but preserving the edges and structure of the image. The chapter introduces the mathematical foundations of the model and their estimation via a \ac{MAP}-\ac{EM} scheme. We present an evaluation of the performance of the model in a synthetic medical imaging corpus and with a reference dataset of real-world images.

\medskip

\emph{The contents of this chapter were published in the journal publication \citep{JuanAlbarracin2019b}---thesis contributions C2 and P3.}

\section{Introduction}
\label{section:nlsvfmm_intro}
Unsupervised learning has historically played a key role in the image segmentation task, constituting one of the first paradigms to automatically identify objects and structures in an image \citep{Zhang2008}. Specifically, clustering has gathered most of the efforts in unsupervised image segmentation research. Clustering is the task of finding natural groupings of data within a population, sharing a similar set of properties \citep{Rokach2005}. Many clustering techniques have been proposed in the literature during the past decades \citep{Saxena2017}, ranging from distance based techniques such as partitional clustering or hierarchical clustering; density-based techniques such as DBSCAN \citep{Ester1996} or Mean Shift \citep{Cheng1995}; graph based algorithms such as graph-cuts \citep{Boykov2001}; or probabilistic models such as \acfp{FMM}\citep{Pal1993}.

Specifically, probabilistic models intend to learn the \ac{pdf} of an image by means of fitting a multi-parametric statistical model to the data. In particular, \acp{FMM} fit a weighted sum of probabilistic distributions, each one representing a component of the image, to capture the heterogeneity nature of the image information. \acp{GMM} are the most extended \acp{FMM}, being widely employed for image segmentation, as they have proven to successfully capture the complexity of an image \citep{JuanAlbarracin2015a}. Moreover, \acp{GMM} can be efficiently estimated by means of \ac{MLE} via the \acf{EM} algorithm \citep{Dempster1977}.

However, learning from an image has several particularities that must be taken into account. Images are structured arrangements of data in which, in addition to the pixel intensities, the location of these intensities provide important information to properly understand its content. Images show patterns of local regularity and spatial intensity redundancy that enclose the idea that adjacent pixels tend to belong to the same semantic object. Conventional \acp{FMM}, by the opposite, do not inherently take into account this information. \acp{FMM} make the heavy assumption that data in an image is \ac{i.i.d.}, ignoring the spatial information that has demonstrated to be useful to generate more accurate and realistic models.

To overcome this limitation, several solutions have been proposed in the literature \citep{Blake2011}. Most of them rely on the inclusion of a \ac{MRF} to model the local dependencies between pixels in an image. Specifically, a variant to the \ac{FMM} called \ac{SVFMM} was proposed by \cite{Sanjay1998}, which replaces the classics mixing coefficients of the \ac{FMM} by \emph{contextual mixing coefficients} for each pixel of the image. This approximation allows to introduce a continuous \ac{MRF} over these contextual mixing coefficients to incorporate the idea that neighboring pixels tends to share the same intensity properties.

Many variants of \acp{MRF} have been proposed in the literature to capture the local information contained in an image. \cite{Nikou2007} proposed a family of Gauss-\acp{MRF}, successfully achieving better results than the classic \acp{FMM}. However, such approximation introduces a local isotropic smoothing over the contextual mixing coefficients, that ignores the presence of edges in the image. Therefore, the contextual mixing coefficients estimated under the Gauss-\ac{MRF} approximation are iteratively smoothed, yielding prior probability maps that lose the information of image edges. \cite{Sfikas2008} proposed a t-Student \ac{MRF} that allowed to regulate the smoothing between pixels in an edge. However, this approximation introduces new parameters to be estimated in the model, yielding a non closed-form analytic solution for it.

In this chapter a fully Bayesian \ac{SVFMM} model, called \ac{NLSVFMM}, that combines the \ac{SVFMM} framework with the \ac{NLM} filtering schema is proposed. The model has 2 variants: the pixel-wise version (NLv-\ac{SVFMM}) and the patch-wise version (NLp-\ac{SVFMM}). The proposed model introduces a Gauss-\ac{MRF} weighted by the probabilistic \ac{NLM} function proposed by \cite{Wu2013} to adaptively adjust the spatial regularization depending on the structure of the image. Such approximation avoids the introduction of new parameters, reducing the degrees of freedom of the model and the number of samples required for a reliable estimation of the parameters.

\section{Background on Spatially Varying Finite Mixture Models}
\label{section:nlsvfmm_svfmm}
The \ac{SVFMM} is a modification of the classic \ac{FMM}, focused mainly on imaging data, in which the coefficients of the mixture are related to each other through a structured graph that defines the statistical dependencies between them. \acp{SVFMM} are thoroughly introduced in section \ref{section:rationale_mixture_models}, however, in order for this chapter to be self-contained, a short remainder will be made.

Let $X=\left(\vec{x}^1,\dots,\vec{x}^N\right)$ a set of observations corresponding to the pixels of an image, where $\vec{x}^i \in \Reals^D$ and represents a vector of $D$ features for the $i^{th}$ pixel. The \ac{SVFMM} is defined as:

\begin{equation}
p\left(X | \Theta, \Pi\right) = \prod_{i=1}^N\sum_{j=1}^K \pi_j^i\phi\left(\vec{x}^i;\Theta_j\right)
\end{equation}

\noindent where $\phi\left(\vec{x}^i|\Theta_j\right)$ is a \ac{pdf} used to model the data (typically a Normal or t-Student distribution) and $\Theta=\left\lbrace \Theta_1,\dots,\Theta_K,\vec{\pi}^1,\dots,\vec{\pi}^N\right\rbrace$ the set of parameters of the model, with $\Pi=\left\lbrace\vec{\pi}^1,\dots,\vec{\pi}^N\right\rbrace$ called the \emph{contextual mixing coefficients}, which must comply with:

\begin{equation}
\label{eq:nlsvfmm_pi_restriction}
\forall \vec{\pi}^i, ~~ 0 \leq\pi_j^i\leq1, ~~ \sum_{j=1}^K \pi_j^i=1
\end{equation}

A \ac{MAP} estimate of $\left(\Theta, \Pi\right)$ is typically conducted to impose a proper prior over $\Pi$ to introduce the idea that neighboring pixels in an image tend to belong to the same semantic object.

\begin{equation}
\left( \hat{\Theta}, \hat{\Pi} \right)_{MAP} = \argmax_{ \left( \Theta , \Pi \right)} \log p \left(X | \Theta, \Pi \right) + \log p \left(\Pi \right)
\end{equation}

Several variants of $p\left(\Pi\right)$ have been proposed in the literature. Specifically, the \ac{DCAGMRF}, introduced in equation \ref{eq:dcagmrf}, has proven to successfully capture the local redundancy and spatial regularity inherent in images, regularizing many ill-posed inverse problems with successful results (please refer to section \ref{subsection:rationale_mixture_models_svfmm} for a detailed description of the \ac{DCAGMRF} prior).

\medskip

As stated in section \ref{subsection:rationale_mixture_models_svfmm}, inference on \acp{SVFMM} is not analytically tractable. Numerical optimization methods are therefore required to estimate $\left(\Theta, \Pi\right)$ in a tractable manner. Typically, a \ac{MAP}-\ac{EM} algorithm is used to iteratively find the updates of the model parameters, based on the conditional expectation of the log-likelihood function, used to guide the estimation procedure. Nevertheless, no closed-form solution can be obtained for $\vec{\pi}^i$ when considering the restriction stated in \ref{eq:nlsvfmm_pi_restriction}. To overcome this limitation, an alternative solution is to consider $\Pi$ as a random variable governed by a \ac{DCM} distribution. \cite{Nikou2010} demonstrated that, following such approach, $\pi_j^i$ can be computed as: 

\begin{equation}
\pi_j^i = \frac{\alpha_j^i}{\sum\limits_{k=1}^K \alpha_k^i}
\end{equation}

\noindent with $\vec{\alpha}^i$ the parameters of the Dirichlet distribution. It is easy to see that such approach always guarantees that $\sum_j \pi_j^i=1 ~~ \forall i$, hence satisfying the condition settled in \ref{eq:nlsvfmm_pi_restriction}. In addition, $\vec{\alpha}^i$ parameters only require to satisfy $\alpha_j^i > 0 ~~ \forall i,j$, making easier their optimization.

Therefore, following this \ac{DCM}-\ac{SVFMM} approach, the \ac{DCAGMRF} density can be imposed over $A = \left\lbrace\vec{\alpha}^1,\dots,\vec{\alpha}^N\right\rbrace$ (instead of over $\Pi$), to enforce the desired local regularity:

\begin{equation}
p\left(A\right)=\prod_{i=1}^N\prod_{j=1}^K\prod_{d=1}^D\prod_{m\in\mathcal{M}_d^i} \frac{1}{\sqrt{2\pi\beta_{j,d}^2}}\exp\left(-\frac{\left(\alpha^i-\alpha_j^m\right)^2}{2\beta_{j,d}^2}\right)
\end{equation}

\noindent where sub-index $d$ refers to the different spatial adjacency directions (i.e. horizontals, verticals or diagonals), and $\mathcal{M}_d^i$ indicates the set of neighbors of the $i^{th}$ pixel that lies in the $d^{th}$ spatial direction.

The complete step-by-step estimation of the parameters of the \ac{DCM}-\ac{SVFMM} via the \ac{EM} algorithm is described in section \ref{subsection:rationale_mixture_models_dcmsvfmm}. We encourage the reader to review this section for an in-depth explanation of the estimation procedure of the \ac{DCM}-\acp{SVFMM}.

\section{Background on Probabilistic Non Local Means}
\label{section:nlsvfmm_nlm}
The \ac{NLM} filter \citep{Buades2005b} proposes a schema for image filtering where pixels are restored by a weighted sum of similar neighbor patches.

The core of \ac{NLM} schema is the weight function that relates neighboring patches, which has taken a lot of variants in the literature. Specially, \cite{Wu2013} derived the probabilistic version of the \ac{NLM} algorithm and its associated probabilistic weighting function.

In order to relate the description of the probabilistic \ac{NLM} with the \ac{SVFMM} background, let's consider $d_{j,d}^{i,m}$ as the distance between a pair of adjacent Dirichlet parameters in the form:

\begin{equation}
d_{j,d}^{i,m} = \frac{\left(\alpha_j^i - \alpha_j^m\right)^2}{2\beta_{j,d}^2}
\end{equation}

Assuming that local differences are \ac{i.i.d.}, we have $d_{j,d}^{i,m} \sim \chi^2\left(1\right)$. For a patch-based version of the algorithm, the distance between two patches centered at $i^{th}$ and $m^{th}$ locations is defined as:

\begin{equation}
D_{j,d}^{i, m} = \sum_{k\in\mathbb{P}} d_{j,d}^{i+k, m+k}
\end{equation}

\noindent where $\mathbb{P}$ is the set of offsets that define a local patch around a given pixel. If patches are completely disjoint, then $D_{j,d}^{i, m} \sim \chi^2\left(\left|\mathbb{P}\right|\right)$, however, in most cases, overlapping occurs between patches, so the \ac{i.i.d.} assumption does not hold. In such cases, an approximation to the sum of a set of correlated $\chi^2$ distributions can be computed as:

\begin{equation}
D_{j,d}^{i,m} \sim \gamma_m\chi^2\left(\eta_m\right)
\end{equation}

\noindent where

\begin{equation}
\begin{split}
\gamma_m &= \mathrm{var}\left[D_{j,d}^{i,m}\right] / 2\mathrm{E}\left[D_{j,d}^{i,m}\right] \\
\eta_m &= \mathrm{E}\left[D_{j,d}^{i,m}\right] / \gamma_m
\end{split}
\end{equation}

\noindent and

\begin{equation}
\begin{split}
\mathrm{E}\left[D_{j,d}^{i,m}\right] &= \left|\mathbb{P}\right| \\
\mathrm{var}\left[D_{j,d}^{i,m}\right] &= 2\left|\mathbb{P}\right| + \left|\mathbb{O}^{i,m}\right|
\end{split}
\end{equation}

\noindent with $\mathbb{O}^{i,m}$ the set of overlapping pixels between the patches centered at $i^{th}$ and $m^{th}$ pixels.

Hence, the weight function $u_{j,d}^{i,m}$ proposed in the probabilistic \ac{NLM} approach, associated to the \ac{SVFMM} framework, is defined as

\begin{equation}
u_{j,d}^{i,m} = \chi^2\left(D_{j,d}^{i,m} / \gamma_m | \eta_m \right) = \frac{\left(D_{j,d}^{i,m} / \gamma_m\right)^{\left(\eta_m / 2\right) - 1}\exp{\left(-D_{j,d}^{i,m} / 2\gamma_ m\right)}}{2^{\eta_m/2}\Gamma\left(\eta_m / 2\right)}
\end{equation}

\section{The Non Local Spatially Variant Finite Mixture Model}
\label{section:nlsvfmm_nlsvfmm}
One of the main drawbacks of the \ac{SVFMM}-based models is that they enforce a local smoothness on the contextual mixing coefficients (or Dirichlet parameters) without taking into account the structure of the image. In other words, the \ac{SVFMM} iteratively applies an isotropic local Gaussian smoothing to these parameters, leading to a over-smoothed prior probability map that losses the information of the edges and structures in the image.

To overcome this limitation, \cite{Sfikas2008} proposed a variant of the \ac{SVFMM} where local differences between Dirichlet parameters follow a t-Student distribution. Such an approach was intended to exploit the heavy-tailed nature of the t-Student distribution, to perform a robust estimation of the Dirichlet coefficients when edges and structures are present in their local neighborhoods.

\begin{equation}
\alpha_j^i-\alpha_j^m \sim \mathcal{S}t\left(0,\beta_{j,d}^2,\nu_j\right)
\end{equation}

Following the Bishop's development in \citep{Bishop2006}, a $\mathcal{S}t$ distribution can be expressed as:

\begin{equation}
\begin{split}
\alpha_j^i-\alpha_j^m &\sim \mathcal{N}\left(0,\beta_{j,d}^2 / g_{j,d}^{i,m}\right) \\
g_{j,d}^{i,m} &\sim \mathcal{G}\left(\nu_{j,d} / 2, \nu_{j,d} / 2\right)
\end{split}
\end{equation}

This model introduces a new set of latent variables $g_{j,d}^{i,m}$, whose posterior density should be estimated at the E-step, and a new set of parameters $\nu_{j,d}$, with non closed-form analytic estimation. Therefore, numerical optimization methods such as Newton-Raphson or Brent's methods should be employed to estimate $\nu_{j,d}$.

In this sense, and similar in spirits than the t-Student model, in this chapter we propose the \ac{NLSVFMM} as a modification of the Sfikas' t-Student model, by replacing the $g_{j,d}^{i,m}$ random variable by the probabilistic \ac{NLM} $u_{j,d}^{i,m}$ weight. Therefore, we propose to reformulate the local differences between contextual Dirichlet parameters to follow a \ac{NLM}-related distribution, denoted by:

\begin{equation}
\begin{split}
\alpha_j^i-\alpha_j^m &\sim \mathcal{N}\left(0,\beta_{j,d}^2 / \chi^2\left(D_{j,d}^{i,m} / \gamma_m | \eta_m \right)\right) \\
D_{j,d}^{i,m} &\sim \gamma_m \chi^2\left(\eta_m\right)
\end{split}
\end{equation}

\noindent with $D_{j,d}^{i,m}$ being latent variables of the model.

\medskip

Following the conventional \ac{EM} scheme, the posterior densities of $D_{j,d}^{i,m}$ should be calculated at the E-step. However, this leads to a different calculation of $D_{j,d}^{i,m}$ than the proposed by \cite{Wu2013} (see Section \ref{section:nlsvfmm_nlm}). Therefore, in order to preserve the use of the original \ac{NLM} weights, we will follow a \emph{Variational \ac{EM}} approach \citep{Neal1999, Bishop2006}. The Variational \ac{EM} framework introduces the concept of partial E-step, in which a functor of the latent variables can be used when the posterior densities of these variables cannot be calculated, or when it is desirable to calculate them differently for reasons of efficiency or performance. As demonstrated by \cite{Neal1999}, such functor can take any form as long as the log-likelihood function is increased at each iteration, effectively driving the model to a local optimum of the function, and hence to an optimum of the parameters of the model. Therefore, following this framework, the $D_{j,d}^{i,m}$ latent variables are estimated at the E-step as the standard quantitative Chi-squared test proposed by \cite{Wu2013}:

$$
D_{j,d}^{i,m} = \sum_{k\in\mathbb{P}} \frac{\left(\alpha_j^{i+k} - \alpha_j^{m+k}\right)^2}{2\beta_{j,d}^2}
$$ 

Once these latent variables are estimated, the $u_{j,d}^{i,m}$ weights are calculated at the M-step following $u_{j,d}^{i,m} = \chi^2\left(D_{j,d}^{i,m} / \gamma_m | \eta_m \right)$. Since $u_{j,d}^{i,m}$ depends on both the $i^{th}$ and $m^{th}$ observations, this model specifies a different instance of a Gaussian distribution for each $\left(\alpha_j^i - \alpha_j^m\right)$ pair of contextual Dirichlet coefficients in the \ac{MRF}. This allows $u_{j,d}^{i,m}$ to regulate the variance of the corresponding Gaussian between the $i^{th}$ and $m^{th}$ observations, if an edge or an homogeneous area is detected at this location. Thus, as $u_{j,d}^{i,m}$ increases, the Gaussian distribution for the corresponding pair shrinks around zero imposing a hard smoothing between the observations. On the contrary, as $u_{j,d}^{i,m}$ decreases, the variance of the Gaussian distributions increases producing a lower \ac{pdf} value that prevents the smooth.

This approximation avoids the introduction of new parameters since $\eta_m$ and $\gamma_m$ are completely known once $i$ and $m$ are fixed. Therefore, no numerical approximate methods are required, simplifying the model and reducing its degrees of freedom and the number of samples required for its statistically reliable estimation.

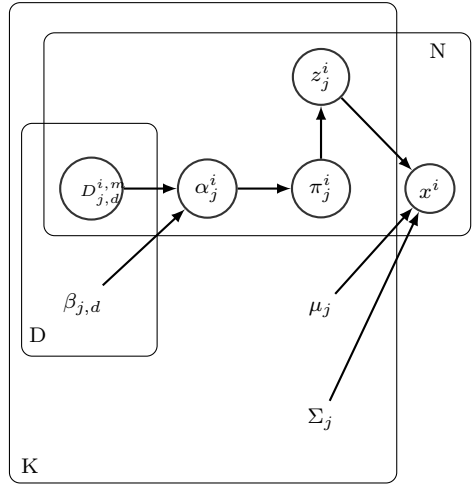
\begin{figure}
\centering
\scriptsize
\begin{tikzpicture}
\tikzstyle{main}=[circle, minimum size = 4mm, thick, draw =black!80, node distance = 7mm]
\tikzstyle{connect}=[-latex, thick]
\node[main, fill = white!100] (alpha) {$\alpha_j^i$};
\node[main] (u)  [left=of alpha, text width=3mm] {\begin{tiny} $D_{j,d}^{i,m}$ \end{tiny}};
\node[main] (pi) [right=of alpha] {$\pi_j^i$};
\node[main] (z)  [above=of pi] {$z_j^i$};
\node[main] (x)  [right=of pi] {$x^i$};
\node(beta) [below left=of alpha] {$\beta_{j,d}$};
\node(mu) [right=of beta,below=of pi] {$\mu_j$};
\node(sigma) [below=of mu] {$\Sigma_j$};

\path (beta) edge [connect] (alpha)
      (u) edge [connect] (alpha)
      (alpha) edge [connect] (pi)
      (pi) edge [connect] (z)
      (z) edge [connect] (x)
      (mu) edge [connect] (x)
      (sigma) edge [connect] (x);

\node[rectangle, rounded corners, inner sep=0mm, fit=(u) (z) (z),label=above right:N, xshift=17mm, yshift=-2.5mm] {};
\node[rectangle, rounded corners, inner sep=2mm, draw=black!100, fit=(u) (x) (z)] {};
\node[rectangle, rounded corners, inner sep=0mm, fit=(beta) (pi) (z) (sigma),label=below left:K, xshift=-1mm, yshift=-8mm] {};
\node[rectangle, rounded corners, inner sep=6mm, draw=black!100, fit=(beta) (pi) (z) (sigma)] {};
\node[rectangle, rounded corners, inner sep=0mm, fit=(u) (beta),label=below left:D, yshift=-6mm] {};
\node[rectangle, rounded corners, inner sep=4.4mm, draw=black!100, fit=(u) (beta)] {};
\end{tikzpicture}
\caption{Graphical model for the non-local spatially varying finite mixture model. Superscripts $i,m \in \left[1,N\right]$ denote pixel indexes, subscript $j \in \left[1, K\right]$ denotes mixture component and subscript $d \in \left[1, D\right]$ denotes neighborhood direction.
}
\label{figure:nlsvfmm_graphical_model}
\end{figure}

The graphical model of the NL-\ac{SVFMM} is shown in Figure \ref{figure:nlsvfmm_graphical_model}.

Imposing the \ac{DCAGMRF} prior to the proposed \ac{NLSVFMM} model, the new density for $p\left(A\right)$ becomes

\begin{equation}
\label{eq:nlsvfmm_pA}
p\left(A\right)=\prod_{i=1}^N\prod_{j=1}^K\prod_{d=1}^D\prod_{m\in\mathcal{M}_d^i} \frac{1}{\sqrt{2\pi\beta_{j,d}^2 / u_{j,d}^{i,m}}}\exp\left(-\frac{\left(\alpha^i-\alpha_j^m\right)^2 u_{j,d}^{i,m}}{2\beta_{j,d}^2}\right)
\end{equation}

\noindent which setting $\partial Q / \partial \alpha_j^i$ yields a third degree equation of the form:

\begin{equation}
\left(\alpha_j^{i(t+1)}\right)^3 + \left(\alpha_j^{i(t+1)}\right)^2 \left(A_{-j}^i - \frac{\widehat{C}_j^i}{\widehat{B}_j^i} \right) - \left(\alpha_j^{i(t+1)}\right)\left(\frac{A_{-j}^i \widehat{C}_j^i}{\widehat{B}_j^i}\right) - \frac{z_j^iA_{-j}^i}{2\widehat{B}_j^i} = 0
\end{equation}

\noindent where

\begin{equation}
\begin{split}
A_{-j}^i    &= \sum\limits_{\substack{k=1 \\ k \neq j}}^K {\alpha_k^i}^{\left( t \right)} \\
\widehat{B}_j^i &= \sum\limits_{d=1}^D\frac{\sum\limits_{m\in\mathcal{M}_d^i}{u_{j,d}^{i,m}}^{\left( t \right)}}{{\beta_{j,d}^2}^{\left( t \right)}} \\
\widehat{C}_j^i &= \sum\limits_{d=1}^D\frac{\sum\limits_{m\in\mathcal{M}_d^i}{\alpha_j^m}^{\left( t \right)} {u_{j,d}^{i,m}}^{\left( t \right)}}{{\beta_{j,d}^2}^{\left( t \right)}}
\end{split}
\end{equation}

Likewise the conventional \ac{DCM}-\ac{SVFMM}, it can be demonstrated that, under polynomial theory, there is always a real non negative solution that satisfies $\alpha_j^i \geq 0$. The Vieta's method is used to algebraically obtain the roots of the proposed third degree equation.

Finally, $\beta_{j,d}^2$ is estimated as:

\begin{equation}
{\beta_{j,d}^2}^{\left( t+1 \right)} = \frac{1}{N} \sum\limits_{i=1}^N\sum\limits_{m\in\mathcal{M}_d^i} \frac{\left({\alpha_j^i}^{\left( t+1 \right)} - {\alpha_j^m}^{\left( t+1 \right)}\right)^2 {u_{j,d}^{i,m}}^{\left( t \right)}}{\left|\mathcal{M}_d^i\right|}
\end{equation}

Hereafter, the pixel-wise $\chi^2\left(1\right)$ version of the proposed \ac{NLSVFMM} will be referred as NLv-\ac{SVFMM}, while the patch-wise $\chi^2\left(\eta_m\right)$ will be referred as NLp-\ac{SVFMM}.

\section{Experimental results}
\label{section:nlsvfmm_results}
Both variants of the proposed \ac{NLSVFMM} algorithm have been evaluated in a simulated and a real-world scenario. First an evaluation on a synthetic high grade glioma dataset of the \ac{BRATS} 2013 Challenge was conducted \citep{Menze2015}. Next, and evaluation over the 300 real-world images of the Berkeley Segmentation dataset was performed \citep{Martin2001}. We have compared our proposed NLv- and NLp-\ac{SVFMM} model with the conventional \ac{FMM}, the \ac{SVFMM} and the $\mathcal{S}$t-\ac{SVFMM}. For the spatially varying algorithms we have employed the \ac{DCM} Bayesian approximation and the \ac{DCAGMRF} prior specified in \ref{eq:nlsvfmm_pA}. All algorithms in all experiments were initialized with a deterministic version of K-means++ \citep{Arthur2007} to ensure a fair comparative. In addition, results on the behavior of the weighting functions and weighting maps of the \ac{NLSVFMM} and t-Student model are shown. The evaluation is presented below.

\subsection[Evaluation on the BRATS 2013 high grade glioma dataset]{Evaluation on the synthetic BRATS 2013 high grade glioma dataset}
\label{subsection:nlsvfmm_results_brats}
The \ac{BRATS} 2013 high grade glioma synthetic dataset is composed of 25 cases, each one segmented into 7 labels: 1) \ac{WM}, 2) \ac{GM}, 3) \ac{CSF}, 4) peripheral edema (ED), 5) tumor core (split into enhancing tumor (5.1) and necrotic core (5.2)) (TC) and 6) vessels (VS). For each voxel, intensities on pre- and post-gadolinium \Ti{}-weighted, \Tii{}-weighted and \ac{FLAIR} \ac{MRI} sequences were employed for the segmentation.

\begin{figure}[ht]
\centering
\includegraphics[width=0.70\linewidth]{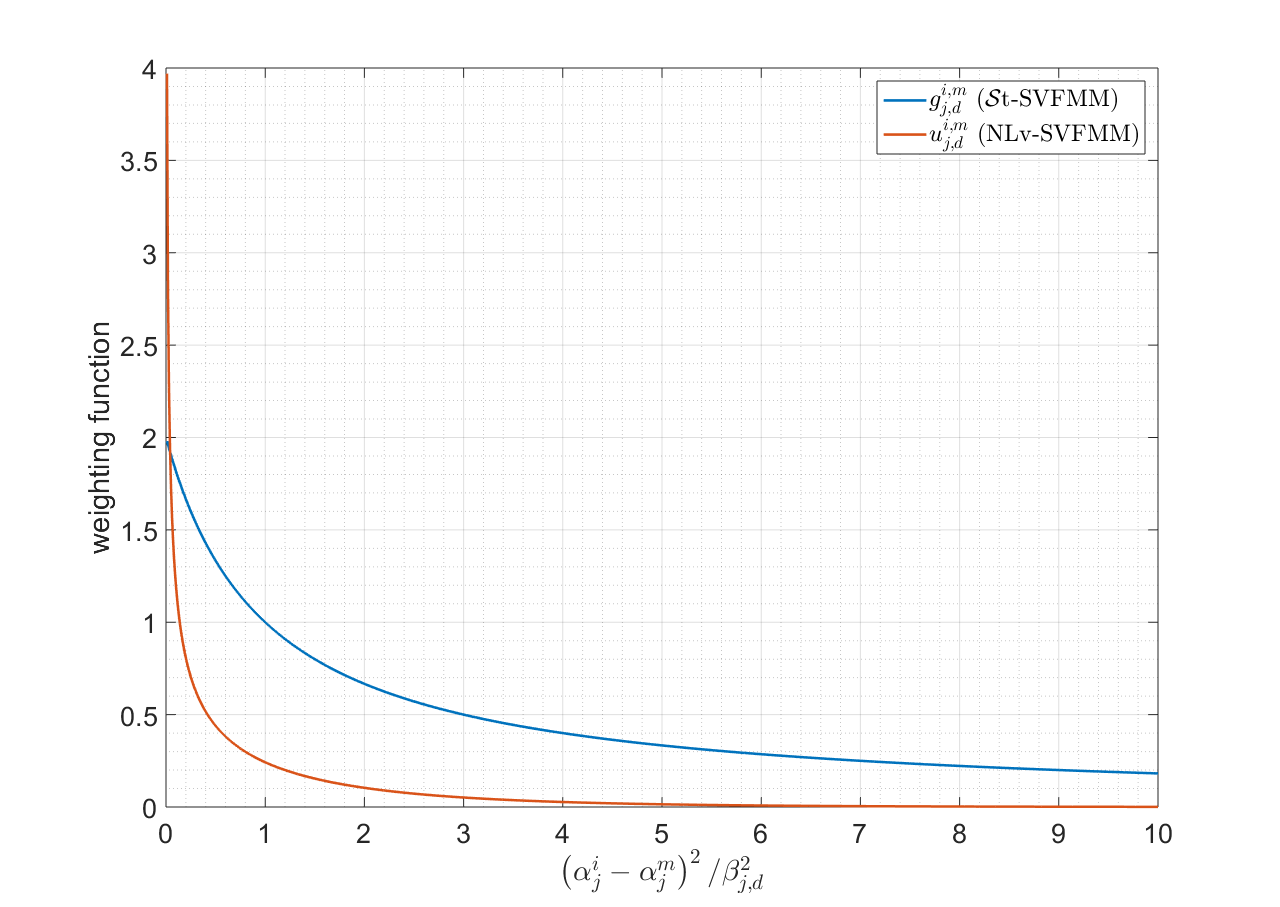}
\caption{Comparison between the behavior of the weighting functions $G = \left\lbrace g_{j,d}^{i,m} \right\rbrace$ for the $\mathcal{S}$t-\ac{SVFMM} model and $U = \left\lbrace u_{j,d}^{i,m} \right\rbrace$ for the NLv-\ac{SVFMM} (NLp-\ac{SVFMM} weighting function is not depicted because is not numerically comparable to the $\mathcal{S}$t and NLv-\ac{SVFMM} functions).}
\label{figure:nlsvfmm_weight_functions}
\end{figure}

Figure \ref{figure:nlsvfmm_weight_functions} compares the behavior of the weighting functions $G = \left\lbrace g_{j,d}^{i,m} \right\rbrace$ for the $\mathcal{S}$t-\ac{SVFMM} model and $U = \left\lbrace u_{j,d}^{i,m} \right\rbrace$ for the NLv-\ac{SVFMM} model (NLp-\ac{SVFMM} weighting function is not depicted because is not numerically comparable to the $\mathcal{S}$t and NLv-\ac{SVFMM} models).

\begin{figure}[ht]
\centering
\includegraphics[width=\linewidth]{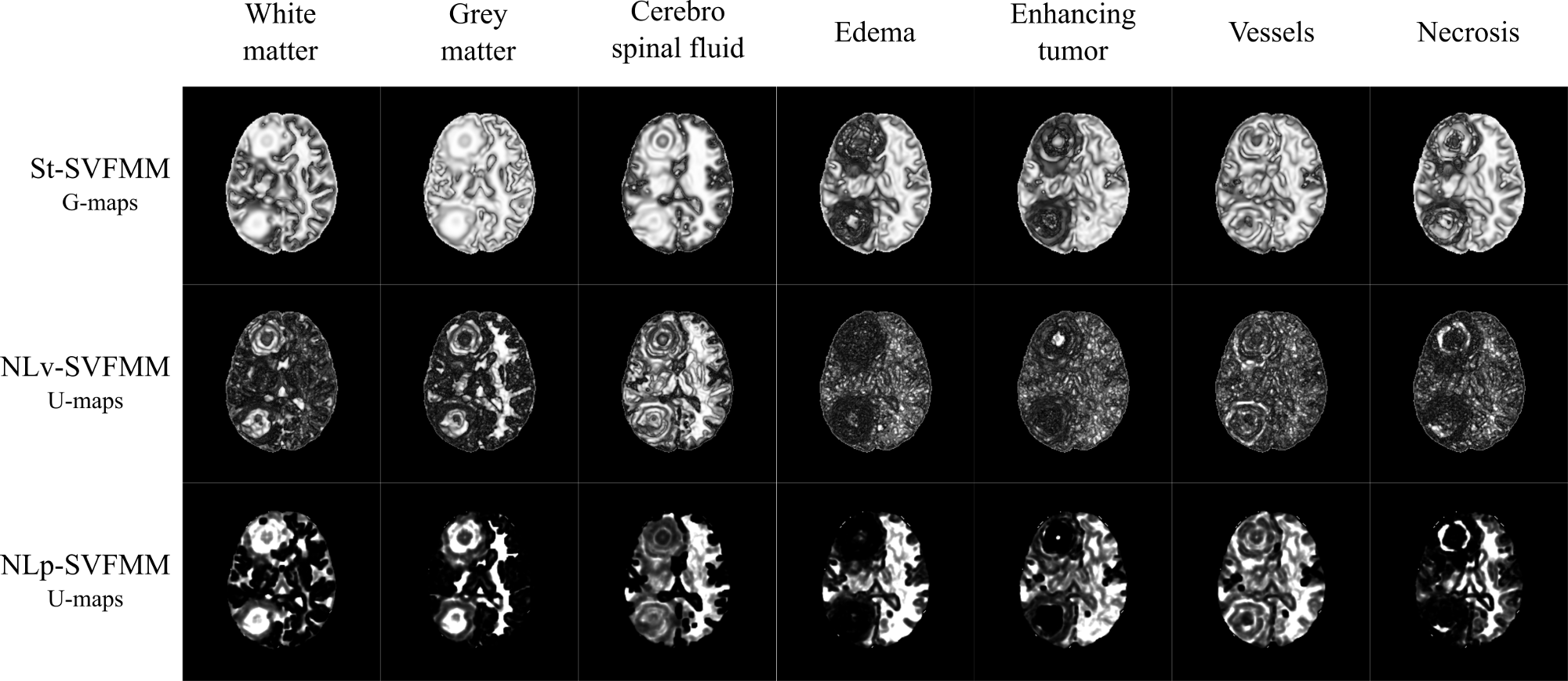}
\caption{Comparison between $G$ maps of the $\mathcal{S}$t-\ac{SVFMM} and $U$ maps for the NLv- and NLp-\ac{SVFMM} models for a case of the \ac{BRATS} 2013 dataset. Each pixel $i$ of the images represent $\sum_{d=1}^D \sum_{m \in \mathcal{M}^i} \lambda_{j,d}^{i,m}$, with $\lambda = g$ or $\lambda = u$ for $\mathcal{S}$t-\ac{SVFMM} and NLv- or NLp-\ac{SVFMM} models respectively.}
\label{figure:nlsvfmm_weight_maps}
\end{figure}

As figure shows, $U$ function behaves more aggressive for differences between observations than the $\mathcal{S}$t-\ac{SVFMM}, hence yielding more dichotomous weighting maps (see Figure \ref{figure:nlsvfmm_weight_maps}). For the shake of simplicity, each pixel of each picture of Figure \ref{figure:nlsvfmm_weight_maps} represent  $\sum_{d=1}^D \sum_{m \in \mathcal{M}^i} \lambda_{j,d}^{i,m}$, with $\lambda = g$ for $\mathcal{S}$t-\ac{SVFMM} and $\lambda = u$ for NLv- or NLp-\ac{SVFMM} models respectively. The weighting maps of Figure \ref{figure:nlsvfmm_weight_maps} demonstrates that the \ac{NLSVFMM}-based algorithms better differentiates among tissues than than the $\mathcal{S}$t-\ac{SVFMM}.

\begin{figure}[ht]
\centering
\includegraphics[width=0.95\linewidth]{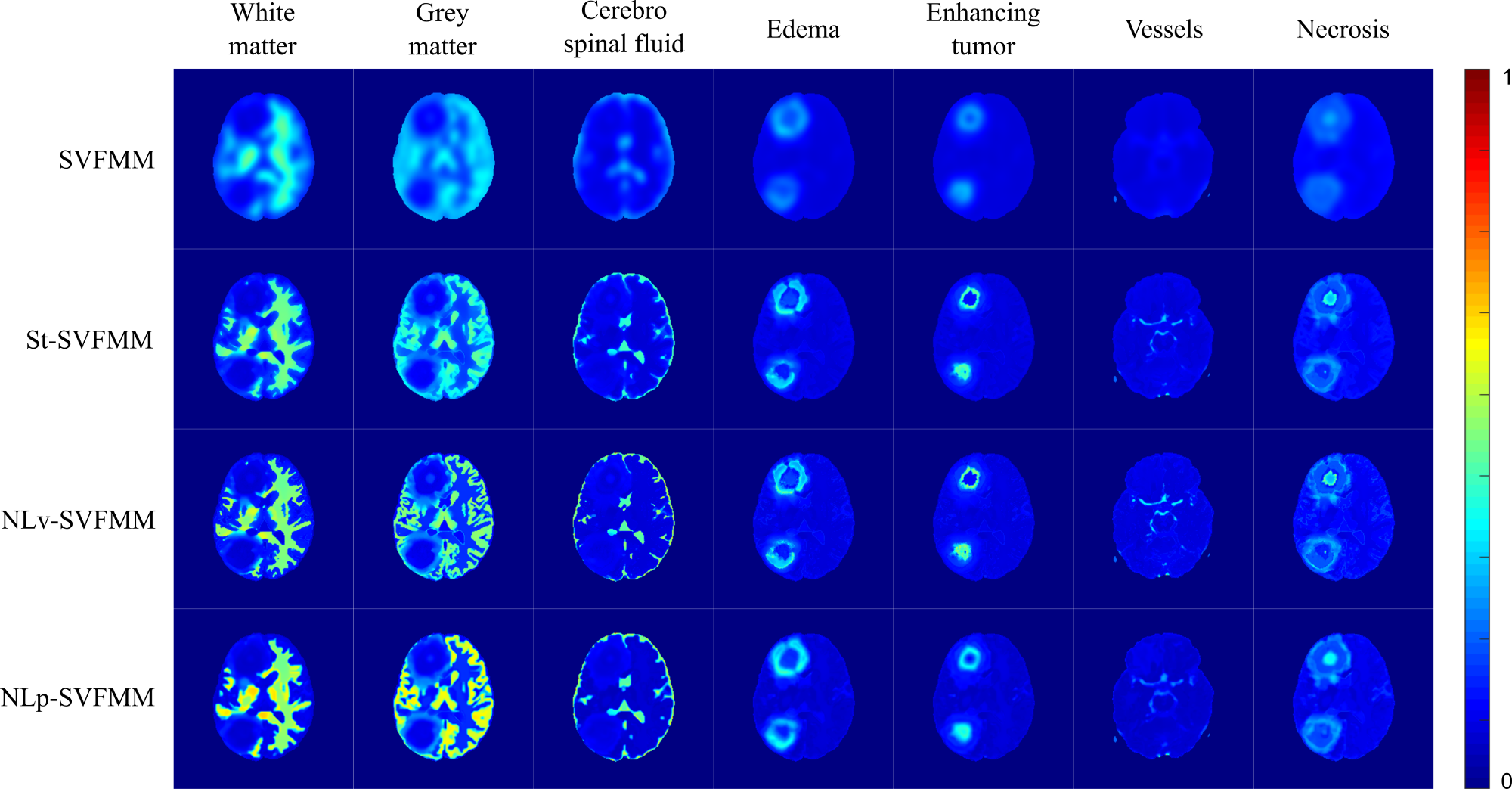}
\caption{Example of contextual mixing coefficient maps for a case of \ac{BRATS} 2013, for each label of the segmentation.}
\label{figure:nlsvfmm_mixing_coefficients_maps}
\end{figure}

Table \ref{table:nlsvfmm_mixing_coefficients} and Figure \ref{figure:nlsvfmm_mixing_coefficients_maps} show the superiority of the proposed NL-\ac{SVFMM} (in both variants) to generate higher confidence prior probability maps for each component. An example of the contextual mixing coefficient maps for the HG0014 case of the \ac{BRATS} 2013 dataset and its associated mixing coefficient values for different pixels obtained by each method is shown. In almost all evaluations, the NLp-\ac{SVFMM} version achieves the best results, indicating that the patch-based probabilistic \ac{NLM} weighting function better captures the local similarities in the images.


\begin{table}[ht]
\caption{Contextual mixing coefficients for different voxels of the HG0014 case of the \ac{BRATS} 2013 challenge. Voxels correspond to coordinates $a=\left(151,127,85\right), b=\left(167,75,85\right), c=\left(151,152,85\right), d=\left(97,89,85\right), e=\left(117,62,85\right), f=\left(110,71,85\right)$ and $f=\left(128,99,85\right)$}
\centering
\small
\rowcolors{2}{gray!10}{white}
\begin{tabular}{lcccc}
	\hline
	\rowcolor{gray!25}
	              & \ac{SVFMM} & $\mathcal{S}$t-\ac{SVFMM} & NLv-\ac{SVFMM} & NLp-\ac{SVFMM} \\
	$\pi^a_{WM}$  & 0.457 & 0.486 & 0.481 & \textbf{0.499} \\
	$\pi^b_{GM}$  & 0.296 & 0.429 & 0.497 & \textbf{0.606} \\
	$\pi^c_{CSF}$ & 0.253 & 0.426 & 0.446 & \textbf{0.447} \\
	$\pi^d_{ED}$  & 0.294 & 0.374 & 0.394 & \textbf{0.411} \\
	$\pi^e_{ET}$  & 0.248 & 0.442 & \textbf{0.453} & 0.381 \\
	$\pi^f_{NC}$  & 0.267 & 0.291 & 0.285 & \textbf{0.361} \\
	$\pi^g_{VS}$  & 0.118 & 0.305 & \textbf{0.355} & 0.265 \\
	\hline
\end{tabular}
\label{table:nlsvfmm_mixing_coefficients}
\end{table}

Table \ref{table:nlsvfmm_dice_BRATS} shows the Dice coefficients obtained for the evaluation based on the \ac{BRATS} 2013 dataset. Consistently with previous results, the NLp-\ac{SVFMM} variant achieves the best results in terms of segmentations based on the maximization of the posterior probabilities (Bayes minimum classification error). An improvement of about 3 points in Dice is obtained when comparing the NLp-\ac{SVFMM} with the standard \ac{SVFMM} and more than 1 point in Dice with respect to the $\mathcal{S}$t-\ac{SVFMM}, thanks to the proposed prior density. Moreover, in order to explore the capabilities of the proposed prior densities to yield accurate segmentations, we have also computed the Dice coefficients for the segmentations based only on the maximization of the prior probability maps generated by each method. As Table \ref{table:nlsvfmm_dice_BRATS} shows, the NLv-\ac{SVFMM} method, followed by the NLp-\ac{SVFMM}, achieves the best results. Of course the Dice coefficients are significantly low because the segmentations do not take into account the pixel intensities. However, the aim is to evaluate which method better captures the local similarities in the images, hence producing more accurate prior information about the different segments in the image.

\begin{table}[ht]
\caption{Results on the Dice coefficient over the 25 synthetic high-grade gliomas of the \ac{BRATS} 2013 dataset for each algorithm evaluated. Segmentations based on the maximization of the posterior and prior probabilities are shown.}
\centering
\resizebox{\textwidth}{!}{
\rowcolors{2}{white}{gray!10}
\begin{tabular}{ccccccccccccc}
	\hline
	\rowcolor{gray!25}
	& \multicolumn{3}{c}{\ac{SVFMM}} & \multicolumn{3}{c}{$\mathcal{S}$t-\ac{SVFMM}} & \multicolumn{3}{c}{NLv-\ac{SVFMM}} & \multicolumn{3}{c}{NLp-\ac{SVFMM}} \\
	\rowcolor{gray!25}
	\multirow{-2}{*}{Segmentation} & Mean & Median & Std. & Mean & Median & Std. & Mean & Median & Std. & Mean & Median & Std. \\
	Posterior & 0.7766 & 0.7680 & 0.0368 & 0.7912 & 0.7817 & 0.0384 & 0.7988 & 0.7833 & 0.0397 & \textbf{0.8044} & \textbf{0.7936} & 0.0351 \\
	Prior     & 0.2231 & 0.2247 & 0.0092 & 0.2467 & 0.2461 & 0.0103 & \textbf{0.2576} & \textbf{0.2578} & 0.0113 & 0.2494          & 0.2495 & 0.0106 \\
	\hline
\end{tabular}
}
\label{table:nlsvfmm_dice_BRATS}
\end{table}

\subsection{Evaluation on the Berkeley Segmentation Dataset}
\label{subsection:nlsvfmm_results_berkeley}
In addition to the synthetic evaluation, we have assessed the performance of the proposed model with the 300 real-world images of the Berkeley Segmentation dataset. In our experimentation, we have employed a 3-dimensional feature vector to represent each each pixel of the images, comprising the 3 channels of the L*a*b color space. We also applied a local median smoothing to each channel using a $5 \times 5$ window centered at each pixel. We have evaluated the performance of each algorithm for different values of $K = \left\lbrace 3,5,7,10,15,20 \right\rbrace$.

Table \ref{table:nlsvfmm_rand_index} shows the results for the evaluation of the 300 images of the Berkeley dataset. \ac{RI} is employed to measure the degree of concordance between the automated segmentation and the manual segmentations. Our experiments show that the proposed methods perform favorably in terms of \ac{RI} to the other approaches in almost all situations. The NLv-\ac{SVFMM} performs comparable to the $\mathcal{S}$t variant in most of cases, achieving very similar results. However, the NLv-\ac{SVFMM} requires less parameters, hence reducing the degrees of freedom of the model. Nevertheless, the NLp-\ac{SVFMM} method achieves, both in average and median cases, the best results in most cases. Only in the $K = 3$ case (the simplest segmentation), the \ac{SVFMM} method outperforms the rest of the models. However, as segmentation complexity increases, the models including edge preserving priors performs better in all cases.

\begin{table}[ht]
\caption{Results on the \ac{RI} over the 300 images of the Berkeley dataset for each algorithm evaluated.}
\centering
\resizebox{\textwidth}{!}{
\rowcolors{2}{white}{gray!10}
\begin{tabular}{cccccccccccccccc}
	\hline
	\rowcolor{gray!25}
	\multicolumn{4}{c}{\ac{SVFMM}} & \multicolumn{4}{c}{$\mathcal{S}$t-\ac{SVFMM}} & \multicolumn{4}{c}{NLv-\ac{SVFMM}} & \multicolumn{4}{c}{NLp-\ac{SVFMM}} \\
	\rowcolor{gray!25}
	K  & Mean   & Median & Std. & K  & Mean & Median & Std. & K  & Mean & Median & Std. & K  & Mean & Median & Std. \\
	3  & \textbf{0.6952} & \textbf{0.6915} & 0.0986 & 3  & 0.6941 & 0.6891 & 0.0988 & 3  & 0.6940 & 0.6891 & 0.0988 & 3  & 0.6944 & 0.6897 & 0.0987 \\
	5  & 0.7274 & 0.7478 & 0.1086 & 5  & 0.7284 & \textbf{0.7482} & 0.1086 & 5  & 0.7284 & \textbf{0.7482} & 0.1085 & 5  & \textbf{0.7288} & 0.7480 & 0.1086 \\
	7  & 0.7283 & 0.7585 & 0.1208 & 7  & 0.7312 & 0.7596 & 0.1207 & 7  & 0.7313 & 0.7597 & 0.1206 & 7  & \textbf{0.7316} & \textbf{0.7599} & 0.1207 \\
	10 & 0.7250 & 0.7618 & 0.1335 & 10 & 0.7281 & 0.7632 & 0.1333 & 10 & 0.7283 & 0.7634 & 0.1334 & 10 & \textbf{0.7288} & \textbf{0.7639} & 0.1334 \\
	15 & 0.7184 & 0.7585 & 0.1431 & 15 & 0.7215 & 0.7594 & 0.1428 & 15 & 0.7214 & 0.7595 & 0.1428 & 15 & \textbf{0.7221} & \textbf{0.7612} & 0.1429 \\
	20 & 0.7136 & 0.7495 & 0.1479 & 20 & 0.7161 & 0.7538 & 0.1478 & 20 & 0.7162 & 0.7538 & 0.1478 & 20 & \textbf{0.7166} & \textbf{0.7545} & 0.1479 \\
	\hline
\end{tabular}
}
\label{table:nlsvfmm_rand_index}
\end{table}

\begin{figure}[ht]
\centering
\includegraphics[width=0.95\linewidth]{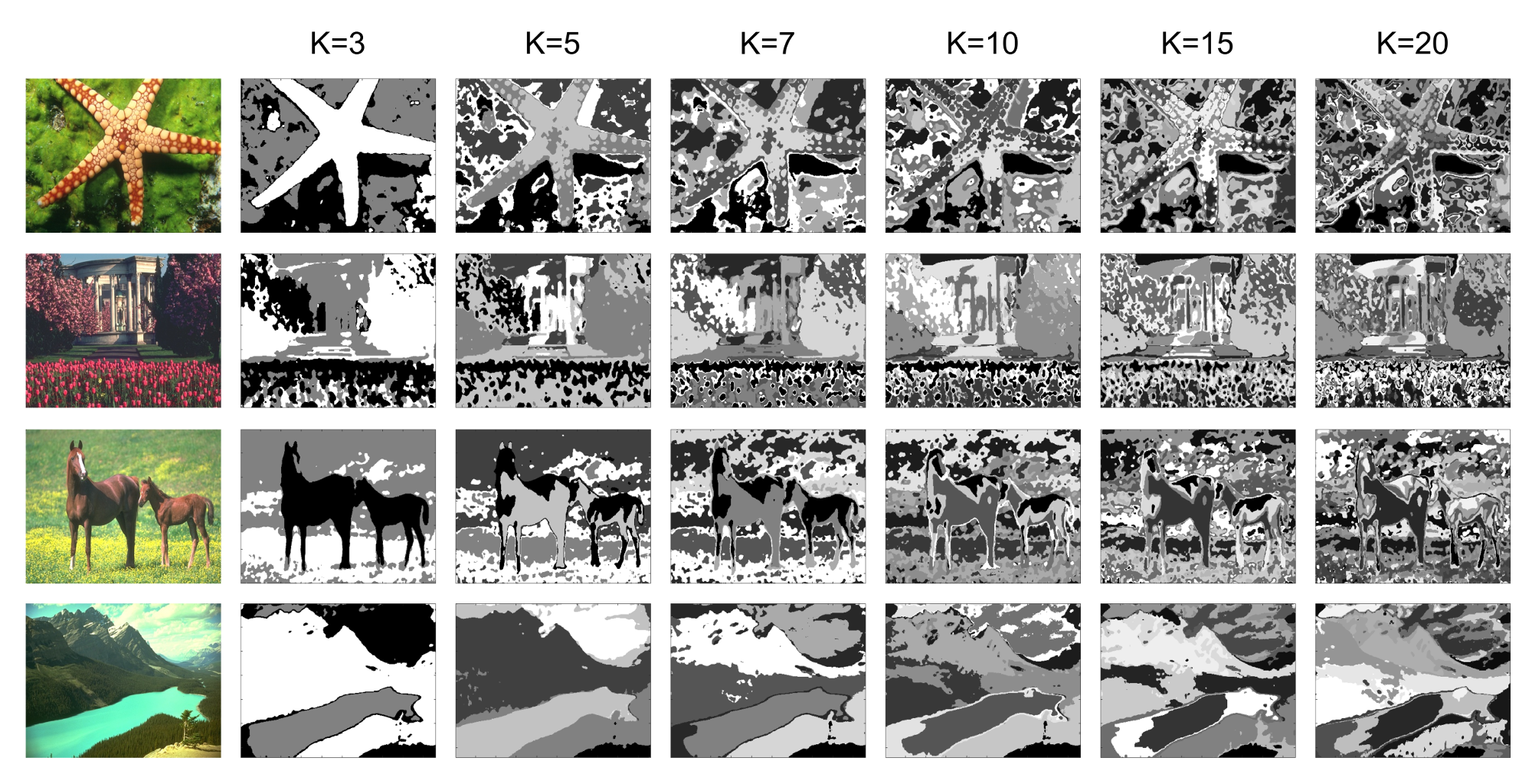}
\caption{Example of segmentation maps for $K \in \left\lbrace 3,5,7,10,15,20 \right\rbrace$ obtained with the NLp-\ac{SVFMM} for 4 images of the Berkeley dataset.}
\label{figure:nlsvfmm_examples}
\end{figure}

Figure \ref{figure:nlsvfmm_examples} shows several examples of segmentations of images of the Berkeley dataset obtained with the NLp-\ac{SVFMM} method.

\subsection{Evaluation of computational time requirements}
\label{subsection:nlsvfmm_results_time}
Additionally, a comparison in terms of the computational time required by each method has been performed. Table \ref{table:nlsvfmm_times} shows the average times (in seconds) and the std. deviation of each method evaluated in the Berkeley 300 dataset for different number of segments calculated in the images.

\begin{table}[!htpb]
\caption{Average and std. deviation time comparison (in seconds) for each algorithm evaluated in the study on the Berkeley 300 dataset for each number of segments computed in the images.}
\centering
\rowcolors{2}{white}{gray!10}
\begin{tabular}{lcccc}
	\hline
	\rowcolor{gray!25}
	K   & \ac{SVFMM}      & $\mathcal{S}$t-\ac{SVFMM} & NLv-\ac{SVFMM}  & NLp-\ac{SVFMM} \\
	3   & $0.91 \pm 0.04$ & $1.48 \pm 0.06$      & $1.73 \pm 0.06$ & $2.01 \pm 0.09$ \\
	5   & $1.39 \pm 0.08$ & $2.27 \pm 0.12$      & $2.56 \pm 0.13$ & $2.99 \pm 0.13$ \\
	7   & $1.87 \pm 0.08$ & $3.05 \pm 0.15$      & $3.37 \pm 0.14$ & $3.98 \pm 0.16$ \\
	10  & $2.69 \pm 0.14$ & $4.40 \pm 0.20$      & $4.92 \pm 0.21$ & $5.75 \pm 0.17$ \\
	15  & $3.91 \pm 0.17$ & $6.32 \pm 0.22$      & $6.96 \pm 0.26$ & $8.23 \pm 0.30$ \\
	20  & $5.18 \pm 0.47$ & $8.39 \pm 0.75$      & $9.27 \pm 0.82$ & $10.92 \pm 0.96$ \\
	\hline
\end{tabular}
\label{table:nlsvfmm_times}
\end{table}

As expected, the \ac{SVFMM} is the fastest method since it doesn't carry the extra computation of the weights for constrain the $\beta_{j,d}^2$ variances. It should be noted that only the NLv-\ac{SVFMM} and the $\mathcal{S}$t-\ac{SVFMM} are directly comparable since both perform the calculation of the $u$ and $g$ weights respectively, and those weights are computed pixel-wise. It can be seen that both methods perform very similar, with no significant difference between them. Although the $\mathcal{S}$t-\ac{SVFMM} model requires a numerical iterative approximation of the $\nu_{j,d}$ parameters, which is often a slow procedure, the complexity in the computation of the $g_{j,d}^{i,m}$ weights is lighter than the $u_{j,d}^{i,m}$ weights. That is the reason why the NLv-\ac{SVFMM} is a bit slower than the $\mathcal{S}$t-\ac{SVFMM}. The calculation of $u_{j,d}^{i,m}$ weights requires the computation of $NKD\left|\mathcal{M}_d\right|$ $\chi^2\left(\eta_m \right)$ pdf values, which ultimately equals or even slightly increases the computational time with respect the $g$ weights. The NLp-\ac{SVFMM} performs the best in terms of Dice and \ac{RI} scores, but also requires more time to compute the segmentation since it carries the extra computation of the patch-based similarity.

\section{Discussion}
\label{section:nlsvfmm_discussion}
In this study we have proposed a new unsupervised image clustering algorithm that successfully merges the \ac{SVFMM} framework with the well-known \ac{NLM} filtering scheme. The main advantage of this algorithm is the proposed new \ac{MRF} density over the contextual mixing proportions, which enforces local smoothness while preserving edges and the structure of the image. This \ac{MRF} improves the previously proposed $\mathcal{S}$t-\ac{MRF} both in performance and in complexity of the model by reducing the number of parameters to be estimated. Experimental results demonstrated the superiority of the proposed method with respect to previous state-of-the-art algorithms proposed in the literature when evaluated in a public reference dataset.

As figure \ref{figure:nlsvfmm_weight_functions} shows, the proposed probabilistic \ac{NLM} weighting function behaves more aggressive in the \ac{NLSVFMM} than its analogous in the $\mathcal{S}$t-\ac{SVFMM}. This shrinks or widens the covariance matrices more dramatically when there are differences (even small ones) between adjacent contextual mixing coefficients, leading to a sharper prior density. This behavior tends to estimate more radical prior maps, which hypothetically should provide stronger information during the learning and inference of the model. On the contrary, uniform flat prior probability maps rapidly fade into an uninformative element in the model. Figure \ref{figure:nlsvfmm_mixing_coefficients_maps} and table \ref{table:nlsvfmm_mixing_coefficients} corroborates this hypothesis. In figure \ref{figure:nlsvfmm_mixing_coefficients_maps}, it can be seen that the prior probability maps obtained by the \ac{NLSVFMM} variants achieve the highest degrees of confidence for the different labels among all models compared. Table \ref{table:nlsvfmm_mixing_coefficients} compares the contextual mixing coefficient values obtained by each method for different voxels corresponding to different tissues of a \ac{BRATS} case. As it can be seen, the highest values are always obtained by the proposed \ac{NLSVFMM} variants, leading to a highest degree of confidence about the presence of a specific tissue.

Results on the \ac{BRATS} synthetic dataset allowed to evaluate the performance of the methods in a controlled environment where the number of labels existing in the data is known. This enabled us to directly measure the impact of the proposed \ac{NLM}-based prior density in the results. Table \ref{table:nlsvfmm_dice_BRATS} compares the Dice scores of the segmentations obtained by each method, based on the posterior probabilities (Bayes decision rule) and only on the contextual mixing coefficient maps estimated by each method. Regarding the posterior probabilities case, which is the optimal decision rule, an improvement of about 3 points in Dice with respect to the \ac{SVFMM} and about 1 point with respect to the $\mathcal{S}$t-\ac{SVFMM} was obtained only by changing the prior density. Considering that the prior density acts as a guide during the learning and segmentation process, this improvement in final results is highly valuable given its limited impact. Similarly, results obtained by the \ac{NLSVFMM} variants only based on the prior probability maps (contextual mixing coefficient maps) outperform the other methods in a similar manner. Obviously, Dice results are significantly lower than those obtained based on the posterior probability (which takes into account also the intensities of the images), however the comparison among them demonstrates the superiority of the \ac{NLSVFMM} approach.

The evaluation performed on the Berkeley dataset also yielded favorably results in terms of \ac{RI} for the \acp{NLSVFMM} methods. In almost all cases, the \acp{NLSVFMM} variants performed superior or comparable to the $\mathcal{S}$t model, with the advantage that they introduce less parameters that must be estimated. Moreover, as expected, the patch-based NLp-\ac{SVFMM} method achieved the best results, confirming that the probabilistic \ac{NLM} weighting function better captures the local redundancy of the images. Only in the $K = 3$ case (the simplest segmentation), the \ac{SVFMM} method outperforms the rest of the models. However, as segmentation complexity increases, the models including edge preserving priors perform better in all cases.

It is worth noting that standard deviations are high and similar for all methods. This is due to several images that are intrinsically difficult to segment and present poor \ac{RI} results across all methods uniformly, rather than to a high variability in the methods themselves. To corroborate it, we measured the percentage of cases of the Berkeley300 dataset that showed \ac{RI} improvement when segmented by our methods with respect to the \ac{SVFMM} and the $\mathcal{S}$t-\ac{SVFMM}. For most of the $K$ states, our NLp-\ac{SVFMM} approach showed \ac{RI} improvement in approximately more than the 85\% of cases compared with the \ac{SVFMM}, and more than the 80\% of cases compared with the $\mathcal{S}$t-\ac{SVFMM}. This behavior indicates that there is a systematic improvement of our algorithms with respect to the previous approaches in the literature, which is not a product of random fluctuations due to the high standard deviations.

On the other side, it is also worth noting that differences in \ac{RI} are not significantly large between methods, which is a reasonably behavior. Under the Bayes' decision rule, the prior probability $p\left( \Pi \right)$ (or $p\left( A \right)$) acts as an initial \emph{degree of belief} of each label at each location of the image before observing the image, which ultimately represents a less informative distribution compared to the class conditional $p\left( X | \Theta, \Pi \right)$. Therefore, the impact on the final results of changing the prior distributions is limited, and thus the segmentation results are less affected. In addition, prior distributions become weaker as the number of observations in a problem increase, which is the case of pixel classification in an image. In those cases, variations in the prior densities also have a lesser impact in final results, which is observed in our experimentation.

In future work, we plan to study the inclusion of prior distributions for other parameters of the model, i.e. $\mu_j$ and $\Sigma_j, \quad \forall j$, to introduce prior knowledge and constraints that can help in the estimation of more accurate and realistic models.

\chapter{Vascular heterogeneity assessment of glioblastoma through the Hemodynamic Tissue Signature}
\label{chapter:hts}
Understanding glioblastoma intra- and inter-tumoral heterogeneity represents one of the most important challenges in advancing the fight against this lethal cancer. Over the years, much evidence has accumulated to suggest that this heterogeneity is highly responsible for the poor prognosis of the tumor and its resistance to effective therapies. Specifically, vascular heterogeneity has been identified as one of the most important pathological hallmarks of glioblastoma. The study of the aberrant vasculature of this lethal tumor, its hemodynamic local behavior and its angiogenesis mechanism is crucial to design new effective therapies that improve patient prognosis. However, to date, determining the extent and characteristics of this intra-tumor heterogeneity is still poorly understood. 

In this chapter we present the \acf{HTS}, an unsupervised machine learning method to describe the vascular heterogeneity of glioblastoma by means of perfusion \ac{MRI} analysis. The method analyzes the perfusion markers to automatically draw four reproducible habitats that describe the tumor vascular heterogeneity: the \acf{HAT} and \acf{LAT} habitats of the enhancing tumor, the potentially tumor \acf{IPE} and the \acf{VPE}.

The purpose of the work presented in this chapter was to assess if preoperative vascular heterogeneity of glioblastoma predicts \ac{OS} of patients undergoing standard of care treatment by using the \ac{HTS} method. To do so we conducted Kaplan-Meier and Cox proportional hazard analyses to study the prognostic potential of the \ac{HTS} habitats on a cohort of 50 retrospective patients from a local hospital. Additionally, we explored the ability of the \ac{HTS} habitats to improve the conventional prognostic models based on clinical, morphological, and demographic features.

\medskip

\emph{The contents of this chapter were published in the journal publications \citep{JuanAlbarracin2018, FusterGarcia2018}---thesis contributions C3, C4, P5 and P6.}

\section{Introduction}
\label{section:hts_introduction}
Glioblastoma heterogeneity has been identified as one of the factors responsible for the high aggressiveness of these neoplasms \citep{Lemee2015} and as a key hallmark to understanding their resistance to effective therapies \citep{Soeda2015}. Molecular characterization of glioblastomas has advanced the understanding of the biology and heterogeneity of these tumors, improving routine diagnosis, prognosis, and response to therapy \citep{Parsons2008, Verhaak2010}. However, significant interest has been placed in the past years in the analysis of glioblastoma heterogeneity based on medical imaging, to discover non-invasive tumor features related to different clinical outcomes such as \ac{OS}, tumor grading or molecular sub-typing \cite{Wangaryattawanich2015}.

Glioblastoma is characterized by highly infiltrative and deeply invasive behavior \citep{Dang2010}. Strong vascular proliferation, robust angiogenesis, and extensive microvasculature heterogeneity are major pathological features that differentiate glioblastomas from low-grade gliomas \citep{Alves2011, Hardee2012, Kargiotis2006}. Such factors have been shown to have a direct effect on prognosis \citep{Hardee2012}. Therefore, the early assessment of the highly heterogeneous vascular architecture of glioblastomas could provide powerful information to improve therapeutic decision making.

\acf{DSC} \ac{MRI} has been used widely to retrieve physiologic information on glioblastoma vasculature \citep{Shah2010, Knopp1999, Lupo2005}. \ac{DSC} quantification involves the computation of the hemodynamic indexes obtained from the kinetic analysis of the \Tiis{} concentration time curves retrieved from the first pass of an intravenously injected paramagnetic contrast agent \citep{Ostergaard2005}. Section \ref{subsubsection:rationale_mri_qmri_perfusion} performs an in depth description of the techniques employed in this thesis to estimate the perfusion parametric maps from the \ac{DSC} sequence. Such perfusion indexes have demonstrated powerful capabilities for a wide range of applications such as tumor grading \citep{Law2003, Emblem2008}, neovascularization assessment \citep{Thompson2011, Tykocinski2012}, early response to treatment assessment \citep{Vidiri2012, Elmghirbi2017}, recurrence versus radionecrosis \citep{Hu2009, Barajas2009} and prediction of clinical outcome \citep{Mangla2010, Jain2014}.

Numerous studies have been focused on the analysis of pretreatment perfusion indexes to assess tumor heterogeneity \citep{Jackson2007, Liu2017a, SanzRequena2013, Ulyte2016}. The most common practice is the manual definition of \acp{ROI} within the tumor to study vascular properties that correlate with clinical outcomes. However, these manual approaches impair the reproducibility studies and the analysis of high-dimensional multiparametric \ac{MRI} data \citep{Young2007}.

An alternative novel approach to describe the heterogeneity of glioblastomas is by means of the definition of lesion sub-compartments or radiological \emph{habitats} that express a specific biological behavior observable from \ac{MRI}. Several attempts have been made in the literature to describe the glioblastoma heterogeneity through this technique \citep{Dextraze2017, Zhou2014, Lee2015, Zhou2017}. However most of them are based on morphological \ac{MRI} and classical image processing techniques such as histogram analysis, intensity thresholding, texture measurements or histogram derived features. The preoperative characterization of the vascular heterogeneity of glioblastomas through a multiparametric search of habitats drawn from an unsupervised machine learning process has not previously been well established in the literature.

In this sense, we hypothesize that vascular-related habitats obtained in the preoperative evaluation of glioblastoma are early predictors of \ac{OS} in patients who subsequently undergo standard-of-care treatment. In this work, we present the \acf{HTS} method: an unsupervised machine learning-based algorithm that delineates a set of vascular habitats within the glioblastoma obtained through a multiparametric structured clustering of morphologic and \ac{DSC} \ac{MRI} features. \ac{HTS} includes consideration of four habitats: the \acf{HAT} (the more perfused area of the enhancing tumor), the \acf{LAT} (the area of the enhancing tumor with a lower angiogenic profile), the potentially \acf{IPE} (the surrounding non-enhancing region adjacent to the tumor with elevated perfusion indexes), and the \acf{VPE} (the remaining edema with a lower perfusion profile). 

To determine whether the preoperative vascular heterogeneity of glioblastoma allows early prediction of \ac{OS} of patients who undergo standard-of-care treatment, we conducted a survival analysis on the basis of perfusion measures obtained from the \ac{HTS} habitats. In addition, we also studied the contribution of the \ac{HTS} habitats to improve the estimation of \ac{OS} with respect to models based solely on clinical, morphological and demographic variables; and models including perfusion markers measured from the conventional enhancing tumor and edema \acp{ROI} instead of \ac{HTS} habitats.

\section{Materials}
\label{section:hts_materials}

\subsection{Patient selection}
\label{subsection:hts_patient_selection}
Our institutional review board approved this retrospective study, and the requirement for patient informed consent was waived. Eighty-four patients from January 2012 to December 2016 with suspected glioblastoma were included. The inclusion criteria were: (a) confirmation of glioblastoma through biopsy; (b) access to preoperative \ac{MRI} examinations, including unenhanced and \ac{GBCA}–enhanced \Ti{}-weighted, \Tii{}-weighted, \ac{FLAIR}, and \ac{DSC} sequences; and (c) patients who underwent standard Stupp treatment \citep{Stupp2005}.

Of the 84 initial patients, six were excluded because of an incomplete \ac{MRI} study, three were excluded because of motion or spike artifacts on the \ac{DSC} images that prevented the quantification (gamma variate $R^2$ goodness of fit $<$ 0.95), 10 patients were excluded because of unconfirmed or unconventional glioblastomas (giant cell glioblastoma and glioblastoma with oligodendroglioma component), and 10 patients were excluded because they did not undergo resection because of their tumor location (only biopsy results available) or they did not undergo radiation therapy and chemotherapy treatment. In addition, five patients who presented with glioblastomas with contiguous leptomeningeal extensions were excluded from the study because of the inability to accurately differentiate the tumor vascularity from the reactive meningeal enhancement in the perfusion signal intensity. Finally, 50 patients constituted the study group, including 33 men with an average age of 60.94 years (range, 25–80 years); 17 women with an average age of 62.53 years (range, 36–75 years); and an overall mean age of 60.08 years (range, 25–80 years).

\subsection{Magnetic Resonance Imaging}
\label{subsection:hts_mri}
Standard-of-care examinations were obtained with 1.5-T or 3-T imagers (Signa HDxt; GE Healthcare, Waukesha, Wisconsin) with an eight-channel-array head coil. \ac{MRI} examinations included unenhanced and \ac{GBCA}–enhanced \Ti{}-weighted three-dimensional spoiled gradient-echo sequences with inversion recovery (repetition times msec/echo times msec, 6–10/2–4; matrix, $256 \times 256$; section thickness, $1.5$ mm; field of view, $24 \times 24$ cm; inversion time, $400$ msec; flip angle, $70^{\circ}–80^{\circ}$), fast spin-echo \Tii{}-weighted imaging ($3000–4000/100–110$; matrix, $256 \times 256$; section thickness, $5$ mm; field of view, $21.9 \times 21.9$ cm; one signal acquired; intersection gap, $2$ mm) and a \ac{FLAIR} sequence ($8000–9000/140–165$; matrix, $256 \times 192$; section thickness, $5$ mm; field of view, $22 \times 22$ cm; one signal acquired; intersection gap, $2$ mm; inversion time, $2200$ msec).

\begin{table}[ht!]
\caption{Summary of the most relevant parameters of the \ac{MR} studies employed in the study.}
\centering
\rowcolors{2}{gray!10}{white}
\resizebox{\textwidth}{!}{
\begin{tabular}{lccccc}
	\hline
	\rowcolor{gray!25}
	\cellcolor{gray!25} & \Ti{} & \Tic{} & \Tii{} & \ac{FLAIR} & \ac{DSC} \\
	\cellcolor{gray!25} \acs{MR} techinque & \cellcolor{white}\thead{Spoiled \\ gradient-echo} & \cellcolor{white}\thead{Spoiled \\ gradient-echo} & \cellcolor{white}\thead{Fast \\ spin-echo} & \cellcolor{white}\thead{Fluid Attenuated \\ Inversion Recovery} & \cellcolor{white}Gradient-echo \\
	\cellcolor{gray!25} \acs{TE} (msec) & $6-10$ & $6-10$ & $3000-4000$ & $8000-9000$ & $25$ \\
	\cellcolor{gray!25} \acs{TR} (msec) & $2-4$ & $2-4$ & $100-110$ & $140-165$ & $2000$ \\
	\cellcolor{gray!25} Matrix (mm) & $256 \times 256$ & $256 \times 256$ & $256 \times 256$ & $256 \times 192$ & $128 \times 128$ \\
	\cellcolor{gray!25} Sect. thickness (mm) & $1.5$ & $1.5$ & $5$ & $5$ & $7$ \\
	\cellcolor{gray!25} Field of view (cm) & $24 \times 24$ & $24 \times 24$ & $21.9 \times 21.9$ & $22 \times 22$ & $14 \times 14$  \\
	\cellcolor{gray!25} Flip angle & $70^{\circ} - 80^{\circ}$ & $70^{\circ} - 80^{\circ}$ & $90^{\circ}$ & $90^{\circ}$ & $60^{\circ}$ \\
	\cellcolor{gray!25} Inv. time (msec) & $400$ & $400$ & & $2200$ & \\
	\cellcolor{gray!25} Bolus-conct. (mmol/kg) & & & & & $0.1$ \\
	\cellcolor{gray!25} Bolus-speed (mL/sec) & & & & & $5$ \\
	\cellcolor{gray!25} Number of dynamics & & & & & $40$ \\
	\hline
\end{tabular}
}
\label{table:hts_mri}
\end{table}

The \ac{DSC} \Tiis{}-weighted gradient-echo perfusion study was performed during the injection of \ac{GBCA} (Multihance; Bracco, Milan, Italy). A bolus injection of $0.1$ mmol/kg of \ac{GBCA} was administered at $5$ mL/sec by using a power injector (no pre-bolus administration). Saline solution was injected after the bolus injection. The study was performed with the following parameters: $2000/25$; matrix, $128 \times 128$ ($1.8 \times 1.8$ mm in-plane resolution); section thickness, $7$ mm; flip angle, $60^{\circ}$; $14$ cm full-coverage cranio-caudal ($20$ sections), $40$ sequential temporally equidistant volumes, each one with an acquisition time of $2$ seconds. The baseline before injection of the bolus was five dynamics. Table \ref{table:hts_mri} summarizes the aforementioned parameters.

\section{Methods}
\label{section:hts_methods}

\subsection{Quantification of DSC parametric maps}
\label{subsection:hts_qpwi}
\ac{DSC} \Tiis{}-weighted quantification involves the computation of the hemodynamic indices obtained from a kinetic analysis of the first pass of a intravenously injected paramagnetic contrast agent \citep{Ostergaard2005}. A detailed explanation of the calculation of perfusion parametric maps is performed in section \ref{subsection:rationale_mri_qmri}. In order for this chapter to be self-contained, a short remainder will be made.

Previous to \ac{DSC} \Tiis{}-weighted quantification, registration of the sequence was performed to the morphologic \Tii{}-weighted image of the patient. Next, signal intensity curves were converted to concentration time curves by using the following equation \ref{eq:rationale_qpwi_AR2}. Contrast material leakage correction was performed by using the technique proposed by \cite{Boxerman2006} and recirculation was corrected by means of gamma-variate curve fitting (please refer to section \ref{subsubsection:rationale_mri_qmri_perfusion} for more details). 

\ac{rCBV} and \ac{rCBF} maps were obtained using standard algorithms previously described \citep{Knutsson2010}. \ac{rCBV} was computed according to numerical integration of the area under the curve of the gamma-variate fittings \citep{Knutsson2010}; while \ac{rCBF} was calculated by means of the block-circulant \ac{SVD} deconvolution technique proposed by \cite{Wu2003}. Both perfusion maps were normalized against the contra-lateral unaffected white matter value.

The \ac{AIF} was automatically selected by following a divide-and-conquer approach. The method recursively dichotomizes the set of concentration time curves of the perfusion study into two groups, selecting those curves with higher peak height, earliest time to peak, and lowest full-width at half maximum. We used the median as a threshold to split the groups. The process is repeated until 10 or fewer curves are conserved, finally fixing the \ac{AIF} as the average of those curves.

\subsection{Enhancing tumor and edema segmentation}
\label{subsection:hts_segmentation}
In this work the enhancing tumor and edema \ac{ROI} delineation was performed by using an unsupervised segmentation method based on a variant of the work proposed in a study by \cite{JuanAlbarracin2015a}. This method is based on \ac{DCM}-\ac{SVFMM} \citep{Nikou2007}, which consists of a clustering algorithm that combines \ac{GMM} with continuous \acp{DCAGMRF} to take advantage of the self-similarity and local redundancy of the images. The method includes the unenhanced and \ac{GBCA}–enhanced \Ti{}-weighted sequences, the \Tii{}-weighted sequence, and the \ac{FLAIR} sequence in combination with atlas-based prior knowledge of healthy tissues to perform the segmentation. The automated enhancing tumor and edema \acp{ROI} obtained with the method were manually revised and validated by two experienced radiologists in consensus (F.A., with 14 years of experience; L.M.B., with 25 years of experience).

Nowadays, this method has been replaced by a more robust and powerful approach based on \acp{CNN} \citep{JuanAlbarracin2019a} (to be presented in the next chapter), which has proven to be comparable and competitive with manual segmentations performed by expert radiologists. The new method achieves a Dice score index of 0.89 on a large public real glioblastomas dataset, of more than 300 cases manually annotated by more than 3 expert radiologist each case.

\subsection{Hemodynamic Tissue Signature habitats}
\label{subsection:hts_hts}
The \ac{HTS} consists of a set of vascular habitats detected in glioblastomas and obtained by means of a multiparametric unsupervised analysis of \ac{DSC} \ac{MRI} patterns within the tumor. The technology used to compute the \ac{HTS} of the glioblastoma is publicly accessible for non-commercial research purposes at \url{https://www.oncohabitats.upv.es}.

The \ac{HTS} defines four habitats within the glioblastoma: the \acf{HAT} and \acf{LAT} habitats, and the \acf{IPE} and \acf{VPE} habitats. Table \ref{table:hts_habitats} summarizes the relationships among \ac{HTS} habitats, glioblastoma tissue, and \ac{DSC} observed vascularity.

\begin{table}[ht!]
\caption{Glioblastoma habitats considered by the \ac{HTS} an their relation with the degree of vascularity and the pathological tissue.}
\centering
\rowcolors{2}{white}{gray!10}
\begin{tabular}{lll}
	\hline
	\rowcolor{gray!25}
	\ac{HTS} habitat & Vascularity & Glioblastoma-related tissue \\
	\ac{HAT} & Highest & High-angiogenic enhancing tumor \\
	\ac{LAT} & High    & Low-angiogenic enhancing tumor \\
	\ac{IPE} & Low     & Potentially tumor infiltrated peripheral edema \\
	\ac{VPE} & Lowest  & Vasogenic peripheral edema \\
	\hline
\end{tabular}
\label{table:hts_habitats}
\end{table}

\begin{figure}[ht!]
\centering
\includegraphics[width=\linewidth]{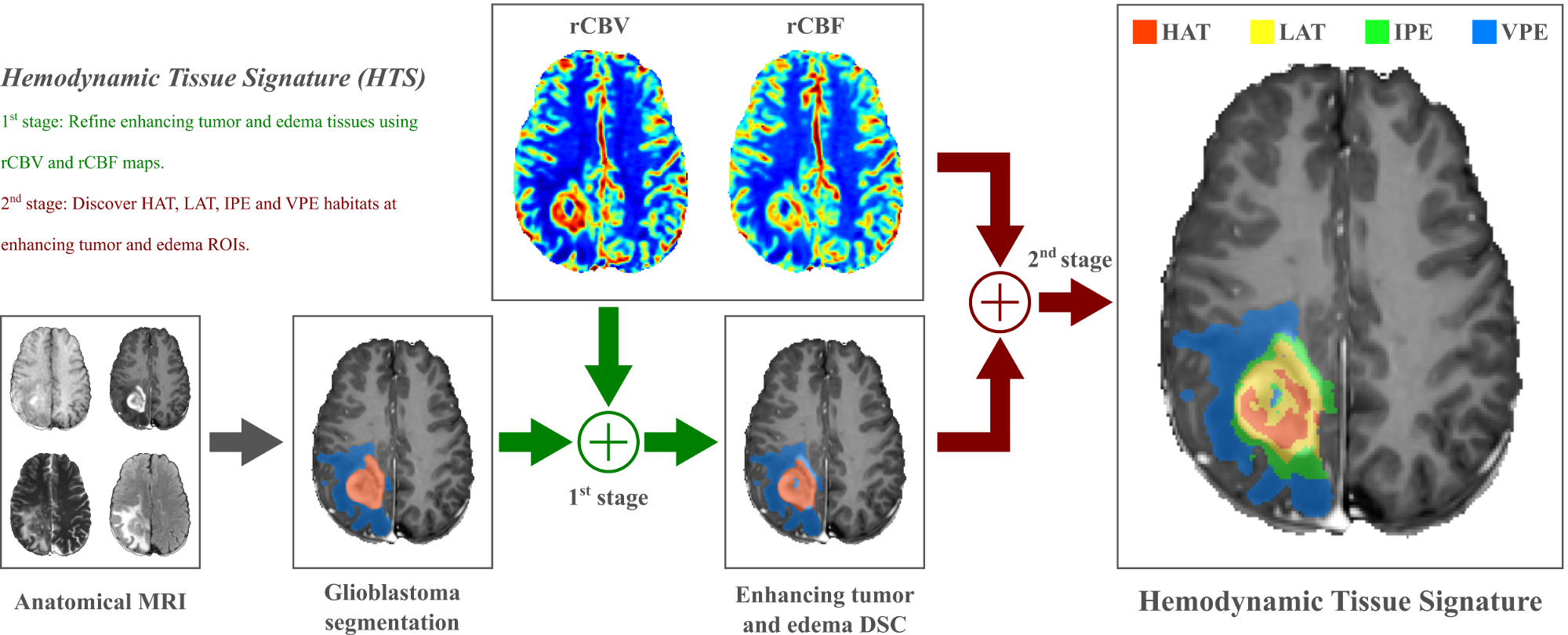}
\caption{Schema of the methodology used to compute the \ac{HTS} of the glioblastoma. \ac{HAT} = high-angiogenic tumor; \ac{LAT} = low-angiogenic tumor, \ac{IPE} = inflitrated peripheral edema, \ac{VPE} = vasogenic peripheral edema}
\label{figure:hts_hts}
\end{figure}

\ac{HTS} habitats are obtained by means of a \ac{DCM}-\ac{SVFMM} structured clustering (with the \ac{DCAGMRF} prior) of \ac{rCBV} and \ac{rCBF} maps. The clustering consists of two stages (see figure \ref{figure:hts_hts}): (a) a two-class clustering of the \ac{rCBV} and \ac{rCBF} data at the whole enhancing tumor and edema \acp{ROI} defined morphologically; and (b) a two-class clustering also of the \ac{rCBV} and \ac{rCBF} data within each \ac{ROI} obtained in stage a). To ensure the reproducibility of the \ac{HTS}, both stages are initialized with a deterministic seed method. We fix the seeds of every two-class clustering to the extremes of the \ac{rCBV} and \ac{rCBV} distributions (5\% and 95\% percentiles, respectively, for each class).

The aim of the first stage is to refine the enhancing tumor and the edema \acp{ROI}, previously delineated through the anatomical \ac{MRI} segmentation, but introducing the perfusion information. Therefore, a spatially varying mixture of two components is fit to the distributions of \ac{rCBV} and \ac{rCBF} observed in the regions previously labeled as enhancing tumor and edema. We named these two components: enhancing tumor at \ac{DSC} ($ET_{DSC}$) and edema at \ac{DSC} ($ED_{DSC}$). During the fitting process, we introduce several constraints to avoid misclassifications of nearby healthy vascular structures $ET_{DSC}$. Thus, we constrained the apparition of the $ET_{DSC}$ class to a neighborhood of less than 1 cm around the enhancing tumor observed on the \ac{GBCA}–enhanced \Ti{}-weighted \ac{MRI} \citep{Guo2016}. This constraint allows the correction of misalignments during the \ac{DSC} registration and the removal of healthy vascular structures far from the enhancing area of the tumor, which may distort the \ac{HTS}. Moreover, the also enforce the $ET_{DSC}$ class to explain at least the 80\% of the enhancing tumor \ac{ROI} obtained from the anatomical segmentation.

The second stage includes \ac{DCM}-\ac{SVFMM} clustering within the $ET_{DSC}$ and $ED_{DSC}$ \acp{ROI}, to delineate the two potential hemodynamic habitats inside each tissue. Likewise, the first stage, a spatially varying mixture of two components is fit to the distributions of \ac{rCBV} and \ac{rCBF} for both the $ET_{DSC}$ and $ED_{DSC}$ \acp{ROI}. In this stage, several constraints are also introduced. First, we force a minimum size of the habitats of at least 10\% of the whole lesion \ac{ROI} to avoid habitat vanishing. Second, infiltrated peripheral edema habitat is also constrained to a nearby region around the $ET_{DSC}$ class. Following the definition of the clinical target volume proposed by \cite{Guo2016}, we fix a 2 cm margin around the enhancing tumor observed at \ac{GBCA}-enhanced \Ti{}-weighted \ac{MRI} as the maximum distance where the infiltrated peripheral edema habitat could appear. Voxels classified as infiltrated peripheral edema outside this region are automatically removed from the \ac{HTS}, since they have a similar vascular pattern to that of infiltrated peripheral edema but far from the plausible region of tumor infiltration.

\subsection{Statistical analysis}
\label{subsection:hts_statistical_analysis}
All analyses were performed with software (Matlab R2015a; MathWorks, Natick, Massachusetts) on a personal workstation.

First, an evaluation of the statistical differences between habitats was conducted to assess the degree of separability of the \ac{rCBV} and \ac{rCBF} distributions within each habitat to confirm their different hemodynamic activity. To do so, global probabilistic deviation \citep{Saez2017} was used as a metric to control the degree of concordance among several statistical distributions. Such a metric is bounded to a $\left[0, 1\right]$ range, with 0 referring to absolute overlapping between distributions, and 1 indicating non-overlapped, completely separated distributions. Therefore, for each perfusion parameter, the distributions of the four proposed habitats and their global probabilistic deviation metric were calculated.

Second, Cox proportional hazards modeling was conducted to investigate the relationship between patient survival and the $rCBV_{max}$ and $rCBF_{max}$ at the \ac{HTS} habitats. We used the maximum of the perfusion parameters because it has been reported to be the most reliable measure for inter-observer and intra-observer reproducibility \citep{Wetzel2002}. Proportional hazard ratios with 95\% confidence intervals were reported, while the Wald test was used to determine the significance of the Cox regression model results.

Finally, Kaplan-Meier survival analyses between populations dichotomized according to the median value of each perfusion biomarker at each \ac{HTS} habitat were also conducted. A log-rank test was used to determine the statistical significance of the differences in observed population survival. Average survival for each population was also reported. Benjamini-Hochberg false discovery rate correction at an $\alpha$ level of .05 was used to correct for multiple hypothesis testing \citep{Benjamini1995} in all analyses.

On the other side, we also conducted a study to determine the added value of \ac{HTS} habitats for predicting patient \ac{OS} when these are added to classical models based on clinical, demographic and morphological variables. To do so, first we conducted an analysis to measure the importance of each clinical, demographic or \ac{MRI}-related variable independently to predict \ac{OS}. For the categorical variables we conducted Kaplan-Meier survival analysis, while for the continuous variables Cox proportional hazards modeling was used.

Next, we studied the added value of the \ac{HTS} habitats by adding its information to the models built with the variables that showed significant association with \ac{OS} in the aforementioned independent study. To do so we constructed three models:

\begin{description}[itemsep=0.5pt, topsep=0pt]
\item[\textbf{Model 1:}] Clinical variables + demographic variables
\item[\textbf{Model 2:}] Clinical variables + demographic variables + $rCBV_{max}$ and $rCBF_{max}$ at enhancing tumor
\item[\textbf{Model 3:}] Clinical variables + demographic variables + $rCBV_{max}$ and $rCBF_{max}$ at \ac{HTS} habitats
\end{description}

Cox proportional hazard regression models were fit for each model and a comparison between the predicted \ac{OS} and the real patient's \ac{OS} was performed. A Kaplan-Meier survival study was next performed by using Cox the predicted \ac{OS} of each model to split the population in two groups: long-survivors, whose Cox predicted \ac{OS} was greater than the real average \ac{OS} of the population (402 days); and short-survivors, whose Cox predicted survival was less or equal than the real average \ac{OS} of the population (402 days). \ac{RMSE} was employed to measure the deviance between the predicted \ac{OS} and the real \ac{OS} of the population. Single-tailed Wilcoxon paired signed rank test was used to determine if there are stastistically significant differences between predicted \acp{RMSE} among models.


\section{Results}
\label{section:hts_results}
\begin{figure}[ht!]
\centering
\includegraphics[width=0.9\linewidth]{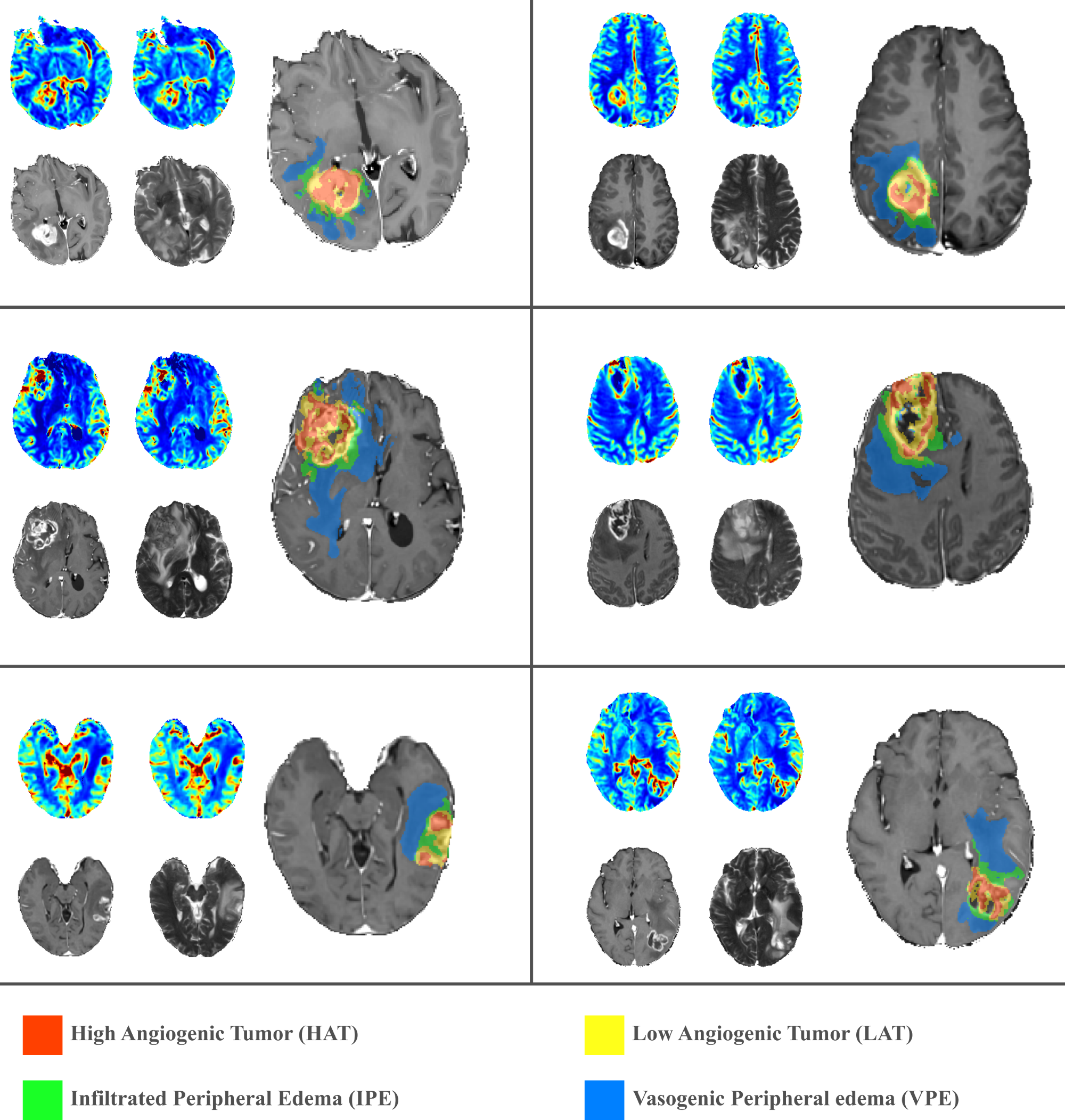}
\caption{Examples of the \ac{HTS} habitat's map placed over the \ac{GBCA}-enhanced \Ti{}-weighted \ac{MR} image for six patients. In addition, for each patient, \ac{GBCA}-enhanced \Ti{}-weighted \ac{MR} image, \Tii{}-weighted \ac{MR} image, \ac{rCBV} map, \ac{rCBF} map, are also shown in small images (left to right, top to bottom).}
\label{figure:hts_examples}
\end{figure}

\begin{figure}[ht!]
\centering
\includegraphics[width=0.8\linewidth]{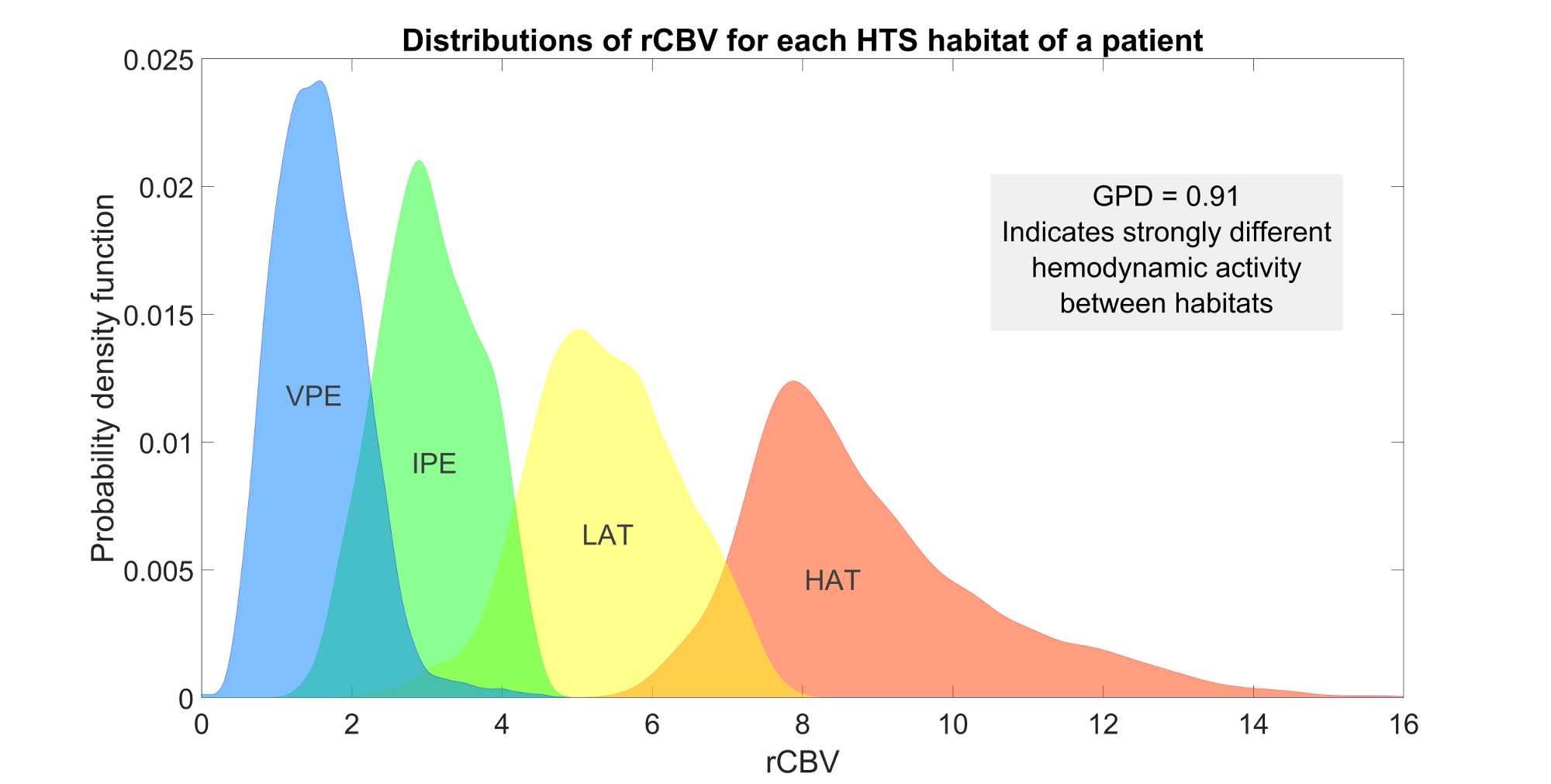}
\caption{Example of the \ac{rCBV} distributions at each \ac{HTS} habitat for a 62-year-old man and the global probabilistic deviation score obtained in the assessment of the separability of the distributions. \ac{HAT} = high-angiogenic tumor, \ac{LAT} = low-angiogenic tumor, \ac{IPE} = infiltrated peripheral edema, \ac{VPE} = vasogenic peripheral edema.}
\label{figure:hts_gpd}
\end{figure}

Figure \ref{figure:hts_examples} shows examples of \ac{HTS} maps. \ac{GBCA}-enhanced \Ti{}-weighted and \Tii{}-weighted images, as well as \ac{rCBV} and \ac{rCBF} maps are shown with the \ac{HTS} map of the patient. The global probabilistic deviation analysis of the different hemodynamic activity among habitats yielded the following average results: $0.88 \pm 0.03$ for \ac{rCBV} and $0.86 \pm 0.05$ for \ac{rCBF}. These results indicate separated perfusion distributions between habitats of the patients. Figure \ref{figure:hts_gpd} shows an example of the \ac{rCBV} distributions for each \ac{HTS} habitat. The Cox proportional hazard analysis is presented in Table \ref{table:hts_cox}.

\begin{table}[ht!]
\caption{Cox proportional hazard analysis for maximum perfusion indexes at the \ac{HTS} habitats to predict patient's \ac{OS}.}
\centering
\rowcolors{2}{white}{gray!10}
\begin{threeparttable}
\begin{tabular}{lll}
	\hline
	\rowcolor{gray!25}
	Habitat and perfusion parameter & Hazard ratio & P-value\textsuperscript{*} \\
	High-angiogenic tumor & & \\
	\hspace{1em}$rCBV_{max}$ & $1.22 \left( 1.10, 1.35 \right)$ & $.0004^{\dagger}$ \\
	\hspace{1em}$rCBF_{max}$ & $1.20 \left( 1.09, 1.32 \right)$ & $.0004^{\dagger}$ \\
	Low-angiogenic tumor & & \\
	\hspace{1em}$rCBV_{max}$ & $1.62 \left( 1.31, 2.01 \right)$ & $.0001^{\dagger}$ \\
	\hspace{1em}$rCBF_{max}$ & $1.89 \left( 1.35, 2.66 \right)$ & $.0005^{\dagger}$ \\
	Infiltrated peripheral edema & & \\
	\hspace{1em}$rCBV_{max}$ & $1.67 \left( 1.05, 2.66 \right)$ & $.0498^{\dagger}$ \\
	\hspace{1em}$rCBF_{max}$ & $2.07 \left( 1.02, 4.20 \right)$ & $.0579$ \\
	Vasogenic peripheral edema & & \\
	\hspace{1em}$rCBV_{max}$ & $1.59 \left( 0.94, 2.70 \right)$ & $.0962$ \\
	\hspace{1em}$rCBF_{max}$ & $1.58 \left( 0.71, 3.54 \right)$ & $.2657$ \\
	\hline
\end{tabular}
\begin{tablenotes}
	\small
	\item Data in parentheses are 95\% confidence intervals.
	\item * False discovery rate corrected.
	\item $\dagger$ Indicates a significant difference.
\end{tablenotes}
\end{threeparttable}
\label{table:hts_cox}
\end{table}

\begin{figure}[ht!]
\centering
\includegraphics[width=0.9\linewidth]{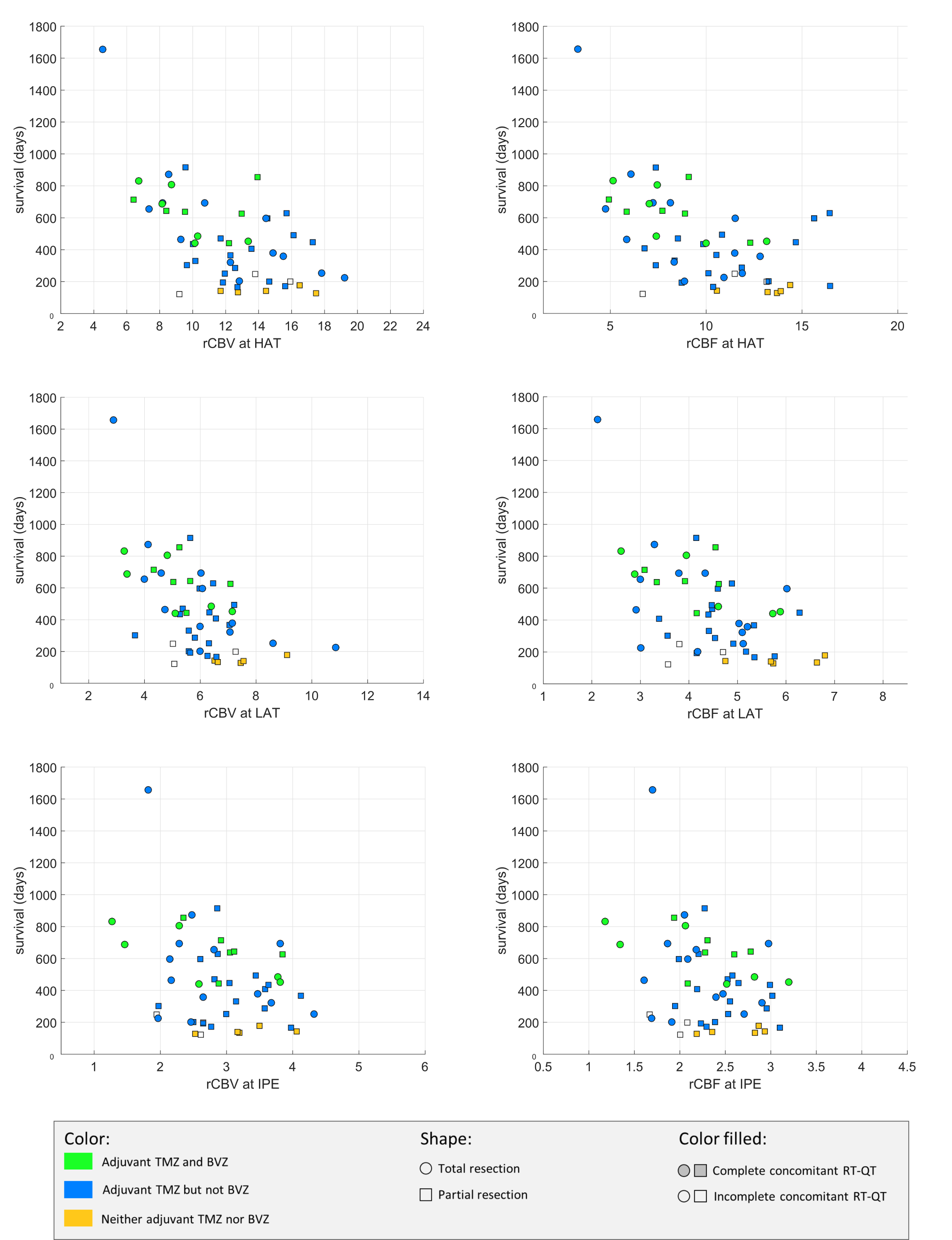}
\caption{Scatterplots of the relation between patient survival and perfusion biomarkers in \acf{HAT}, \acf{LAT} and \acf{IPE} habitats. Treatment undergone by each patient is also shown, to allow differentiation between total versus partial resection, complete versus incomplete concomitant radiation therapy and chemotherapy, adjuvant temozolomide and bevacizumab administration. BVZ = bevacizumab, RT-QT = radiation therapy-chemotherapy, TMZ = temozolomide.}
\label{figure:hts_cox}
\end{figure}

Significant results were obtained for $rCBV_{max}$ and $rCBF_{max}$ in the high-angiogenic habitat (hazard ratios, $1.22$ $\left[P = .0004\right]$ and $1.20$ $\left[P = .0004\right]$, respectively), $rCBV_{max}$ and $rCBF_{max}$ in the low-angiogenic habitat (hazard ratios, $1.62$ $\left[P = .0001\right]$ and $1.89$ $\left[P = .0005\right]$, respectively) and $rCBV_{max}$ in the infiltrated peripheral edema habitat (hazard ratio, $1.67$ $\left[P = .498\right]$). Non-significant results were obtained for $rCBF_{max}$ in the infiltrated peripheral edema (hazard ratio, $2.07$ $\left[P = .0579\right]$), and $rCBV_{max}$ in the vasogenic peripheral edema habitat (hazard ratio, $1.59$ $\left[P = .0962\right]$) and $rCBF_{max}$ at vasogenic peripheral edema (hazard ratio, $1.58$ $\left[P = .2657\right]$). Figure \ref{figure:hts_cox} shows the scatterplots of the combinations of perfusion biomarkers and \ac{HTS} habitats that yielded significant correlation in the Cox survival analysis. Total versus partial maximum safe resection, complete versus incomplete concomitant radiation therapy and chemotherapy and adjuvant temozolomide plus bevacizumab administration are also shown.

\begin{figure}[ht!]
\centering
\includegraphics[width=0.9\linewidth]{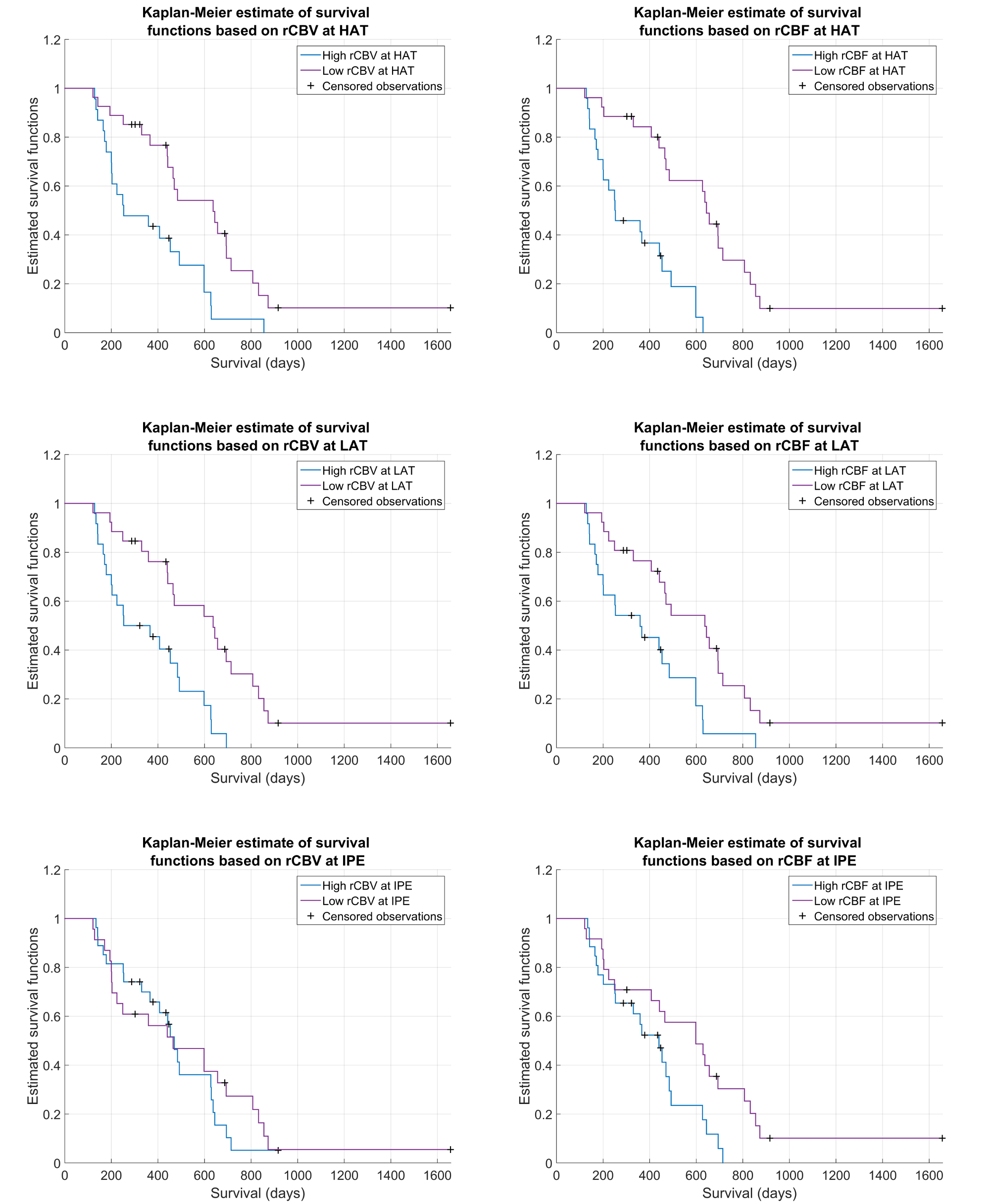}
\caption{Kaplan-Meier estimated survival functions for the populations divided by the median $rCBV_{max}$ or $rCBF_{max}$ in \acf{HAT}, \acf{LAT} and \acf{IPE} habitats.}
\label{figure:hts_kaplan}
\end{figure}

Kaplan-Meier survival analysis yielded significant differences for the survival times observed for the populations dichotomized by low and high $rCBV_{max}$ in the high-angiogenic habitat (log-rank test $P = .0104$), $rCBF_{max}$ in the high-angiogenic habitat (log-rank test $P = .0003$), $rCBV_{max}$ at low-angiogenic habitat (log-rank test $P = .0048$) and $rCBF_{max}$ in the low-angiogenic habitat (log-rank test $P = .0128$). An average difference of 230 days in overall survival between populations was observed. Mean survival of the population was 459 days $\pm$ 286.15 (range, 121–1656 days). Table \ref{table:hts_kaplan} shows the average observed survival times for each population and the corrected P-values for the log-rank survival test. Figure \ref{figure:hts_kaplan} demonstrates the Kaplan-Meier estimated survival functions for the different populations dichotomized according to the $rCBV_{max}$ and $rCBF_{max}$ at different \ac{HTS} habitats.

\begin{table}[ht!]
\caption{Kaplan-Meier survival analysis for populations with high versus low maximum perfusion parameters at the different habitats.}
\centering
\rowcolors{2}{white}{gray!10}
\begin{threeparttable}
\begin{tabular}{llll}
	\hline
	\rowcolor{gray!25}
	& \multicolumn{2}{c}{Average survival (days)} & \\ \cline{2-3}
	\rowcolor{gray!25}
	Habitat and perfusion parameter & Low & High & P-value\textsuperscript{*} \\
	High-angiogenic tumor & & & \\
	\hspace{1em}$rCBV_{max}$ & $550.33$ & $351.78$ & $.0104^{\dagger}$ \\
	\hspace{1em}$rCBF_{max}$ & $594.73$ & $311.96$ & $.0003^{\dagger}$ \\
	Low-angiogenic tumor & & & \\
	\hspace{1em}$rCBV_{max}$ & $571.62$ & $337$    & $.0048^{\dagger}$ \\
	\hspace{1em}$rCBF_{max}$ & $554.96$ & $355.04$ & $.0128^{\dagger}$ \\
	Infiltrated peripheral edema & & & \\
	\hspace{1em}$rCBV_{max}$ & $500.52$ & $423.63$ & $.7986$ \\
	\hspace{1em}$rCBF_{max}$ & $557.17$ & $368.38$ & $.0641$ \\
	Vasogenic peripheral edema & & & \\
	\hspace{1em}$rCBV_{max}$ & $532.56$ & $385.44$ & $.1300$ \\
	\hspace{1em}$rCBF_{max}$ & $500.92$ & $420.31$ & $.8992$ \\
	\hline
\end{tabular}
\begin{tablenotes}
	\small
	\item * False discovery rate corrected.
	\item $\dagger$ Indicates a significant difference.
\end{tablenotes}
\end{threeparttable}
\label{table:hts_kaplan}
\end{table}

Tables \ref{table:hts_uniparametric_categorical} and \ref{table:hts_uniparametric_continuous} show the log-rank test and the Cox-Wald test to measure the potential of each categorical and additional continuous variable to independently correlate with patient \ac{OS}.

\begin{table}[ht!]
\caption{Kaplan-Meier survival analysis for clinical and demographic variables included in the study.}
\centering
\rowcolors{2}{white}{gray!10}
\begin{threeparttable}
\begin{tabular}{llll}
	\hline
	\rowcolor{gray!25}
	Variable & \# Population & Average survival & P-value\textsuperscript{*} \\
	Gender & $\left[36, 24 \right]$ & $\left[437, 349 \right]$ & $0.522$ \\
	Laterality & $\left[24, 33 \right]$ & $\left[487, 371 \right]$ & $0.525$ \\
	Resection & & & \\
	\hspace{1em}Total & $\left[20, 40 \right]$ & $\left[562, 322 \right]$ & $.025^{\dagger}$ \\
	\hspace{1em}Subtotal & $\left[34, 26 \right]$ & $\left[356, 462 \right]$ & $.262$ \\
	\hspace{1em}Biopsy & $\left[6, 54 \right]$ & $\left[129, 432 \right]$ & $.001^{\dagger}$ \\
	Distance to ventricles & & & \\	
	\hspace{1em}Long & $\left[20, 40 \right]$ & $\left[575, 316 \right]$ & $.002^{\dagger}$ \\
	\hspace{1em}Mid & $\left[21, 39 \right]$ & $\left[391, 408 \right]$ & $.957$ \\
	\hspace{1em}Short & $\left[19, 41 \right]$ & $\left[233, 481 \right]$ & $3\mathrm{e}{-5}^{\dagger}$ \\
	Location & & & \\	
	\hspace{1em}Frontal & $\left[14, 46 \right]$ & $\left[299, 433 \right]$ & $.028^{\dagger}$ \\
	\hspace{1em}Parietal & $\left[14, 46 \right]$ & $\left[465, 383 \right]$ & $.217$ \\
	\hspace{1em}Temporal & $\left[27, 33 \right]$ & $\left[391, 411 \right]$ & $.962$ \\
	\hspace{1em}Occipital & $\left[5, 55 \right]$ & $\left[573, 386 \right]$ & $.525$ \\
	Radiochemotherapy & & & \\	
	\hspace{1em}Complete & $\left[49, 11 \right]$ & $\left[465, 122 \right]$ & $3\mathrm{e}{-11}^{\dagger}$ \\
	\hspace{1em}Incomplete & $\left[5, 55 \right]$ & $\left[173, 423 \right]$ & $.006^{\dagger}$ \\
	\hline
\end{tabular}
\begin{tablenotes}
	\small
	\item Note that this study was carried out with 10 patients more than the previous studies.
	\item * False discovery rate corrected.
	\item $\dagger$ Indicates a significant difference.
\end{tablenotes}
\end{threeparttable}
\label{table:hts_uniparametric_categorical}
\end{table}

\begin{table}[ht!]
\caption{Cox proportional hazard analysis for continuous variables included in the study.}
\small
\centering
\rowcolors{2}{white}{gray!10}
\begin{threeparttable}
\begin{tabular}{lll}
	\hline
	\rowcolor{gray!25}
	Variable & Hazard ratio & P-value\textsuperscript{*} \\
	Age & $1.08 \left( 1.01, 1.07 \right)$ & $.007^{\dagger}$ \\
	Enhancing tumor & & \\
	\hspace{1em}$rCBV_{max}$ & $1.23 \left( 1.10, 1.37 \right)$ & $.0004^{\dagger}$ \\
	\hspace{1em}$rCBF_{max}$ & $1.24 \left( 1.11, 1.37 \right)$ & $8\mathrm{e}{-5}^{\dagger}$ \\
	\hspace{1em}Volumetry & $1.02 \left( 1.00, 1.03 \right)$ & $.012^{\dagger}$ \\
	Edema & & \\
	\hspace{1em}$rCBV_{max}$ & $1.20 \left( 0.94, 1.54 \right)$ & $.134$ \\
	\hspace{1em}$rCBF_{max}$ & $1.28 \left( 0.93, 1.75 \right)$ & $.127$ \\
	\hspace{1em}Volumetry & $1.00 \left( 0.99, 1.01 \right)$ & $.979$ \\
	\ac{HAT} & & \\
	\hspace{1em}$rCBV_{max}$ & $1.14 \left( 1.06, 1.23 \right)$ & $6\mathrm{e}{-4}^{\dagger}$ \\
	\hspace{1em}$rCBF_{max}$ & $1.16 \left( 1.08, 1.39 \right)$ & $4\mathrm{e}{-5}^{\dagger}$ \\
	\hspace{1em}Volumetry & $1.06 \left( 1.01, 1.11 \right)$ & $.011^{\dagger}$ \\
	\ac{LAT} & & \\
	\hspace{1em}$rCBV_{max}$ & $1.28 \left( 1.07, 1.52 \right)$ & $.007^{\dagger}$ \\
	\hspace{1em}$rCBF_{max}$ & $1.44 \left( 1.07, 1.93 \right)$ & $.015^{\dagger}$ \\
	\hspace{1em}Volumetry & $1.03 \left( 1.01, 1.06 \right)$ & $.006^{\dagger}$ \\
	\ac{IPE} & & \\
	\hspace{1em}$rCBV_{max}$ & $1.89 \left( 1.07, 3.34 \right)$ & $.027^{\dagger}$ \\
	\hspace{1em}$rCBF_{max}$ & $2.57 \left( 1.12, 5.91 \right)$ & $.027^{\dagger}$ \\
	\hspace{1em}Volumetry & $1.01 \left( 0.98, 1.05 \right)$ & $.401$ \\
	\ac{VPE} & & \\
	\hspace{1em}$rCBV_{max}$ & $1.84 \left( 0.99, 3.42 \right)$ & $.052$ \\
	\hspace{1em}$rCBF_{max}$ & $2.31 \left( 0.95, 5.64 \right)$ & $.065$ \\
	\hspace{1em}Volumetry & $1.00 \left( 0.99, 1.01 \right)$ & $.917$ \\
	\hline
\end{tabular}
\begin{tablenotes}
	\small
	\item Note that this study was carried out with 10 patients more than the previous studies.
	\item Data in parentheses are 95\% confidence intervals.
	\item * False discovery rate corrected.
	\item $\dagger$ Indicates a significant difference.
\end{tablenotes}
\end{threeparttable}
\label{table:hts_uniparametric_continuous}
\end{table}

\begin{figure}[ht!]
\centering
\includegraphics[width=0.95\linewidth]{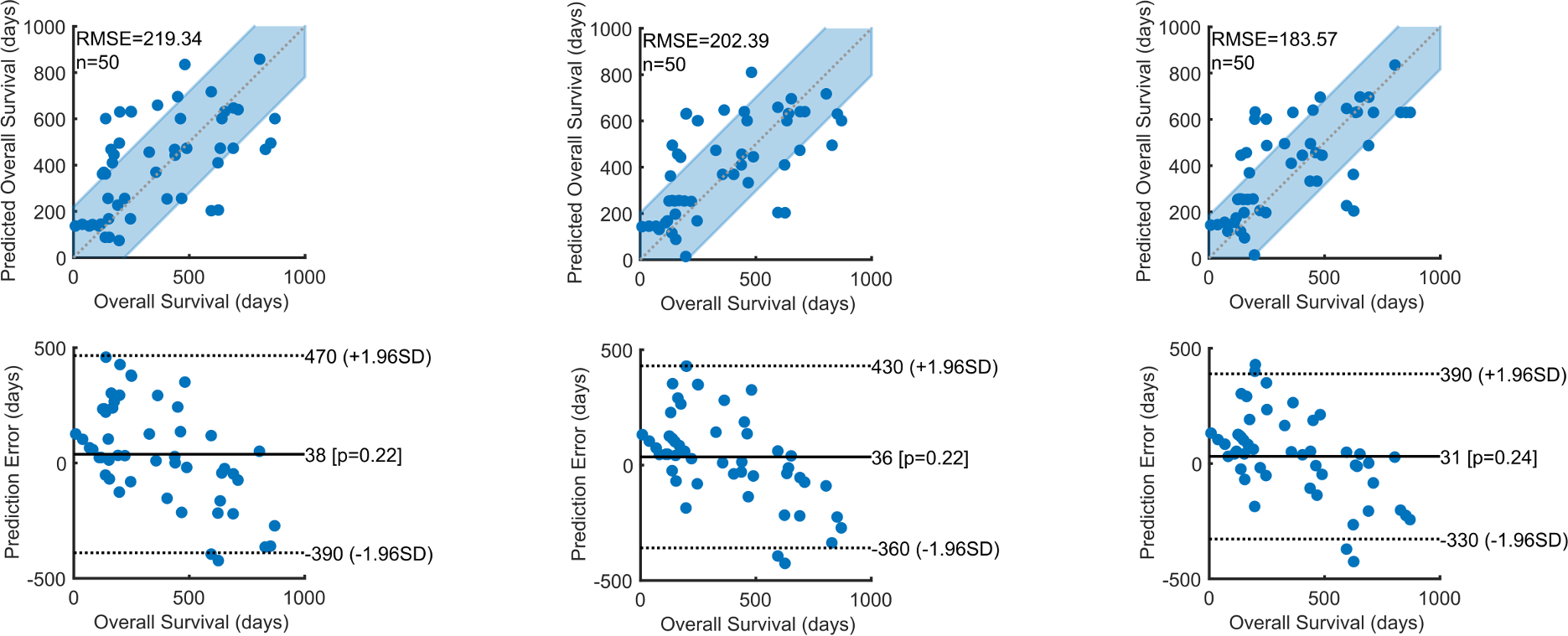}
\caption{Scatter plot for predicted versus real \ac{OS} (top) and Bland-Altman plot (bottom) for the proposed aforementioned models: Model 1 (left), Model 2 (middle), and Model 3 (right). Red dashed line indicates the hypothetical perfect prediction, while shadow blue bands indicate the confidence intervals of the prediction.}
\label{figure:hts_prediction}
\end{figure}

Figure \ref{figure:hts_prediction} shows the \ac{OS} predictions generated by Models 1, 2 and 3 during the leave-one-out evaluation. Model 1, based solely on clinical and demographic variables, obtained a $RMSE = 219.33$; Model 2, based on clinical, demographic and perfusion markers at enhancing tumor region, obtained a $RMSE = 202.39$; while Model 3, which was based on clinical, demographic and perfusion markers measured from the \ac{HTS} habitats, obtained a $RMSE = 183.57$.

The improvement achieved in the prognostic estimation obtained by the Model 2 with respect the Model 1 was 7.7\%, while the improvement obtained by the Model 3, which incorporates the \ac{HTS} habitats, reached 16.3\% in terms of \ac{RMSE}. Wilcoxon paired signed test yielded no significant differences between Model 1 and Model 2, however it did obtained statistically significant differences between Model 3 and Model 1 $\left( P < .05\right)$.

\section{Discussion}
\label{section:hts_discussion}
In this study, we investigated whether the perfusion heterogeneity in the four vascular habitats of the \ac{HTS} is predictive of survival in untreated glioblastomas. Our results demonstrate that the preoperative perfusion heterogeneity contains relevant information about patient survival, even considering the effect of other known relevant factors such as standard-of-care treatment. Gradually longer survival times were found for patients who presented with lower preoperative perfusion indexes in different \ac{HTS} habitats. The influence of standard-of-care treatment on patient survival was also directly observed. As expected, patients who underwent maximal safe resection plus concomitant adjuvant chemotherapy and radiation therapy showed better survival times. However, a tendency of longer survival times within subgroups of patients who underwent the same specific treatment and had lower perfusion indexes at several habitats was also observed. This indicates that preoperative perfusion heterogeneity contains early important information about patient survival.

Cox proportional hazard analysis substantiated these conclusions. High- and low-angiogenic habitats arose as those with the highest prognostic abilities, yielding significant correlations between survival and $rCBV_{max}$ and $rCBF_{max}$ (with multiple-test false discovery rate correction). $rCBV_{max}$ in the infiltrated peripheral edema habitat also was significantly correlated with survival, while $rCBF_{max}$ in the infiltrated peripheral edema was significantly correlated without multiple-test correction $\left(P = .0434\right)$. These results suggest that relevant information about patient survival is also contained in the peripheral edema \citep{Jain2014, Akbari2014, Artzi2014}.

Significant differences also were observed in the Kaplan-Meier estimated survival functions for populations divided according to the median $rCBV_{max}$ and $rCBF_{max}$ at several \ac{HTS} habitats. An improvement of approximately 230 days in overall survival was observed for patients who had lower $rCBV_{max}$ and $rCBF_{max}$ in the high- and low-angiogenic habitats. These results support the potential of \ac{HTS} to accurately describe the preoperative vascular heterogeneity of glioblastomas and its prognostic abilities at early stages.

Regarding the added value provided by the \ac{HTS} to predict \ac{OS} in combination with clinical and demographic variables, we found that, in agreement with the literature, variables such as age, biopsy instead of resection, long or short distance of the lesion to the ventricles, frontal tumor location, complete vs incomplete radiochemotherapy or enhancing tumor volumetry presented strong association with patient \ac{OS}.

The comparison of the prognostic capabilities of the three proposed regression models also confirmed the added value of the \ac{HTS} markers in the prognosis estimation of the patient. As expected, the models including perfusion \ac{MRI} performed better than those based solely on clinical and demographic variables. However, it is worth noting that the inclusion of perfusion related measurements from the enhancing tumor \ac{ROI} improved the prognosis estimation by 7.7\% in terms of \ac{RMSE}, however, the improvement obtained when including the \ac{HTS} perfusion-based measurements outperformed the previous model by 16.3\%. Additionally, only this latter improvement proved to be statistically significant. These results reinforce the evidence of the need for a better characterization of the heterogeneity of glioblastoma in order to improve the current management of the disease.

Several studies have been conducted to analyze the vascular heterogeneity of the glioblastoma, many of them focusing on the enhancing tumor region. \cite{Law2008} found that patients who presented with an $rCBV_{max}$ of less than 1.75 in the enhancing tumor had longer progression-free survival times; however, they did not find a significant correlation with overall survival. \cite{Sawlani2010} also correlated time to progression with several hyper-perfused regions delineated in the $rCBV$. However, they also observed no significant correlation between patients' overall survival and $rCBV_{max}$ in the enhancing tumor. \cite{Hirai2008} and \cite{Jain2013}, studied the potential for prediction of survival of $rCBV$ in the enhancing tumor of high-grade gliomas. They showed that patients who presented with an $rCBV_{max}$ of greater than 2.3 had significantly shorter survival times. These results are consistent with our findings, hence aligning \ac{HTS} with the results of previous studies in the literature. However, these studies were only based on the $rCBV_{max}$; other perfusion indexes such as $rCBF_{max}$, which may also add important information about instantaneous capillary flow in the tissues to the analysis, were not considered. Moreover, manual delineation of \acp{ROI} based on $rCBV_{max}$ was used, which may affect reproducibility and may not fully capture the tumor information available in a multiparametric study.

Authors of other studies focused on peripheral edema of glioblastoma. \cite{Akbari2014}, \cite{Jain2014} and \cite{Artzi2014} studied the peritumoral region of the glioblastoma to account for heterogeneity and possible tumor infiltration in the peripheral edema. Akbari et al used \acp{ROI} to train a support vector machine, which was then used to generate heterogeneity maps. Jain et al analyzed the association of Visually Accessible Rembrandt Images, or VASARI, features and molecular data with overall survival and progression-free survival, while Artzi et al used diffusion, perfusion, and morphologic \ac{MR} imaging with an unsupervised segmentation algorithm to analyze the edema region. Their results correlate with our findings in the infiltrated peripheral edema habitat, because they also found that vascular heterogeneity in the peripheral edema correlate with overall patient survival. However, \acp{ROI} to describe tumor heterogeneity in these studies were also delineated manually. Moreover, statistical tests were conducted without multiple comparison correction, which decreases the statistical power of the conclusions.

One of the main limitations of our study and similar studies in which authors attempted to describe the vascular heterogeneity of glioblastoma by means of discovery of new habitats was the unavailability of a ground truth for validating the habitat's segmentation \citep{Akbari2014, Hirai2008, Jain2013}. Multiple biopsy or pathological sampling could confirm the accuracy of the habitats; however, such techniques cannot always be performed in clinical practice. To overcome this limitation, alternative validation should be conducted to demonstrate the clinical relevance of the habitats. In this study, we have analyzed the relationship between the preoperative vascular heterogeneity of glioblastomas described through \ac{HTS} and patient survival.

Another important limitation was the lack of molecular markers in the population of our study. Molecular markers are currently considered a standard of care for \ac{WHO} glioblastoma classification and are also known to affect prognosis of patients with glioblastoma. Positive correlation of genetic markers with the \ac{HTS} habitats would strengthen the study and the predictive potential of the proposed method and should be performed in the future.

Finally, a limitation of the \ac{HTS} arises in the presence of highly vascularized healthy structures close to the glioblastoma, such as nearby vessels or arteries. In such cases, these structures can be misidentified as high- and low-angiogenic or infiltrated peripheral edema habitats depending on their degree of vascularity. Although \ac{HTS} implements several constraints to remove these healthy structures, nearby vessels may influence the \ac{HTS}, modifying measures obtained from the habitats. Results of future studies should improve vessel detection by using a vascular probability atlas to weight the \ac{HTS} inference process.

In conclusion, preoperative vascular heterogeneity of glioblastomas demonstrated by the habitats of \ac{HTS} is associated with patient survival. \ac{HTS} separates glioblastomas into four vascular habitats with early prognostic capabilities, offering an opportunity to define refined imaging biomarkers surrogated to clinical outcomes.

\chapter{Multi-center international validation of the \acl{HTS} for glioblastoma}
\label{chapter:validation}
In the previous chapter it was shown that early-stage vascular heterogeneity of glioblastoma has a direct effect on prognosis and a strong association with tumor aggressiveness. The proposed \acf{HTS} method provides an unsupervised \ac{ML} solution to study the vascular heterogeneity of glioblastomas by analyzing patterns of local hemodynamic activity in perfusion \ac{MRI}. The four habitats delineated by the \ac{HTS} method have demonstrated strong associations with patient \ac{OS} and early prognostic capabilities. However, the validation of the method was conducted on a single-center cohort of patients, thus avoiding the variability inherent in real-life multi-center heterogeneous scenarios.

In this chapter we present a multi-center retrospective international validation of the \ac{HTS} method. The validation was performed under the umbrella of the clinical study \href{https://clinicaltrials.gov/ct2/show/NCT03439332}{NCT03439332}, which involved seven international centers with more than 180 patients. The purpose of this chapter is to validate the association between the hemodynamic markers obtained from the \ac{HTS} habitats and the patient \ac{OS}, considering the inter-center variability of \ac{MRI} acquisition protocols, patient demographics and lesion heterogeneity. Kaplan-Meier and Cox proportional hazard analyses were conducted to study the prognostic potential of the \ac{HTS} habitats under the proposed environment.

\medskip

\emph{The contents of this chapter were published in the journal publications \citep{JuanAlbarracin2018, Alvarez2019}---thesis contributions C4, C5, P5, P6 and P7.}

\section{Introduction}
\label{section:hts_introduction}
Glioblastoma is the most aggressive malignant primary brain tumor in adults with a median survival rate of 12-15 months \citep{Louis2016, Gately2017}. It still carries a poor prognosis despite aggressive treatment, which includes tumor resection followed by chemo-radiotherapy \citep{Bae2018, Akbari2014}. One of the main factors thought to be responsible of glioblastoma aggressiveness is its vascular heterogeneity \citep{Akbari2014, Soeda2015}, mainly defined by a strong angiogenesis that supplies the glioblastoma metabolic requirements and accounts for its rapid progression \citep{Weis2011, Palma2017}. The early-vascular profile of the tumor is strongly associated with molecular characteristics of the lesion \citep{Palma2017}, which in combination with the local micro-environment are both directly related to the glioblastoma progression \citep{Weis2011}.

The negative association between patient survival rates and vascular markers extracted from perfusion \ac{MRI} has been widely demonstrated in the literature \citep{Akbari2014, Jain2014, Jensen2014}. Perfusion indexes such as \ac{rCBV} or capillary heterogeneity were found to be associated with prognosis and patient survival rates. Dozens of methodologies are proposed in the literature to assess these perfusion indexes, ranging from manually defined \acp{ROI}, which introduce high uncertainty and lack of repeatability; to more up-to-date techniques based on artificial intelligence methods able to analyze imaging patterns to describe tumor heterogeneity \citep{Demerath2017, Jena2016, Price2016, Chang2017, Cui2016}.

In 2018, \cite{JuanAlbarracin2018} proposed the \acf{HTS} method able to characterize the vascular heterogeneity of glioblastomas by means of delineating vascular habitats obtained from perfusion \ac{MRI}. The \ac{HTS} method draws four habitats within the lesion related to: the \acf{HAT} region, the \acf{LAT} region, the potentially \acf{IPE} and the \acf{VPE} habitats. The \ac{HTS} method is publicly accessible at ONCOhabitats website \url{https://www.oncohabitats.upv.es} for non-commercial research purposes.

The study conducted by \cite{JuanAlbarracin2018} found statistically significant correlations between \ac{OS} and several measures obtained from the \ac{HTS} markers. In 2018, \cite{FusterGarcia2018} demonstrated the ability of these imaging markers to improve the prognosis of conventional models based on clinical, morphological and demographic features. Both studies were conducted on a single-center cohort of 50 patients from a local institution.

However, the current road map to validate an imaging marker into clinical routine requires to overcome two translational gaps \citep{Abramson2015, Oconnor2017}: the marker validation with pre-clinical or clinical datasets from a single or a few expert centers, and the subsequent extension of the evaluation to multiple centers, along with the biological validation of the biomarkers. The aforementioned studies of \cite{JuanAlbarracin2018} and \cite{FusterGarcia2018} addressed the first translational gap, however it is still necessary to validate the \ac{HTS} markers in a multi-center heterogeneous cohort with the purpose of demonstrating their robustness and stability under highly variable clinical conditions.

The purpose of this work is to determine if the habitats obtained by the \ac{HTS} method are predictive of the \ac{OS} of glioblastoma patients undergoing standard-of-care treatment. To this end, we have involved the \ac{HTS} technology in an international multi-center observational retrospective clinical study registered at the \href{https://clinicaltrials.gov/}{ClinicalTrial.gov} official platform with name \emph{``Multicentre Validation of How Vascular Biomarkers From Tumor Can Predict the Survival of the Patient With Glioblastoma (ONCOhabitats)"} and identifier \href{https://clinicaltrials.gov/ct2/show/NCT03439332}{NCT03439332}. We have analyzed the possible association between the \ac{HTS} markers and patients \ac{OS}, as well as their capability to stratify groups of patients according to these markers, in a large heterogeneous international cohort. Additionally, we also have assessed the robustness of the \ac{HTS} method operating under a highly variable \ac{MRI} acquisition protocols from multiple centers.

\section{Materials}
\label{section:validation_materials}

\subsection{Patient selection}
\label{subsection:validation_patient_selection}
Seven European clinical centers participated in the clinical study \href{https://clinicaltrials.gov/ct2/show/NCT03439332}{NCT03439332}: the Hospital Universitario de La Ribera, Alzira, Spain; Hospital de Manises, Manises, Spain; Hospital Clinic, Barcelona, Spain; Hospital Universitario Vall d'Hebron, Barcelona, Spain; Azienda Ospedaliero-Universitaria di Parma, Parma, Italy; Centre Hospitalier Universitaire de Liege, Liege, Belgium and the Oslo University Hospital, Oslo, Norway.

A material transfer agreement document was approved by all the participating centers and an acceptance report was issued by the ethical committee of each center. The institution review board of each center also approved this retrospective study and the requirement for patient-informed consent was waived.

The inclusion criteria for patients participating in the study were: (a) adult patients (age $>$ 18 y.o.) with histopathological confirmation of glioblastoma diagnosed between January 1, 2012 and January 1, 2018; (b) access to preoperative \ac{MRI} studies, including: pre- and post-gadolinium \Ti{}-weighted, \Tii{}-weighted, \ac{FLAIR} and \ac{DSC} \Tiis{}-weighted perfusion sequences; and (c) patients who underwent standard Stupp treatment \citep{Stupp2005} with a minimum survival of 30 days.

From the initial cohort consisting of 196 patients, two cases were excluded due to incomplete \ac{DSC} perfusion acquisitions; five cases were excluded due to excessive noise in \ac{DSC} concentration curves that prevented quantification (gamma variate goodness of fit $R^2 < 0.95$); four cases were excluded due to \ac{MRI} processing errors; and one case was excluded due to inability to differentiate between tumor vascularity and reactive meningeal enhancement. Table \ref{table:validation_patients} summarizes the number of patients initially contributed by each center and the number of patients finally excluded due to noncompliance with the inclusion criteria.

\begin{table}[ht!]
\caption{Number of patients contributed by each center with their corresponding excluded due to noncompliance of the inclusion criteria.}
\centering
\rowcolors{2}{gray!10}{white}
\resizebox{\textwidth}{!}{
\begin{tabular}{lcccccccc}
	\hline
	\rowcolor{gray!25}
	& H. Ribera & H. Manises & C. Barcelona & H. Vall d'Hebron & AO Parma & CH Liege & Oslo UH & Total \\
	Initial & 10 & 14 & 28 & 34 & 42 & 34 & 34 & 196 \\
	Excluded & 3 &  0 &  3 &  1 &  2 &  1 &  2 &  15 \\
	Enrolled & 7 & 14 & 25 & 33 & 40 & 33 & 32 & 184 \\
	\hline
\end{tabular}
}
\label{table:validation_patients}
\end{table}

The final cohort enrolled in the study was of 184 patients. Those who were still alive during the study were considered as censored observations. The date of censorship was the last date of contact with the patient or, if was not available, the date of the last \ac{MRI} exam. Table \ref{table:validation_demographics} summarizes the most important demographic and clinical characteristics of the population.

\begin{table}[ht!]
\caption{Summary of demographic and clinical variables of the cohort of 184 patients enrolled in the study.}
\centering
\rowcolors{2}{gray!10}{white}
\resizebox{\textwidth}{!}{
\begin{tabular}{lcccccccc}
	\hline
	\rowcolor{gray!25}
	& H. Ribera & H. Manises & C. Barcelona & H. Vall d'Hebron & AO Parma & CH Liege & Oslo UH & Total \\
	\multicolumn{-8}{l}{Gender (F/M)} & & & & & & & & \\
	\hspace{1em} \# of patients & $6/1$ & $5/9$ & $10/15$ & $14/19$ & $12/28$ & $11/22$ & $8/24$ & $66/118$ \\
	\multicolumn{-8}{l}{Age at diagnosis (years)} & & & & & & & & \\
	\hspace{1em} Mean & $49$ & $65$ & $56$ & $60$ & $61$ & $58$ & $63$ & $60$ \\
	\hspace{1em} Range & $\left[ 24,67 \right]$ & $\left[ 39,79 \right]$ & $\left[ 35,74 \right]$ & $\left[ 30,81 \right]$ & $\left[ 35,76 \right]$ & $\left[ 32,77 \right]$ & $\left[ 40,81 \right]$ & $\left[ 24,81 \right]$ \\
	\multicolumn{-8}{l}{Survival (months)} & & & & & & & & \\
	\hspace{1em} Mean & $14.6$ & $14.4$ & $10.3$ & $15.2$ & $11.7$ & $15.3$ & $15.4$ & $13.7$ \\
	\hspace{1em} Median & $9.1$ & $12.8$ & $9.6$ & $13.0$ & $12.9$ & $14.5$ & $12.6$ & $12.6$ \\
	\hspace{1em} Range & $\left[ 3.4,52.6 \right]$ & $\left[ 3.4,38.4 \right]$ & $\left[ 1.3,26.9 \right]$ & $\left[ 4.1,40.0 \right]$ & $\left[ 1.1,30.7 \right]$ & $\left[ 2.5,41.0 \right]$ & $\left[ 3.0,36.9 \right]$ & $\left[ 1.1,52.6 \right]$ \\
	\multicolumn{-8}{l}{Resection (\# of patients)} & & & & & & & & \\
	\hspace{1em} Total & $3$ & $3$ & $0$ & $12$ & $19$ & $22$ & $11$ & $70$ \\
	\hspace{1em} Sub-total & $1$ & $4$ & $1$ & $10$ & $15$ & $6$ & $21$ & $67$ \\
	\hspace{1em} Biopsy & $1$ & $7$ & $6$ & $11$ & $2$ & $5$ & $0$ & $32$ \\
	\hspace{1em} Unknown & $2$ & $0$ & $9$ & $0$ & $4$ & $0$ & $0$ & $15$ \\
	\multicolumn{-8}{l}{Tumor location (\# of patients)} & & & & & & & & \\
	\hspace{1em} Frontal & $2$ & $4$ & $7$ & $10$ & $18$ & $11$ & $12$ & $64$ \\
	\hspace{1em} Parietal & $2$ & $0$ & $5$ & $7$ & $4$ & $9$ & $3$ & $30$ \\
	\hspace{1em} Temporal & $3$ & $7$ & $11$ & $13$ & $12$ & $9$ & $14$ & $69$ \\
	\hspace{1em} Occipital & $0$ & $2$ & $1$ & $2$ & $2$ & $0$ & $1$ & $8$ \\
	\hspace{1em} Other/Unknown & $0$ & $1$ & $1$ & $1$ & $4$ & $4$ & $2$ & $13$ \\
	\multicolumn{-8}{l}{IDH1 (\# of patients)} & & & & & & & & \\
	\hspace{1em} Mutated & $2$ & $0$ & $4$ & $0$ & $0$ & $0$ & $1$ & $6$ \\
	\hspace{1em} Wild type & $2$ & $0$ & $4$ & $32$ & $30$ & $34$ & $31$ & $99$ \\
	\hspace{1em} Unknown & $3$ & $14$ & $17$ & $1$ & $10$ & $0$ & $1$ & $79$ \\
	\hline
\end{tabular}
}
\label{table:validation_demographics}
\end{table}

\subsection{\acl{MRI}}
\label{subsection:validation_mri}
Standard-of-care MR examinations were obtained with 1.5-T or 3-T imagers. Pre- and post-\ac{GBCA} \Ti{}-weighted \ac{MRI}, as well as \Tii{}-weighted, \ac{FLAIR} and \ac{DSC} perfusion \ac{MRI} sequences were collected from each center. Table \ref{table:validation_mri} summarizes the MRI acquisition protocol employed by each center.

\begin{table}[ht!]
\caption{Summary of the most relevant parameters of the \ac{MRI} studies collected by each center. (MFS: Magnetic Field Strength)}
\centering
\rowcolors{2}{gray!10}{white}
\resizebox{\textwidth}{!}{
\begin{tabular}{lclcccccc}
	\hline
	\rowcolor{gray!25}
	Center & MFS & Sequence & TR (msec) & TE (msec) & Matrix (mm) & Sect. thickness (mm) & FOV (cm\textsuperscript{2}) & \# Dynamics \\
	\hline
	\cellcolor{gray!10} & \cellcolor{white} & \Ti{} & $25$ & $4.6$ & $268 \times 268$ & $0.9$ & $24 \times 24$ & - \\
	\cellcolor{gray!10} & \cellcolor{white} & \Tii{} & $2000$ & $120$ & $320 \times 199$ & $5.0$ & $23 \times 18.3$ & - \\
	\cellcolor{gray!10} & \cellcolor{white} & \ac{FLAIR} & $11000$ & $140$ & $256 \times 164$ & $6.0$ & $23 \times 18.3$ & - \\
	\multirow{-4}{*}{\cellcolor{gray!10}H. Ribera} & \multirow{-4}{*}{\cellcolor{white}$1.5$T} & \ac{DSC} & $1650$ & $40$ & $116 \times 116$ & $2.2$ & $24 \times 24$ & 80 \\
	\hline
	\cellcolor{gray!10} & \cellcolor{white} & \Ti{} & $500$ & $20$ & $304 \times 241$ & $5.0$ & $24 \times 24$ & - \\
	\cellcolor{gray!10} & \cellcolor{white} & \Tii{} & $2000$ & $120$ & $304 \times 228$ & $5.0$ & $24 \times 24$ & - \\
	\cellcolor{gray!10} & \cellcolor{white} & \ac{FLAIR} & $11000$ & $140$ & $256 \times 209$ & $6.0$ & $24 \times 24$ & - \\
	\multirow{-4}{*}{\cellcolor{gray!10}H. Manises} & \multirow{-4}{*}{\cellcolor{white}$1.5$T} & \ac{DSC} & $836$ & $30$ & $128 \times 128$ & $5.0$ & $24 \times 24$ & 40 \\
	\hline
	\cellcolor{gray!10} & \cellcolor{white} & \Ti{} & $12$ & $4.68$ & $256 \times 256$ & $1.0$ & $24 \times 24$ & - \\
	\cellcolor{gray!10} & \cellcolor{white} & \Tii{} & $3000$ & $80$ & $256 \times 256$ & $5.0$ & $24 \times 24$ & - \\
	\cellcolor{gray!10} & \cellcolor{white} & \ac{FLAIR} & $9000$ & $164$ & $256 \times 256$ & $5.0$ & $24 \times 24$ & - \\
	\multirow{-4}{*}{\cellcolor{gray!10}C. Barcelona} & \multirow{-4}{*}{\cellcolor{white}$3.0$T} & \ac{DSC} & $1550$ & $32$ & $128 \times 128$ & $5.0$ & $24 \times 24$ & 50 \\
	\hline
	\cellcolor{gray!10} & \cellcolor{white} & \Ti{} & $253$ & $2.64$ & $320 \times 180$ & $4.0$ & $22 \times 16.5$ & - \\
	\cellcolor{gray!10} H. Vall & \cellcolor{white} & \Tii{} & $6100$ & $91$ & $512 \times 326$ & $4.0$ & $22 \times 17.5$ & - \\
	\cellcolor{gray!10} d'Hebron & \cellcolor{white} & \ac{FLAIR} & $9000$ & $68$ & $320 \times 288$ & $4.0$ & $22 \times 19.8$ & - \\
	\cellcolor{gray!10} & \multirow{-4}{*}{\cellcolor{white}$3.0$T} & \ac{DSC} & $1450$ & $45$ & $128 \times 128$ & $5.0$ & $23 \times 23$ & 60 \\
	\hline
	\cellcolor{gray!10} & \cellcolor{white} & \Ti{} & $8.18$ & $8.18$ & $256 \times 256$ & $1.0$ & $24 \times 24$ & - \\
	\cellcolor{gray!10} & \cellcolor{white} & \Tii{} & $6500$ & $65.90$ & $160 \times 160$ & $4.0$ & $24 \times 24$ & - \\
	\cellcolor{gray!10} & \cellcolor{white} & \ac{FLAIR} & $12000$ & $96.72$ & $384 \times 224$ & $4.0$ & $24 \times 24$ & - \\
	\multirow{-4}{*}{\cellcolor{gray!10}AO. Parma} & \multirow{-4}{*}{\cellcolor{white}$3.0$T} & \ac{DSC} & $1500$ & $30$ & $128 \times 128$ & $4.0$ & $24 \times 24$ & 60 \\
	\hline
	\cellcolor{gray!10} & \cellcolor{white} & \Ti{} & $13$ & $4.76$ & $256 \times 218$ & $1.0$ & $25 \times 25$ & - \\
	\cellcolor{gray!10} & \cellcolor{white} & \Tii{} & $5000$ & $109$ & $384 \times 384$ & $5.0$ & $23 \times 23$ & - \\
	\cellcolor{gray!10} & \cellcolor{white} & \ac{FLAIR} & $9000$ & $90$ & $256 \times 173$ & $5.0$ & $23 \times 23$ & - \\
	\multirow{-4}{*}{\cellcolor{gray!10}CH. Liege} & \multirow{-4}{*}{\cellcolor{white}$1.5$T} & \ac{DSC} & $1460$ & $47$ & $128 \times 128$ & $5.0$ & $23 \times 25$ & 50 \\
	\hline
	\cellcolor{gray!10} & \cellcolor{white} & \Ti{} & $5.2$ & $2.3$ & $512 \times 512$ & $1.0$ & $25.6 \times 25.6$ & - \\
	\cellcolor{gray!10} & \cellcolor{white} & \Tii{} & $3800$ & $84$ & $896 \times 896$ & $3.0$ & $22 \times 22$ & - \\
	\cellcolor{gray!10} & \cellcolor{white} & \ac{FLAIR} & $4800$ & $325$ & $512 \times 512$ & $0.9$ & $25.6 \times 25.6$ & - \\
	\multirow{-4}{*}{\cellcolor{gray!10}Oslo UH} & \multirow{-4}{*}{\cellcolor{white}$3.0$T} & \ac{DSC} & $1500$ & $25$ & $128 \times 128$ & $5.0$ & $25.6 \times 25.6$ & 100 \\
	\hline
\end{tabular}
}
\label{table:validation_mri}
\end{table}

\section{Methods}
\label{section:validation_methods}

\subsection[Vascular heterogeneity assessment based on HTS habitats]{Vascular heterogeneity assessment of glioblastoma based on \acs{HTS} habitats}
\label{section:validation_hts}
The \ac{HTS} method, available at ONCOhabitats (\url{https://www.oncohabitats.upv.es}), was used to describe the vascular heterogeneity of the glioblastomas enrolled in the multicenter study. The methodology comprises the following stages:

\begin{enumerate}[itemsep=0.5pt, topsep=0pt]
\item \textbf{\ac{MRI} preprocessing:} including denoising, magnetic field inhomogeneity correction, multi-modal registration, brain extraction, motion correction and intensity standardization.
\item \textbf{Glioblastoma segmentation:} implementing a state-of-the-art deep learning 3D \ac{CNN} that delineates the enhancing tumor, edema and necrotic tissues.
\item \textbf{Perfusion quantification:} to calculate the parametric \ac{rCBV}, \ac{rCBF}, \ac{MTT} and K2 maps derived from the \ac{DSC} perfusion sequence.
\item \textbf{\ac{HTS} habitats:} in which an unsupervised segmentation algorithm performs the detection of the \ac{HAT}, \ac{LAT}, \ac{IPE} and \ac{VPE} habitats to describe the vascular heterogeneity within the lesion.
\end{enumerate}

According to \cite{Wetzel2002}, for each habitat we defined the \emph{\ac{HTS} marker} as the maximum \ac{rCBV} $\left(rCBV_{max}\right)$, computed as the $95^{th}$ percentile of the \ac{rCBV} distribution within the region defined by the corresponding habitat.

\subsection{Association among \acs{OS} and \acs{HTS} markers - whole cohort study}
\label{subsection:validation_whole_cohort}
Cox proportional hazard regression analysis was used to quantify the associations between patient \ac{OS} and \ac{HTS} markers. The proportional \acp{HR} with their 95\% confidence intervals were reported, as well as the associated P-values corrected for multiple-test with Benjamini-Hochberg false discovery rate correction at an $\alpha$ level of $.05$. 

Kaplan-Meier analyses were also conducted to study the survival evolution of the population stratified into two groups according to \ac{HTS} markers: the \emph{high-vascular} and the \emph{low-vascular} groups. We defined the high-vascular and low-vascular groups as the set of patients with a $rCBV_{max}$ higher or lower than an optimal cut-off threshold calculated with the C-index method. Log-rank test was used to determine the statistical differences between the estimated survival functions of the aforementioned groups.

\subsection{Association among \acs{OS} and \acs{HTS} markers - inter-center study}
\label{subsection:validation_whole_cohort}
In order to determine the degree of agreement in describing the vascular heterogeneity of glioblastomas, we conducted an study measuring the similarities of the \ac{HTS} marker distributions among the clinical centers enrolled in the study. To this end we conducted a pair-wise Mann-Whitney U-test $\left(\alpha=.05\right)$, followed by a post-hoc Tukey's honest significant difference criterion test.

Cox regression analyses were also conducted to assess whether the association between patient \ac{OS} and \ac{HTS} markers differed among the centers. Kaplan-Meier analyses were conducted after dividing the population of each center using the same cut-off thresholds previously calculated for the whole cohort study.

All the statistical analyses were performed on Matlab R2017b (MathWorks, Natick, MA).

\section{Results}
\label{section:validation_results}

\subsection{Association among \acs{OS} and \acs{HTS} markers - whole cohort study}
\label{subsection:validation_results_whole_cohort}
Table \ref{table:validation_cox_whole_cohort} summarizes the Cox proportional hazard analysis between \ac{HTS} markers and patient \ac{OS}. Statistically significant negative associations were found for $rCBV_{max}$ at \ac{HAT}, \ac{LAT} and \ac{IPE} habitats and patient \ac{OS}, with the \ac{IPE} marker showing the highest \ac{HR} $\left(1.28\right)$. Kaplan-Meier results are presented in Table \ref{table:validation_kaplan_whole_cohort}, including estimated optimal cut-off thresholds, the number of patients assigned to each group, the estimated C-index area under the curve, the median \ac{OS} calculated per group, and the log-rank P-values.

Significant differences in \ac{OS} between low and high vascular groups divided by \ac{HTS} marker values were found. Consistently with previous results in the literature \citep{JuanAlbarracin2018}, patients with a low $rCBV_{max}$ at \ac{HAT}, \ac{LAT} and \ac{IPE} habitats presented a higher median survival rate. 

\begin{table}[ht!]
\caption{Cox regression analysis for $rCBV_{max}$ at each vascular habitat and patient \ac{OS}.}
\centering
\rowcolors{2}{white}{gray!10}
\begin{threeparttable}
\begin{tabular}{ccccc}
	\hline
	\rowcolor{gray!25}
	& & 95\% confidence & & FDR-corrected \\
	\rowcolor{gray!25}
	& \multirow{-2}{*}{Hazard Ratio} & interval & \multirow{-2}{*}{P-value} & P-value \\
	$rCBV_{max}$ & & & & \\
	\hspace{1em}\ac{HAT} & $1.05$ & $\left[ 1.01,1.09 \right]$ & $0.0115^{\dagger}$ & $0.0174^{\dagger}$ \\
	\hspace{1em}\ac{LAT} & $1.11$ & $\left[ 1.02,1.20 \right]$ & $0.0131^{\dagger}$ & $0.0174^{\dagger}$ \\
	\hspace{1em}\ac{IPE} & $1.28$ & $\left[ 1.05,1.55 \right]$ & $0.0122^{\dagger}$ & $0.0174^{\dagger}$ \\
	\hspace{1em}\ac{VPE} & $1.19$ & $\left[ 0.89,1.60 \right]$ & $0.2502^{\dagger}$ & $0.2502$ \\
	\hline
\end{tabular}
\begin{tablenotes}
	\small
	\item $\dagger$ Indicates a significant difference.
\end{tablenotes}
\end{threeparttable}
\label{table:validation_cox_whole_cohort}
\end{table}

\begin{table}[ht!]
\caption{Kaplan Meier and log-rank test results for patient stratification in low- and high-vascular groups according to $rCBV_{max}$ value at the \ac{HTS} habitats.}
\centering
\rowcolors{2}{white}{gray!10}
\resizebox{\textwidth}{!}{
\begin{threeparttable}
\begin{tabular}{cccccc}
	\hline
	\rowcolor{gray!25}
	& Cut-off  & \# Patients per group & AUC & \ac{OS} per group & \\
	\rowcolor{gray!25}
	& threshold & [low, high] & C-Index & [low, high] & \multirow{-2}{*}{P-value} \\
	$rCBV_{max}$ & & & & & \\
	\hspace{1em}\ac{HAT} & $11.06$ & $\left[ 97,87 \right]$ & $0.606$ & $\left[ 14.3,11.3 \right]$ & $0.0014^{\dagger}$ \\
	\hspace{1em}\ac{LAT} & $5.31$ & $\left[ 91,93\right]$ & $0.605$ & $\left[ 13.9,11.3\right]$ & $0.0085^{\dagger}$ \\
	\hspace{1em}\ac{IPE} & $1.92$ & $\left[ 59,125 \right]$ & $0.634$ & $\left[ 14.3,11.4 \right]$ & $0.0101^{\dagger}$ \\
	\hspace{1em}\ac{VPE} & $1.67$ & $\left[ 100,84 \right]$ & $0.599$ & $\left[ 13.8,11.2 \right]$ & $0.1356$ \\
	\hline
\end{tabular}
\begin{tablenotes}
	\small
	\item AUC: Area under the curve.
	\item $\dagger$ Indicates a significant difference.
\end{tablenotes}
\end{threeparttable}
}
\label{table:validation_kaplan_whole_cohort}
\end{table}

\begin{figure}[htbp!]
\centering
\includegraphics[width=0.9\linewidth]{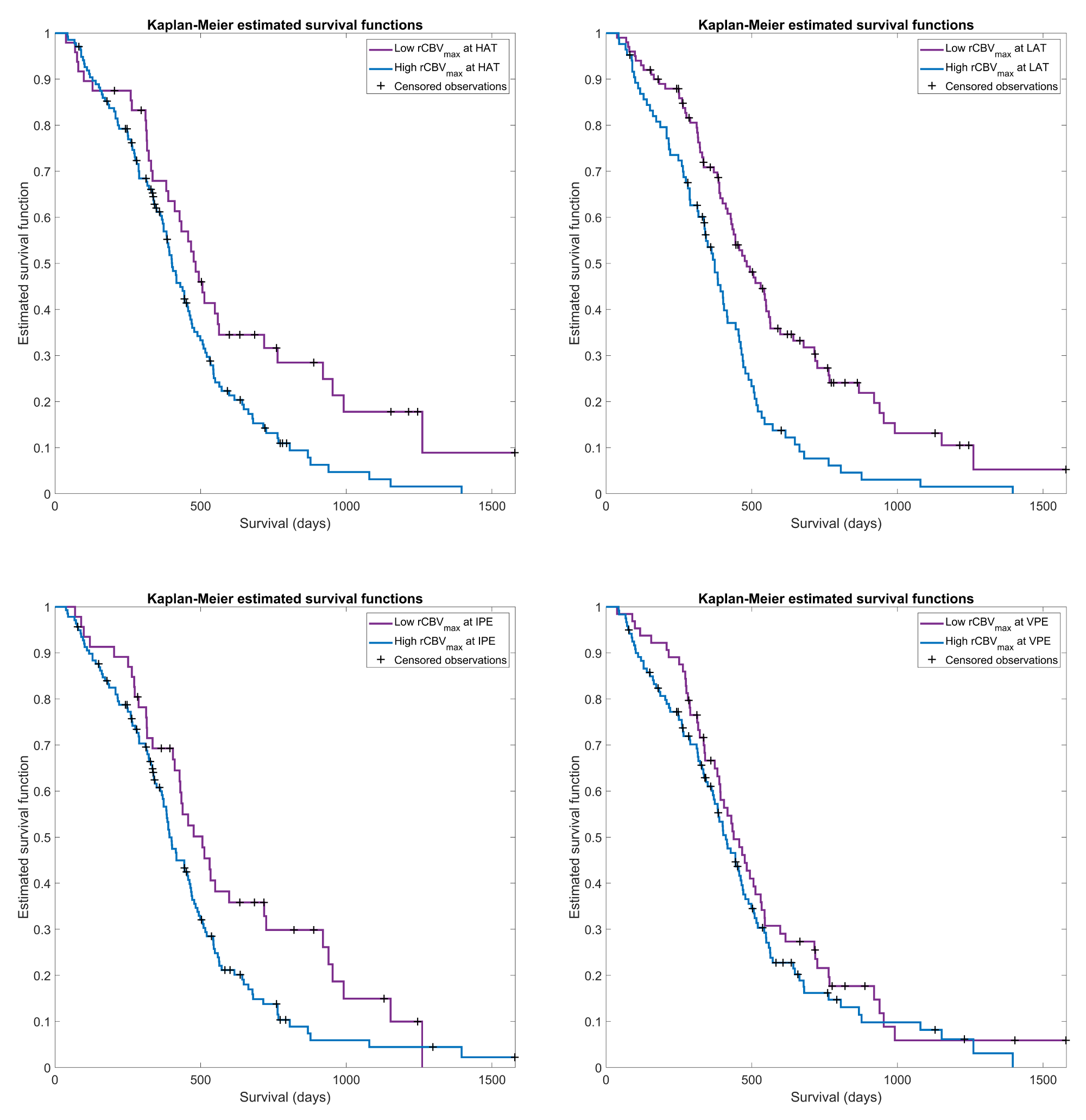}
\caption{Kaplan-Meier estimated survival functions for the populations stratified into groups according to high and low $rCBV_{max}$ at \ac{HAT} (top left), \ac{LAT} (top right), \ac{IPE} (bottom left) and \ac{VPE} (bottom right) \ac{HTS} habitats.}
\label{figure:validation_kaplan_whole_cohort}
\end{figure}

Figure \ref{figure:validation_kaplan_whole_cohort} shows the Kaplan-Meier estimated survival functions for the populations divided in high-vascular and low-vascular groups according to the optimal-cutoff threshold for the $rCBV_{max}$ estimated with the C-Index method.

\subsection{Association among \acs{OS} and \acs{HTS} markers - inter-center study}
\label{subsection:validation_results_whole_cohort}
No statistical differences were found between the $rCBV_{max}$ values at the different \ac{HTS} habitats among most of the centers, specially for the \ac{IPE} habitat, which was the most correlated with \ac{OS} in the whole cohort study of section \ref{subsection:validation_results_whole_cohort} (see tables \ref{table:validation_mann_whitney_HAT}, \ref{table:validation_mann_whitney_LAT}, \ref{table:validation_mann_whitney_IPE}, \ref{table:validation_mann_whitney_VPE}; $\dagger$ indicates statistical significant difference).

\begin{table}[ht!]
\caption{Pair-wise Mann Whitney U-test results for $rCBV_{max}$ at \ac{HAT} habitat.}
\centering
\rowcolors{2}{gray!10}{white}
\resizebox{\textwidth}{!}{
\begin{tabular}{lcccccc}
	\hline
	\rowcolor{gray!25}
	& H. Manises & C. Barcelona & H. Vall d'Hebron & AO Parma & CH Liege & Oslo UH \\
	\cellcolor{gray!25} H. Ribera & $0.3139$ & $0.6816$ & $0.1762$ & $0.1937$ & $0.4765$ & $0.0959$ \\
	\cellcolor{gray!25} H. Manises & & $0.0363^{\dagger}$ & $0.6170$ & $0.7899$ & $0.1111$ & $0.7292$ \\
	\cellcolor{gray!25} C. Barcelona & & & $0.0258^{\dagger}$ & $0.0124^{\dagger}$ & $0.2091$ & $0.0017^{\dagger}$ \\
	\cellcolor{gray!25} H. Vall d'Hebron & & & & $0.6063$ & $0.1302$ & $0.1584$ \\
	\cellcolor{gray!25} AO Parma & & & & & $0.0504$ & $0.3557$ \\
	\cellcolor{gray!25} CH Liege & & & & & & $0.0085$ \\
	\hline
\end{tabular}
}
\label{table:validation_mann_whitney_HAT}
\end{table}

\begin{table}[ht!]
\caption{Pair-wise Mann Whitney U-test results for $rCBV_{max}$ at \ac{LAT} habitat.}
\centering
\rowcolors{2}{gray!10}{white}
\resizebox{\textwidth}{!}{
\begin{tabular}{lcccccc}
	\hline
	\rowcolor{gray!25}
	& H. Manises & C. Barcelona & H. Vall d'Hebron & AO Parma & CH Liege & Oslo UH \\
	\cellcolor{gray!25} H. Ribera & $0.0932$ & $0.3384$ & $0.1445$ & $0.0914$ & $0.2401	$ & $0.0135^{\dagger}$ \\
	\cellcolor{gray!25} H. Manises & & $0.1243$ & $0.6170$ & $0.7003$ & $0.2794$ & $0.5586$ \\
	\cellcolor{gray!25} C. Barcelona & & & $0.1317$ & $0.932$ & $0.5300$ & $0.0103^{\dagger}$ \\
	\cellcolor{gray!25} H. Vall d'Hebron & & & & $0.7690$ & $0.3832$ & $0.0844$ \\
	\cellcolor{gray!25} AO Parma & & & & & $0.2423$ & $0.1392$ \\
	\cellcolor{gray!25} CH Liege & & & & & & $0.0228$ \\
	\hline
\end{tabular}
}
\label{table:validation_mann_whitney_LAT}
\end{table}

\begin{table}[ht!]
\caption{Pair-wise Mann Whitney U-test results for $rCBV_{max}$ at \ac{IPE} habitat.}
\centering
\rowcolors{2}{gray!10}{white}
\resizebox{\textwidth}{!}{
\begin{tabular}{lcccccc}
	\hline
	\rowcolor{gray!25}
	& H. Manises & C. Barcelona & H. Vall d'Hebron & AO Parma & CH Liege & Oslo UH \\
	\cellcolor{gray!25} H. Ribera & $0.0932$ & $0.0917$ & $0.1016$ & $0.2500$ & $0.1349$ & $0.1383$ \\
	\cellcolor{gray!25} H. Manises & & $0.8262$ & $0.6170$ & $0.2910$ & $0.7011$ & $0.3457$ \\
	\cellcolor{gray!25} C. Barcelona & & & $0.8752$ & $0.3771$ & $0.9624$ & $0.5253$ \\
	\cellcolor{gray!25} H. Vall d'Hebron & & & & $0.2949$ & $0.9387$ & $0.6227$ \\
	\cellcolor{gray!25} AO Parma & & & & & $0.2848$ & $0.7296$ \\
	\cellcolor{gray!25} CH Liege & & & & & & $0.6698$ \\
	\hline
\end{tabular}
}
\label{table:validation_mann_whitney_IPE}
\end{table}

\begin{table}[ht!]
\caption{Pair-wise Mann Whitney U-test results for $rCBV_{max}$ at \ac{VPE} habitat.}
\centering
\rowcolors{2}{gray!10}{white}
\resizebox{\textwidth}{!}{
\begin{tabular}{lcccccc}
	\hline
	\rowcolor{gray!25}
	& H. Manises & C. Barcelona & H. Vall d'Hebron & AO Parma & CH Liege & Oslo UH \\
	\cellcolor{gray!25} H. Ribera & $0.0676$ & $0.0754$ & $0.0299^{\dagger}$ & $0.1937$ & $0.0462^{\dagger}$ & $0.4103$ \\
	\cellcolor{gray!25} H. Manises & & $1.0000$ & $0.8615$ & $0.2030$ & $0.8433$ & $0.0678$ \\
	\cellcolor{gray!25} C. Barcelona & & & $0.8260$ & $0.2275$ & $0.9374$ & $0.0546$ \\
	\cellcolor{gray!25} H. Vall d'Hebron & & & & $0.0931$ & $0.8174$ & $0.0160^{\dagger}$ \\
	\cellcolor{gray!25} AO Parma & & & & & $0.1117$ & $0.5595$ \\
	\cellcolor{gray!25} CH Liege & & & & & & $0.0280^{\dagger}$ \\
	\hline
\end{tabular}
}
\label{table:validation_mann_whitney_VPE}
\end{table}

\begin{figure}[htbp!]
\centering
\includegraphics[width=0.9\linewidth]{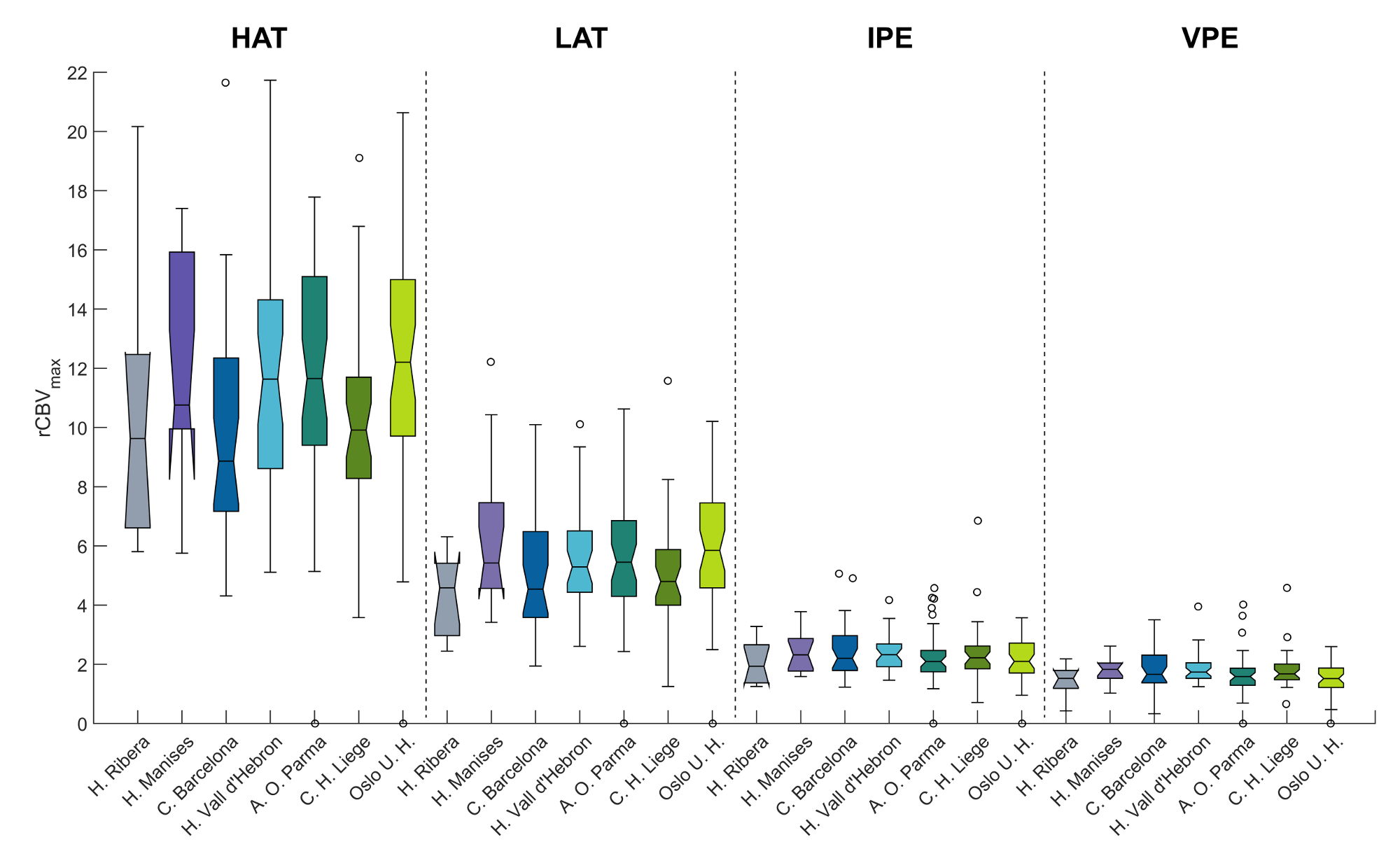}
\caption{Box-whisker plots per center of the $rCBV_{max}$ distributions at \ac{HAT}, \ac{LAT}, \ac{IPE} and \ac{VPE} habitats.}
\label{figure:validation_boxplot}
\end{figure}

Box-whisker plot shown in Figure \ref{figure:validation_boxplot} summarizes the information contained in tables \ref{table:validation_mann_whitney_HAT}, \ref{table:validation_mann_whitney_LAT}, \ref{table:validation_mann_whitney_IPE}, \ref{table:validation_mann_whitney_VPE}. Significant overlapping among the distributions of $rCBV_{max}$ of each hospital can be observed, indicating no statistical differences in the \ac{HTS} markers among centers.

Table \ref{table:validation_cox_per_center} shows the results of the Cox regression analysis grouped per hospital, to investigate the association of the \ac{HTS} markers with patient \ac{OS} at each center. Due to the small sample sizes of some centers, confidence intervals are wider, so the results of this analysis are more uncertain. However, overall, results are consistent with those obtained in the whole cohort study, consolidating the significant association between \ac{HTS} habitats and patient \ac{OS} in highly heterogeneous scenarios.

\begin{table}[ht!]
\caption{Cox regression analysis for $rCBV_{max}$ at each \ac{HTS} habitat and patient \ac{OS} per center.}
\centering
\rowcolors{2}{gray!10}{white}
\resizebox{\textwidth}{!}{
\begin{threeparttable}
\begin{tabular}{llccccccc}
	\hline
	\rowcolor{gray!25}
	& & H. Ribera & H. Manises & C. Barcelona & H. Vall d'Hebron & AO Parma & CH Liege & Oslo UH \\
	\cellcolor{gray!25} $rCBV_{max}$ & & & & & & & & \\
	\cellcolor{gray!25} & \ac{HR} & $1.00$ & $1.04$ & $1.10$ & $1.10$ & $1.07$ & $0.98$ & $1.09$ \\
	\cellcolor{gray!25} \hspace{1em}\multirow{-2}{*}{\ac{HAT}} & CI & $\left[ 0.84,1.20 \right]$ & $\left[ 0.96, 1.13 \right]$ & $\left[ 0.96, 1.28 \right]$ & $\left[ 0.98, 1.23 \right]$ & $\left[ 0.98, 1.18 \right]$ & $\left[ 0.86,1.12 \right]$ & $\left[ 1.00, 1.20 \right]$ \\
	\cellcolor{gray!25} & \ac{HR} & $1.08$ & $1.11$ & $1.07$ & $1.33$ & $1.11$ & $0.96$ & $1.15$ \\
	\cellcolor{gray!25} \hspace{1em}\multirow{-2}{*}{\ac{LAT}} & CI & $\left[ 0.64, 1.81 \right]$ & $\left[ 0.89, 1.37 \right]$ & $\left[ 0.79, 1.44 \right]$ & $\left[ 0.98, 1.80 \right]$ & $\left[ 0.95, 1.30 \right]$ & $\left[ 0.77, 1.28 \right]$ & $\left[ 0.97, 1.36 \right]$ \\
	\cellcolor{gray!25} & \ac{HR} & $1.95$ & $1.76$ & $1.01$ & $1.73$ & $1.40$ & $1.10$ & $1.54$ \\
	\cellcolor{gray!25} \hspace{1em}\multirow{-2}{*}{\ac{IPE}} & CI & $\left[ 0.5, 7.65 \right]$ & $\left[ 0.80, 3.89 \right]$ & $\left[ 0.61, 1.65 \right]$ & $\left[ 0.65, 4.61 \right]$ & $\left[ 0.97, 2.01 \right]$ & $\left[ 0.64, 1.90 \right]$ & $\left[ 0.92, 2.57 \right]$ \\
	\cellcolor{gray!25} & \ac{HR} & $2.13$ & $1.83$ & $0.92$ & $1.20$ & $1.31$ & $1.11$ & $1.67$ \\
	\cellcolor{gray!25} \hspace{1em}\multirow{-2}{*}{\ac{VPE}} & CI & $\left[ 0.12, 35.9 \right]$ & $\left[ 0.42, 7.89 \right]$ & $\left[ 0.39, 2.18 \right]$ & $\left[ 0.39, 3.68 \right]$ & $\left[ 0.81, 2.13 \right]$ & $\left[ 0.45, 2.67 \right]$ & $\left[ 0.80, 3.48 \right]$ \\
	\cellcolor{gray!25} \# patients & & 7 & 14 & 25 & 33 & 40 & 33 & 32 \\
	\hline
\end{tabular}
\begin{tablenotes}
	\small
	\item CI Indicates a Confidence Interval.
\end{tablenotes}
\end{threeparttable}
}
\label{table:validation_cox_per_center}
\end{table}

\begin{figure}[htbp!]
\centering
\includegraphics[width=0.9\linewidth]{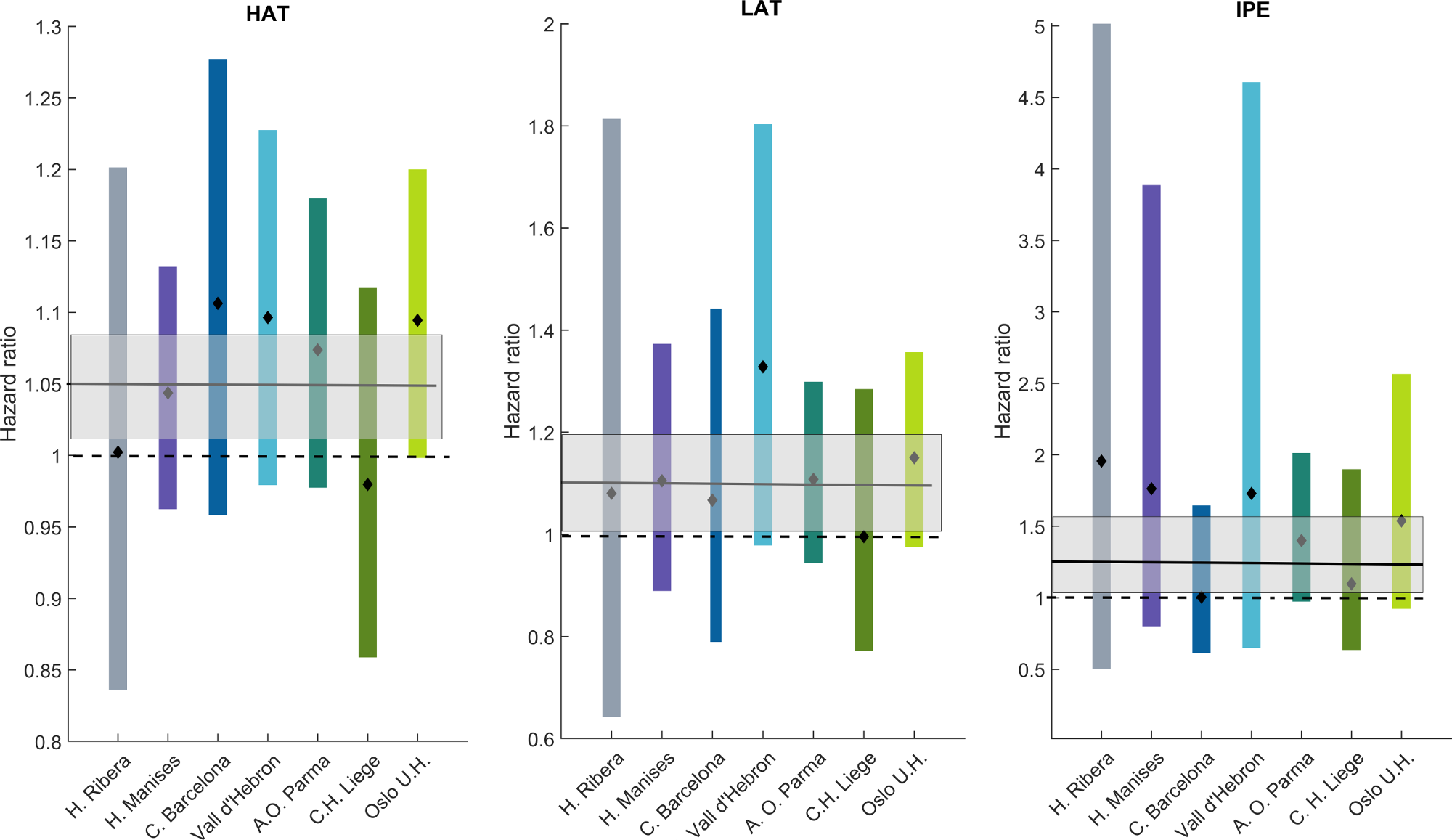}
\caption{Diagram with the \acp{HR} and 95\% confidence intervals per center, to investigate the association between \ac{OS} and \ac{HTS} markers at \ac{HAT}, \ac{LAT} and \ac{IPE} habitats. The continuous black lines and the grey bands correspond to the \acp{HR} and their associated confidence intervals using the data from all centers. Black diamond markers and colored bars represent the \ac{HR} with its confidence interval for each respective center.}
\label{figure:validation_hr}
\end{figure}

Figure \ref{figure:validation_hr} shows a diagram of the \acp{HR} and confidence intervals of the \ac{HAT}, \ac{LAT} and \ac{IPE} habitats per center, and using all the data together in the analysis. The figure shows a significant overlap between confidence intervals for most of the centers, suggesting no significant differences among them, and consolidating the results obtained in the whole cohort study.

\begin{figure}[htbp!]
\centering
\includegraphics[width=0.9\linewidth]{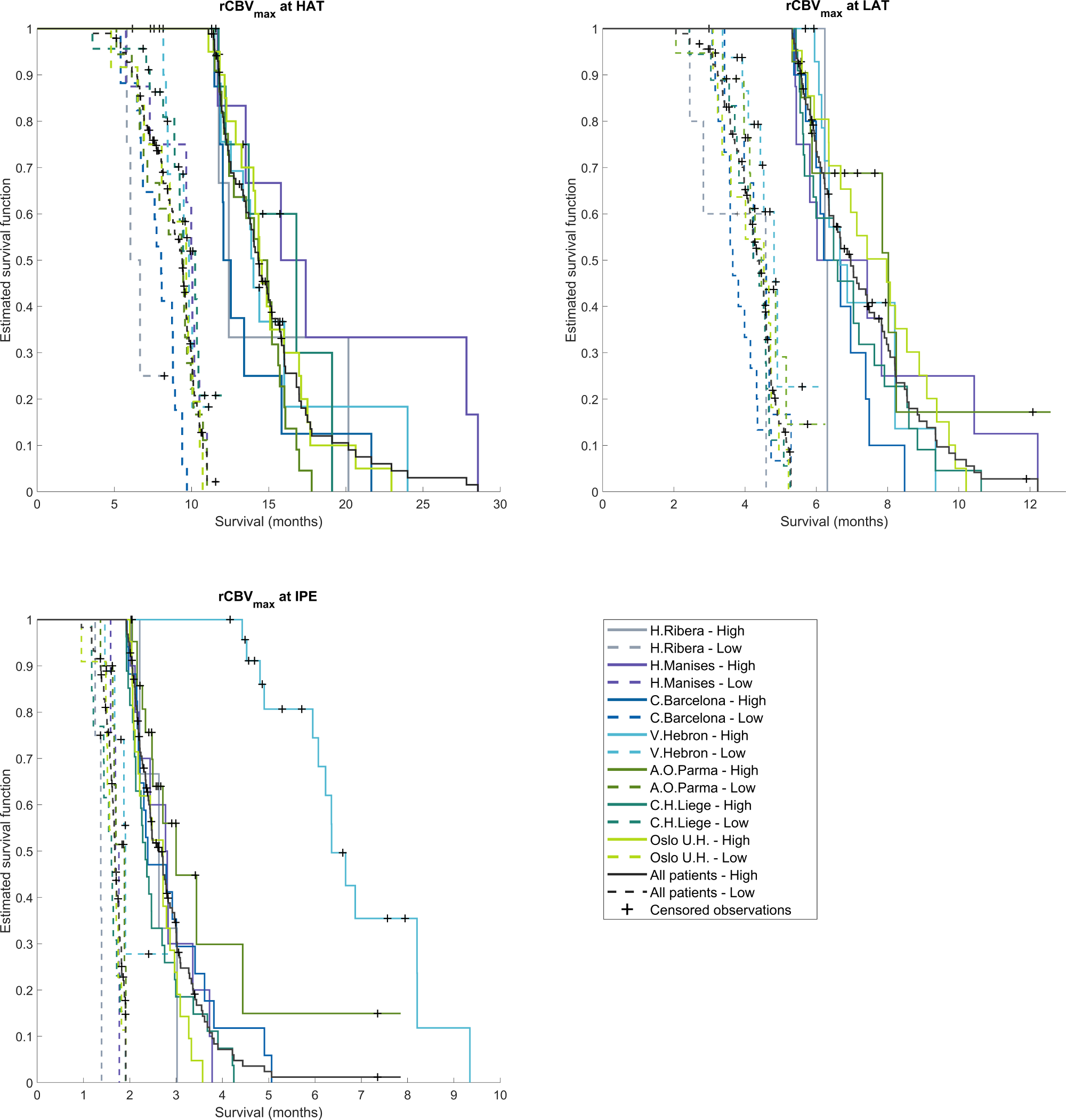}
\caption{Kaplan-Meier estimated survival functions for the populations of each center stratified into groups according to high or low $rCBV_{max}$ at \ac{HAT} (top left), \ac{LAT} (top right) and \ac{IPE} (bottom left), using the optimal thresholds calculated with the C-index method.}
\label{figure:validation_km_centers}
\end{figure}

Figure \ref{figure:validation_km_centers} show the Kaplan-Meier estimated survival functions per center, dividing the population of each center in high and low $rCBV_{max}$ values at \ac{HAT}, \ac{LAT} and \ac{IPE} habitats according to the optimal thresholds obtained with the C-index method.

\section{Discussion}
In this work we have conducted an international multi-center validation of the \ac{HTS} method published in \citep{JuanAlbarracin2018}. Using data from seven European centers, significant negative associations have been found between patient \ac{OS} and the \ac{HTS} markers at \ac{HAT}, \ac{LAT} and \ac{IPE} habitats, consolidating the results obtained in the aforementioned single-center study conducted by \cite{JuanAlbarracin2018}.
 
Addressing heterogeneity between centers in the estimation of \ac{MRI} markers is not an easy task. Several authors in the bibliography have pointed out the uncertainty and low reproducibility of \ac{MRI} markers, especially across multiple centers \citep{Abramson2015, Oconnor2017, Schnack2004, Deguio2016}. The non-quantitative nature of several \ac{MRI} acquisitions and the manual procedure for obtaining \ac{MR}-based biomarkers introduce important sources of variability, making it difficult to validate new robust and stable \ac{MRI} markers \citep{Schnack2004}.

The \ac{HTS} method focuses its efforts on the automated delineation of habitats related to perfusion patterns within the lesion in a robust and reliable manner. In the current study, a cohort with large variations in terms of patient demographics as well as \ac{MRI} acquisition protocols was used to measure the robustness of the method. The experiments conducted in our study did not show relevant differences among the distributions of the \ac{HTS} markers obtained from \ac{MRI} studies acquired at the different centers. Only for a small number of cases, significant differences were found for the \ac{HAT} and \ac{LAT} markers among centers. These results strongly suggest that the \ac{HTS} method is robust against inter-center variability in the task of describing the vascular heterogeneity of the glioblastoma. Furthermore, the results of the Cox and Kaplan-Meier analyses per center showed robust associations between patient \ac{OS} and the \ac{HTS} markers, regardless of the center of origin. The proposed thresholds were also effective in stratifying patients from different centers into low- and high-vascular groups, presenting different \ac{OS} tendencies. 

Consistently with the literature, the \ac{HTS} method strongly correlates with patient \ac{OS} when observing the $rCBV_{max}$ values at \ac{HAT} habitat. Such values represent the most hyper-perfused measures of the glioblastoma, which aligns with the measurements proposed in previous studies in the literature \citep{Jain2014, Liu2017a, Hirai2008}. As expected, shorter \ac{OS} survival rates were found for patients with higher $rCBV_{max}$ values at the \ac{HAT} habitat. Similarly, we also found that \ac{LAT} habitat present strong association with \ac{OS} and high stratification abilities. Both results have been replicated under highly variable conditions of \ac{MRI} acquisition protocols and patient demographics, hence demonstrating the robustness of the \ac{HTS} method in describing the vascular arrangement of the glioblastoma.

One of the most important finding presented in \cite{JuanAlbarracin2018} was the correlation between long-term \acp{OS} and lower $rCBV_{max}$ values in the \ac{IPE} habitat. The peritumoral region of the glioblastoma is the most heterogeneous area of the tumor, in which uncontrolled infiltration occurs. Moreover, the inter-patient variability and the inter-center heterogeneity significantly increases the uncertainty in this region, obscuring the important information that it contains. However, in the present study we found statistical association between perfusion markers at this region and patient \ac{OS}, even under the large heterogeneous nature of the proposed cohort. Moreover, effective stratification capabilities were also found when employing this marker as indicator to divide the population between high- and low-vascular glioblastomas.

Having demonstrated the influence of early-stage vascularity on the prognosis of glioblastoma, we suggest the use of this factor in any clinical study that includes population randomization. Authors consider that the \ac{HTS} method will help overcome current limitations and improve patient recruitment and randomization by initiating a route map to avoid the second translational gap cited previously in \citep{Abramson2015}.

One of the most important limitations of our study is the imbalance between the number of patients in each center. Although the whole cohort size is large enough for a powerful statistical study, some of the participating centers provided a low number of patients (less than 15 patients), which introduces limitations and uncertainties into the studies conducted per center. On the other hand, since the influence of the molecular markers in patient prognosis has been clearly demonstrated \citep{Louis2016, Verhaak2010}, it may be of interest to add them as co-factors in the regression survival models. In future studies we plan to analyze the possible association between molecular and imaging markers and their prognostic possibilities.

\chapter{ONCOhabitats: A system for glioblastoma heterogeneity assessment through \acl{MRI}}
\label{chapter:oncohabitats}
Neuroimaging analysis is currently crucial for an early assessment of glioblastoma, to help improving treatment and tumor follow-up. To this end, multiple quantitative and morphological \ac{MRI} sequences are usually employed, requiring the development of automated tools capable to extract the relevant information contained in these sources. Despite major advances in \ac{MRI} brain tumor technology, the latter is generally private and inaccessible to the research community. This significantly slows down the advances in tumor understanding, as many researchers continually have to re-implement many software pieces for the typical \ac{MRI} analysis pipeline to conduct their investigation. This task is often arduous or even unreachable to many research groups specialized on the clinical aspects of the tumor, and with less knowledge about state-of-the-art technology for \ac{MRI} analysis. As a result, many efforts are lost in developing this necessary technology before research begins.

In this thesis, several methods have been developed to analyze glioblastoma through \ac{MRI}. One of the aspects significantly taken into consideration since the beginning of this thesis was to, parallel to the academic and theoretical research, develop the required infrastructure to facilitate public access to the technology developed in the thesis. In this sense, this chapter presents ONCOhabitats (\url{https://www.oncohabitats.upv.es}): an online open access system for glioblastoma heterogeneity assessment by \ac{MRI} data. ONCOhabitats provides two services for untreated glioblastomas: 1) malignant tissue segmentation, and 2) vascular heterogeneity assessment of the tumor. The segmentation service was validated against the \ac{BRATS} 2017 reference dataset, showing comparable results with current state-of-the-art methods (\ac{WT} Dice score: 0.89). The vascular heterogeneity assessment service was validated in a retrospective cohort of 50 patients, in a study focused on predicting patient \ac{OS}. Cox proportional hazard regression analysis and Kaplan-Meier survival study showed significant positive correlations (p-value $<$ .05) between the \ac{HTS} habitats and patient \ac{OS}. ONCOhabitats system also generates radiological reports for each service, including volumetries and perfusion measurements of the different regions of the lesion. Additionally, ONCOhabitats gives access to the scientific community to a computational cluster capable to process about 300 cases per day.

\medskip

\emph{The contents of this chapter were published in the journal publication \citep{JuanAlbarracin2019b}---thesis contributions C6 and P4, P8, P9 and P10.}

\section{Introduction}
\label{section:oncohabitats_introduction}
Glioblastoma is a primary brain tumor presumed to arise from neuroglial cells. It is the most malignant and frequent astrocytoma, accounting for more than 60\% of all brain tumors in adults. Glioblastoma has a global incidence of 4.67 to 5.73 per 100,000 people, and presents a poor prognosis of 14-15 months under the best treatment \citep{Stupp2005}.

Heterogeneity is a hallmark that has been identified as crucial to understand the tumor aggressiveness and its resistance against therapies \citep{Lemee2015, Soeda2015}. Specifically, glioblastoma is characterized by a high heterogeneity both at macroscopic tissue level, with co-existence of different malignant tissues within the neoplasm \citep{Liu2017b}; as well as at microscopic cellular level, with different molecular sub-types and genetic alterations \citep{Inda2014}. Such heterogeneity rises this tumor as one of the deadliest malignant primary brain tumor in adults \citep{Ostrom2015}.

Molecular analysis of glioblastoma has largely improved the understanding of the biological heterogeneity of these tumors. Molecular profiling of glioblastoma has allowed the identification of different tumor sub-types, helping in the development of more efficient drugs \citep{Parsons2008, Verhaak2010}. However, in the past years, significant interest has been placed in the analysis of glioblastoma heterogeneity based on medical imaging, to discover non-invasive tumor features related to different outcomes such as overall survival, tumor grading or glioblastoma molecular sub-typing \citep{Wangaryattawanich2015}.

Characterization of glioblastoma heterogeneity based on \ac{MRI} has been addressed from a wide range of approaches. Glioblastoma tissue segmentation has gathered most of these efforts. Automated identification of the different tissues that co-exist in the lesion, such as enhancing tumor, non-enhancing tumor, edema and necrosis, has been largely addressed by the scientific community. This has lead to initiatives such as the \ac{BRATS} challenge, which was born in 2012 and has become the reference benchmark \citep{Menze2015} to evaluate the state-of-the-art of automated high-grade and low-grade glioma segmentation algorithms. Nowadays, with the advent of novel deep learning techniques, the current state-of-the art is mostly dominated by \ac{CNN} classifiers. \acp{CNN} are a class of deep feed-forward neural networks whose architecture is particularly well suited for computer vision recognition tasks. In medical image analysis field, \acp{CNN} have outperformed most of algorithms in many problems, arising as the winner technique in most challenges such as \ac{BRATS}, ISLES or PROMISE12 challenges \citep{BRATS2016, BRATS2017, BRATS2018}.

In addition to glioblastoma tissue segmentation, \ac{PWI} has played a key role in the advanced characterization of the tumor heterogeneity based on \ac{MRI} \citep{Shah2010, Lupo2005, Knopp1999}. Glioblastoma is characterized by a robust angiogenesis, strong vascular proliferation and an aberrant microvasculature \citep{Alves2011, Hardee2012, Kargiotis2006}. Numerous studies have focused on the analysis of perfusion indices to assess tumor grading \citep{Law2003, Emblem2008}, early response to treatment assessment \citep{Elmghirbi2017, Vidiri2012}, recurrence vs radionecrosis \citep{Hu2009, Barajas2009} or clinical outcome prediction \citep{Mangla2010, Jain2014}. More recent studies addressed the local characterization of sub-regions within the glioblastoma using \ac{DSC}, \ac{DCE} or \ac{MRSI} \citep{Artzi2014, Akbari2014, Sawlani2010, Raschke2019}. Specifically, in \citep{JuanAlbarracin2018} an unsupervised method called \ac{HTS} was proposed to characterize the vascular heterogeneity of glioblastoma based on \ac{DSC}. This method combined perfusion biomarkers and glioblastoma tissue segmentation to discover habitats within the neoplasm, showing significant correlation with patient overall survival.

However, despite the great advances in novel methods to describe glioblastoma heterogeneity, most of them are based on private algorithms and in-house technology developed by the authors, non-accessible for the scientific community. \ac{MRI}-dedicated libraries such as \ac{ANTs} \citep{Avants2011, Tustison2014}, FSL \citep{Jenkinson2012} or ITK \citep{Avants2014}, as well as modern toolkits for deep learning such as \href{https://www.tensorflow.org/}{Tensorflow\textsuperscript{\tiny\texttrademark}} or \href{https://pytorch.org/}{PyTorch}, are provided to develop such technologies. However, these libraries are just the pieces to build the complex state-of-the-art models to analyze glioblastoma, whose development requires considerable efforts, resources and arduous learning curves, which are not often accessible to many researchers or institutions. In this regard, open-access public platforms that implement state-of-the-art techniques in a user-transparent manner are highly desirable to bring to the scientific community the possibility to conduct advanced multiparametric analysis of glioblastoma.

In this work we present \href{https://www.oncohabitats.upv.es}{ONCOhabitats}: an online system aimed to provide state-of-the-art analysis services for glioblastoma. ONCOhabitats provides two main services for untreated glioblastoma: 1) High-grade glioma tissue segmentation based on \ac{CNN}; and 2) glioblastoma vascular heterogeneity assessment by means of the \ac{HTS} method proposed in \citep{JuanAlbarracin2018}. For each service, ONCOhabitats returns the preprocessed images, the tissue segmentation and habitats maps, and automatically generates a radiological report summarizing all the findings of the study.

\section{Materials}
\label{section:oncohabitats_materials}
To validate ONCOhabitats technology several datasets were employed for the different services.

ONCOhabitats high-grade glioma segmentation service was evaluated with the public \ac{BRATS} 2017 challenge dataset, provided for the international MICCAI 2017 conference. The training corpus of the \ac{BRATS} 2017 dataset consists of multi-parametric \ac{MR} scans of 210 high-grade gliomas: 20 patients from the \ac{BRATS} 2013 dataset, 88 from the \ac{CBICA} and 102 from the \ac{TCIA} corpus. The validation corpus consists of 46 multi-contrast MR scans of high-grade glioma patients distributed in 16 cases from the \ac{CBICA} institution, 24 cases from the \ac{TCIA} corpus and 6 cases from the \ac{UAB} department.

For each patient, pre- and post-gadolinium \Ti{}-weighted, \Tii{}-weighted and \ac{FLAIR} \ac{MR} exams were provided. All images were linearly co-registered to the post-gadolinium \Ti{}-weighted exam, skull stripped, and interpolated to 1mm\textsuperscript{3} isotropic resolution.

Manual expert annotations of this dataset comprise 4 classes: Class 1) necrosis, cyst, hemorrhage and non-enhancing tumor; class 2) surrounding edema; class 4) enhancing tumor core; and class 0) for everything else. Evaluation is assessed for 3 different compartments, whose composition is shown in table \ref{table:oncohabitats_evaluation_subcompartments}.

\begin{table}[h]
\caption{Labels composing each sub-compartment evaluated in the \ac{BRATS} 2013 challenge.}
\centering
\rowcolors{2}{gray!10}{white}
\begin{tabular}{lccccc}
	\hline
	\rowcolor{gray!25}
	\cellcolor{gray!25} & Label 0 & Label 1 & Label 2 & Label 4 \\
	\cellcolor{gray!25}\ac{WT} & & \ding{53} & \ding{53} & \ding{53} \\
	\cellcolor{gray!25}\ac{TC} & & \ding{53} & & \ding{53} \\
	\cellcolor{gray!25}\ac{ET} & & & & \ding{53} \\ \hline
\end{tabular}
\label{table:oncohabitats_evaluation_subcompartments}
\end{table}

The glioblastoma vascular heterogeneity service was validated with a retrospective local dataset of 50 patients, including 33 men with an average age of 60.94 years (range, 25–80 years) and 17 women with an average age of 62.53 years. \ac{MRI} included pre- and post-gadolinium \Ti{}-weighted, \Tii{}-weighted and \ac{FLAIR} \ac{MR} exams, and \ac{DSC} \Tiis{}-weighted perfusion study. The institutional review board approved this retrospective study, and the requirement for patient informed consent was waived. The patient inclusion criteria and \ac{MRI} protocol are extensively detailed in section \ref{section:hts_materials}.

\section{Methods}
\label{section:oncohabitats_methods}
\href{https://www.oncohabitats.upv.es}{ONCOhabitats} provides two main services: 1) High-grade glioma segmentation, and 2) glioblastoma vascular heterogeneity assessment.

\begin{figure}[ht!]
\centering
\includegraphics[width=0.8\linewidth]{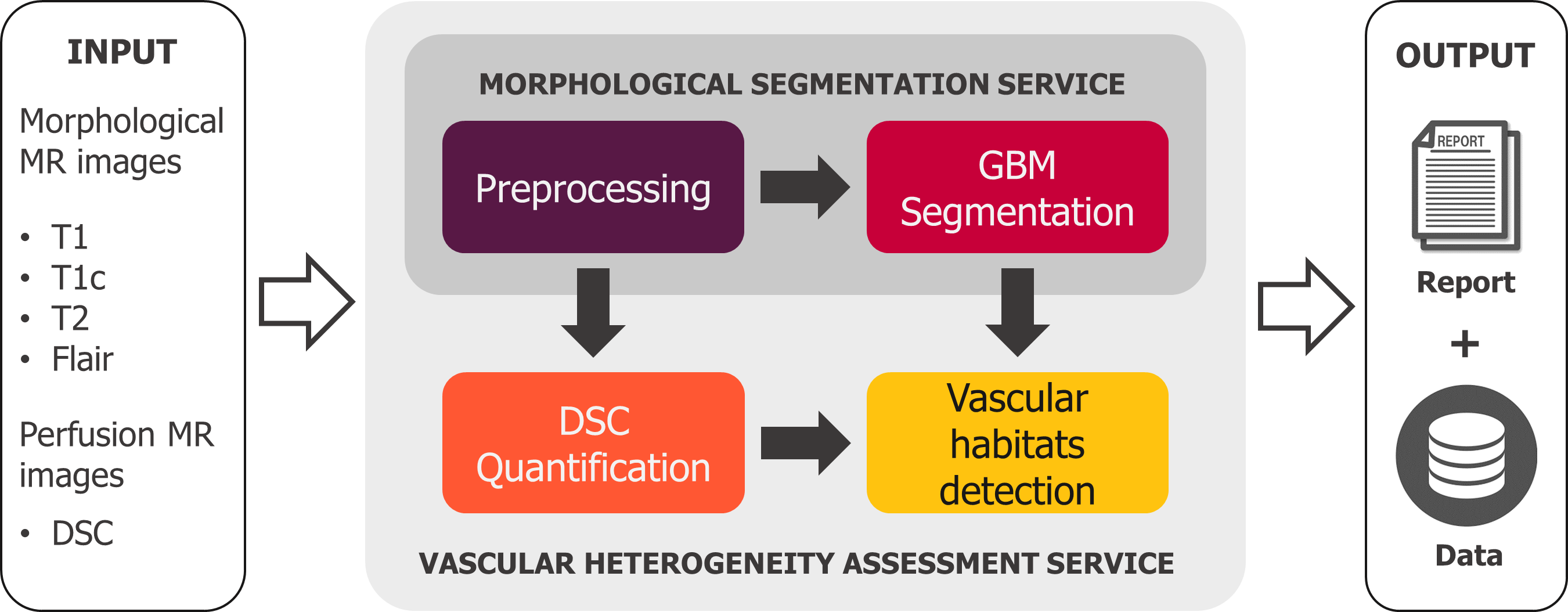}
\caption{Diagram of the different modules comprising the high-grade glioma segmentation and glioblastoma vascular heterogeneity assessment services of ONCOhabitats.}
\label{figure:oncohabitats_modules}
\end{figure}

Figure \ref{figure:oncohabitats_modules} shows an outline of the different sub-processes involved in each service. The first pipeline performs a morphological segmentation of high-grade glioma tumors by first pre-processing the \ac{MRI} and then using a 3D U-Net \ac{CNN} classifier. The second pipeline extends the morphological segmentation pipeline by incorporating \ac{DSC} perfusion pre-processing and quantification, and the \ac{HTS} method to detect regions within the glioblastoma with different hemodynamic activity.

We will first describe both services and then the ONCOhabitats on-line system will be presented.

\subsection{High-grade glioma segmentation service}
\label{subsection:oncohabitats_morphologicalservice}
ONCOhabitats considers three tissues to morphologically describe high-grade gliomas: 1) enhancing tumor, 2) edema and 3) necrotic and non-enhancing regions of the tumor. The service is composed of two stages: 1) \ac{MRI} preprocessing and 2) Segmentation based on \acp{CNN}.

\subsubsection{\acs{MRI} preprocessing}
\label{subsubsection:oncohabitats_mripreprocessing}
Our preprocessing module includes the following steps: (1) voxel isotropic resampling of all \ac{MR} images, (2) denoising, (3) rigid intra-patient \ac{MRI} registration, (4) affine registration of all sequences to \ac{MNI} \ac{ICBM} space, (5) skull-stripping and (6) magnetic field inhomogeneity correction. Voxel resampling is performed at 1mm\textsuperscript{3} by means of linear interpolation. Denoising is carried out using the adaptive non local means filter \citep{Manjon2010a} with a search window of $7 \times 7 \times 7$ voxels and a patch window of $3 \times 3 \times 3$ voxels. Registration is conducted with the \ac{ANTs} software \citep{Avants2008}, taking the \Tic{} sequence as reference and using Mutual Information metric. In a previous version of ONCOhabitats, skull-stripping was performed with an in-house pipeline based on a non-linear registration of the \Tic{} sequence to a template with a known intra-cranial mask. Nowadays, skull-stripping if performed with a patch-based U-net \ac{CNN} working with \Tic{} patches of $32 \times 32 \times 32$ trained on 120 manually segmented glioblastomas. The network has a performance of $0.94$ Dice score on an independent test set. Finally, magnetic field inhomogeneities are corrected with the N4 software using the previously computed intra-cranial mask \citep{Tustison2010}.
 
\subsubsection{High-grade glioma segmentation}
\label{subsubsection:oncohabitats_cnnsegmentation}
ONCOhabitats \ac{CNN} takes as input the \Tic{}, \Tii{} and \ac{FLAIR} \ac{MRI} and works with 3D patches of size $32 \times 32 \times 32$. We followed a U-net architecture \citep{Ronneberger2015, Soltaninejad2018a, Dong2017} of 5 levels, with a contracting and expanding paths of 4 \emph{residual-blocks} preceded of 4 \emph{simple-blocks}. A \emph{simple-block} consists of the following sequence of operations: convolution + batch normalization + ReLu activation function, while a \emph{residual-block} implements the proposal of He et al \citep{He2016}: convolution + batch normalization + ReLu + convolution + batch normalization + residual connection + ReLu activation function. Max-pooling of size 2 is employed to down-sample patches at each level, while transpose convolutions are employed to up-sample patches in the expanding path. The number of filters per level are: 16 at first level (native patch resolution of $32\times32\times32$), 32 filters at second level (patch resolution of $16\times16\times16$), 64 filters at third level (patch resolution of $8\times8\times8$), 128 filters at fourth level (patch resolution of $4\times4\times4$) and 256 filters at fifth level (patch resolution of $2\times2\times2$). Long-term concatenations are also employed to connect blocks at each level.

Isotropic kernels of $3\times3\times3$ were employed for all convolutions. The network was trained using Adam Optimizer with an initial learning rate of $1e-3$ and cross-entropy was used as loss function. $L2$ regularization with penalty $1e-3$ was employed to avoid for over-fitting. We employed a batch size of 64 individuals, forcing an equal representation of enhancing tumor, edema, necrosis and healthy patches to compensate for class imbalance. The network was trained for $50k$ iterations.

\begin{figure}[ht!]
\centering
\includegraphics[width=0.95\linewidth]{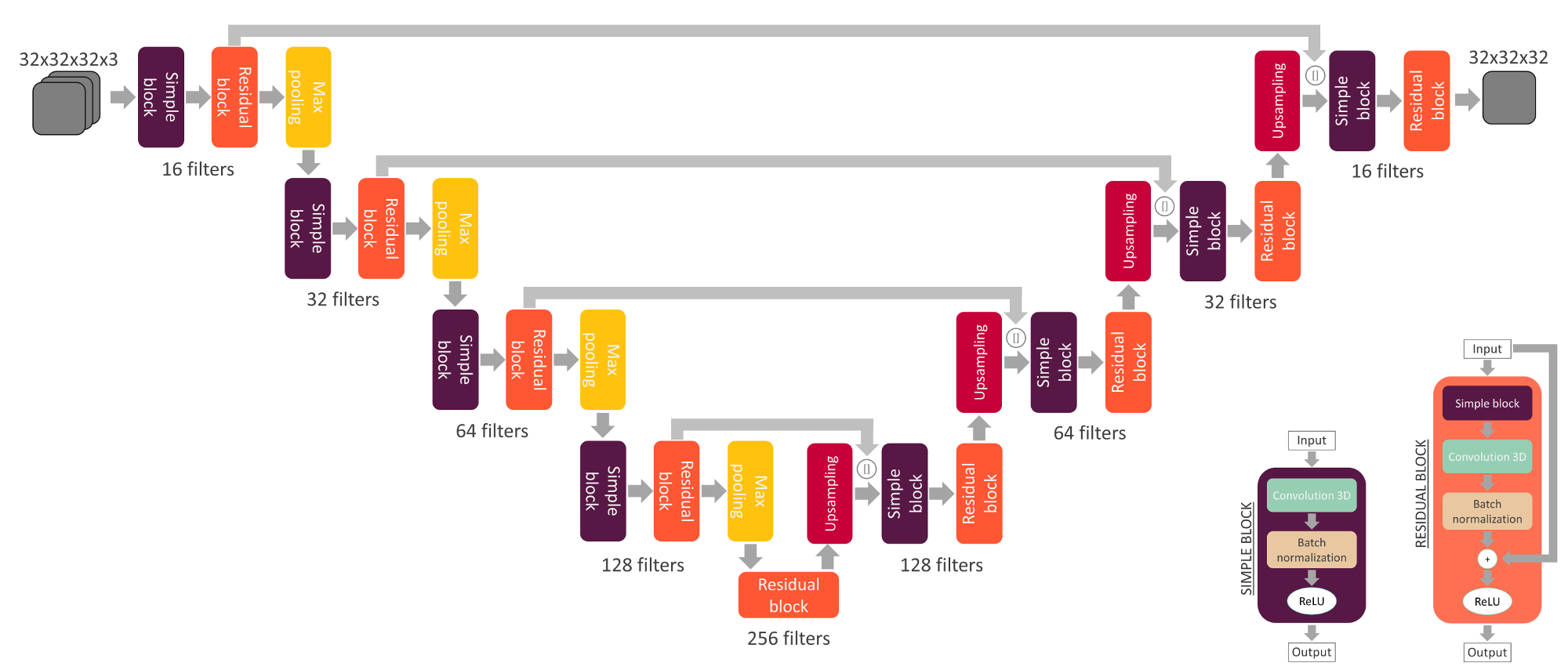}
\caption{High-grade glioma \acl{CNN} architecture for the ONCOhabitats morphological segmentation service.}
\label{figure:oncohabitats_CNN}
\end{figure}

Figure \ref{figure:oncohabitats_CNN} summarizes the network architecture and the internal design of the \emph{simple} and \emph{residual} blocks.

\subsection{Glioblastoma vascular heterogeneity assessment service}
\label{subsection:oncohabitats_htsservice}
This service extends the morphological segmentation service by introducing perfusion information into the study. The service implements the \ac{HTS} method presented in \citep{JuanAlbarracin2018}, which aims to describe the vascular heterogeneity of glioblastoma. The \ac{HTS} combines the glioblastoma morphological segmentation with perfusion indexes such as \ac{rCBV} and \ac{rCBF} to discover habitats within each tissue with different patterns of vascularity. We found that these habitats provide relevant information to early predict patient survival, even taking into account the variations in treatment. 

The service includes four stages: 1) \ac{MRI} preprocessing, 2) Segmentation based on \acp{CNN}, 3) Perfusion quantification and 4) Vascular habitats detection. The \ac{MRI} preprocessing and glioblastoma segmentation are inherited from the high-grade glioma segmentation service.

\subsubsection{Perfusion quantification}
\label{subsubsection:oncohabitats_perfusionquantification}
ONCOhabitats \ac{rCBV} and \ac{rCBF} maps are quantified by means of standard techniques proposed in the literature \citep{Knutsson2010}. For a detailed explanation of the calculation of perfusion parametric maps, please refer to section \ref{subsection:rationale_mri_qmri}. In order for this chapter to be self-contained, a short remainder will be made.

\Ti{}-weighted leakage effects are automatically corrected using the Boxerman method \citep{Boxerman2006}, while gamma-variate curve fitting is employed to correct for \Tii{} extravasation phase. \ac{rCBV} is computed by numerical integration of the area under the gamma-variate curve \citep{Knutsson2010}, while \ac{rCBF} is calculated based on the block-circulant \ac{SVD} devolution technique proposed in \citep{Wu2003}. The \ac{AIF} is automatically calculated using a \emph{divide and conquer} algorithm, which recursively dichotomizes the gadolinium concentration-time curves into two groups, selecting those curves with higher peak height, earliest time to peak and quickest wash-out (i.e. lowest full width at half maximum). The \ac{AIF} is finally computed as the average of the curves of the final group that contains 10 or fewer curves.

\subsubsection{Vascular habitats detection}
\label{subsubsection:oncohabitats_vascularhabitatsdetection}
The \ac{HTS} method \citep{JuanAlbarracin2018} describes the vascular heterogeneity of the glioblastoma by means of an unsupervised analysis of the perfusion patterns detected within the lesion. Such analysis is designed to yield four habitats: the \ac{HAT}, the \ac{LAT}, the potentially \ac{IPE} and the \ac{VPE}.

The unsupervised analysis of perfusion patterns is carried out through the \ac{DCM}-\ac{SVFMM} algorithm \citep{Sfikas2008} (with \ac{DCAGMRF} prior \citep{Nikou2007}). Such algorithm is an extension of the classic \ac{FMM} specially focused on image data, which incorporates a continuous \ac{MRF} on the spatial coefficients of the model to capture the self-similarity and local redundancy of the images. The \ac{HTS} method consists of two stages: (a) an initial re-definition of the enhancing tumor and edema \acsp{ROI} obtained by the morphological segmentation using perfusion information, and (b) a cluster analysis of the perfusion heterogeneity within each previously mentioned \acs{ROI} to detect the different vascular behaviors expressed by the neoplasm. A comprehensive detailed explanation of the \ac{HTS} method is performed in section \ref{subsection:hts_hts}.

\subsection{Clinical report}
\label{subsection:oncohabitats_report}
\begin{figure}[htbp!]
\centering
\includegraphics[width=\linewidth]{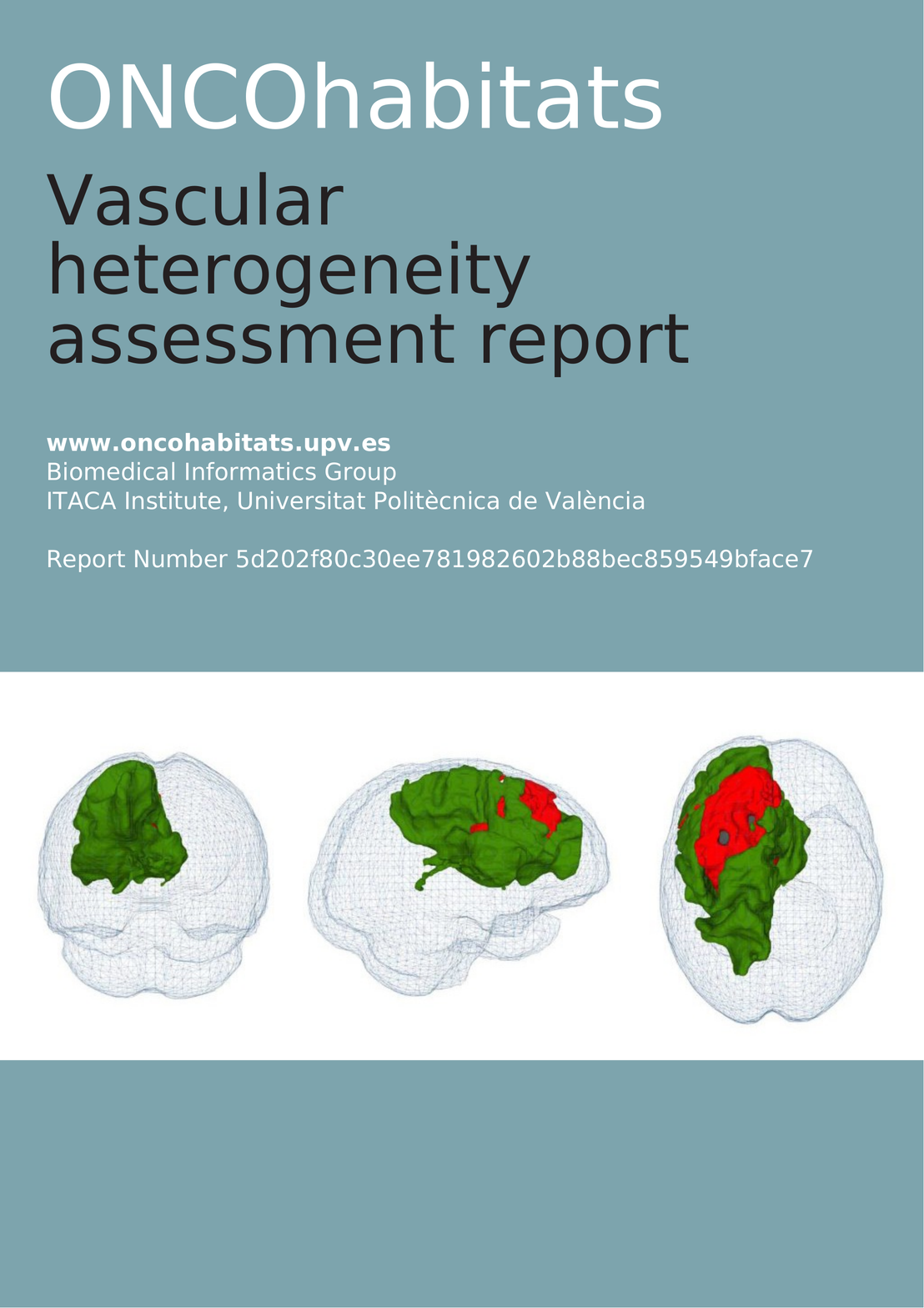}
\caption{Example of the first page of a report for a vascular heterogeneity assessment analysis of a glioblastoma.}
\label{figure:oncohabitats_report_1}
\end{figure}

\begin{figure}[htbp!]
\centering
\includegraphics[width=\linewidth]{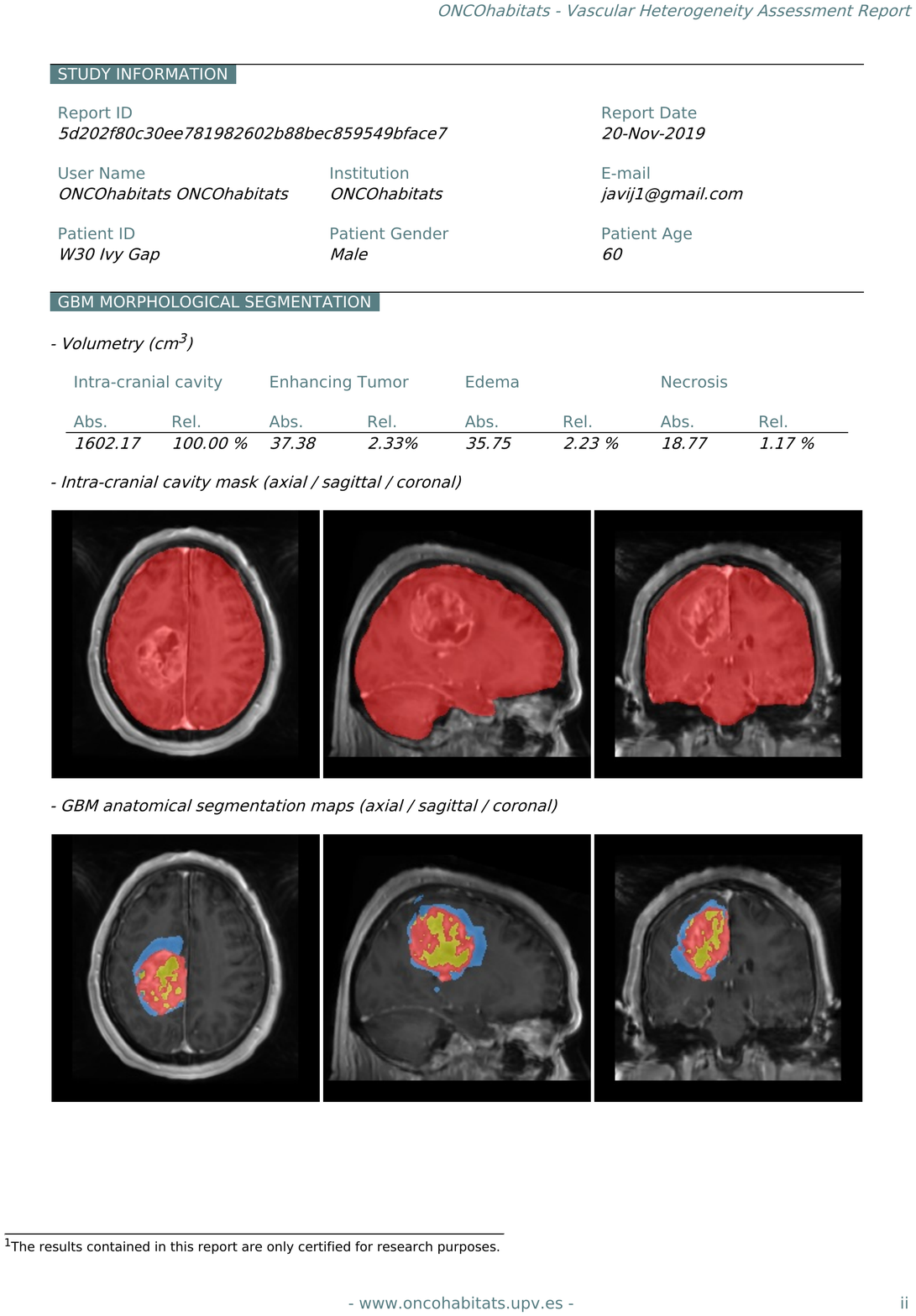}
\caption{Example of the second page of a report for a vascular heterogeneity assessment analysis of a glioblastoma.}
\label{figure:oncohabitats_report_2}
\end{figure}

\begin{figure}[htbp!]
\centering
\includegraphics[width=\linewidth]{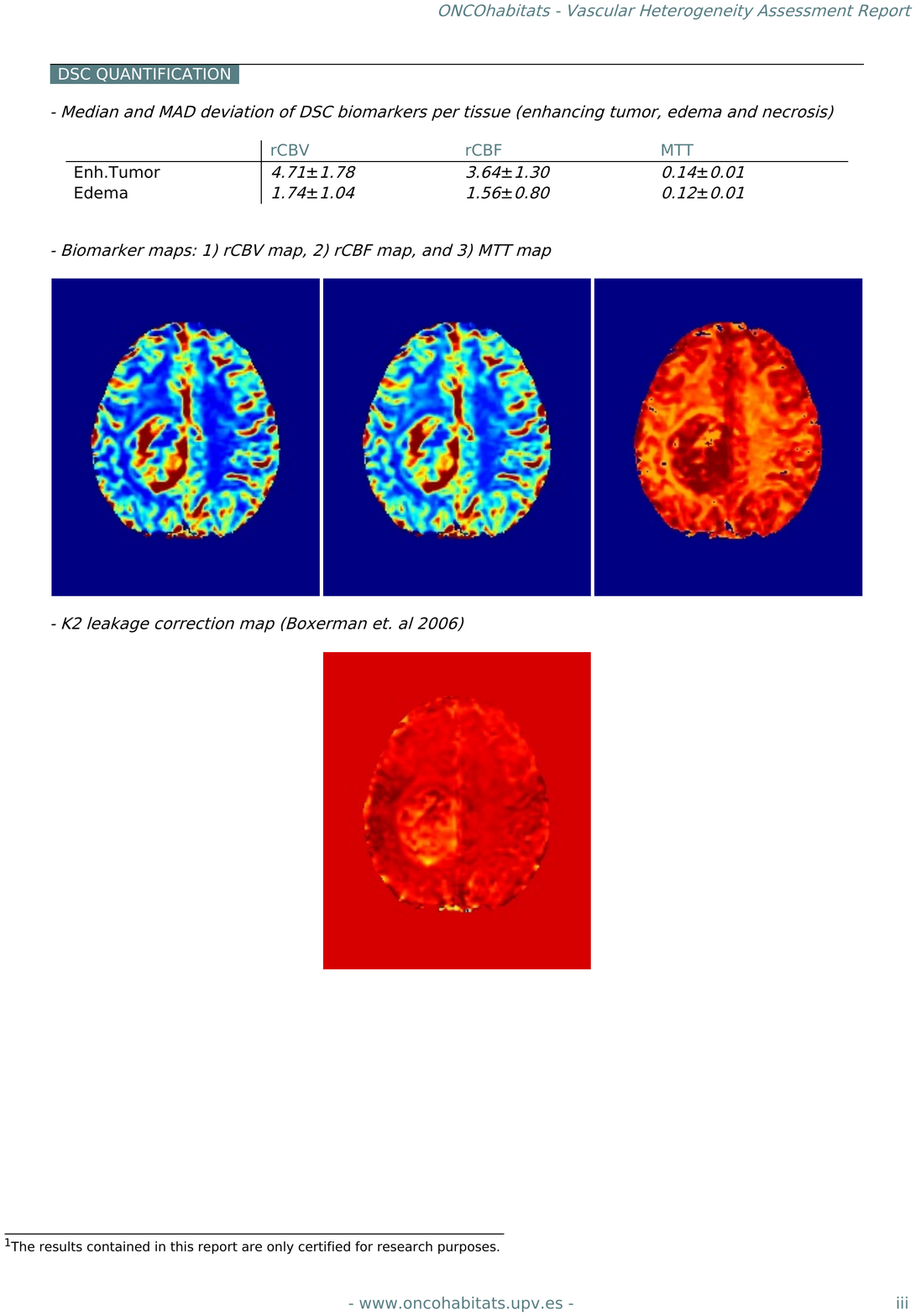}
\caption{Example of the third page of a report for a vascular heterogeneity assessment analysis of a glioblastoma.}
\label{figure:oncohabitats_report_3}
\end{figure}

\begin{figure}[htbp!]
\centering
\includegraphics[width=\linewidth]{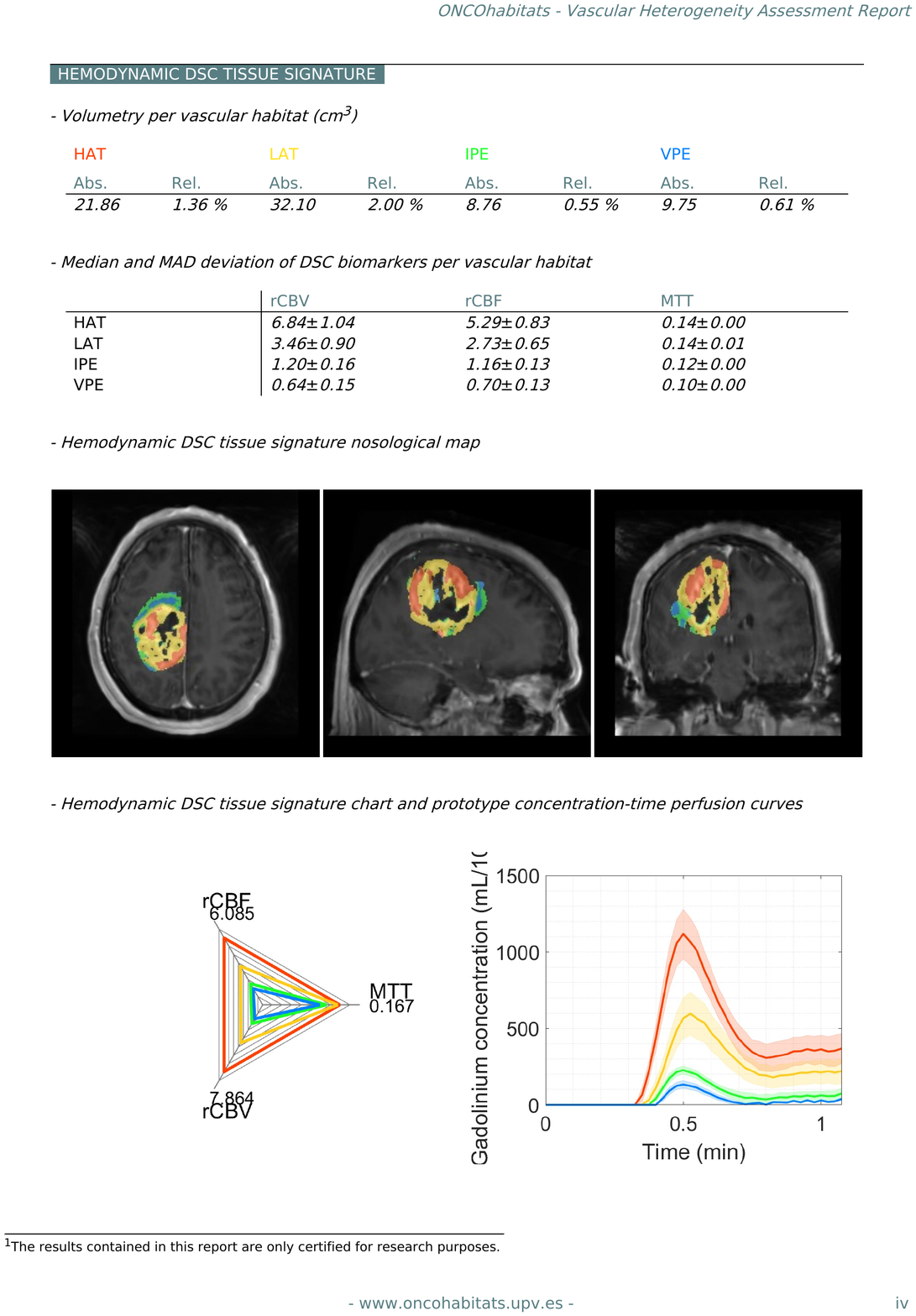}
\caption{Example of the fourth page of a report for a vascular heterogeneity assessment analysis of a glioblastoma.}
\label{figure:oncohabitats_report_4}
\end{figure}

\begin{figure}[htbp!]
\centering
\includegraphics[width=\linewidth]{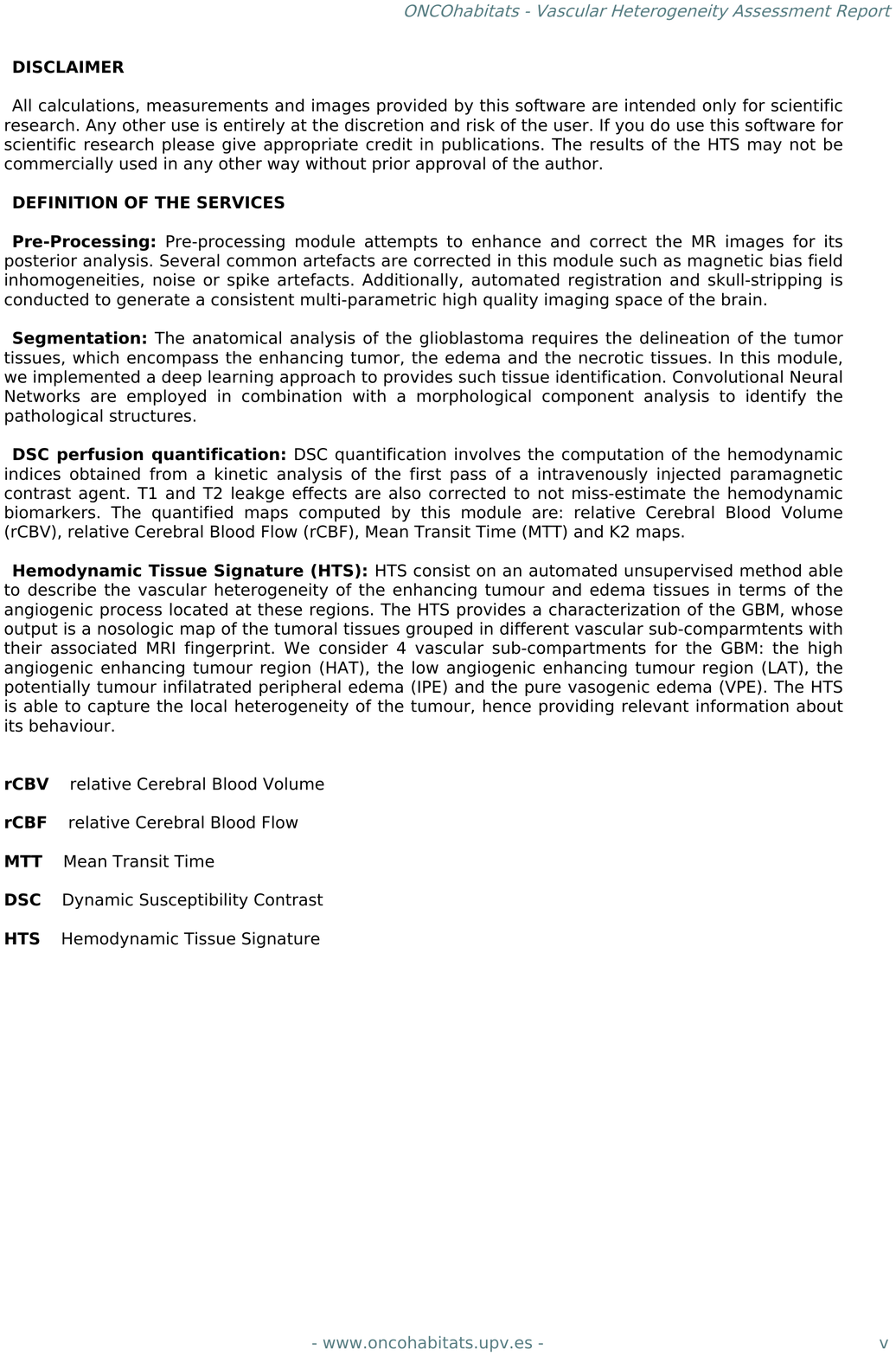}
\caption{Example of the fifth page of a report for a vascular heterogeneity assessment analysis of a glioblastoma.}
\label{figure:oncohabitats_report_5}
\end{figure}

Clinical reports are automatically generated after the finalization of each ONCOhabitats job. The reports summarize all the findings of the studies, including morphological and functional measurements of each glioblastoma tissue and habitat. Regarding the morphological segmentation service jobs, absolute tissue volumetry in cm\textsuperscript{3}, as well as relative volumetry with respect to the intra-cranial cavity are calculated for the enhancing tumor, necrosis and edema tissues. Concerning the vascular heterogeneity assessment service jobs, in addition to the tissue volumetries, habitat's absolute and relative volumetries are also calculated. Moreover, a tendency analysis of the perfusion biomarkers confined within each tissue and habitat is also included in the report. Median perfusion values of each \acs{ROI} as well as Median Absolute Deviations are calculated to robustly determine the vascular tendency of each sub-comparment of the lesion. Finally, perfusion prototypical curves in combination with a radar chart of the perfusion biomarkers at each region are also included in the report for a visual representation of the functional behavior of the glioblastoma. Figures \ref{figure:oncohabitats_report_1}, \ref{figure:oncohabitats_report_2}, \ref{figure:oncohabitats_report_3}, \ref{figure:oncohabitats_report_4}, \ref{figure:oncohabitats_report_5} show an example of a vascular heterogeneity assessment report.

\subsection{ONCOhabitats system}
\label{subsection:oncohabitats_webservice}
\begin{figure}[ht!]
\centering
\includegraphics[width=0.95\linewidth]{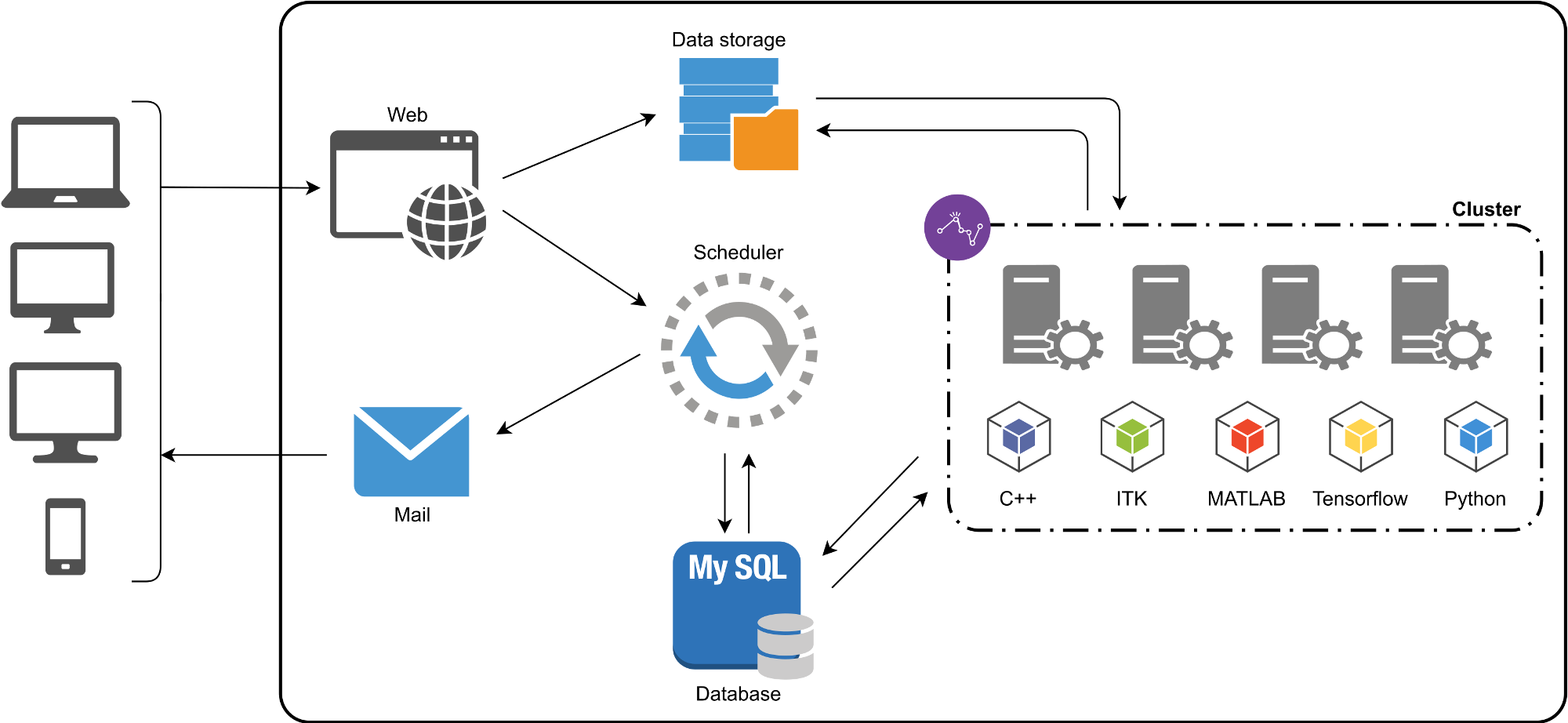}
\caption{ONCOhabitats system architecture.}
\label{figure:oncohabitats_architecture}
\end{figure}

\href{https://www.oncohabitats.upv.es}{ONCOhabitats} is a web service solution to carry out the previously presented analyses. The platform implements a \ac{SaaS} model to automatically analyze glioblastoma cases. Figure \ref{figure:oncohabitats_architecture} shows the diagram scheme of ONCOhabitats system.

The system implements a Wordpress\textsuperscript{\tiny\textregistered} landing web-page as front-end for the user. Before applying for a job, user must be registered in ONCOhabitats system. Registration requires a username, the first and last names, the institution of provenance and a valid email only used to inform the user about the status of their jobs. After registration, the user can upload the \ac{MR} images to a data storage secure server to launch the jobs.

The system implements secure encrypted communication via HTTPS protocol with a trusted certificate to enhance data protection and privacy. The data storage includes file encryption and secure transfer protocol via SFTP. ONCOhabitats currently supports DICOM and NIfTI (compressed and uncompressed) medical imaging formats. An automated de-identification is carried out for all DICOM files using the \emph{gdcmanon} tool from the \emph{Grassroots} DICOM library. Once DICOM files are de-identified, they are converted to compressed NIfTI format and automatically removed from the server. The de-identified NIfTI files are stored in a secure server, non-accessible through the ONCOhabitats website, to enhance security and data protection. Once the analysis is complete, the pre-processed images as well as the resulting segmentation masks and biomarker maps are stored in a separated data storage, only accessible for 15 days by the user owner of the job. After this period, all the data is completely removed from the ONCOhabitats servers, unless the user explicitly specifies through his account web-page that his data can be used for research purposes, in which case the data is kept on a private server. This procedure follows the guidelines recommended by the Data Protection Officer (DPO) of the Universitat Politècnica de Valencia (UPV) and has been approved by the ethical committee of our institution.

\begin{figure}[ht!]
\centering
\includegraphics[width=0.8\linewidth]{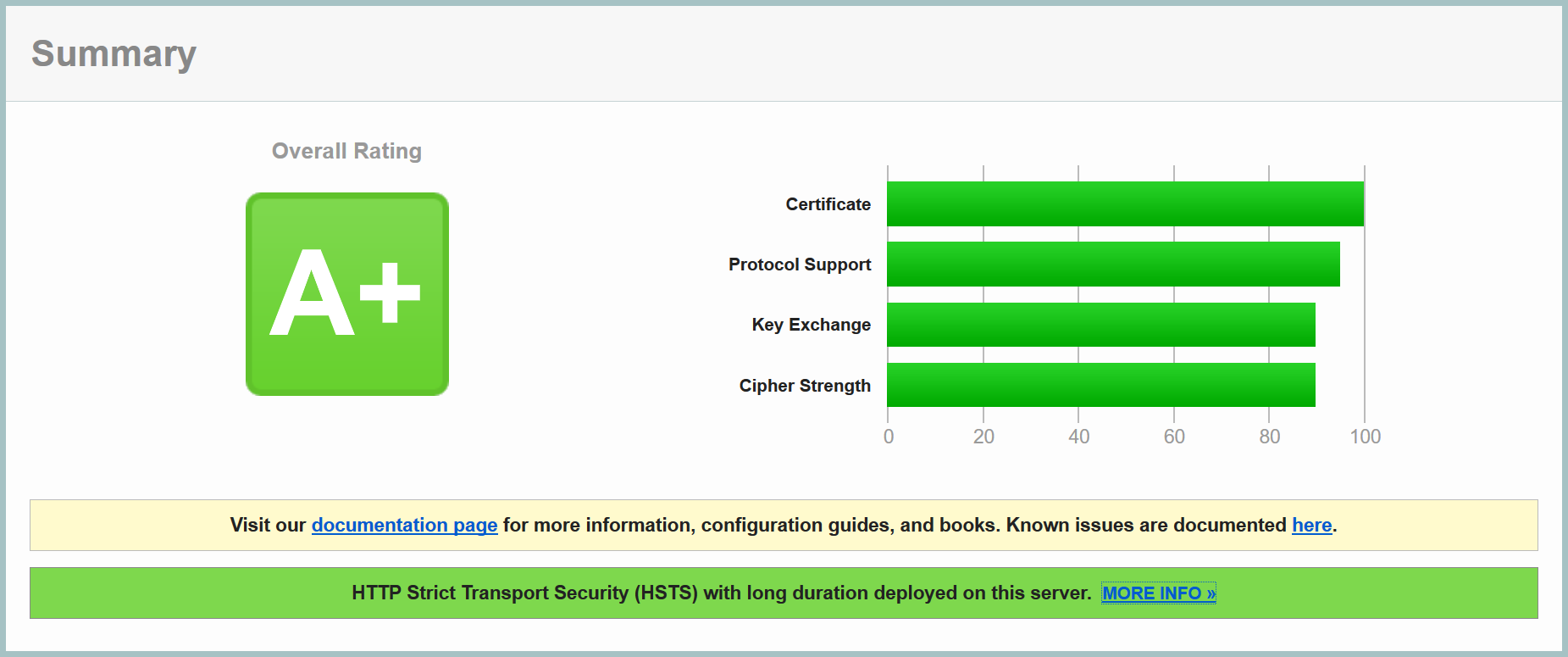}
\caption{ONCOhabitats system security rating evaluated by SSL Labs by Qualy.}
\label{figure:oncohabitats_security}
\end{figure}

ONCOhabitats system equips 7 DELL PowerEdge R720\textsuperscript{\tiny\textregistered} dedicated servers, each one shipping two Intel Xeon E5-2620 CPUs with a total of 12 cores and 64 GB of RAM. Two NVidia Titan Xp with 3840 CUDA\textsuperscript{\tiny\textregistered} cores and 12 GB of RAM supports the cluster for the deep learning tasks.

The ONCOhabitats pipelines are mostly implemented in C++ using \href{https://itk.org/}{ITK} and \href{http://eigen.tuxfamily.org/}{Eigen} libraries. MATLAB\textsuperscript{\tiny\textcopyright} and PHP are also employed as scripting languages for different tasks. \href{https://www.tensorflow.org/}{Tensorflow\textsuperscript{\tiny\texttrademark}} is used to develop the deep learning segmentation models.

ONCOhabitats has been designed to deal with up to 14 concurrent jobs, each job taking approximately one hour, which yields a theoretical limit of 336 processed cases per day. The terms of use of the system are available in the landing web page, paying particular attention to the GDPR compliance and the non-commercial research purpose of the system.

\section{Results}
\label{section:oncohabitats_results}
Since each ONCOhabitats service performs a different task, they have been evaluated separately with different datasets and methodologies. The results of the evaluations are presented below.

\subsection{High-grade glioma segmentation service}
\label{subsection:oncohabitats_results_segmentation}
High-grade glioma tissue segmentation performance was evaluated according to \ac{BRATS} evaluation guidelines. Such evaluation comprises the assessment of the segmentation quality of the \ac{WT} region, the \ac{TC} region and the \ac{ET} area. Dice metric, as well as sensitivity and specificity of each region was computed to compare ONCOhabitats results with several state-of-the-art methods.

\begin{table}[ht!]
\caption{Segmentation results obtained by the ONCOhabitats high-grade glioma segmentation service for the \ac{BRATS} 2017 validation set. \emph{ET}: Enhancing Tumor, \emph{WT}: Whole Tumor, \emph{TC}: Tumor Core}
\centering
\rowcolors{2}{gray!10}{white}
\begin{tabular}{lccc}
	\hline
	\rowcolor{gray!25}
	              & \ac{ET} & \ac{WT} & \ac{TC} \\
	Dice          & 0.73 & 0.89 & 0.73 \\
	Sensitivity   & 0.77 & 0.87 & 0.72 \\
	Specificity   & 0.99 & 0.99 & 0.99 \\
	\hline
\end{tabular}
\label{table:oncohabitats_results_segmentation}
\end{table}

\begin{figure}[ht!]
\centering
\includegraphics[width=0.9\linewidth]{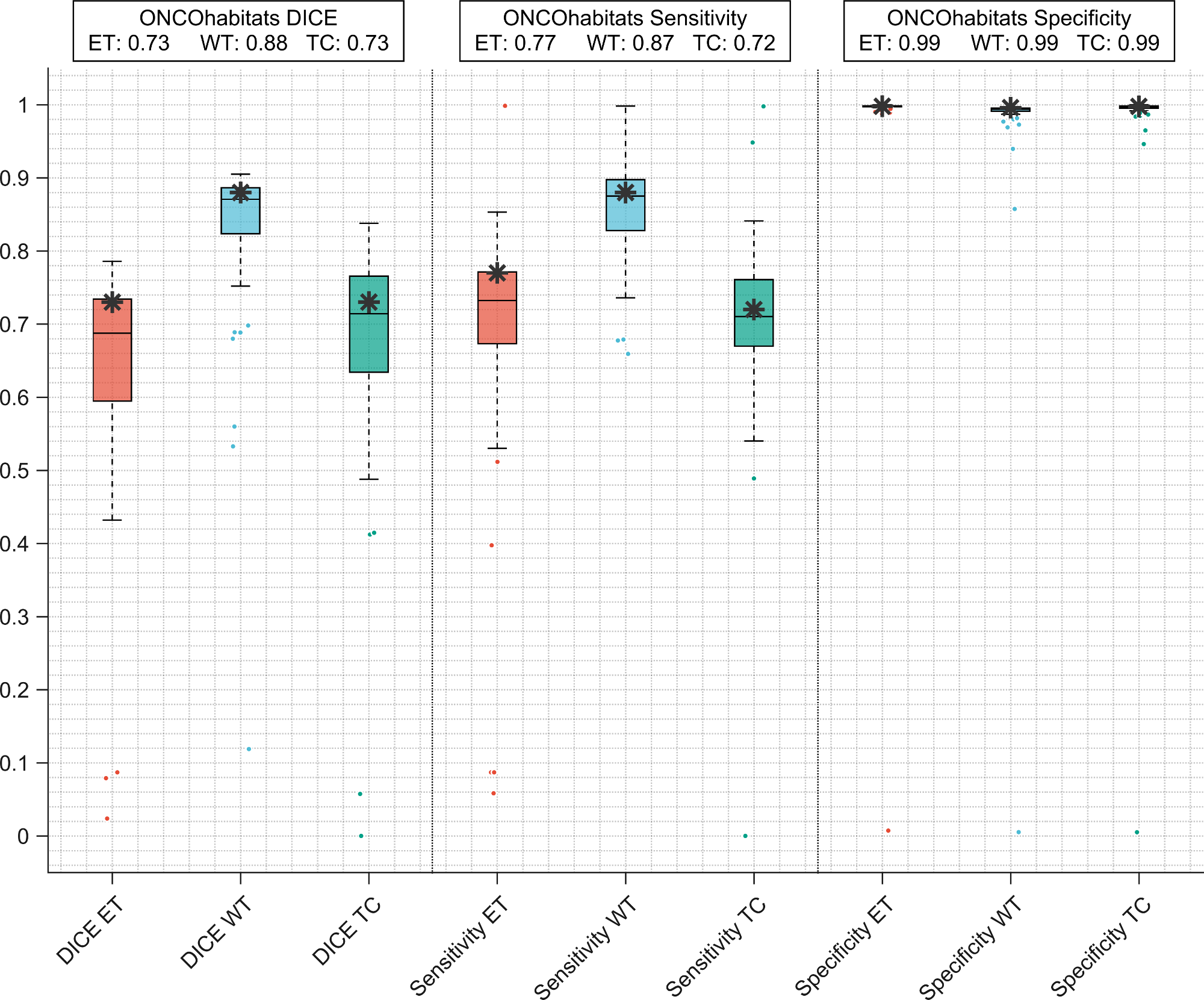}
\caption{Distributions of Dice scores, sensitivities and specificities for the \ac{BRATS} 2017 validation set results obtained by the participants of the challenge. ONCOhabitats results are indicated with a * marker. \emph{ET}: Enhancing Tumor, \emph{WT}: Whole Tumor, \emph{TC}: Tumor Core}
\label{figure:oncohabitats_boxplot}
\end{figure}

The results of our high-grade glioma segmentation service are presented in Table \ref{table:oncohabitats_results_segmentation}. A comparison between the results obtained by ONCOhabitats system and the ones obtained by each participant of the challenge for the \ac{BRATS} 2017 validation set are presented in Figure \ref{figure:oncohabitats_boxplot}. As the figure shows, ONCOhabitats offers competitive comparable segmentations with the state-of-the-art algorithms, always achieving results above the median of the participants and in most cases close to the third quartile. Moreover, \acs{CNN} yield segmentations of high specificity, which is a highly desirable property for a medical tool, ensuring a very low false positive rate.

\subsection{Glioblastoma vascular heterogeneity assessment service}
An extensive evaluation of the \ac{HTS} method at different levels is presented in section \ref{section:hts_results}.

First, a statistical evaluation to assess the degree of similarity among the \ac{rCBV} and \ac{rCBF} distributions for the \ac{HAT}, \ac{LAT}, \ac{IPE} and \ac{VPE} habitats was conducted. Global probabilistic deviation metric \citep{Saez2017} was employed as a multi-dimensional extension of Jensen-Shannon divergence to measure distances between distributions. The analysis yielded the following average results for the population: $0.88 \pm 0.03$ for \ac{rCBV} and $0.86 \pm 0.05$ for \ac{rCBF}. Such results indicated that the perfusion distributions of the habitats for each patient of the study were statistically significantly separated, hence corroborating that the \ac{HTS} habitats describe regions within the glioblastoma with different hemodynamic behavior.

Second, the prognostic capabilities of the \ac{HTS} habitats were studied. Cox proportional hazard regression analysis and a Kaplan-Meier study were conducted to measure the degree of correlation of the \ac{HTS} habitats with patient \ac{OS}. The maximum \ac{rCBV} and \ac{rCBF} value at each habitat (computed as the 95\% percentile of the distribution) was the marker with better results in concordance with previous studies in the literature \citep{Wetzel2002}. \ac{HAT} and \ac{LAT} habitats, as well as \ac{IPE} yielded positive correlations with overall survival for both Cox and Kaplan-Meier studies. Benjamini-Hochberg false discovery rate at $\alpha$ level of 0.05 was employed to correct for multiple hypothesis testing, increasing confidence of the statistical results.

\section{Discussion}
In this work we presented ONCOhabitats: an on-line open-access system to study different aspects of glioblastoma such as the tumor morphology and the vascular tumor heterogeneity patterns. ONCOhabitats provides the user with consolidated state-of-the-art techniques based on both deep learning and unsupervised structured learning algorithms previously published in the literature. The system also generates automated radiological reports summarizing the findings of the analysis, in a formal document easily integrable in clinical routine.

The ONCOhabitats high-grade glioma segmentation was compared against the current state-of-the-art methods presented at \ac{BRATS} 2017 challenge. Our system yields comparable results with these approaches, demonstrating competitive comparable results with no significant differences between them. Our method is completely deterministic and reproducible, which is a highly desirable property to conduct large population studies or clinical trials. The higher reproducibility of a system, the greater the likelihood to detect changes in the disease. Additionally, ONCOhabitats implements the \acl{HTS} method presented in \citep{JuanAlbarracin2018}, to study the vascular heterogeneity of the tumor. Glioblastoma vascular heterogeneity has been demonstrated to be a key hallmark to understand the behavior of this masses. The \ac{HTS} analysis allows to study the different vascular patterns of the lesion, detecting functional habitats within the tissues that have been demonstrated to contain relevant information about patient's survival at a very early stage of the disease.

ONCOhabitats offers these analyses by means of a free web-based solution. We give access to the scientific community not only to our software services but also to our computational resources, avoiding the requirement for medical imaging experts, expensive computational labs and arduous learning curves to develop the technology. We provide a system capable to process about 300 cases per day including \ac{MRI} preprocessing and standardization, tissue segmentation, \ac{DSC} perfusion quantification and vascular heterogeneity assessment of the lesion. The system is designed to be immediately scalable by adding new computing machines and also cloud-based services as the workload increases.

The major limitation of our system is the time to process a case. Glioblastoma study involves the analysis of a considerable amount of \ac{MRI} to capture all the information contained in the lesion, requiring to process a huge amount of data. Currently, \ac{MRI} preprocessing accounts for the most part of the processing time, so we are currently developing new approaches to optimize this module.

Currently, ONCOhabitats system is participating in the international clinical trial \href{https://clinicaltrials.gov/ct2/show/NCT03439332?cond=Glioblastoma&cntry=ES&city=Valencia&rank=1}{NCT03439332}, aimed to validate the ONCOhabitats technology in a multi-centre observational study with 300 patients from hospitals from Spain, Italy, Belgium and Norway. In future work, we plan to extend ONCOhabitats system to handle \ac{DWI} and \ac{DTI} sequences, as well as post-surgery and follow-up studies for a longitudinal assessment of the glioblastoma.

\chapter{Concluding remarks and recommendations}
\label{chapter:conclusion}

This chapter finalizes the work conducted in this thesis and summarizes the main concluding remarks and recommendations derived from it. Additionally, in this chapter we provide the guidelines for continuing the scientific research and development based on this work.

\section{Concluding remarks}
\label{section:concluding_remarks}
\ac{AI} medical image analysis is a cornerstone in the future of modern precision medicine. The ability to non-invasively measure morphological and quantitative characteristics of complicated diseases such as glioblastoma is an invaluable aid in successfully combating these lethal lesions. In the particular case of glioblastoma, to date, this tumor still remains a major challenge, as there is no satisfactory therapy for it. Understanding its high heterogeneity, and in particular its vascular heterogeneity, constitutes a key element in advancing the design of effective therapies. Therefore, it is essential to continue the study and research of this neoplasm from different perspectives: its pathology, its molecular and genetic processes, its immunology and, of course, through neuroimaging. The latter is rapidly evolving towards richer and more complex multiparametric acquisitions that require increasingly advanced computational models to harness the raw information they contain. In this sense, this thesis has contributed to the assessment and characterization of the vascular heterogeneity of the glioblastoma by means of unsupervised \ac{ML} techniques applied to \ac{MRI} data, able to discover habitats within the lesion with early prognostic capabilities.

This thesis has contributed to the state-of-the-art in the fields of Medical Informatics, Statistics and Probability, Radiology and Nuclear Medicine, Machine Learning and Data Mining and Biomedical Engineering. The scientific publications in top-ranked journals and international conferences derived from this thesis endorse the research carried out in these fields. Furthermore, the methods and technology developed in this thesis have been integrated in a public open-access platform for its use by the medical and research community, or for its posterior industrialization.

The specific concluding remarks of this thesis are listed as follows.

\begin{itemize}
\item[\textbf{CR1}] Unsupervised learning is confirmed as a viable tool for \ac{MRI} analysis and pathological pattern detection. We found that, although supervised learning normally achieves better results in well-known tasks such as image segmentation, unsupervised learning is also able to accurately capture \ac{MRI} patterns related to morphological and physiological characteristics of the lesion. The study conducted in chapter \ref{chapter:comparative_unsupervised_learning} demonstrated the potential of unsupervised learning approaches, showing a consistent behavior across the different datasets, which is a highly desirable property when dealing with heterogeneous data.

This settled the basis for the subsequent contributions carried out in the thesis, where no ground-truth exist from where to learn supervised methods. The performance obtained in this preliminary study in segmenting well-known tissues based on \ac{MR} intensity patterns gave us evidences and confidence in the following studies conducted in the thesis.

\emph{This concluding remark responds to the research question RQ1, covers the objectives O1 and O2 and was derived from the works in publications P1, P2 and P3.}

\item[\textbf{CR2}] The \ac{SVFMM} is a powerful and robust state-of-the-art framework for unsupervised learning of imaging data. The Bayesian nature of this model provides great flexibility to inject existing knowledge into the learning process, to successfully capture spatial redundancy of the images and introduce local regularization. Moreover, it also provides mechanisms to guide the learning process towards plausible solutions aligned with task-specific constraints.

The \ac{SVFMM} has been the basis for the \ac{HTS} method presented in this thesis to describe the vascular heterogeneity of glioblastomas. Particularly, we have proposed a variant of the \ac{SVFMM} combined with the probabilistic \ac{NLM} scheme that achieves better results compared to the alternative approaches in their family. The proposed approach also simplifies the previous models since the probabilistic \ac{NLM} weighting function does not introduce additional parameters to the model.

\emph{This concluding remark responds to the research question RQ2, covers the objective O3 and was derived from the work in publication P3.}

\item[\textbf{CR3}] Early-stage vascular heterogeneity provide crucial information about expected survival of patients with glioblastoma undergoing standard-of-care treatment. The \ac{HTS} habitats successfully capture this heterogeneity by analyzing the perfusion patterns within the lesion, showing improved association with \ac{OS} with respect to alternative approaches. Consistently with the literature, habitats related to the enhancing tumor presented strong correlation with patient \ac{OS}. However, the most important finding is the positive association of the $rCBV_{max}$ in the \ac{IPE} habitat with \ac{OS}. The infiltrated edema today catches all the attention since it is identified as the critical region for many decisive treatments such as surgical resection or radiotherapy, thus increasing the importance of this finding.

Moreover, the \ac{HTS} relies on a conceptual framework to describe the heterogeneity of a lesion by means of detecting habitats with differentiated \ac{MRI} functional profiles. This enables a new perspective in the characterization of complex lesions under the medical imaging paradigm. This approach introduces a concept-shift in the segmentation of lesions towards the delineation of regions sharing a similar physiological behavior rather than a common morphological appearance.

\emph{This concluding remark responds to the research questions RQ3 and RQ4, covers the objectives O4 and O5 and was derived from the works in publications P5, P6 and P7.}

\item[\textbf{CR4}] ONCOhabitats (\url{https://www.oncohabitats.upv.es}) platform encapsulates all the original methods and algorithms developed in this thesis, and several state-of-the-art algorithms for medical image analysis, in a public open-access system free for the medical and research community. ONCOhabitats is a reliable system for the study of glioblastoma that provides an end-to-end analysis of the lesion, from the preprocessing of the raw \ac{MRI} obtained from the scanner to the final measurement of volumetries and quantitative biomarkers of habitats representing regions with a specific physiological behavior.

ONCOhabitats is designed modularly to allow an easy reuse of the technology, a sustainable development cycle and a high scalability. It is mainly written using open-access state-of-the-art libraries with the aim to positioning the system as a reference platform for the analysis of brain tumors through medical imaging. In this sense, ONCOhabitats not only offers cutting edge technology, but also provides access to their computational resources, allowing a case-analysis-rate of about 300 cases per day.

The software is registered in the technological offer of the \ac{UPV} and is protected under the patents ES201431289A in Spain, EP3190542A1 in Europe and US20170287133A1 in United States.

\emph{This concluding remark responds to the research question RQ5, covers the objective O6 and was derived from the works in publications P4, P8 and P9.}
\end{itemize}

\section{Recommendations}
\label{section:recommendations}
Glioblastoma tumor still remains a lethal disease that requires a tireless multi-disciplinary effort to understand its behavior, evolution, proliferation and survival mechanisms that grant its uncontrollable aggressiveness. Under this scenario, the future of the analysis of such lethal diseases lies in the inclusion of \ac{ML} techniques, capable of analyzing the vast amount of complex multi-disciplinary medical information available to a patient, with the aim of developing personalized therapies that exploit the particularities of each individual.

The developed methods and research findings performed in this thesis points to the aforementioned direction and can serve as a starting point for further research. In this sense, the following recommendations are suggested.

\begin{itemize}
\item[\textbf{R1}] Despite the unquestionable power, utility and performance that supervised learning is demonstrating nowadays, its ability to discover new knowledge from biomedical data is severely limited by its learning mechanism. This task is, by the opposite, perfectly suited for unsupervised learning. The exploratory nature inherent in unsupervised learning provides it with the ability to detect hidden patterns within the data, often imperceptible to the human.

In this regard, we encourage the use of unsupervised learning for medical image analysis and biomedical data mining in general, to build descriptive models of the data capable of capturing subtle patterns that lead to undetectable important findings for the human being. Therefore, unsupervised learning must assume a relevant role in modern medicine to situate \ac{ML} as an essential tool for the future of personalized therapies.

\item[\textbf{R2}] Learning from structured data, such as \ac{MR} images, requires \ac{ML} models capable to exploit the conditional dependencies and spatial correlations associated to these data. Historically, \acp{MRF} have proven to be powerful mathematical models to capture the local dependencies encoded in the images. Specifically, \acp{SVFMM} constitute a versatile statistical framework to describe heterogeneous structured imaging data under a strong mathematical foundation. On the other hand, the \ac{NLM} image processing scheme has also proven to achieve state-of-the-art results in many image-related tasks such as denoising, super-resolution, in-painting or patch-based segmentation. We encourage the use of the \ac{NLSVFMM} algorithm, as it brings together the potential of both approaches in a fully Bayesian statistical model that has demonstrated comparable state-of-the-art performance in a model mathematically less complex than those of its family. 

Nevertheless, the current trend in \ac{ML} focuses on the use of \acp{CNN}, since they have demonstrated to be the state-of-the-art models for most of image analysis related tasks. \acp{CNN}, however, are mathematical models mainly oriented to the supervised learning paradigm. Therefore, we advocate the need to investigate new architectures and learning schemes to exploit the potential of \acp{CNN} in an unsupervised learning scheme. Nowadays, \acp{CAE} with latent space clustering losses are being raised as the most powerful alternatives for performing unsupervised learning segmentation in images based on \acp{CNN}.

\item[\textbf{R3}] The \ac{HTS} method settles an innovative approach to characterize the vascular heterogeneity of glioblastomas by means of detecting functional habitats within the lesion with differentiated \ac{MRI} profile. We consider that this conceptual framework provides a powerful tool to explore the internal behavior of a tumor in an objective non-biased data-driven manner. The underlying unsupervised \ac{ML} approach behind the method allows the heterogeneity of the lesion to be easily explored from different points of view, ranging from varying the number of habitats to find within the lesion, to adding additional \ac{MRI} sequences, such \ac{DWI} or \ac{NMR} relaxometry maps, to enrich the imaging profile of the tumor.

\item[\textbf{R4}] Enhancing tumor vascularity has historically demonstrated strong association with \ac{OS} of patients affected by glioblastoma. However, there is a lack of consensus in the literature on the $rCBV_{max}$ quantities that correlate with this outcome. This is probably due to the large variability in the perfusion \ac{MRI} protocols and quantification methods employed, in addition to the manual \ac{ROI} selection for tumor measurements and perfusion relative normalization techniques.

In this sense, we encourage the use of the \ac{HTS} method, as it has proven to be robust to highly variable \ac{MRI} acquisition protocols and manufacturers from different international centers, yielding $rCBV_{max}$ distributions for the different \ac{HTS} habitats with no statistically significant differences among most of the centers. This is a highly desirable property to conduct large cross-sectional population studies, where the conclusions of the experiments must be extrapolable to the entire population.

\item[\textbf{R5}] Diffuse infiltration is one of the most crucial aspects of the glioblastoma, as it renders total resection impossible and, therefore, progression after surgery is almost inevitable. Thus, detecting the areas of potentially infiltrated tumor cells is of clinical significance for many targeted interventions such as surgery or radiotherapy. The findings of this thesis related to the \ac{IPE} habitat strongly positions the \ac{HTS} method as a primary tool to study the morphology, distribution, profile and characteristics of the infiltration region. The positive association of the \ac{IPE} habitat with patient \ac{OS} confirms the importance of this habitat and enhances the \ac{HTS} as a method to study the heterogeneity of the glioblastoma.

\item[\textbf{R6}] Glioblastoma heterogeneity is evidenced at multiple levels, ranging from macroscopic co-existence of malignant tissues, to genetic alterations that derive in different glioblastoma molecular sub-types and to longitudinal evolution of mutant glioblastomas. Molecular analysis is nowadays crucial to investigate the response of the tumor to different therapies and its mechanisms of proliferation. Nevertheless, multi-parametric medical image analysis of the tumor has gained a lot of attention in the past years since it is demonstrating strong associations with relevant clinical outcomes such as \ac{OS}, tumor grading or genetic mutations.

In this sense, we recommend that further research on glioblastoma using medical imaging takes into account molecular and genetic alterations to enhance and complement the imaging information, with the goal of designing more accurate models capable of predicting clinical outcomes more reliably.

\item[\textbf{R7}] Multi-disciplinary cross-sectional research is nowadays necessary to find effective therapies for such a lethal disease as the glioblastoma. From the technological point of view, it is almost impossible for each research group to develop from scratch all the necessary state-of-the-art technology needed to analyze the different sources of information available for glioblastoma, i.e. \ac{MRI}, genetic profiling, monitoring events, electronic health records, etc.

In this regard, we encourage research groups specialized in particular topics must make an effort to provide open access to their methods to facilitate the cross-sectional research between different disciplines, since it will accelerate and improve the research on these complex and lethal diseases.
\end{itemize}

\bibliographystyle{spbasic}
{\footnotesize
\bibliography{bibliography}
}

\end{document}